\documentclass{article}

\usepackage{arxiv}

\usepackage[utf8]{inputenc} 
\usepackage[T1]{fontenc}    
\usepackage{hyperref}       
\usepackage{url}            
\usepackage{booktabs}       
\usepackage{amsfonts}       
\usepackage{nicefrac}       
\usepackage{microtype}      
\usepackage{lipsum}         
\usepackage[pdftex]{graphicx}

\usepackage{textcase}

\usepackage{amssymb}
\usepackage{amsfonts}
\usepackage{mathtools}

\PassOptionsToPackage{american}{babel} 
    \usepackage{babel}

\usepackage{csquotes}
\PassOptionsToPackage{%
  backend=biber,bibencoding=utf8, 
  language=auto,%
  style=numeric-comp,%
  giveninits=true,  
  sorting=none,
  maxbibnames=10, 
  backref=true,%
  natbib=true 
}{biblatex}
    \usepackage{biblatex}

\PassOptionsToPackage{printonlyused,smaller}{acronym}
	\usepackage{acronym} 

\usepackage[version=4]{mhchem}

\usepackage{array}
\newcolumntype{T}{>{\tiny}l} 

\usepackage{amssymb}
\usepackage{pifont}
\usepackage{physics}

\usepackage{bm}

\usepackage{listings}

\usepackage{xcolor}

\definecolor{codegreen}{rgb}{0,0.6,0}
\definecolor{codegray}{rgb}{0.5,0.5,0.5}
\definecolor{codepurple}{rgb}{0.58,0,0.82}
\definecolor{backcolour}{rgb}{0.95,0.95,0.92}

\usepackage{color}
\usepackage{nth}

\usepackage{xurl}

\usepackage[T1]{fontenc}

\usepackage{caption}
\usepackage{subcaption}
\usepackage[space]{grffile}

\usepackage{float}
\usepackage{longtable}            
\usepackage{rotating}
\usepackage{pdflscape}
\usepackage{flafter}
\usepackage{afterpage}
\usepackage{placeins}
\usepackage{geometry}

\usepackage[english]{isodate}

\usepackage[acronyms,toc=false]{glossaries-extra}
\makeglossaries
\glsdisablehyper

\newcommand{\matr}[1]{\mathbf{#1}} 
\renewcommand{\vec}[1]{\bm{#1}} 

\newcommand{\mean[1]}{\bar{#1}}

\DeclareMathOperator*{\E}{E}
\newcommand{\seeref}[2]{{#1}, \emph{see \cref{#2}};}

\newcommand{\myTitle}{Three Decades of Activations: A Comprehensive Survey of 400 Activation Functions for Neural Networks}
\newcommand{\myKeywords}{adaptive activation functions, deep learning, neural networks}

\usepackage{cleveref}       

\usepackage{kvsetkeys}

\makeatletter
\let\org@@cref\@cref
\renewcommand*{\@cref}[2]{%
  \edef\process@me{%
    \noexpand\org@@cref{#1}{\zap@space#2 \@empty}%
  }\process@me
}
\makeatother

\DeclareUnicodeCharacter{3B1}{\ensuremath{\alpha}}

\title{\myTitle}

\newif\ifuniqueAffiliation
\uniqueAffiliationtrue

\ifuniqueAffiliation 
\author{ \href{https://orcid.org/0000-0002-5130-4384}{\includegraphics[scale=0.06]{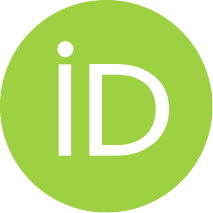}\hspace{1mm}Vladimír Kunc}
\\
\href{https://cs.fel.cvut.cz/}{\color{black}{Department of Computer Science}}\\
\href{https://fel.cvut.cz/}{\color{black}{Faculty of Electrical Engineering}}\\
		\href{https://www.cvut.cz/}{\color{black}{Czech Technical University in Prague}}\\
	\texttt{kuncvlad@fel.cvut.cz} \\
	\And
	\href{https://orcid.org/0000-0003-1753-9435}{\includegraphics[scale=0.06]{orcid.pdf}\hspace{1mm}Jiří Kléma} \\
	\href{https://cs.fel.cvut.cz/}{\color{black}{Department of Computer Science}}\\
	\href{https://fel.cvut.cz/}{\color{black}{Faculty of Electrical Engineering}}\\
	\href{https://www.cvut.cz/}{\color{black}{Czech Technical University in Prague}}\\
	\texttt{klema@fel.cvut.cz} \\
}
\else
\usepackage{authblk}

\newbox{\orcid}\sbox{\orcid}{\includegraphics[scale=0.06]{orcid.pdf}} 
\author[1]{%
	\href{https://orcid.org/0000-0002-5130-4384}{\usebox{\orcid}\hspace{1mm}Vladimír Kunc\thanks{\texttt{kuncvlad@fel.cvut.cz}}}%
}
\author[1]{%
	\href{https://orcid.org/0000-0003-1753-9435}{\usebox{\orcid}\hspace{1mm}Jiří Kléma\thanks{\texttt{klema@fel.cvut.cz}}}%
}
\affil[1]{\href{https://cs.fel.cvut.cz/}{\color{black}{Department of Computer Science}}, \href{https://fel.cvut.cz/}{\color{black}{Faculty of Electrical Engineering}}, \href{https://www.cvut.cz/}{\color{black}{Czech Technical University in Prague}}}
\fi

\renewcommand{\shorttitle}{A Comprehensive Survey of 400 Activation Functions}

\hypersetup{
pdftitle={\myTitle},
pdfauthor={Vladimír Kunc, Jiří Kléma},
pdfkeywords={\myKeywords},
}

\setabbreviationstyle[acronym]{long-short}

\newcommand{\newglossaryentrywithacronym}[4]{
    \newglossaryentry{#2_gls}{
        name={#2},
        long={#3},
        description={#4}
    }

    \newglossaryentry{#1}{
        type=\acronymtype,
        name={#1},
        short={{#1}\glsadd{#2_gls}},
        text={{#1}\glsadd{#2_gls}},
        long={{#3}\glsadd{#2_gls}},
        shortplural={{#1s}\glsadd{#2_gls}},
        longplural={{#3s}\glsadd{#2_gls}},
        description={#3},
        first={#3 (#1)\glsadd{#2_gls}},
        firstplural={#3s (#1s)\glsadd{#2_gls}},
        see=[Glossary:]{#2_gls}
    }

}

\newcommand{\newgacronymwithref}[3]{
    \newglossaryentry{#1}{
        type=\acronymtype,
        name={#1},
        short={#1},
        long={#2},
        first={#2 (#1)},
        shortplural={#1s},
        longplural={#2s},
        firstplural={#2s (#1s)},
        description={\seeref{#2}{#3}},
    }
}

\newcommand{\newacronymwithrefdesc}[4]{
    \newglossaryentry{#1}{
        type=\acronymtype,
        name={#1},
        short={#1},
        long={#2},
        first={#2 (#1)},
        shortplural={#1s},
        longplural={#2s},
        firstplural={#2s (#1s)},
        description={\seeref{#3}{#4}},
    }
}

\newcommand{\newgacronymwithdesc}[3]{
    \newglossaryentry{#1}{
        type=\acronymtype,
        name={#1},
        short={#1},
        long={#2},
        first={#2 (#1)},
        shortplural={#1s},
        longplural={#2s},
        firstplural={#2s (#1s)},
        description={#3},
    }
}

\newcommand{\newgacronymwithcustomshortdesc}[4]{
    \newglossaryentry{#1}{
        type=\acronymtype,
        name={#2},
        short={#2},
        long={#3},
        first={#3 (#2)},
        shortplural={#2s},
        longplural={#3s},
        firstplural={#3s (#2s)},
        description={#4},
    }
}

\newcommand{\newglossaryentrywithacronymandref}[5]{
    \newglossaryentry{#2_gls}{
        name={#2},
        long={#3},
        description={\seeref{#4}{#5}},
    }

    \newglossaryentry{#1}{
        type=\acronymtype,
        name={#1},
        short={{#1}\glsadd{#2_gls}},
        shortplural={{#1}s\glsadd{#2_gls}},
        text={{#1}\glsadd{#2_gls}},
        long={{#3}\glsadd{#2_gls}},
        longplural={{#3}s\glsadd{#2_gls}},
        description={\seeref{#3}{#5}},
        first={#3 (#1)\glsadd{#2_gls}},
        firstplural={#3s (#1s)\glsadd{#2_gls}},
        see=[Glossary:]{#2_gls}
    }
}

\begingroup
  \boolfalse{citerequest}

\newglossaryentry{WoS}{
    type=\acronymtype,
    name={WoS},
    short={WoS},
    long={Web of Science},
    description={\href{https://www.webofscience.com/wos}{Web of Science}},
    first={Web of Science (WoS)},
}

\newglossaryentrywithacronym{JIF}{Journal Impact Factor}{Journal Impact Factor}{a journal citation metric calculated by \href{https://clarivate.com/}{Clarivate} using items indexed in \glsxtrshort{WoS}; it is equal to the mean number of citations for articles published in the last two years\footnote{see \url{https://incites.help.clarivate.com/Content/Indicators-Handbook/ih-journal-citation-reports.htm}}}
\newglossaryentrywithacronym{JCI}{Journal Citation Indicator}{Journal Citation Indicator}{a journal citation metric calculated by \href{https://clarivate.com/}{Clarivate} using items indexed in \glsxtrshort{WoS}; it is equal to the average \glsxtrshort{CNCI} published in the last three years}
\newglossaryentrywithacronym{CNCI}{Category Normalized Citation Impact}{Category Normalized Citation Impact}{a document citation metric calculated by \href{https://clarivate.com/}{Clarivate}; it is equal to the number of a document's citations normalized by the expected number of citations for the same document type, publication year, and subject area \footnote{see \url{https://incites.help.clarivate.com/Content/Indicators-Handbook/ih-normalized-indicators.htm}}}

\newglossaryentry{CiteScore}{
    name={CiteScore},
    description={a journal citation metric calculated by \href{https://www.elsevier.com/}{Elsevier} using items indexed in \href{https://www.scopus.com/}{Scopus}; it is equal to the average number of citations per document published in the last four years \footnote{see \url{https://service.elsevier.com/app/answers/detail/a_id/14880/supporthub/scopus/}}},
}

\newglossaryentrywithacronym{A}{adenine}{adenine}{a purine nucleobase. One of the four bases in the \glsxtrshort{DNA}.}
\newglossaryentrywithacronym{G}{guanine}{guanine}{a purine nucleobase. One of the four bases in the \glsxtrshort{DNA}.}
\newglossaryentrywithacronym{C}{cytosine}{cytosine}{a pyrimidine nucleobase. One of the four bases in the \glsxtrshort{DNA}.}
\newglossaryentrywithacronym{T}{thymine}{thymine}{a pyrimidine nucleobase. One of the four bases in the \glsxtrshort{DNA}.}
\newglossaryentrywithacronym{DNA}{DNA}{deoxyribonucleic acid}{an extended molecular structure for storing hereditary information}
\newglossaryentrywithacronym{RNA}{RNA}{ribonucleic acid}{an extended molecular structure; stores hereditary information but also can catalyze biological reactions and control \glsxtrlong{GE} among other things}
\newglossaryentrywithacronym{GE}{gene expression}{gene expression}{process of synthesizing gene product (e.g., a protein) using information encoded in a gene}
\newglossaryentrywithacronym{cDNA}{cDNA}{copy DNA}{copy \glsxtrshort{DNA}, also called complementary \glsxtrshort{DNA}; synthetic DNA transcribed from \glsxtrshort{mRNA} \cite{Jaksik2015} using reverse transcriptase \cite[19]{BolonCanedo2019}}
\newglossaryentrywithacronym{cRNA}{cRNA}{copy RNA}{copy \glsxtrshort{RNA}, transcribed from \glsxtrshort{cDNA} during amplification phase}
\newglossaryentrywithacronym{mRNA}{messenger RNA}{messenger RNA}{copied from \glsxtrshort{DNA} during transcription; used for protein synthesis during translation}
\newglossaryentrywithacronym{miRNA}{microRNA}{microRNA}{a small, non-coding \glsxtrshortpl{RNA} containing 21 -- 28 \glsxtrlongpl{nt}}
\newglossaryentrywithacronym{ERC}{external RNA control}{external RNA control}{an approach for \textit{microarray performance assessment}; see \cite{Tong2006} for more details}

\newacronym{GO}{GO}{gene ontology}
\newacronym{DE}{DE}{differentially expressed}
\newacronym{GRN}{GRN}{gene regulatory network}
\newgacronymwithdesc{GRNN}{gene regulatory \glsentrylong{NN}}{gene regulatory\glsxtrlong{NN}}
\newgacronymwithdesc{QNN}{quadratic \glsentrylong{NN}}{quadratic \glsxtrlong{NN}}

\newacronym{ATAC-seq}{ATAC-seq}{Assay for Transposase-Accessible Chromatin using sequencing}
\newgacronymwithdesc{scATAC-Seq}{single-cell \glsentryshort{ATAC-seq}}{single-cell \glsxtrshort{ATAC-seq}}
\newacronym{snmC-Seq}{snmC-Seq}{single-nucleus methylcytosine sequencing}
\newacronym{scVI}{scVI}{single-cell variational inference}

\newglossaryentrywithacronym{GeNN}{genetic neural network}{genetic neural network}{a neural network architecture for \glsxtrlong{GE} tasks presented in \cite{Eetemadi2018}}
\newglossaryentrywithacronym{LinGeNN}{linear GeNN}{linear GeNN}{a variant of \glsxtrshort{GeNN} with linear activation function \cite{Eetemadi2018}}

\newglossaryentry{dgea_gls}{
    name={differential gene expression analysis},
    short={DGE analysis},
    long={differential gene expression analysis},
    description={a commonly used computational approach for identifying genes whose expressions are significantly different between two phenotypes \cite{Abbas2020}},
}

\newglossaryentry{dgea}{
    type=\acronymtype,
    name={DGE analysis},
    short={DGE analysis},
    long={differential gene expression analysis},
    description={differential gene expression analysis},
    first={differential gene expression (DGE) analysis\glsadd{dgea_gls}},
    plural={differential gene expression analyses},
    shortplural={DGE analyses},
    firstplural=differential gene expression (DGE) analyses\glsadd{dgea_gls},
    see=[Glossary:]{dgea_gls}
}

\newglossaryentry{DGEX}{
    name={D--GEX},
    description={\seeref{a \glsxtrlong{NN} for \glsxtrlong{GE} inference \cite{Chen2016b}}{sec:dgex}},
}
\newglossaryentry{dnamicroarray}{
    name={DNA microarray},
    description={used for measuring \glsxtrshort{DNA} levels \cite{Sealfon2010}; mainly for \glsxtrlong{GE} analysis},
}

\newglossaryentry{rnamicroarray}{
    name={RNA microarray},
    description={used for measuring \glsxtrshort{RNA} levels \cite{Sealfon2010}},
}
\newglossaryentry{microarray}{
    name={microarray},
    description={used for measuring, usually \glsxtrshort{DNA} or \glsxtrshort{RNA} levels \cite{Sealfon2010}; called \gls{dnamicroarray} when measuring \glsxtrshort{DNA} levels and \gls{rnamicroarray} when measuring \glsxtrshort{RNA} levels  \cite{Sealfon2010}; see \cite{Aparna2023} for other types},
}
\newglossaryentry{hybridization}{
    name={hybridization},
    description={a process in which a \glsxtrshort{cDNA} binds to the probes on the \gls{microarray} surface},
}

\newglossaryentry{L1000}{
    name={L1000},
    description={a cost-efficient \glsxtrlong{GE} assay, see \cite{Subramanian2017}},
}

\newglossaryentry{Hi-C}{
    name={Hi-C},
    description={high-throughput method for detection of chromatin interactions},
}

\newglossaryentry{k-means}{
    name={k-means},
    description={a greedy clustering algorithm},
}

\newglossaryentry{lasso}{
    name={lasso},
    description={a \glsxtrlong{LR} variant with $l_1$ regularization; more in \cite{Tibshirani1996}},
}

\newglossaryentrywithacronym{BP}{backpropagation}{backpropagation}{a method for calculation of the gradient of the loss function w.r.t. the network's weight}
\newglossaryentrywithacronym{SGD}{stochastic gradient descent}{stochastic gradient descent}{a method for optimization of an objective function; it is a variant of gradient descent that uses stochastic batches of data instead of the entire dataset to calculate the gradient}

\newglossaryentry{Adam}{
    name={Adam},
    description={a popular variant of the \glsxtrshort{SGD} optimization algorithm; more in \cite{Kingma2014}},
}

\newglossaryentrywithacronym{rRNA}{ribosomal RNA}{ribosomal RNA}{a primary component of ribosomes; non-coding \glsxtrshort{RNA}}
\newglossaryentrywithacronym{RNA-Seq}{RNA-Seq}{RNA sequencing}{ a \glsxtrshort{NGS} \glsxtrshort{RNA} sequencing method; allows for measuring \glsxtrlong{GE} levels}
\newgacronymwithdesc{scRNA-Seq}{single-cell \glsentryshort{RNA-Seq}}{single-cell \glsxtrshort{RNA-Seq}}
\newglossaryentrywithacronym{GEO}{Gene Expression Omnibus}{Gene Expression Omnibus}{a central and public repository of high-throughput \glsxtrlong{GE} data by \glsxtrshort{NCBI}, more in \cite{Edgar2002}}
\newglossaryentrywithacronym{PCA}{principal component analysis}{principal component analysis}{a dimensionality reduction method by linear transformation into a new coordinate system that respects the variance in the data}

\newglossaryentrywithacronym{RNN}{recurrent neural network}{recurrent neural network}{a type of \glsxtrlong{NN} that contains loop in the information flow}

\newglossaryentrywithacronym{BRNN}{bidirectional recurrent neural network}{bidirectional recurrent neural network}{a type of \glsxtrlong{RNN} where the outputs can use information from both past and future states, more in \cite{Schuster1997}}
\newacronym{gcd}{GCD}{Greatest Common Divisor}
\newglossaryentrywithacronymandref{ANN}{artificial neural network}{artificial neural network}{a biologically inspired computational model, interchangeably used with the term \glsxtrlong{NN}}{ch:background_nn}
\newglossaryentrywithacronymandref{NN}{neural network}{neural network}{a biologically inspired computational model, interchangeably used with the term \glsxtrlong{ANN}}{ch:background_nn}
\newglossaryentrywithacronym{GPU}{GPU}{graphics processing unit }{a specialized piece of hardware initially designed to accelerate image processing and computer graphics in general; often used for acceleration of training and inference of \glsxtrlongpl{NN}}

\newacronym{CPU}{CPU}{central processing unit}
\newacronym{LSTM}{LSTM}{long short-term memory}
\newacronym{BiLSTM}{BiLSTM}{bidirectional long short-term memory}
\newacronym{CNN}{CNN}{convolutional neural network}
\newacronym{GMDH}{GMDH}{group method of data handling}
\newacronym{KNN}{KNN}{k--nearest neighbor}
\newacronym{MNN}{MNN}{mutual nearest neighbor}
\newacronym[see=NN]{DKNN}{DKNN}{Deep Kronecker neural network}
\newacronym{FISH}{FISH}{fluorescence in situ hybridization}
\newacronym{nt}{nt}{nucleotide}
\newacronym{NGS}{NGS}{next-generation sequencing}
\newacronym{NCBI}{NCBI}{National Center for Biotechnology Information}
\newacronym{DL}{DL}{deep learning}
\newacronym[see=NN]{GNN}{GNN}{graph neural networks}
\newacronym{MR}{MR}{master regulator}
\newacronym{LP}{LP}{linear programming}
\newacronym{LR}{LR}{linear regression}
\newacronym{TRN}{TRN}{transcriptional regulatory network}
\newacronym{MLP}{MLP}{multi-layer perceptron}
\newacronym{NLP}{NLP}{natural language processing}
\newacronym{CDF}{CDF}{cumulative distribution function}
\newacronym{ML}{ML}{machine learning}
\newacronym{MP}{MP}{max-pooling}
\newacronym{ILSVRC}{ILSVRC}{ImageNet Large-Scale Visual Recognition Challenge}
\newacronym{NO}{NO}{neural operator}
\newacronym{NFT}{NFT}{non-fungible token}

\newglossaryentrywithacronym{SHAP}{Shapley additive explanations}{Shapley additive explanations}{a game theoretic approach for explaining the output of a \glsxtrlong{ML} model}

\newacronym{AF}{AF}{activation function}
\newglossaryentrywithacronym{AAF}{adaptive activation function}{adaptive \glsxtrlong{AF}}{an \glsxtrlong{AF} that can adapt to the data; often, it has a parameter whose value is data-dependent}
\newglossaryentrywithacronym{TAF}{trainable activation function}{trainable \glsxtrlong{AF}}{an \glsxtrlong{AF}; another name for the \glsxtrlong{AAF} (\glsxtrshort{AAF})}

\newglossaryentry{sigmoid}{
    name={sigmoid},
    long={sigmoid},
    description={a mathematical function having \textit{S-shaped} curve, \gls{logisticsigmoid} is the most known example},
}

\newglossaryentry{probit}{
    name={probit},
    description={\seeref{another name for the cumulative standard distribution function when used as \glsxtrlong{AF}}{sec:logistic_sigmoid}},
}
\newglossaryentry{logisticsigmoid}{
    name={logistic sigmoid},
    description={\seeref{one of most common \glsxtrlongpl{AF}  used in \glsxtrshortpl{NN}}{sec:logistic_sigmoid}},
}

\newglossaryentry{improvedlogisticsigmoid}{
    name={improved logistic sigmoid},
    description={\seeref{an \glsxtrlong{AF} based on the \gls{logisticsigmoid}}{sec:improved_logistic_sigmoid}},
}

\newglossaryentry{SigLin}{
    name={SigLin},
    description={\seeref{an \glsxtrlong{AF}; a combination of the \gls{logisticsigmoid} and a linear function}{sec:siglin}},
}

\newglossaryentry{scaledlogisticsigmoid}{
    name={scaled logistic sigmoid},
    description={\seeref{an \glsxtrlong{AF} based on the \gls{logisticsigmoid}}{sec:scaled_logistic_sigmoid}},
}
\newacronymwithrefdesc{LRTanh}{linearized \glsentrylong{tanh} unit}{linearized \glsxtrlong{tanh}}{sec:logistic_sigmoid}
\newglossaryentrywithacronymandref{tanh}{hyperbolic tangent}{hyperbolic tangent}{one of most common \glsxtrlongpl{AF}  used in \glsxtrshortpl{NN}}{sec:logistic_sigmoid}
\newgacronymwithref{SSS}{shifted and scaled sigmoid}{sec:sss}
\newacronymwithrefdesc{VSF}{variant \gls{sigmoid} function}{variant \gls{sigmoid} function}{sec:vsf}
\newglossaryentrywithacronymandref{stanh}{scaled hyperbolic tangent}{scaled hyperbolic tangent}{a scaled variant of \glsxtrlong{tanh}}{sec:stanh}
\newacronymwithrefdesc{pRPPSG}{p-recursive piecewise polynomial sigmoid generator}{p-recursive piecewise polynomial \gls{sigmoid} generator}{sec:logistic_sigmoid}

\newglossaryentrywithacronymandref{arctan}{arctangent}{arctangent}{the inverse of the tangent function, also used as an \glsxtrlong{AF}}{sec:arctan}

\newglossaryentry{rectifiedhyperbolicsecant}{
    name={rectified hyperbolic secant},
    description={\seeref{an \glsxtrlong{AF} based on the hyperbolic secant}{sec:resech}},
}
\newgacronymwithref{binary AF}{binary activation function}{sec:binaryaf}
\newgacronymwithref{LiSHT}{linearly scaled hyperbolic tangent}{sec:lisht}
\newgacronymwithref{PSGU}{Parameterized self-circulating gating unit}{sec:psgu}
\newglossaryentry{PATS}{
    name={PATS},
    description={\seeref{an \glsxtrlong{AF}; not an abbreviation}{sec:pats}},
}
\newgacronymwithref{SiLU}{sigmoid-weighted linear unit}{sec:silu}
\newgacronymwithref{WiG}{weighted sigmoid gate unit}{sec:silu}
\newacronymwithrefdesc{TS-swish}{triple-state \gls{swish} unit}{triple-state \gls{swish} unit}{sec:tsswish}
\newacronymwithrefdesc{TS-sigmoid}{triple-state \gls{sigmoid} unit}{triple-state \gls{sigmoid} unit}{sec:tssigmoid}
\newgacronymwithref{GLU}{gated linear unit}{sec:glu}
\newacronymwithrefdesc{GTU}{gated \glsentryshort{tanh} unit}{gated \glsxtrshort{tanh} unit}{sec:gtu}
\newacronymwithrefdesc{ReGLU}{gated \glsentryshort{ReLU}}{gated \glsxtrshort{ReLU}}{sec:reglu}
\newacronymwithrefdesc{GEGLU}{gated \glsentryshort{GELU}}{gated \glsxtrshort{GELU}}{sec:geglu}
\newacronymwithrefdesc{SwiGLU}{gated \gls{swish}}{gated \gls{swish}}{sec:swiglu}
\newacronymwithrefdesc{dSiLU}{derivative of  \glsentrylong{SiLU}}{derivative of  \glsxtrlong{SiLU}}{sec:dsilu}
\newacronymwithrefdesc{DoubleSiLU}{double \glsentrylong{SiLU}}{double \glsxtrlong{SiLU}}{sec:double_silu}
\newacronymwithrefdesc{TSiLU}{\glsentrylong{tanh} \glsentrylong{SiLU}}{\glsxtrlong{tanh} \glsxtrlong{SiLU}}{sec:tsilu}
\newacronymwithrefdesc{ATSiLU}{arctan \glsentrylong{SiLU}}{arctan \glsxtrlong{SiLU}}{sec:atsilu}
\newglossaryentry{SwAT}{
    name={SwAT},
    description={\seeref{an \glsxtrlong{AF} combining \gls{logisticsigmoid} and arctan; not an abbreviation}{sec:swat}},
}
\newglossaryentry{E-swish}{
    name={E-swish},
    description={\seeref{an \glsxtrlong{AF} based on the \gls{swish}}{sec:eswish}},
}
\newacronymwithrefdesc{T-swish}{tunable \gls{swish}}{arctan \gls{swish}}{sec:tswish}
\newacronymwithrefdesc{RePSU}{rectified parametric \gls{sigmoid} unit}{rectified parametric \gls{sigmoid} unit}{sec:repsu}
\newacronymwithrefdesc{RePSKU}{rectified parametric \gls{sigmoid} shrinkage unit}{rectified parametric \gls{sigmoid} shrinkage unit}{sec:repsu}
\newacronymwithrefdesc{RePSHU}{rectified parametric \gls{sigmoid} stretchage unit}{rectified parametric \gls{sigmoid} stretchage unit}{sec:repsu}
\newacronymwithrefdesc{SSBS}{smooth \gls{sigmoid}-based shrinkage}{smooth \gls{sigmoid}-based shrinkage}{sec:repsu}
\newgacronymwithref{AQuLU}{adaptive quadratic linear unit}{sec:aqulu}
\newgacronymwithref{SinLU}{sinu-sigmoidal linear unit}{sec:sinlu}
\newglossaryentry{ErfAct}{
    name={ErfAct},
    description={\seeref{an \glsxtrlong{AAF} based on the \gls{swish}}{sec:erfact}},
}

\newacronymwithrefdesc{pserf}{parametric serf}{parametric \gls{serf}}{sec:pserf}
\newglossaryentry{swim}{
    name={swim},
    description={\seeref{an \glsxtrlong{AAF} similar to the \gls{swish}}{sec:swim}},
}

\newgacronymwithref{MSiLU}{modified sigmoid-weighted linear unit}{sec:msilu}
\newgacronymwithref{ptanh}{penalized hyperbolic tangent}{sec:penalized_hyperbolic_tangent}
\newgacronymwithref{SLS-SS}{scaled logistic sigmoid with scaled sine}{sec:scaled_logistic_sigmoid}
\newgacronymwithref{SRS}{soft-root-sign}{sec:srs}
\newgacronymwithref{SC}{soft clipping}{sec:sc}
\newglossaryentry{hexpo}{
    name={Hexpo},
    description={\seeref{an \glsxtrlong{AF}}{sec:hexpo}},
}
\newglossaryentry{Elliott}{
    name={Elliott},
    description={\seeref{an \glsxtrlong{AF}}{sec:elliott}},
}
\newglossaryentrywithacronymandref{ReLU}{ReLU}{rectified linear unit}{one of the most popular \glsxtrlongpl{AF}}{sec:relu}

\newglossaryentry{softmax}{
    name={softmax},
    description={\seeref{a popular \glsxtrlong{AF} for classification problems; outputs a soft argmax of outputs of a given layer}{sec:softmax}},
}
\newacronymwithrefdesc{tsoftmax}{tuned softmax}{tuned \gls{softmax}}{sec:tsoftmax}
\newacronymwithrefdesc{glsoftmax}{generalized Lehmer softmax}{generalized Lehmer \gls{softmax}}{sec:glsoftmax}
\newacronymwithrefdesc{gpsoftmax}{generalized power softmax}{generalized power \gls{softmax}}{sec:gpsoftmax}
\newglossaryentry{betasoftmax}{
    name={\texorpdfstring{$\beta$}{Beta}-softmax},
    description={\seeref{an extension of the \gls{softmax} \gls{AF}}{sec:betasoftmax}},
}
\newglossaryentry{shifted_relu}{
    name={Shifted ReLU},
    description={\seeref{an \glsxtrlong{AF}; translated \glsxtrshort{ReLU}}{sec:shifted_relu}},
}
\newglossaryentrywithacronymandref{ShiLU}{ShiLU}{adaptive shifted ReLU}{a \glsxtrshort{ReLU} based \glsxtrlong{AF}; not to be confused with \gls{shifted_relu}}{sec:shilu}
\newglossaryentrywithacronymandref{LReLU}{leaky ReLU}{leaky ReLU}{a popular \glsxtrshort{ReLU} based \glsxtrlong{AF}; allows information flow for negative inputs unlike \glsxtrshort{ReLU}}{sec:lrelu}
\newglossaryentrywithacronymandref{RReLU}{randomized leaky ReLU}{randomized leaky ReLU}{a \glsxtrshort{LReLU} based \glsxtrlong{AF} with stochastic leakiness during training}{sec:rrelu}
\newglossaryentrywithacronymandref{S-RReLU}{softsign randomized leaky ReLU}{softsign randomized leaky ReLU}{a \glsxtrshort{RReLU} based \glsxtrlong{AF} combined with \gls{softsign}}{sec:srrelu}
\newacronymwithrefdesc{MarReLU}{margin \glsentryshort{PReLU}}{margin\glsxtrshort{PReLU}}{sec:marrelu}
\newacronymwithrefdesc{FunPReLU}{funnel \glsentrylong{PReLU}}{funnel \glsxtrlong{PReLU}}{sec:funprelu}
\newacronymwithrefdesc{FunReLU}{funnel \glsentrylong{ReLU}}{funnel \glsxtrlong{ReLU}}{sec:funprelu}
\newacronymwithrefdesc{RPReLU}{react-\glsentryshort{PReLU}}{react-\glsxtrshort{PReLU}}{sec:rprelu}

\newgacronymwithref{SAU}{smooth activation unit}{sec:sau}
\newgacronymwithref{SMU}{smooth maximum unit}{sec:smu}
\newglossaryentry{SMU1}{
    name={SMU-1},
    description={\seeref{an \glsxtrlong{AF}; a variant of the \glsxtrshort{SMU} using a different smoothing approach}{sec:lreluplus}},
}
\newacronymwithrefdesc{VLReLU}{very \glsentrylong{LReLU}}{very \glsxtrlong{LReLU}}{sec:lrelu}
\newacronymwithrefdesc{OLReLU}{optimized \glsentrylong{LReLU}}{optimized \glsxtrlong{LReLU}}{sec:lrelu}
\newacronymwithrefdesc{SlReLU}{Sloped ReLU}{Sloped \glsxtrshort{ReLU}}{sec:slrelu}
\newacronymwithrefdesc{NReLU}{noisy ReLU}{noisy \glsxtrshort{ReLU}}{sec:nrelu}
\newglossaryentry{SineReLU}{
    name={SineReLU},
    description={\seeref{an \glsxtrlong{AF}; extension of \glsxtrshort{ReLU}}{sec:sinerelu}},
}
\newglossaryentry{minsin}{
    name={minsin},
    description={\seeref{an \glsxtrlong{AF}}{sec:minsin}},
}
\newgacronymwithref{VLU}{variational linear unit}{sec:vlu}
\newacronymwithrefdesc{PVLU}{parametric \glsentrylong{VLU}}{parametric \glsxtrlong{VLU}}{sec:pvlu}
\newgacronymwithref{SCAA}{spatial context-aware activation}{sec:scaa}
\newacronymwithrefdesc{FReLU}{flexible ReLU}{flexible \glsxtrshort{ReLU}}{sec:frelu}
\newglossaryentry{StarReLU}{
    name={StarReLU},
    description={\seeref{an \glsxtrlong{AF}; extension of \glsxtrshort{ReLU}}{sec:starrelu}},
}
\newacronymwithrefdesc{AReLU}{Attention-based ReLU}{Attention-based \glsxtrshort{ReLU}}{sec:arelu}

\newacronymwithrefdesc{DPReLU}{dual parametric ReLU}{dual parametric \glsxtrshort{ReLU}}{sec:dprelu}
 \newglossaryentry{dual_line}{
    name={Dual Line},
    description={\seeref{an \glsxtrlong{AF}; extension of \glsxtrshort{DPReLU}}{sec:dual_line}},
}
\newgacronymwithref{PiLU}{piecewise linear unit}{sec:pilu}
\newglossaryentrywithacronymandref{DPAF}{dual parametric activation function}{dual parametric activation function}{an \glsxtrlong{AF} inspired by \glsxtrshort{DPReLU}}{sec:dpaf}
\newglossaryentrywithacronymandref{FPAF}{fully parameterized activation function}{fully parameterized activation function}{an \glsxtrlong{AF} similar to \glsxtrshort{DPAF}}{sec:fpaf}
\newacronymwithrefdesc{EPReLU}{Elastic \glsentryshort{PReLU}}{Elastic \glsxtrshort{PReLU}}{sec:eprelu}
\newglossaryentry{RT-ReLU}{
    type=\acronymtype,
    name={RT--ReLU},
    short={RT--ReLU},
    long={randomly translational ReLU},
    first={randomly translational ReLU (RT--ReLU)},
    description={\seeref{randomly translational ReLU}{sec:rtrelu}},
}
\newgacronymwithref{NLReLU}{natural-logarithm-ReLU}{sec:nlrelu}
\newgacronymwithref{SLU}{softplus linear unit}{sec:slu}
\newgacronymwithref{ReSP}{rectified softplus}{sec:resp}
\newgacronymwithref{PReNU}{parametric rectified non-linear unit}{sec:prenu}
\newglossaryentrywithacronymandref{BReLU}{bounded ReLU}{bounded ReLU}{an \glsxtrlong{AF}; \glsxtrshort{ReLU} variant with bounds}{sec:brelu}
\newglossaryentry{hard_sigmoid}{
    name={Hard sigmoid},
    description={\seeref{an \glsxtrlong{AF} similar to \glsxtrshort{BReLU} with \gls{sigmoid}-like shape}{sec:hard_sigmoid}},
}
\newglossaryentry{hard_tanh}{
    name={HardTanh},
    description={\seeref{an \glsxtrlong{AF} similar to \gls{hard_sigmoid} with \glsxtrshort{tanh}-like shape}{sec:hard_tanh}},
}
\newglossaryentry{SvHardTanh}{
    type=\acronymtype,
    name={SvHardTanh},
    short={SvHardTanh},
    long={SvHardTanh},
    description={\seeref{\gls{hard_tanh} with a fixed vertical shift}{sec:shifted_hardtanh}},
}
\newglossaryentry{ShHardTanh}{
    type=\acronymtype,
    name={ShHardTanh},
    short={ShHardTanh},
    long={ShHardTanh},
    description={\seeref{\gls{hard_tanh} with a fixed horizontal shift}{sec:shifted_hardtanh}},
}
\newglossaryentry{hardshrink}{
    name={Hardshrink},
    description={\seeref{an \glsxtrlong{AF} similar to \gls{hard_sigmoid}}{sec:hardshrink}},
}
\newglossaryentry{softshrink}{
    name={Softshrink},
    description={\seeref{an \glsxtrlong{AF} similar to \gls{hard_sigmoid}}{sec:softshrink}},
}
\newgacronymwithref{TRec}{truncated rectified}{sec:trec}
\newglossaryentry{hardswish}{
    name={Hard-Swish},
    description={\seeref{an \glsxtrlong{AF} similar to \gls{hard_sigmoid} related to the \gls{swish} function}{sec:hardswish}},
}
\newglossaryentrywithacronymandref{BLReLU}{bounded leaky ReLU}{bounded leaky ReLU}{an \glsxtrlong{AF}; \glsxtrshort{ReLU} variant combining \glsxtrshort{BReLU} and \glsxtrshort{LReLU}}{sec:blrelu}
\newgacronymwithref{PLU}{piecewise linear unit}{sec:plu}
\newgacronymwithref{AdaLU}{adaptive linear unit}{sec:adalu}
\newacronymwithrefdesc{TSAF}{trapezoid-shaped \glsentrylong{AF}}{trapezoid-shaped \glsxtrlong{AF}}{sec:tsaf}
\newgacronymwithref{ARiA}{Adaptive Richard's curve weighted activation}{sec:aria}
\newglossaryentrywithacronymandref{ARiA2}{ARiA2}{Adaptive Richard's curve weighted activation 2}{an \glsxtrlong{AF}; special case of \glsxtrshort{ARiA}}{sec:aria}
\newgacronymwithref{MTLU}{multi-bin trainable linear unit}{sec:mtlu}
\newgacronymwithref{PWLU}{piecewise linear unit}{sec:mtlu}
\newacronymwithrefdesc{N-PWLU}{non-uniform \glsentrylong{PWLU}}{non-uniform \glsxtrlong{PWLU}}{sec:mtlu}
\newacronymwithrefdesc{CPN}{continuous piecewise nonlinear activation function}{continuous piecewise nonlinear \glsxtrlong{AF}}{sec:cpn}
\newgacronymwithref{LuTU}{look-up table unit}{sec:lutu}
\newglossaryentry{maxout}{
    name={maxout unit},
    description={\seeref{an \glsxtrlong{AF} returning the maximum of several linear functions}{sec:maxout}},
}
\newgacronymwithref{KAF}{kernel activation function}{sec:kaf}
\newacronym{SAVE}{SAVE}{structured activation of vertex entropy}
\newgacronymwithref{vReLU}{V-shaped ReLU}{sec:vrelu}
\newglossaryentry{pan}{
    name={pan},
    description={\seeref{a piecewise linear \glsxtrlong{AF}}{sec:pan}},
}
\newgacronymwithref{AbsLU}{absolute linear unit}{sec:abslu}
\newacronymwithrefdesc{mReLU}{mirrored \glsentrylong{ReLU}}{mirrored \glsxtrlong{ReLU}}{sec:mrelu}
\newgacronymwithref{LRTLU}{leaky rectified triangle linear unit}{sec:lsptlu}
\newgacronymwithref{LSPTLU}{leaky single-peaked triangle linear unit}{sec:lsptlu}
\newglossaryentry{tent}{
    name={tent},
    description={\seeref{an \glsxtrshort{ReLU}-based \glsxtrlong{AF}}{sec:tent}},
}
\newglossaryentry{hat}{
    name={hat},
    description={\seeref{an \glsxtrshort{ReLU}-based \glsxtrlong{AF}}{sec:hat}},
}
\newglossaryentry{softmodulusq}{
    name={SoftModulusQ},
    description={\seeref{an \glsxtrlong{AF}; a quadratic approximation of the \glsxtrshort{vReLU}}{sec:softmodulusq}},
}
\newglossaryentry{softmodulust}{
    name={SoftModulusT},
    description={\seeref{an \glsxtrlong{AF}; a \glsxtrshort{tanh} based approximation of the \glsxtrshort{vReLU}}{sec:softmodulust}},
}
\newglossaryentry{SignReLU}{
    name={SignReLU},
    description={\seeref{an \glsxtrlong{AF}; a combination of \glsxtrshort{ReLU} and \gls{softsign}}{sec:signrelu}},
}
\newglossaryentry{Li-ReLU}{
    name={Li-ReLU},
    description={\seeref{an \glsxtrlong{AF}; a combination of a linear function and \glsxtrshort{ReLU} }{sec:lirelu}},
}
\newglossaryentry{DLU}{
    name={DLU},
    description={\seeref{different name for \gls{SignReLU} used in \cite{Li2022ANewActivation,Pan2023Smoothing}}{sec:signrelu}},
}
\newacronymwithrefdesc{CReLU}{concatenated \glsentryshort{ReLU}}{concatenated \glsxtrshort{ReLU}}{sec:crelu}
\newacronymwithrefdesc{NCReLU}{negative \glsentryshort{CReLU}}{negative \glsxtrshort{CReLU}}{sec:ncrelu}
\newacronymwithrefdesc{BAF}{bipolar activation function}{bipolar \glsxtrlong{AF}}{sec:ncrelu}
\newglossaryentry{dualrelu}{
    name={DualReLU},
    description={\seeref{an \glsxtrlong{AF}; a two-dimensional \glsxtrshort{ReLU} variant}{sec:dualrelu}},
}
\newgacronymwithref{OPLU}{orthogonal permutation liner unit}{sec:oplu}
\newglossaryentry{power_activation_function}{
    name={power activation function},
    description={\seeref{an \glsxtrlong{AF}; also known as \glsxtrshort{RePU}}{sec:repu}},
}

\newacronymwithrefdesc{EReLU}{elastic \glsentryshort{ReLU}}{elastic \glsxtrshort{ReLU}}{sec:erelu}
\newgacronymwithref{RePU}{rectified power unit}{sec:repu}
\newacronymwithrefdesc{AppReLU}{approximate \glsentryshort{ReLU}}{approximate \glsxtrshort{ReLU}}{sec:apprelu}
\newacronymwithrefdesc{PLAF}{power linear \glsentrylong{AF}}{power linear \glsxtrlong{AF}}{sec:plaf}
\newacronymwithrefdesc{EPLAF}{even \glsentrylong{PLAF}}{even \glsxtrlong{PLAF}}{sec:plaf}
\newacronymwithrefdesc{OPLAF}{odd \glsentrylong{PLAF}}{odd \glsxtrlong{PLAF}}{sec:plaf}
\newacronymwithrefdesc{ABReLU}{average biased \glsentryshort{ReLU}}{average biased \glsxtrshort{ReLU}}{sec:abrelu}
\newacronymwithrefdesc{DRLU}{delay \glsentryshort{ReLU}}{delay \glsxtrshort{ReLU}}{sec:drlu}
\newgacronymwithref{AOAF}{adaptive offset activation function}{sec:aoaf}
\newgacronymwithref{DLReLU}{dynamic leaky ReLU}{sec:dlrelu}
\newgacronymwithref{DReLU}{dynamic ReLU}{sec:drelu}
\newgacronymwithref{DisReLU}{displaced ReLU}{sec:disrelu}
\newacronymwithrefdesc{MLReLU}{modified \glsentryshort{LReLU}}{modified \glsxtrshort{LReLU}}{sec:mlrelu}
\newgacronymwithref{FTS}{flatted-T swish}{sec:fts}
\newglossaryentry{ReLUSwish}{
    name={ReLU-Swish},
    description={\seeref{an \glsxtrlong{AF}; special case of \glsxtrshort{FTS}}{sec:fts}},
}
\newgacronymwithref{OAF}{Optimal Activation Functio}{sec:oaf}
\newglossaryentrywithacronymandref{ELU}{exponential linear unit}{exponential linear unit}{a popular \glsxtrlong{AF} extending \glsxtrshort{ReLU}}{sec:elu}
\newgacronymwithref{REU}{rectified exponential unit}{sec:reu}
\newgacronymwithref{ADA}{apical dendrite activation}{sec:ada}
\newacronymwithrefdesc{LADA}{leaky \glsentrylong{ADA}}{leaky \glsxtrlong{ADA}}{sec:lada}
\newgacronymwithref{SigLU}{sigmoid linear unit}{sec:siglu}
\newacronymwithrefdesc{SaRa}{\gls{swish} and \glsentryshort{ReLU} activation}{\gls{swish} and \glsxtrshort{ReLU} activation}{sec:sara}
\newglossaryentry{maxsig}{
    name={maxsig},
    description={\seeref{an \glsxtrlong{AF}}{sec:maxsig}},
}
\newacronymwithrefdesc{ThLU}{\glsentryshort{tanh} linear unit}{\glsxtrshort{tanh} linear unit}{sec:thlu}
\newglossaryentry{maxtanh}{
    name={maxtanh},
    description={\seeref{an \glsxtrlong{AF}}{sec:thlu}},
}
\newglossaryentry{dualelu}{
    name={DualELU},
    description={\seeref{an \glsxtrlong{AF}; an \glsxtrshort{ELU} variant similar to \gls{dualrelu}}{sec:dualelu}},
}
\newacronymwithrefdesc{DiffELU}{difference \glsentrylong{ELU}}{difference \glsxtrlong{ELU}}{sec:diffelu}
\newgacronymwithref{PolyLU}{polynomial linear unit}{sec:polylu}
\newglossaryentry{L-ReLU}{
    type=\acronymtype,
    name={L--ReLU},
    short={L--ReLU},
    long={Lipschitz ReLU},
    description={\seeref{Lipschitz \glsxtrshort{ReLU}}{sec:l-relu}},
}
\newgacronymwithref{IpLU}{polynomial linear unit}{sec:iplu}
\newgacronymwithref{PoLU}{power linear unit}{sec:polu}
\newgacronymwithref{PFLU}{power function linear unit}{sec:pflu}
\newacronymwithrefdesc{FPFLU}{faster \glsentrylong{PFLU}}{faster \glsxtrlong{PFLU}}{sec:fpflu}
\newgacronymwithref{EACU}{elastic adaptively parametric compounded unit}{sec:eacu}
\newacronymwithrefdesc{SELU}{scaled ELU}{scaled \glsxtrshort{ELU}}{sec:selu}
\newgacronymwithref{SERLU}{scaled exponentially-regularized linear unit}{sec:serlu}
\newglossaryentry{ASERLU}{
    name={ASERLU},
    description={\seeref{an \glsxtrlong{AF}; an extension of the \glsxtrshort{SERLU} for \glsxtrshort{BiLSTM}  architectures}{sec:serlu}},
}
\newgacronymwithref{LSELU}{leaky scaled exponential linear unit}{sec:lselu}
\newgacronymwithref{sSELU}{scaled scaled exponential linear unit}{sec:sselu}
\newglossaryentry{rsigelu}{
    name={RSigELU},
    description={\seeref{an \glsxtrlong{AF}; a parametric \glsxtrshort{ELU}}{sec:rsigelu}},
}
\newglossaryentry{HardSReLUE}{
    name={HardSReLUE},
    description={\seeref{an \glsxtrlong{AF}; a parametric \glsxtrshort{ELU}}{sec:hardsrelue}},
}
\newgacronymwithref{ELiSH}{exponential linear sigmoid squashing}{sec:elish}
\newgacronymwithref{HardELiSH}{hard exponential linear sigmoid squashing}{sec:hardelish}
\newglossaryentry{rsigelud}{
    name={RSigELUD},
    description={\seeref{an \glsxtrlong{AF}; a variant of \gls{rsigelu} with two parameters}{sec:rsigelud}},
}
\newglossaryentry{LSReLU}{
    name={LS--ReLU},
    description={\seeref{an \glsxtrlong{AF} inspired by \gls{ReLU}; not an abbreviation}{sec:lsrelu}},
}
\newglossaryentry{SQNL}{
    name={SQNL},
    description={\seeref{an \glsxtrlong{AF}; not an abbreviation}{sec:sqnl}},
}
\newgacronymwithref{SQLU}{square linear unit}{sec:sqlu}
\newgacronymwithref{squish}{square swish}{sec:squish}
\newgacronymwithref{SqSoftplus}{square softplus}{sec:sqsoftplus}
\newgacronymwithref{LogSQNL}{square logistic sigmoid}{sec:logsqnl}
\newgacronymwithref{SQMAX}{square softmax}{sec:sqmax}
\newgacronymwithref{ISRLU}{inverse square root linear unit}{sec:isrlu}
\newgacronymwithref{ISRU}{inverse square root unit}{sec:isru}
\newacronymwithrefdesc{MEF}{modified Elliott function}{modified \gls{Elliott} function}{sec:mef}
\newgacronymwithref{LinQ}{linear quadratic activation}{sec:linq}
\newacronymwithrefdesc{SQRT}{square-root-based activation function}{square-root-based \glsxtrlong{AF}}{sec:sqrt}
\newglossaryentry{bent_identity}{
    name={bent identity},
    description={\seeref{an \glsxtrlong{AF}}{sec:bent_identity}},
}
\newacronymwithrefdesc{SSAF}{S-shaped activation function}{S-shaped \glsxtrlong{AF}}{sec:sqrt}
\newglossaryentry{Mishra}{
    name={Mishra},
    description={\seeref{an \glsxtrlong{AF}; unnamed in the original paper}{sec:mishra}},
}
\newacronymwithrefdesc{SBAF}{Saha-Bora activation function}{Saha-Bora \glsxtrlong{AF}}{sec:sbaf}
\newacronymwithrefdesc{LAF}{logarithmic activation function}{logarithmic \glsxtrlong{AF}}{sec:laf}
\newgacronymwithref{SPOCU}{scaled polynomial constant unit}{sec:spocu}
\newacronymwithrefdesc{PUAF}{polynomial \glsentrylong{UAF}}{polynomial \glsxtrlong{UAF}}{sec:puaf}
\newglossaryentry{softplus}{
    name={softplus},
    description={\seeref{an \glsxtrlong{AF}}{sec:softplus}},
}
\newgacronymwithref{PSoftplus}{parametric softplus}{sec:psoftplus}
\newgacronymwithref{RSP}{rand softplus}{sec:rsp}
\newglossaryentry{softsign}{
    name={softsign},
    description={\seeref{an \glsxtrlong{AF}}{sec:softsign}},
}
\newglossaryentry{smooth_step}{
    name={smooth step},
    description={\seeref{an \glsxtrlong{AF}}{sec:smooth_step}},
}
\newglossaryentry{mish}{
    name={mish},
    description={\seeref{an \glsxtrlong{AF}; combination of \glsxtrshort{tanh} and \gls{softplus}}{sec:mish}},
}
\newglossaryentry{smish}{
    name={smish},
    description={\seeref{an \glsxtrlong{AF}; combination of \glsxtrshort{tanh}, logarithm, and \gls{logisticsigmoid}}{sec:smish}},
}
\newacronymwithrefdesc{SC-mish}{soft clipping mish}{soft clipping \gls{mish}}{sec:scmish}
\newacronymwithrefdesc{SC-swish}{soft clipping swish}{soft clipping \gls{swish}}{sec:scswish}
\newacronymwithrefdesc{p-swish}{parametric swish}{parametric\gls{swish}}{sec:pswish}
\newacronymwithrefdesc{SCL-mish}{soft clipping learnable mish}{soft clipping learnable \gls{mish}}{sec:scmish}
\newglossaryentry{TanhExp}{
    name={TanhExp},
    description={\seeref{an \glsxtrlong{AF}; combination of \glsxtrshort{tanh} and exponential function}{sec:mish}},
}
\newglossaryentry{serf}{
    name={serf},
    description={\seeref{an \glsxtrlong{AF}; combination of the Gauss error function and \gls{softplus}}{sec:serf}},
}
\newgacronymwithref{EANAF}{efficient asymmetric nonlinear activation function}{sec:eanaf}
\newglossaryentry{sinsig}{
    name={SinSig},
    description={\seeref{an \glsxtrlong{AF}; uses \gls{logisticsigmoid} and is similar to \gls{mish} and \gls{swish}}{sec:sinsig}},
}
\newacronymwithrefdesc{SiELU}{\glsentrylong{GELU} with \gls{sigmoid} \glsentrylong{AF}}{\glsxtrlong{GELU} with \gls{sigmoid} \glsxtrlong{AF}}{sec:sielu}
\newglossaryentry{Aranda-Ordaz}{
    name={Aranda-Ordaz},
    description={\seeref{an \glsxtrlong{AF}}{sec:arandaordaz}},
}
\newgacronymwithref{bfire}{bi-firing activation function}{sec:bfire}
\newgacronymwithref{bbfire}{bounded bi-firing activation function}{sec:bbfire}
\newacronymwithrefdesc{PMAF}{piecewise Mexican-hat activation function}{piecewise Mexican-hat \glsxtrlong{AF}}{sec:pmaf}
\newacronymwithrefdesc{PRBF}{piecewise \glsentrylong{RBF}}{piecewise \glsxtrlong{RBF}}{sec:prbf}
\newglossaryentrywithacronymandref{GELU}{GELU}{Gaussian error linear unit}{a popular \glsxtrlong{AF} based on the cumulative distribution function of the normal distribution}{sec:gelu}
\newgacronymwithref{SGELU}{symmetrical Gaussian error linear unit}{sec:sgelu}
\newgacronymwithref{CaLU}{Cauchy linear unit}{sec:calu}
\newgacronymwithref{LaLU}{Laplace linear unit}{sec:lalu}
\newgacronymwithref{CoLU}{Collapsing linear unit}{sec:colu}
\newgacronymwithref{ASSF}{adaptive slope sigmoidal function}{sec:assf}
\newgacronymwithref{SG}{Sigmoid-Gumbel}{sec:sg}
\newglossaryentry{comb-H-sine}{
    name={comb-H-sine},
    description={\seeref{an \glsxtrlong{AF}}{sec:combhsine}},
}
\newgacronymwithref{m-arcsinh}{modified arcsinh}{sec:marcsinh}
\newglossaryentry{hyper-sinh}{
    name={hyper-sinh},
    description={\seeref{an \glsxtrlong{AF}}{sec:hypersinh}},
}

\newglossaryentry{arctid}{
    name={arctid},
    description={\seeref{an \glsxtrlong{AF}}{sec:arctid}},
}
\newglossaryentry{sinc}{
    name={sinc},
    description={\seeref{an \glsxtrlong{AF}}{sec:sinc}},
}
\newglossaryentry{polyexp}{
    name={polyexp},
    description={\seeref{an \glsxtrlong{AF}}{sec:polyexp}},
}
\newglossaryentry{E-Tanh}{
    name={E-Tanh},
    description={\seeref{an \glsxtrlong{AF}}{sec:etanh}},
}
\newglossaryentry{wave}{
    name={wave},
    description={\seeref{an \glsxtrlong{AF}}{sec:wave}},
}
\newglossaryentry{cosid}{
    name={cosid},
    description={\seeref{an \glsxtrlong{AF}}{sec:cosid}},
}
\newglossaryentry{sinp}{
    name={sinp},
    description={\seeref{an \glsxtrlong{AF} based on the sine function}{sec:sinp}},
}
\newgacronymwithref{GCU}{growing cosine unit}{sec:gcu}
\newgacronymwithref{ASU}{amplifying sine unit}{sec:asu}
\newgacronymwithref{SSU}{shifted sine unit}{sec:sinc}
\newgacronymwithref{DSU}{decaying sine unit}{sec:dsu}
\newgacronymwithref{NCU}{non-monotonic cubic unit}{sec:ncu}
\newglossaryentry{triple}{
    name={triple},
    description={\seeref{an \glsxtrlong{AAF}}{sec:triple}},
}
\newgacronymwithref{SQU}{shifted quadratic unit}{sec:squ}
\newgacronymwithref{HcLSH}{hyperbolic cosine linearized squashing function}{sec:hclsh}
\newgacronymwithref{KDAC}{knowledge discovery activation function}{sec:kdac}
\newacronym{WTA}{WTA}{winner-take-all}
\newglossaryentrywithacronymandref{k-WTA}{k-\glsentrylong{WTA}}{k-\glsentrylong{WTA}}{an \glsxtrlong{AF} based on the \glsentrylong{WTA} principle}{sec:kwta}
\newacronymwithrefdesc{VBAF}{volatility-based activation function}{volatility-based  \glsxtrlong{AF}}{sec:vbaf}
\newacronymwithrefdesc{CAF}{chaotic activation function}{chaotic \glsxtrlong{AF}}{sec:caf}
\newacronymwithrefdesc{HCAF}{hybrid chaotic activation function}{hybrid \glsxtrlong{CAF}}{sec:hcaf}
\newacronymwithrefdesc{FCAF}{fusion of chaotic activation function}{fusion of \glsxtrlong{CAF}}{sec:fcaf}
\newacronymwithrefdesc{CCAF}{cascade chaotic activation function}{cascade \glsxtrlong{CAF}}{sec:ccaf}
\newglossaryentry{sincos}{
    name={sincos},
    description={\seeref{an \glsxtrlong{AF}}{sec:sincos}},
}
\newgacronymwithref{CSS}{combination of sine and logistic sigmoid}{sec:css}
\newacronymwithrefdesc{CatAF}{catalytic activation function}{catalytic \glsxtrlong{AF}}{sec:cataf}
\newglossaryentry{expcos}{
    name={expcos},
    description={\seeref{an \glsxtrlong{AF}}{sec:expcos}},
}
\newglossaryentry{rootsig}{
    name={rootsig},
    description={\seeref{an \glsxtrlong{AF}}{sec:rootsig}},
}
\newglossaryentry{nsigmoid}{
    name={n-sigmoid},
    description={\seeref{a \gls{sigmoid}-based \glsxtrlong{AF}}{sec:logistic_sigmoid}},
}
\newglossaryentrywithacronymandref{PReLU}{PReLU}{parametric rectified linear unit}{a popular \glsxtrlong{AAF}; a \glsxtrshort{LReLU} variant with trainable leakiness}{sec:prelu}
\newglossaryentry{PPReLU}{
    type=\acronymtype,
    name={positive PReLU},
    short={PReLU\textsuperscript{+}},
    long={positive PReLU},
    description={\seeref{positive \gls{PReLU}}{sec:pprelu}},
}
\newacronymwithrefdesc{LeLeLU}{leaky learnable \glsentryshort{ReLU}}{leaky learnable \glsxtrshort{ReLU}}{sec:lelelu}
\newgacronymwithref{PREU}{parametric rectified exponential unit}{sec:preu}
\newglossaryentry{RT-PReLU}{
    type=\acronymtype,
    name={RT--PReLU},
    short={RT--PReLU},
    long={randomly translational PReLU},
    description={\seeref{randomly translational \glsxtrshort{PReLU}}{sec:rtprelu}},
    first={randomly translational PReLU (RT--PReLU)}
}
\newglossaryentry{ProbAct}{
    type=\acronymtype,
    name={ProbAct},
    short={ProbAct},
    long={ProbAct},
    description={\seeref{probabilistic activation}{sec:probact}},
}
\newglossaryentry{paired_relu}{
    name={paired ReLU},
    description={\seeref{paired \glsxtrshort{ReLU}}{sec:paired_relu}},
}
\newacronymwithrefdesc{RMAF}{ReLU memristor-like activation function}{\glsxtrshort{ReLU} memristor-like \glsxtrlong{AF}}{sec:rmaf}

\newacronymwithrefdesc{PTELU}{parametric tanh linear unit}{parametric \glsxtrshort{tanh} linear unit}{sec:ptelu}
\newacronymwithrefdesc{TaLU}{tanh linear unit}{\glsxtrshort{tanh} linear unit}{sec:talu}
\newglossaryentry{PTaLU}{
    name={PTaLU},
    description={\seeref{an \glsxtrshort{AAF}, \glsxtrshort{TaLU} variant with another parameter}{sec:ptalu}},
}\newglossaryentry{tanhLU}{
    name={tanhLU},
    description={\seeref{an \glsxtrshort{AAF}, combination of \glsxtrshort{tanh} and a linear function}{sec:tanhlu}},
}
\newacronymwithrefdesc{TReLU}{tanh based ReLU}{\glsxtrshort{tanh} based \glsxtrshort{ReLU}}{sec:trelu}
\newacronymwithrefdesc{TReLU2}{TReLU variant 2}{\glsxtrshort{TReLU} variant 2}{sec:trelu}
\newacronymwithrefdesc{TeLU}{tanh \glsentrylong{ELU}}{\glsxtrshort{tanh} \glsxtrlong{ELU}}{sec:telu}
\newacronymwithrefdesc{ReLTanh}{rectified linear tanh}{rectified linear \glsxtrshort{tanh}}{sec:reltanh}
\newacronymwithrefdesc{PELU}{parametric \glsentrylong{ELU}}{parametric \glsxtrlong{ELU}}{sec:pelu}
\newacronymwithrefdesc{FELU}{fast \glsentrylong{ELU}}{fast \glsxtrshort{ELU}}{sec:felu}
\newglossaryentry{PFELU}{
    name={P+FELU},
    description={\seeref{trainable \glsxtrshort{FELU} based \glsxtrshort{AF}}{sec:pfelu}},
}
\newacronymwithrefdesc{MPELU}{multiple \glsentrylong{PELU}}{multiple \glsxtrlong{PELU}}{sec:mpelu}
\newglossaryentry{P-E2-X}{
    name={P-E2-XU},
    description={\seeref{family of \glsxtrlongpl{AAF}}{sec:pe2relu}},
}
\newglossaryentry{P-E2-ReLU}{
    name={P-E2-ReLU},
    description={\seeref{an \glsxtrlong{AAF} combining two \glsxtrshortpl{ELU} and \glsxtrshort{ReLU}}{sec:pe2relu}},
}
\newglossaryentry{P-E2-Id}{
    name={P-E2-Id},
    description={\seeref{an \glsxtrlong{AAF} based on \gls{P-E2-ReLU}}{sec:pe2relu}},
}
\newglossaryentry{P-E2-ReLU-1}{
    name={P-E2-ReLU-1},
    description={\seeref{an \glsxtrlong{AAF} based on \gls{P-E2-ReLU}}{sec:pe2relu}},
}
\newgacronymwithref{BLU}{bendable linear unit}{sec:blu}
\newacronymwithrefdesc{ReBLU}{rectified \glsentryshort{BLU}}{rectified \glsxtrshort{BLU}}{sec:reblu}
\newglossaryentry{DELU}{
    name={DELU},
    description={\seeref{an \glsxtrlong{AF} proposed in \cite{Pishchik2023}; not an abbreviation}{sec:delu}},
}
\newglossaryentry{soft_exponential}{
    name={soft exponential},
    description={\seeref{an \glsxtrlong{AF} interpolating between logarithmic, linear, and exponential functions}{sec:soft_exponential}},
}
\newacronymwithrefdesc{CELU}{continuously differentiable \glsentrylong{ELU}}{continuously differentiable \glsxtrlong{ELU}}{sec:celu}
\newacronymwithrefdesc{ErfReLU}{Erf-based \glsentryshort{ReLU}}{Erf-based \glsxtrshort{ReLU}}{sec:erfrelu}
\newacronymwithrefdesc{PSELU}{parametric scaled \glsentrylong{ELU}}{parametric scaled \glsxtrlong{ELU}}{sec:pselu}
\newacronymwithrefdesc{LPSELU}{leaky \glsentrylong{PSELU}}{leaky \glsxtrlong{PSELU}}{sec:lpselu}
\newglossaryentry{LPSELURP}{
    type=\acronymtype,
    name={LPSELU\_RP},
    short={LPSELU\_RP},
    long={\glsentrylong{LPSELU} with reposition parameter},
    description={\seeref{\glsxtrlong{LPSELU} with reposition parameter}{sec:lpselurp}},
}
\newglossaryentry{shifted_elu}{
    name={shifted \glsentrylong{ELU}},
    short={shifted \glsentryshort{ELU}},
    long={shifted \glsentrylong{ELU}},
    shortplural={shifted \glsentryshort{ELU}s},
    longplural={shifted \glsentrylong{ELU}s},
    description={\seeref{an \glsxtrlong{AF}; an \glsxtrshort{ELU} with a vertical shift (\glsxtrshort{SvELU}, \glsxtrshort{PSvELU}) or an \glsxtrshort{ELU} with  ahorizontal shift (\glsxtrshort{ShELU}, \glsxtrshort{PShELU})}{sec:shifted_elus}},
}
\newglossaryentry{SvELU}{
    type=\acronymtype,
    name={SvELU},
    short={SvELU},
    long={SvELU},
    description={\seeref{\glsxtrlong{ELU} with a fixed vertical shift}{sec:shifted_elus}},
}
\newglossaryentry{ShELU}{
    type=\acronymtype,
    name={ShELU},
    short={ShELU},
    long={ShELU},
    description={\seeref{\glsxtrlong{ELU} with a fixed horizontal shift}{sec:shifted_elus}},
}
\newglossaryentry{PSvELU}{
    type=\acronymtype,
    name={PSvELU},
    short={PSvELU},
    long={PSvELU},
    description={\seeref{\glsxtrlong{ELU} with a trainable vertical shift}{sec:shifted_elus}},
}
\newglossaryentry{PShELU}{
    type=\acronymtype,
    name={PShELU},
    short={PShELU},
    long={PShELU},
    description={\seeref{\glsxtrlong{ELU} with a trainable horizontal shift}{sec:shifted_elus}},
}
\newacronymwithrefdesc{PDELU}{parametric deformable \glsentrylong{ELU}}{parametric deformable \glsxtrlong{ELU}}{sec:pdelu}
\newacronymwithrefdesc{EDELU}{extended \glsentrylong{ELU}}{extended \glsxtrlong{ELU}}{sec:edelu}
\newacronymwithrefdesc{EELU}{elastic \glsentrylong{ELU}}{elastic \glsxtrlong{ELU}}{sec:eelu}
\newgacronymwithref{PFPLUS}{parametric first power linear unit with sign}{sec:pfplus}
\newgacronymwithref{FPLUS}{first power linear unit with sign}{sec:polylu}
\newglossaryentry{generalized hyperbolic tangent}{
    name={generalized hyperbolic tangent},
    description={\seeref{an \glsxtrlong{AAF}}{sec:generalized_hyperbolic_tangent}},
}
\newacronymwithrefdesc{SVAF}{slope varying \glsentrylong{AF}}{slope varying \glsxtrlong{AF}}{sec:svaf}
\newglossaryentry{TanhSoft}{
    name={TanhSoft},
    description={\seeref{a family of \glsxtrlongpl{AAF} proposed in \cite{Biswas2021TanhSoft}}{sec:tanhsoft}},
}
\newacronymwithrefdesc{psigmoid}{parametric sigmoid}{parametric \gls{sigmoid}}{sec:psigmoid}
\newglossaryentry{swish}{
    name={swish},
    description={\seeref{an \glsxtrlong{AAF}; an adaptive variant of \glsxtrshort{SiLU}}{sec:swish}},
}
\newacronymwithrefdesc{AHAF}{adaptive hybrid \glsentrylong{AF}}{adaptive hybrid \glsxtrlong{AF}}{sec:ahaf}
\newglossaryentrywithacronymandref{PSiLU}{PSiLU}{parametric SiLU}{another name for the \gls{swish} \glsxtrshort{AF}}{sec:swish}
\newacronymwithrefdesc{PSSiLU}{parametric shifted \glsentryshort{SiLU}}{parametric shifted \glsxtrshort{SiLU}}{sec:pssilu}
\newacronymwithrefdesc{ACON}{activate or not \glsentrylong{AF}}{activate or not \glsxtrlong{AF}}{sec:swish}
\newglossaryentry{ACON-A}{
    name={ACON-A},
    description={\seeref{an \glsxtrlong{AAF} from the \glsxtrshort{ACON} family; another name for the \gls{swish}}{sec:swish}},
}
\newglossaryentry{ACON-B}{
    name={ACON-B},
    description={\seeref{an \glsxtrlong{AAF} from the \glsxtrshort{ACON} family; extension of the \gls{ACON-A}}{sec:aconb}},
}
\newglossaryentry{ACON-C}{
    name={ACON-C},
    description={\seeref{an \glsxtrlong{AAF} from the \glsxtrshort{ACON} family; extension of the \gls{ACON-B}}{sec:aconc}},
}
\newglossaryentry{MetaACON}{
    name={MetaACON},
    description={\seeref{an extension of the \glsxtrshort{ACON} family where the parameter $a_i$ is determined by a small \glsxtrshort{NN}}{sec:aconc}},
}
\newglossaryentry{MetaACON-C}{
    name={MetaACON-C},
    description={\seeref{an \gls{ACON-C} variant where the parameter $a_i$ is determined by a small \glsxtrshort{NN}}{sec:aconc}},
}
\newglossaryentry{1Dmeta-ACON}{
    name={1Dmeta-ACON},
    description={\seeref{a \gls{MetaACON} variant}{sec:aconc}},
}
\newglossaryentry{gish}{
    name={gish},
    description={\seeref{an \glsxtrlong{AF} based on the \glsxtrshort{SiLU}}{sec:gish}},
}
\newglossaryentry{logish}{
    name={logish},
    description={\seeref{an \glsxtrlong{AF} based on the \glsxtrshort{SiLU}}{sec:logish}},
}
\newglossaryentry{LogLogish}{
    name={LogLogish},
    description={\seeref{an \glsxtrlong{AF} based on the \glsxtrshort{SiLU} and the \gls{LogLog}}{sec:loglogish}},
}
\newglossaryentry{ExpExpish}{
    name={ExpExpish},
    description={\seeref{an \glsxtrlong{AF} based on the \glsxtrshort{SiLU}}{sec:expexpish}},
}
\newglossaryentry{selfarctan}{
    name={self arctan},
    description={\seeref{an \glsxtrlong{AF} based on the \glsxtrshort{SiLU}}{sec:selfarctan}},
}
\newglossaryentry{phish}{
    name={phish},
    description={\seeref{an \glsxtrlong{AF} based on the \glsxtrshort{SiLU} and \glsxtrshort{GELU}}{sec:phish}},
}
\newglossaryentry{suish}{
    name={suish},
    description={\seeref{an \glsxtrlong{AF}; proposed as the alternative to the \glsxtrshort{SiLU} and \gls{swish}}{sec:suish}},
}
\newacronymwithrefdesc{TSReLU}{tangent-sigmoid ReLU}{\glslink{tanh}{tangent}-\gls{sigmoid} \glsxtrshort{ReLU}}{sec:tsrelu}
\newacronymwithrefdesc{TBSReLU}{tangent-bipolar-sigmoid ReLU}{\glslink{tanh}{tangent}-bipolar-\gls{sigmoid} \glsxtrshort{ReLU}}{sec:tbsrelu}
\newacronymwithrefdesc{TSReLUl}{TSReLU learnable}{\glsxtrshort{TSReLU} learnable}{sec:psgu}
\newacronymwithrefdesc{TBSReLUl}{TBSReLU learnable}{\glsxtrshort{TBSReLU} learnable}{sec:tbsrelul}
\newacronymwithrefdesc{pLogish}{parametric logish}{parametric \gls{logish}}{sec:plogish}
\newacronymwithrefdesc{ARBF}{adaptive \glsentrylong{RBF}}{parametric \glsxtrlong{RBF}}{sec:arbf}
\newacronymwithrefdesc{PGELU}{parametric \glsentrylong{GELU}}{parametric \glsxtrlong{GELU}}{sec:pgelu}
\newacronymwithrefdesc{PSF}{parametric \glsentrylong{sigmoid} function}{parametric \glsxtrlong{sigmoid} function}{sec:psf}
\newacronymwithrefdesc{STAC-tanh}{slope and threshold \glsentrylong{AAF} function with \glsentryshort{tanh} function}{slope and threshold \glsxtrlong{AAF} function with \glsxtrshort{tanh} function}{sec:stactanh}
\newgacronymwithref{GRA}{generalized Riccati activation}{sec:gra}
\newacronymwithrefdesc{PFTS}{parametric \glsentrylong{FTS}}{parametric \glsxtrlong{FTS}}{sec:pfts}
\newacronymwithrefdesc{PFPM}{parametric flatten-p \gls{mish}}{parametric flatten-p \gls{mish}}{sec:pfpm}
\newgacronymwithref{GEU}{Gaussian error unit}{sec:geu}
\newacronymwithrefdesc{SGT}{scaled-gamma-\glsentryshort{tanh}}{scaled-gamma-\glsxtrshort{tanh}}{sec:sgt}
\newgacronymwithref{RSign}{react-sign}{sec:rsign}
\newglossaryentry{P-SIG-RAMP}{
    name={P-SIG-RAMP},
    description={\seeref{an \glsxtrlong{AAF} combining \gls{logisticsigmoid} and \glsxtrshort{ReLU}}{sec:psigramp}},
}
\newacronymwithrefdesc{PSTanh}{parametric scaled \glsentrylong{tanh}}{parametric scaled \glsxtrlong{tanh}}{sec:pstanh}
\newacronymwithrefdesc{SSinH}{scaled sine-hyperbolic function}{scaled sine-hyperbolic function}{sec:ssinh}
\newacronymwithrefdesc{SExp}{scaled exponential function}{scaled exponential function}{sec:sexp}
\newacronymwithrefdesc{MWF}{modified Weibull function}{modified Weibull function}{sec:mwf}
\newgacronymwithref{LAU}{logmoid activation unit}{sec:lau}
\newglossaryentry{symlog}{
    name={symlog},
    description={\seeref{an alternative name of the \glsxtrshort{LAU}}{sec:lau}},
}
\newglossaryentry{symexp}{
    name={symexp},
    description={\seeref{an \glsxtrlong{AF} inverse of the \glsxtrshort{LAU}}{sec:symexp}},
}
\newgacronymwithref{CosLU}{cosinu-sigmoidal linear unit}{sec:coslu}
\newgacronymwithref{AGumb}{adaptive Gumbel}{sec:AGumb}
\newglossaryentry{NewSigmoid}{
    name={NewSigmoid},
    description={\seeref{an \glsxtrlong{AF} similar to \gls{logisticsigmoid}}{sec:newsigmoid}},
}
\newglossaryentry{generalized_swish}{
    name={generalized swish},
    description={\seeref{an \glsxtrlong{AF} related to the \gls{SiLU}}{sec:generalized_swish}},
}
\newglossaryentry{exponential_swish}{
    name={exponential swish},
    description={\seeref{an \glsxtrlong{AF} related to the \gls{SiLU}}{sec:exponential_swish}},
}
\newglossaryentry{Sigmoid-Algebraic}{
    name={Sigmoid-Algebraic},
    description={\seeref{an \glsxtrlong{AF} similar to \gls{logisticsigmoid}}{sec:sigmoidalgebraic}},
}
\newglossaryentry{SincSigmoid}{
    name={Sinc-Sigmoid},
    description={\seeref{an \glsxtrlong{AF}}{sec:sincsigmoid}},
}
\newglossaryentry{root2sigmoid}{
    name={root2sigmoid},
    description={\seeref{an \glsxtrlong{AF} similar to \gls{logisticsigmoid}}{sec:root2sigmoid}},
}
\newglossaryentry{LogLog}{
    name={LogLog},
    description={\seeref{an \glsxtrlong{AF}}{sec:loglog}},
}
\newacronymwithrefdesc{cLogLog}{complementary LogLog}{complementary \gls{LogLog}}{sec:cloglog}
\newacronymwithrefdesc{cLogLogm}{modified \glsentryshort{cLogLog}}{modified \glsxtrshort{cLogLog}}{sec:cloglog}
\newglossaryentry{SechSig}{
    name={SechSig},
    description={\seeref{an \glsxtrlong{AF}}{sec:sechsig}},
}
\newacronymwithrefdesc{pSechSig}{parametric SechSig}{parametric \gls{SechSig}}{sec:sechsig}
\newglossaryentry{TanhSig}{
    name={TanhSig},
    description={\seeref{an \glsxtrlong{AF}}{sec:tanhsig}},
}
\newacronymwithrefdesc{pTanhSig}{parametric TanhSig}{parametric \gls{TanhSig}}{sec:tanhsig}
\newacronymwithrefdesc{MSAF}{multistate \glsentrylong{AF}}{multistate \glsxtrlong{AF}}{sec:msaf}
\newacronymwithrefdesc{SymMSAF}{symmetrical \glsentryshort{MSAF}}{symmetrical \glsxtrshort{MSAF}}{sec:msaf}
\newglossaryentry{scaled_softsign}{
    name={scaled softsign},
    description={\seeref{an \glsxtrlong{AAF}; an adaptive variant of \glsxtrlong{softsign}}{sec:scaled_softsign}},
}
\newglossaryentry{parameterized_softplus}{
    type=\acronymtype,
    name={parameterized softplus},
    short={s\textsubscript{+}2L},
    long={parameterized softplus},
    description={\seeref{parametrized \gls{softplus}}{sec:parameterized_softplus}},
}
\newacronymwithrefdesc{UAF}{universal \glsentrylong{AF}}{universal \glsxtrlong{AF}}{sec:uaf}
\newacronymwithrefdesc{LEAF}{learnable extended \glsentrylong{AF}}{learnable extended \glsxtrlong{AF}}{sec:leaf}
\newacronymwithrefdesc{GReLU}{generalized \glsentryshort{ReLU}}{generalized \glsxtrshort{ReLU}}{sec:grelu}
\newacronymwithrefdesc{MAF}{multiquadratic \glsentrylong{AF}}{multiquadratic \glsxtrlong{AF}}{sec:maf}
\newglossaryentry{EIS}{
    name={EIS},
    description={\seeref{a family of \glsxtrlongpl{AF}, not an abbreviation}{sec:eis}},
}
\newgacronymwithref{GLN}{global-local neuron}{sec:gln}
\newacronymwithrefdesc{NAF}{neuron-adaptive \glsentrylong{AF}}{neuron-adaptive \glsxtrlong{AF}}{sec:naf}
\newacronymwithrefdesc{LAAF}{locally \glsentrylong{AAF}}{locally \glsxtrlong{AAF}}{sec:laaf}
\newacronymwithrefdesc{GAAF}{globally \glsentrylong{AAF}}{globally \glsxtrlong{AAF}}{sec:laaf}
\newacronymwithrefdesc{SAAAF}{shape autotuning \glsentrylong{AAF}}{shape autotuning \glsxtrlong{AAF}}{sec:saaaf}
\newacronymwithrefdesc{FAAF}{fractional \glsentrylong{AAF}}{fractional \glsxtrlong{AAF}}{sec:faaf}
\newacronymwithrefdesc{FracReLU}{fractional ReLU}{fractional \glsxtrshort{ReLU}}{sec:fracrelu}
\newacronymwithrefdesc{FracSoftplus}{fractional softplus}{fractional \gls{softplus}}{sec:fracsoftplus}
\newacronymwithrefdesc{FracTanh}{fractional tanh}{fractional \glsxtrshort{tanh}}{sec:fractanh}
\newacronymwithrefdesc{FracLReLU}{fractional LReLU}{fractional \glsxtrshort{LReLU}}{sec:fraclrelu}
\newacronymwithrefdesc{FracPReLU}{fractional PReLU}{fractional \glsxtrshort{PReLU}}{sec:fracprelu}
\newacronymwithrefdesc{FracELU}{fractional ELU}{fractional \glsxtrshort{ELU}}{sec:fracelu}
\newacronymwithrefdesc{FracSiLU}{fractional SiLU}{fractional \glsxtrshort{SiLU}}{sec:fracsilu}
\newacronymwithrefdesc{FracSiLU1}{FracSiLU variant 1}{\glsxtrshort{FracSiLU} variant 1}{sec:fracsilu}
\newacronymwithrefdesc{FracSiLU2}{FracSiLU variant 1}{\glsxtrshort{FracSiLU} variant 2}{sec:fracsilu}
\newacronymwithrefdesc{FracGELU}{fractional GELU}{fractional \glsxtrshort{GELU}}{sec:fracgelu}
\newacronymwithrefdesc{FracGELU1}{FracGELU variant 1}{\glsxtrshort{FracGELU} variant 1}{sec:fracgelu}
\newacronymwithrefdesc{FracGELU2}{FracGELU variant 1}{\glsxtrshort{FracGELU} variant 2}{sec:fracgelu}
\newgacronymwithref{FALU}{fractional adaptive linear unit}{sec:falu}
\newgacronymwithref{APLU}{adaptive piece-wise linear unit}{sec:aplu}
\newacronymwithrefdesc{SAAF}{smooth \glsentrylong{AAF}}{smooth \glsxtrlong{AAF}}{sec:aplu}
\newgacronymwithref{SPLASH}{simple piecewise linear and adaptive function with symmetric hinges}{sec:splash}
\newgacronymwithref{MBA}{multi-bias activation}{sec:mba}
\newacronymwithrefdesc{MeLU}{Mexican \glsentryshort{ReLU}}{Mexican \glsxtrshort{ReLU}}{sec:melu}
\newacronymwithrefdesc{MMeLU}{modified \glsentrylong{MeLU}}{modified \glsxtrlong{MeLU}}{sec:mmelu}
\newacronymwithrefdesc{GaLU}{Gaussian \glsentryshort{ReLU}}{Gaussian \glsxtrshort{ReLU}}{sec:galu}
\newacronymwithrefdesc{SReLU}{S-shaped \glsentryshort{ReLU}}{S-shaped \glsxtrshort{ReLU}}{sec:srelu}
\newglossaryentry{N-activation}{
    name={N-activation},
    description={\seeref{an \glsxtrlong{AAF} resembling the letter N}{sec:nactivation}},
}
\newgacronymwithref{LiSA}{linearized sigmoidal activation}{sec:alisa}
\newacronymwithrefdesc{ALiSA}{adaptive \glsentryshort{LiSA}}{adaptive \glsxtrlong{LiSA}}{sec:alisa}
\newacronymwithrefdesc{All-ReLU}{alternated left \glsentryshort{ReLU}}{alternated left \glsxtrlong{ReLU}}{sec:allrelu}
\newgacronymwithref{ABU}{adaptive blending unit}{sec:abu}
\newacronymwithrefdesc{TCA}{trained compound activation function}{trained compound \glsxtrlong{AF}}{sec:tca}
\newacronymwithrefdesc{TCAv2}{\glsxtrlong{TCA} variant 2}{\glsxtrlong{TCA} variant 2}{sec:tca}
\newacronymwithrefdesc{APAF}{average of a pool of activation functions}{average of a pool of \glsxtrlongpl{AF}}{sec:apaf}
\newacronymwithrefdesc{GABU}{gating \glsentrylong{ABU}}{gating \glsxtrlong{ABU}}{sec:gabu}

\newacronymwithrefdesc{SLAF}{self-learnable \glsentrylong{AF}}{self-learnable \glsxtrlong{AF}}{sec:slaf}
\newacronymwithrefdesc{ChPAF}{Chebyshev polynomial-based \glsentrylong{AF}}{Chebyshev polynomial-based \glsxtrlong{AF}}{sec:chpaf}
\newacronymwithrefdesc{LPAF}{Legendre polynomial-based \glsentrylong{AF}}{Legendre polynomial-based \glsxtrlong{AF}}{sec:lpaf}
\newacronymwithrefdesc{HPAF}{Hermite polynomial-based \glsentrylong{AF}}{Hermite polynomial-based \glsxtrlong{AF}}{sec:hpaf}
\newgacronymwithref{MoGU}{mixture of Gaussian unit}{sec:mogu}
\newgacronymwithref{FSA}{Fourier series activation}{sec:fsa}
\newgacronymwithref{PAU}{Padé activation unit}{sec:pau}
\newacronymwithrefdesc{RPAU}{randomized \glsentrylong{PAU}}{randomized \glsxtrlong{PAU}}{sec:rpau}
\newacronymwithrefdesc{OPAU}{orthogonal \glsentrylong{PAU}}{orthogonal \glsxtrlong{PAU}}{sec:opau}
\newgacronymwithref{ERA}{enhanced rational activation}{sec:era}
\newacronymwithrefdesc{SAF}{spline interpolating \glsentrylong{AF}}{spline interpolating \glsxtrlong{AF}}{sec:saf}
\newacronymwithrefdesc{PPAF}{piecewise polynomial \glsentrylong{AF}}{piecewise polynomial \glsxtrlong{AF}}{sec:saf}
\newgacronymwithref{TruG}{truncated gaussian unit}{sec:trug}
\newgacronymwithref{MSRF}{mollified square root function}{sec:msrf}
\newglossaryentry{SquarePlus}{
    name={SquarePlus},
    description={\seeref{an \glsxtrlong{AF} proposed in \cite{Barron2021}}{sec:squareplus}},
}
\newglossaryentry{StepPlus}{
    name={StepPlus},
    description={\seeref{an \glsxtrlong{AF} proposed in \cite{Pan2023Smoothing}}{sec:stepplus}},
}
\newglossaryentry{BipolarPlus}{
    name={BipolarPlus},
    description={\seeref{an \glsxtrlong{AF} proposed in \cite{Pan2023Smoothing}}{sec:stepplus}},
}
\newglossaryentry{LReLUPlus}{
    name={LReLUPlus},
    description={\seeref{an \glsxtrlong{AF}; a smoothed variant of \glsxtrshort{LReLU}}{sec:lreluplus}},
}
\newglossaryentry{vReLUPlus}{
    name={vReLUPlus},
    description={\seeref{an \glsxtrlong{AF}; a smoothed variant of \glsxtrshort{vReLU}}{sec:vreluplus}},
}
\newglossaryentry{SoftshrinkPlus}{
    name={SoftshrinkPlus},
    description={\seeref{an \glsxtrlong{AF}; a smoothed variant of \gls{softshrink}}{sec:softshrinkplus}},
}
\newglossaryentry{PanPlus}{
    name={PanPlus},
    description={\seeref{an \glsxtrlong{AF}; a smoothed variant of \gls{pan} function}{sec:panplus}},
}
\newglossaryentry{BReLUPlus}{
    name={BReLUPlus},
    description={\seeref{an \glsxtrlong{AF}; a smoothed variant of \glsxtrshort{BReLU}}{sec:breluplus}},
}
\newglossaryentry{SReLUPlus}{
    name={SReLUPlus},
    description={\seeref{an \glsxtrlong{AF}; a smoothed variant of \glsxtrshort{SReLU}}{sec:sreluplus}},
}
\newglossaryentry{HardTanhPlus}{
    name={HardTanhPlus},
    description={\seeref{an \glsxtrlong{AF}; a smoothed variant of \gls{hard_tanh} function}{sec:hardtanhplus}},
}
\newglossaryentry{HardshrinkPlus}{
    name={HardshrinkPlus},
    description={\seeref{an \glsxtrlong{AF}; a smoothed variant of \gls{hardshrink}}{sec:hardshrinkplus}},
}
\newglossaryentry{MeLUPlus}{
    name={MeLUPlus},
    description={\seeref{an \glsxtrlong{AF}; a smoothed variant of \glsxtrshort{MeLU}}{sec:meluplus}},
}
\newglossaryentry{TSAFPlus}{
    name={TSAFPlus},
    description={\seeref{an \glsxtrlong{AF}; a smoothed variant of \glsxtrshort{TSAF}}{sec:tsafplus}},
}
\newglossaryentry{ELUPlus}{
    name={ELUPlus},
    description={\seeref{an \glsxtrlong{AF}; a mollified variant of \glsxtrshort{ELU}}{sec:eluplus}},
}
\newglossaryentry{SwishPlus}{
    name={SwishPlus},
    description={\seeref{an \glsxtrlong{AF}; a mollified variant of \gls{swish}}{sec:swishplus}},
}
\newglossaryentry{MishPlus}{
    name={MishPlus},
    description={\seeref{an \glsxtrlong{AF}; a mollified variant of \gls{mish}}{sec:mishplus}},
}
\newglossaryentry{LogishPlus}{
    name={LogishPlus},
    description={\seeref{an \glsxtrlong{AF}; a mollified variant of \gls{logish}}{sec:logishplus}},
}
\newglossaryentry{SoftsignPlus}{
    name={SoftsignPlus},
    description={\seeref{an \glsxtrlong{AF}; a mollified variant of \gls{softsign}}{sec:softsignplus}},
}
\newglossaryentry{SignReLUPlus}{
    name={SignReLUPlus},
    description={\seeref{an \glsxtrlong{AF}; a smoothed variant of \glsxtrshort{SignReLU}}{sec:signreluplus}},
}
\newacronym{NIN}{NIN}{network in network}
\newacronym{MIN}{MIN}{maxout-in-network}
\newacronym{WHE}{WHE}{wide hidden expansion}

\newglossaryentry{NPF}{
    name={NPF},
    description={\seeref{an \glsxtrlong{AAF} proposed in \cite{Eisenach2017}; based on Nonparametric Fourier
    Basis Expansion}{sec:complex_approaches}},
}
\newacronymwithrefdesc{VAF}{variable \glsentrylong{AF}}{variable \glsxtrlong{AF}}{sec:vaf}
\newgacronymwithref{FAB}{flexible activation bag}{sec:fab}
\newacronymwithrefdesc{DY--ReLU}{dynamic parameter \glsentryshort{ReLU}}{dynamic parameter \glsxtrshort{ReLU}}{sec:dyrelu}

\newacronym{MCDNN}{MCDNN}{multi-column deep neural network}
\newacronym{PC}{PC}{Parallel Circuit}
\newacronym{SNN}{SNN}{self-normalizing neural network}

\newacronym{RVFLN}{RVFLN}{random vector functional link network}
\newgacronymwithdesc{pRVFLN}{parsimonious \glsentrylong{RVFLN}}{parsimonious \glsxtrlong{RVFLN}}
\newgacronymwithdesc{M-RVFLN}{M-estimation-based \glsentryshort{RVFLN}}{M-estimation-based \glsxtrshort{RVFLN}}
\newgacronymwithcustomshortdesc{epsRVFLN}{\ce{\epsilon}-HRVFLN}{\glsentryshort{RVFLN} with \ce{\epsilon}-insensitive Huber loss function}{\glsxtrshort{RVFLN} with \ce{\epsilon}-insensitive Huber loss function}
\newgacronymwithdesc{MK-RVFLN}{multi-kernel \glsentryshort{RVFLN}}{multi-kernel \glsxtrshort{RVFLN}}
\newgacronymwithdesc{K-RVFLN}{kernel \glsentryshort{RVFLN}}{kernel \glsxtrshort{RVFLN}}
\newgacronymwithdesc{EVWCA-MKRVFLN}{\glsentryshort{MK-RVFLN} with evaporation-based water cycle based parameter optimization}{\glsxtrshort{MK-RVFLN} with evaporation-based water cycle based parameter optimization}
\newgacronymwithdesc{WCRVFLN}{wavelet-coupled \glsentryshort{RVFLN}}{wavelet-coupled \glsxtrshort{RVFLN}}
\newgacronymwithdesc{SP-RVFLN}{sparse pre-trained \glsentryshort{RVFLN}}{sparse pre-trained \glsxtrshort{RVFLN}}
\newgacronymwithdesc{CRVFLN}{convolutional \glsentryshort{RVFLN}}{convolutional \glsxtrshort{RVFLN}}

\newacronym{SCN}{SCN}{stochastic configuration network}

\newgacronymwithdesc{LPSCN}{locality preserving \glsentryshort{SCN}}{locality preserving \glsxtrshort{SCN}}
\newgacronymwithdesc{RS-SCN}{\glsentryshort{SCN} with rough set based attribute reduction}{\glsxtrshort{SCN} with rough set based attribute reduction}
\newgacronymwithdesc{GA-SCN}{\glsentryshort{SCN} based on genetic algorithms}{\glsxtrshort{SCN} based on genetic algorithms}
\newgacronymwithdesc{G-BAPSO-SCN}{\glsentryshort{SCN} with hybrid bat-particle swarm optimization}{\glsxtrshort{SCN} with hybrid bat-particle swarm optimization}
\newgacronymwithdesc{MoGL-SCN}{Bayesian robust \glsentryshort{SCN} based on a mixture of the Gaussian and Laplace distributions}{Bayesian robust \glsxtrshort{SCN} based on a mixture of the Gaussian and Laplace distributions}
\newgacronymwithdesc{OSCN}{orthogonal \glsentryshort{SCN}}{orthogonal \glsxtrshort{SCN}}
\newgacronymwithdesc{FSCN}{fast \glsentryshort{SCN}}{fast \glsxtrshort{SCN}}
\newgacronymwithdesc{ISSA-FSCN}{\glsentryshort{FSCN} with an improved sparrow search algorithm}{\glsxtrshort{FSCN} with an improved sparrow search algorithm}
\newgacronymwithdesc{BSCN}{bidirectional \glsentryshort{SCN}}{bidirectional \glsxtrshort{SCN}}
\newgacronymwithdesc{CSSA-SCN}{chaotic sparrow search algorithm based \glsentryshort{SCN}}{chaotic sparrow search algorithm based \glsxtrshort{SCN}}
\newgacronymwithdesc{DSCN}{deep \glsentryshort{SCN}}{deep \glsxtrshort{SCN}}
\newgacronymwithdesc{PRSCN}{pruning regularization \glsentryshort{SCN}}{pruning regularization \glsxtrshort{SCN}}
\newgacronymwithdesc{SCBNN}{stochastic configured Bayesian \glsentrylong{NN}}{stochastic configured Bayesian \glsxtrlong{NN}}

\newacronym{ELM}{ELM}{extreme learning machine}
\newgacronymwithdesc{DAELM}{domain adaptation \glsentryshort{ELM}}{domain adaptation \glsxtrshort{ELM}}
\newgacronymwithdesc{ML-OCELM}{multilayer \glsentrylong{NN} based one-class classification with \glsentryshort{ELM}}{multilayer \glsxtrlong{NN} based one-class classification with \glsxtrshort{ELM}}
\newgacronymwithdesc{Fuzzy-ELM}{fuzzy \glsentryshort{ELM}}{fuzzy \glsxtrshort{ELM}}
\newgacronymwithdesc{ASELM}{daptive semi-supervised \glsentryshort{ELM}}{daptive semi-supervised \glsxtrshort{ELM}}
\newgacronymwithdesc{K-ELM}{kernel based \glsentryshort{ELM}}{kernel based \glsxtrshort{ELM}}
\newgacronymwithdesc{OS-ELM}{online sequential \glsentryshort{ELM}}{online sequential \glsxtrshort{ELM}}
\newgacronymwithdesc{SAO-ELM}{structure-adjustable \glsentryshort{OS-ELM}}{structure-adjustable \glsxtrshort{OS-ELM}}
\newgacronymwithdesc{DOS-ELM}{dynamic forgetting factor based \glsentryshort{OS-ELM} algorithm}{dynamic forgetting factor based \glsxtrshort{OS-ELM} algorithm}
\newgacronymwithdesc{FOS-ELM}{fuzziness-based \glsentryshort{OS-ELM} algorithm}{fuzziness-based \glsxtrshort{OS-ELM} algorithm}
\newgacronymwithdesc{ML-DOS-ELM}{multilayer \glsentryshort{DOS-ELM}}{\glsxtrshort{DOS-ELM}}
\newgacronymwithdesc{I-ELM}{incremental \glsentryshort{ELM}}{incremental \glsxtrshort{ELM}}
\newgacronymwithdesc{CI-ELM}{convex \glsentrylong{I-ELM}}{convex \glsxtrlong{I-ELM}}
\newgacronymwithdesc{cWOB-ELM}{coiflet wavelet-based optimization method-based \glsentryshort{ELM}}{coiflet wavelet-based optimization method-based \glsxtrshort{ELM}}
\newgacronymwithdesc{ML-ELM}{multi-layer \glsentrylong{ELM}}{multi-layer \glsxtrlong{ELM}}
\newgacronymwithdesc{H-ELM}{hierarchical \glsentrylong{ELM}}{hierarchical \glsxtrlong{ELM}}
\newgacronymwithdesc{D-HELM}{densely connected \glsentryshort{D-HELM}}{densely connected \glsxtrshort{D-HELM}}
\newgacronymwithdesc{ES-ELM}{evolutionary optimized \glsentryshort{ELM}}{evolutionary optimized \glsxtrshort{ELM}}
\newgacronymwithdesc{EM-ELM}{error minimized \glsentrylong{ELM}}{error minimized \glsxtrlong{ELM}}
\newgacronymwithdesc{BELM}{bayesian \glsentrylong{ELM}}{bayesian \glsxtrlong{ELM}}
\newgacronymwithdesc{ADHKELM}{adaptive deep hybrid kernel  \glsentrylong{ELM}}{adaptive deep hybrid kernel  \glsxtrlong{ELM}}

\newacronym{SVM}{SVM}{support vector machine}
\newacronym{RC}{RC}{reservoir computing}
\newacronym{ESN}{ESN}{echo state network}
\newgacronymwithdesc{PESN}{polynomial \glsentryshort{ESN}}{polynomial \glsxtrshort{ESN}}
\newgacronymwithdesc{DeepESN}{deep \glsentrylong{ESN}}{deep \glsxtrlong{ESN}}
\newgacronymwithdesc{S-PESN}{simplified \glsentryshort{PESN}}{simplified \glsxtrshort{PESN}}
\newgacronymwithdesc{VML-ESN}{variable memory length \glsentryshort{ESN}}{variable memory length \glsxtrshort{ESN}}
\newgacronymwithdesc{DRESN}{double-reservoir \glsentryshort{ESN}}{double-reservoir \glsxtrshort{ESN}}
\newgacronymwithdesc{MI-ESN}{mutual information optimized \glsentryshort{ESN}}{mutual information optimized \glsxtrshort{ESN}}
\newgacronymwithdesc{DBEN}{deep belief \glsentrylong{ESN}}{deep belief \glsxtrlong{ESN}}
\newgacronymwithdesc{MR-ESN}{multiple reservoirs \glsentrylong{ESN}}{multiple reservoirs \glsxtrlong{ESN}}
\newgacronymwithdesc{ESN-DE}{differential evolution based \glsentryshort{ESN}}{differential evolution based \glsxtrshort{ESN}}
\newgacronymwithdesc{O-ESN}{particle swarm optimized \glsentryshort{ESN}}{particle swarm optimized \glsxtrshort{ESN}}
\newgacronymwithcustomshortdesc{piESN}{\ce{\pi}-ESN}{probabilistic \glsentryshort{ESN}}{probabilistic \glsxtrshort{ESN}}
\newgacronymwithdesc{HCR-ESN}{hybrid circle reservoir \glsentryshort{ESN}}{hybrid circle reservoir \glsxtrshort{ESN}}
\newgacronymwithdesc{SESN}{sinusoidal \glsentryshort{ESN}}{sinusoidal \glsxtrshort{ESN}}
\newgacronymwithdesc{FSDESN}{fast subspace decomposition \glsentrylong{ESN}}{fast subspace decomposition \glsxtrlong{ESN}}
\newgacronymwithdesc{RESN}{robust \glsentrylong{ESN}}{robust \glsxtrlong{ESN}}
\newgacronymwithdesc{FESN}{functional \glsentryshort{ESN}}{functional \glsxtrshort{ESN}}
\newgacronymwithdesc{TWIESN}{time warp invariant \glsentrylong{ESN}}{time warp invariant \glsxtrlong{ESN}}
\newacronym{SVESM}{SVESM}{support vector echo-state vector machine}
\newgacronymwithdesc{ESGNN}{echo state \glsentrylong{GNN}}{echo state \glsxtrlong{GNN}}
\newgacronymwithdesc{NGRC}{next generation \glsentrylong{RC}}{next generation \glsxtrlong{RC}}

\newacronym{RCN}{RCN}{random convolution node}

\newgacronymwithdesc{DNN}{deep \glsentrylong{NN}}{deep \glsxtrlong{NN}}
\newgacronymwithdesc{FFNN}{feed-forward \glsentrylong{NN}}{feed-forward \glsxtrlong{NN}}
\newacronym{GCN-RW}{GCN-RW}{graph convolutional networks with random weights}
\newacronym{RMDL}{RMDL}{random multimodel deep learning}
\newacronym{STM}{STM}{short-term memory}
\newacronym{GAN}{GAN}{generative adversarial network}
\newacronym{WRN}{WRN}{wide residual network}
\newacronym{RBM}{RBM}{restricted Boltzmann machine}
\newacronym{DBN}{DBN}{deep belief network}
\newacronym{DBM}{DBM}{deep Boltzmann machine}
\newacronym{RBF}{RBF}{radial basis function}
\newgacronymwithdesc{MI-CDBN}{mode isolation \glsentrylong{CDBN}}{mode isolation \glsxtrlong{CDBN}}
\newgacronymwithdesc{CDBN}{convolutional \glsentrylong{DBN}}{convolutional \glsxtrlong{DBN}}
\newgacronymwithdesc{GARBM}{Gaussian \glsentryshort{RBM} with binary auxiliary units}{Gaussian \glsxtrshort{RBM} with binary auxiliary units}
\newacronym{EBM}{EBM}{energy-based model}
\newacronym{AE}{AE}{autoencoder}
\newgacronymwithdesc{DAE}{denoising \glsentrylong{AE}}{denoising \glsxtrlong{AE}}
\newgacronymwithdesc{VAE}{variational \glsentrylong{AE}}{variational \glsxtrlong{AE}}
\newgacronymwithdesc{SAE}{sparse \glsentrylong{AE}}{sparse \glsxtrlong{AE}}
\newgacronymwithdesc{AVAE}{autoencoder \glsentryshort{VAE}}{autoencoder \glsxtrshort{VAE}}
\newgacronymwithdesc{PuVAE}{purifying \glsentryshort{VAE}}{purifying \glsxtrshort{VAE}}
\newacronym{TSP}{TSP}{travelling salesman problem}
\newgacronymwithdesc{EGAN}{edge adversarial  \glsentryshort{GAN}}{edge adversarial  \glsxtrshort{GAN}}
\newgacronymwithdesc{WGAN-GP}{Wasserstein \glsentryshort{GAN} with gradient penalty}{Wasserstein \glsxtrshort{GAN} with gradient penalty}
\newgacronymwithdesc{AC-GAN}{auxiliary classifier \glsentryshort{GAN}}{auxiliary classifier \glsxtrshort{GAN}}
\newgacronymwithdesc{TransGAN}{transformer based \glsentryshort{GAN}}{transformer based \glsxtrshort{GAN}}
\newgacronymwithdesc{SAGAN}{self-attention \glsentryshort{GAN}}{self-attention \glsxtrshort{GAN}}
\newgacronymwithdesc{STGAN}{selective transfer \glsentryshort{GAN}}{selective transfer \glsxtrshort{GAN}}
\newgacronymwithdesc{GCD-GAN}{gradient-guided dual-branch \glsentryshort{GAN}}{gradient-guided dual-branch \glsxtrshort{GAN}}
\newgacronymwithdesc{RI-GAN}{\glsentryshort{GAN} with residual inception modules}{ \glsxtrshort{GAN} with residual inception modules}
\newgacronymwithdesc{AutoGAN}{\glsentryshort{GAN} with neural architecture search}{\glsxtrshort{GAN} with neural architecture search}
\newgacronymwithdesc{PLGAN}{panoptic layout \glsentryshort{GAN}}{panoptic layout \glsxtrshort{GAN}}
\newgacronymwithdesc{DF-GAN}{deep fusion \glsentryshort{GAN}}{deep fusion \glsxtrshort{GAN}}
\newgacronymwithdesc{CMAFGAN}{cross-modal attention gusion based \glsentryshort{GAN}}{cross-modal attention gusion based \glsxtrshort{GAN}}
\newgacronymwithdesc{GGAN}{graph \glsentryshort{GAN}}{graph \glsxtrshort{GAN}}
\newgacronymwithdesc{AM-GAN}{fused \glsentryshort{GAN} with attention mechanism}{fused \glsxtrshort{GAN} with attention mechanism}
\newgacronymwithdesc{DGattGAN}{dual Generator attentional \glsentryshort{GAN}}{dual Generator attentional \glsxtrshort{GAN}}
\newgacronymwithdesc{CML-GAN}{contrastive meta-learning \glsentryshort{GAN}}{contrastive meta-learning \glsxtrshort{GAN}}
\newgacronymwithdesc{D2GAN}{dual discriminator \glsentryshort{GAN}}{dual discriminator \glsxtrshort{GAN}}
\newgacronymwithdesc{D2WMGAN}{dual discriminator weighted mixture \glsentryshort{GAN}}{dual discriminator weighted mixture \glsxtrshort{GAN}}
\newgacronymwithdesc{RoCGAN}{robust conditional \glsentryshort{GAN}}{robust conditional \glsxtrshort{GAN}}
\newgacronymwithdesc{VARGAN}{variance enforcing \glsentryshort{GAN}}{variance enforcing \glsxtrshort{GAN}}
\newgacronymwithdesc{DPGAN}{dual-stream \glsentryshort{GAN} with phase awareness}{dual-stream \glsxtrshort{GAN} with phase awareness}
\newgacronymwithdesc{GAGAN}{geometry-aware \glsentryshort{GAN}}{geometry-aware \glsxtrshort{GAN}}
\newgacronymwithdesc{RePGAN}{\glsentryshort{GAN} with residual partial modules}{\glsxtrshort{GAN} with residual partial modules}
\newgacronymwithdesc{SE-DCGAN}{squeeze-excitation network-deep convolution \glsentryshort{GAN}}{squeeze-excitation network-deep convolution \glsxtrshort{GAN}}
\newgacronymwithdesc{EA-GAN}{example attention \glsentryshort{GAN}}{example attention \glsxtrshort{GAN}}
\newgacronymwithdesc{HGAN}{hyperbolic \glsentryshort{GAN}}{hyperbolic \glsxtrshort{GAN}}
\newgacronymwithdesc{CRRAGAN}{cascading residual--residual attention \glsentryshort{GAN}}{cascading residual--residual attention \glsxtrshort{GAN}}
\newgacronymwithdesc{CycleGAN}{cycle-consistent \glsentryshort{GAN}}{cycle-consistent \glsxtrshort{GAN}}
\newgacronymwithdesc{cscGAN}{conditional single-cell \glsentryshort{GAN}}{conditional single-cell \glsxtrshort{GAN}}

\newgacronymwithdesc{MRI}{magnetic resonance imaging}{magnetic resonance imaging}
\newgacronymwithdesc{sMRI}{structural \glsentryshort{MRI}}{structural \glsxtrshort{MRI}}
\newacronym{CT}{CT}{computed tomography}
\newacronym{SPECT}{SPECT}{single-photon emission computed tomography}
\newacronym{PPGN}{PPGN}{Plug and Play generative networks}
\newacronym{DM}{DM}{diffusion model}
\newacronym{DDPM}{DDPM}{denoising diffusion probabilistic model}
\newacronym{SCIBER}{SCIBER}{single-cell integrator and batch effect remover}

\newacronym{DP}{DP}{dynamic programming}

\newglossaryentrywithacronymandref{TAAF}{transformative adaptive activation function}{transformative adaptive activation function}{a class of \glsxtrlongpl{AAF} allowing for translation and scaling of an activation function; proposed in this work}{sec:taaf}
\newglossaryentrywithacronymandref{ATU}{adaptive transformative unit}{adaptive transformative unit}{a proposed unit, used for the implementation of the proposed \glsxtrshort{TAAF}}{sec:atu}
\newgacronymwithdesc{DTAAF}{dual \glsentrylong{TAAF}}{dual \glsxtrlong{TAAF}}
\newgacronymwithdesc{GDTAAF}{generalized \glsentrylong{DTAAF}}{generalized \glsxtrlong{DTAAF}}

\newacronym{MAE}{MAE}{mean absolute error}
\newacronym{RMS}{RMS}{root mean square}
\newacronym{MSE}{MSE}{mean squared error}
\newacronym{SSE}{SSE}{sum of squared errors}
\newacronym{CI}{CI}{confidence interval}
\newacronym{MCC}{MCC}{Matthew's correlation coefficient}
\newgacronymwithdesc{MMAE}{mean \glsentrylong{MAE}}{mean \glsxtrlong{MAE}}
\newgacronymwithref{MDAE}{mean difference of absolute errors}{eq:mdae}
\newacronymwithrefdesc{MMDAE}{mean \glsentryshort{MDAE}}{mean \glsxtrshort{MDAE}}{eq:mmdae}

\newacronym{FPGA}{FPGA}{field-programmable gate array}

\makeglossaries

\endgroup

\begin{document}
\maketitle

\begin{abstract}
	Neural networks have proven to be a highly effective tool for solving complex problems in many areas of life. Recently, their importance and practical usability have further been reinforced with the advent of deep learning. One of the important conditions for the success of neural networks is the choice of an appropriate activation function introducing non-linearity into the model. Many types of these functions have been proposed in the literature in the past, but there is no single comprehensive source containing their exhaustive overview. The absence of this overview, even in our experience, leads to redundancy and the unintentional rediscovery of already existing activation functions. To bridge this gap, our paper presents an extensive survey involving 400 activation functions, which is several times larger in scale than previous surveys. Our comprehensive compilation also references these surveys; however, its main goal is to provide the most comprehensive overview and systematization of previously published activation functions with links to their original sources. The secondary aim is to update the current understanding of this family of functions.
\end{abstract}
\keywords{\myKeywords}

\section{Introduction}
Neural networks --- and deep learning in particular --- have exhibited remarkable success in addressing diverse challenges across various fields. They stand as state-of-the-art approaches, showcasing their prowess in solving complex and intricate problems. At the heart of these networks, \glspl{AF} play an important role by introducing nonlinearity to neural network layers. In the absence of nonlinear \glspl{AF}, typical neural networks would only model a weighted sum of inputs, limiting their capacity to capture intricate relationships within the data.

The choice of \glsxtrlongpl{AF} profoundly influences a network's ability to learn and generalize, directly impacting its performance across a spectrum of tasks. Effective \glsxtrlongpl{AF} possess several key properties, as outlined by \citeauthor{Dubey2022} in \cite{Dubey2022}: a) introducing non-linear curvature to enhance training convergence within the optimization landscape; b) maintaining an unobstructed gradient flow during training; c) ensuring a minimal increase in the computational complexity of the model; and d) preserving the distribution of data to optimize the network's training.

There are many \glsxtrlongpl{AF} proposed in the literature in the last three decades --- some more computationally complex or with higher performance than others. However, further research of the \glsxtrlongpl{AF} is hampered by the absence of a consolidated list. This gap leads to the inadvertent reinvention of existing \glsxtrlongpl{AF} and the independent proposal of identical or very similar ones, resulting in a wasteful consumption of research resources. Even comprehensive surveys and reviews, such as those by \citeauthor{Dubey2022} \cite{Dubey2022} and \citeauthor{Apicella2021} \cite{Apicella2021}, often omit numerous \glsxtrlongpl{AF} present in the literature; furthermore, these reviews are also a bit older and many new \glsxtrlongpl{AF} emerged since then. This oversight can lead to instances where an \gls{AF} is redundantly proposed as novel, despite its prior introduction in the literature --- e.g., \gls{RePU} (\cref{sec:repu}), \gls{DPReLU} (\cref{sec:dprelu}), \gls{TRec} (\cref{sec:trec}), \gls{ReLUSwish} (\cref{sec:fts}), and \gls{BReLU} (\cref{sec:brelu}). By providing a more extensive list of available \glsxtrlongpl{AF}, we aim to avoid such redundancy and promote faster advances in the research of \glsxtrlongpl{AF} in \glsxtrlongpl{NN}. 

To address this issue, we strive to provide an extensive and consolidated list of available \glspl{AF}. This survey aims to prevent redundancy, eliminate the reinvention of established \glspl{AF} to promote innovation, and accelerate the advancement of research in the field of neural networks. By offering a comprehensive resource, we aim to promote efficiency and innovation in the exploration of \glspl{AF} within the field.

It is important to note that our contribution primarily focuses on providing a comprehensive list of \glspl{AF} rather than conducting extensive benchmarks or in-depth analyses. The breadth of the compilation encompasses a wide array of \glspl{AF}, making a detailed benchmark or a deeper analysis beyond the scope of this work. Our aim is to provide researchers with a foundational resource that facilitates informed decision-making in selecting \glspl{AF} for neural networks, recognizing that a more exhaustive exploration or detailed analysis would necessitate a dedicated and focused effort beyond the scope of this comprehensive listing. The presented overview is limited to real-valued \glsxtrlongpl{AF}; complex-valued \glsxtrlongpl{NN} (e.g., \cite{Celebi2019, Zhang2023GaussianType, Scardapane2020, Ornek2022, Hua2021, Kim2002, Savitha2009, Huang2008, Savitha2012, Hu2020, Tan2018, Kuroe2003, Ozdemir2011, Gao2019Enhanced}, brief overview available in \cite{Lee2022ComplexValued, Jagtap2023}), bicomplex-valued \glsxtrlongpl{NN} (e.g., \cite{Vieira2023}),  quaternion-valued \glsxtrlongpl{NN} (e.g., \cite{GarcaRetuerta2020, Zhu2018, Popa2016, Popa2017, Yu2023Multistability}), photonic \glsxtrlongpl{NN} (e.g., \cite{Xu2022Reconfigurable}), fuzzy \glsxtrlongpl{NN} (e.g., \cite{deCamposSouza2020, Das2020, Bangare2022, Malcangi2021, Fei2022Fractional, Liu2021APath}), \glspl{AF} for probabilistic boolean logic (e.g., \cite{Duersch2022}), quantum \glspl{AF} (e.g., \cite{Parisi2022}) and others are out of the scope of this work.\footnote{While these kinds of \glspl{NN} are not discussed throughout this work, some of these approaches will use \glspl{AF} presented in this work.}

We have chosen to categorize \glspl{AF} into two main classes: fixed \glspl{AF} (\cref{sec:classical_activation_functions}) and \glspl{AAF} (\cref{sec:adaptive_activation_functions}), the latter having a parameter that is trained alongside the other weights in a network. Although instances exist where \glspl{AF} are virtually identical, differing only in the presence of a particular adaptive parameter (e.g., \gls{swish} (see \cref{sec:swish}) and \glsxtrshort{SiLU} (see \cref{sec:silu})), this classification proves valuable. \Glspl{AAF}, by virtue of their parameterization, offer an added layer of flexibility in capturing complex relationships within the data during the training process. 

\section{Literature review}
There are several reviews of \glspl{AF} available in the literature; however, most of them encompass only the most commonly known \glspl{AF}. While this is sufficient for an overview for newcomers to the field, it does not allow for efficient research of \glspl{AF} themselves. Probably the most extensive review is the \cite{Dubey2022} from \citeyear{Dubey2022}, which lists over 70 \glspl{AF} and provides a benchmark for 18 of them. Other reviews works containing list of \glspl{AF} include \cite{Apicella2021, Farzad2017, Bingham2022, Badiger2022, Jagtap2022, Jagtap2023,Nwankpa2018,Duch2000, Raitani2022,Datta2020,Mercioni2023,Rasamoelina2020,Mercioni2020TheMost,Liu2021Analysis,Kiliarslan2021AnOverview}.

While there are several existing works \cite{Jagtap2022, Pandey2023, Noel2021, Dubey2022, Ramachandran2017, Eger2018, Nanni2022, Adem2022, Adu2021, Haq2021, Nguyen2021, Mishra2017, Ratnawati2020, Dureja2019, Vargas2021, Singh2023Impact, Lai2023, Papavasileiou2017, Kang2021, Vargas2023, Jagtap2023, Emanuel2023, Zhang2023Enhancing, Singh2023, Lin2024, Koh2022Solar, Bhojani2021, Hurley2023, Makhrus2023, Mishra2021ANonMonotonic, EssaiAli2022, Rumala2020, Suciningtyas2023, Mercioni2023, Wang2023Learning, Pantal2023, Dung2023, Liu2019Comparison, Ingole2020, Nanni2022Comparisons, Zhang2022Transfer, Chaturvedi2022, Pedamonti2018, Bai2022RELUFunction, Kamalov2021, Lau2018, Dubowski2020, Kandhro2022, Seyrek2023, Bircanoglu2018, Tatraiya2023, Qiu2023, Kayala2021, Xu2020Comparison, Purba2020, Salam2021, Lutakamale2022} that offer benchmarks and empirical comparisons of various \glspl{AF}, it is unfortunate that these studies are often constrained by a limited selection of \glspl{AF}. Typically, the focus is centered on the most well-known and widely used functions, neglecting the broader spectrum of \glspl{AF} available in the literature. 

To avoid the manual selection of \glspl{AF}, many researchers resort to various optimization approaches to find the optimal \gls{AF} for their problems.  e.g. an evolutionary approach was used to evolve the optimal \glsxtrlong{AF} in \cite{Nader2021, Bingham2022, Basirat2018, Basirat2019, Mayer2002, Hagg2017, Knezevic2023, Liu1996, Cui2022, Cui2019, Vijayaprabakaran2022, ONeill2018, Sipper2021,Salimi2023, Salimi2023SARAF, Li2023GAAF,Chen2016Combinatorially,Vijayaprabakaran2021} and grid search using artificial data was used in \cite{Pan2020}. Another search for the optimal \glsxtrlongpl{AF} was presented in \cite{Ramachandran2017} where several simple \glsxtrlongpl{AF} were found to perform remarkably well. These automatic approaches might be used for evolving the \glsxtrlongpl{AF} (e.g., \cite{Nader2021, Knezevic2023}) or for selecting the optimal \glsxtrlong{AF} for a given neuron (e.g., \cite{Cui2019, Marchisio2018}). While evolved \glsxtrlong{AF} may perform well for a given problem, they also might be very complex --- e.g., evolved \glsxtrlongpl{AF} in \cite{Knezevic2023}. The complexity of an \glsxtrlong{AF} is also important characteristic as it significantly influences the computational efficiency of a \glsxtrlong{NN}; however, this might be mitigated by efficient implementations (including hardware implementations) of such \glsxtrlongpl{AF} (e.g., \cite{Tripathi2021, Bouguezzi2021, Li2018AnEfficient, Tsai2015, Namin2009, Pogiri2022, Shakiba2021, Xie2020, Priyanka2019, Lin2023LowArea, Shatravin2022}). An empirical analysis of computational efficiency and power consumption of various \glspl{AF} is available in \cite{Derczynski2020}. 

\section{Classical activation functions}
\label{sec:classical_activation_functions}
First, our discussion delves into fixed \glsxtrlongpl{AF}, devoid of adaptive parameters. This category of \glsxtrlongpl{AF} represents the basic type that was predominantly employed in the initial \glsxtrlong{NN} architectures and continues to be prevalent today. Fixed \glsxtrlongpl{AF}, such as the \gls{logisticsigmoid} and \glsxtrlong{tanh}, are characterized by their predetermined mathematical formulations, where the activation output solely depends on the input without the introduction of any trainable parameters.

\subsection{Binary activation function}
\label{sec:binaryaf}

The \gls{binary AF} --- also called a step function --- is a simple yet important \glsxtrlong{AF} used in \glsxtrlongpl{NN} \cite{Zhang2018Efficient}. It assigns an output value of 1 if the input is positive or zero and an output value of 0 if the input is negative \cite{Apicella2021}. Mathematically, it can be defined as follows:

\begin{equation}
    f(z)=\begin{cases}
                1, \quad &  z \geq 0, \\
                0, \quad & z < 0. \\
        \end{cases}
\end{equation}

Similar to \glsxtrlong{binary AF} is the sign function, which produces an output value of -1 if the input is negative and 1 if it is positive (and 0 for outputs that are exactly zero) \cite{Apicella2021}. Since the sign and the binary \glsxtrlongpl{AF} have nearly exact properties from the point of view of \glsxtrlongpl{NN}, only the \glsxtrlong{binary AF} is mentioned, but the points hold similarly for the sign \glsxtrlong{AF}.

The main advantage of the \glsxtrlong{binary AF} is that it is straightforward and computationally efficient to implement. It does not involve complex mathematical operations, making it suitable for networks with low computational resources or for hardware implementations \cite{Dinu2007, Dinu2010}.
    However, the \glsxtrlong{binary AF} has one glaring disadvantage - the lack of differentiability. The \glsxtrlong{binary AF} is not differentiable at the point of discontinuity (x = 0) and is zero elsewhere. This poses challenges for optimization algorithms that rely on gradients, such as \gls{BP}, since the gradient is noninformative \cite{Zhang2018Efficient, Chen2023Saturated, Noel2021}.
Since the gradient-based methods are used predominantly, the \glsxtrlong{binary AF} is used very rarely and is important mainly for historical reasons as it was used in the original perceptron \cite{Rosenblatt1958, Noel2021}.

\subsection{Sigmoid family of activation functions}
\label{sec:logistic_sigmoid}
Various smoothed variants of the binary \glsxtrlongpl{AF} (\glspl{sigmoid}) are commonly used; the most common is the \glslink{logisticsigmoid}{logistic function} --- the standard \gls{logisticsigmoid} function was dominant in the field prior the introduction of \gls{ReLU} (see \cref{sec:relu}) \cite{Goodfellow2016}, the \glslink{logisticsigmoid}{logistic function} is often called just sigmoid in the literature which is also used throughout this work for brevity (unless specified otherwise, \textit{sigmoid} is equivalent to standard \glslink{logisticsigmoid}{logistic function} in the text).
Standard \glslink{logisticsigmoid}{logistic function} is defined as
\begin{equation}
    f(z) = \sigma(z) = \frac{1}{1+\exp(-z)}.
\end{equation}
The \gls{logisticsigmoid} was a popular choice since its output values can interpreted as the probability that a binary variable is 1 \cite{Goodfellow2016} since it squashes the input to the interval $(0,1)$ \cite{Dubey2022}. The problem of sigmoid \glsxtrlongpl{AF} is that they saturate --- they saturate when their input $z$ is either a large positive number or a large negative number, which makes gradient-based learning difficult \cite{Goodfellow2016, Dubey2022}; therefore their use in feedforward networks is usually discouraged \cite{Goodfellow2016}. Another option, albeit significantly less popular in \glspl{ANN}, is the \gls{probit} \gls{AF} \cite{daSGomes2010}, which is just the cumulative standard normal distribution function used as an \gls{AF} \cite{daSGomes2010}.

Another popular sigmod function is the \gls{tanh} \glsxtrlong{AF} which is just scaled and shifted \gls{logisticsigmoid}
\begin{equation}
    \mathrm{tanh}(z) = \frac{\exp(z) - \exp(-z)}{\exp(z) + \exp(-z)} = 2\sigma(2z)-1.
\end{equation}
Similarly as the \gls{logisticsigmoid}, the \gls{tanh} also squashes the inputs; however, it squashes them to the interval $(-1, 1)$. The \gls{tanh} function is often advantageous over the \gls{logisticsigmoid} function because it is centered around zero and it is similar to the identity function near zero, which makes training of a network easier if the activations are kept small \cite{Goodfellow2016}. Nevertheless, the \gls{tanh} function saturates similarly as does the \gls{logisticsigmoid} and therefore similarly suffers from the vanishing gradients \cite{Dubey2022}. Computationally efficient approximation of the \gls{tanh} \glsxtrlongpl{AF} based on splines were proposed in \cite{Simos2021} -- \textit{tanh36} based on approximation relying on 36 equidistant points and \textit{tanh3} using only 3 points. Scaled variant $\tanh\left(\frac{z}{2}\right)$ was used in \cite{Dudek2021}. The \gls[prereset]{LRTanh} is a \gls{tanh} variant used together with modified \gls{BP} that substitutes a different \glsxtrlong{AF} derivative proposed in \cite{Edwards2019}.  There are also approximations of the \gls{logisticsigmoid} and \gls{tanh} that are meant to speed up the computations; e.g., \glsxtrshort{pRPPSG} \cite{Sunat2006} and other similar piecewise approximations \cite{Kwan1992,MingZhang1996}.

A scaled version of the \gls{logisticsigmoid} function was proposed in \cite{Xu2016} with the motivation to have the same linear regimes as the \gls{tanh} and relu \glsxtrlongpl{AF} when initialized with the popular normalized initialized method proposed in \cite{Glorot2010}. The scaled version used fixed parameters
\begin{equation}
    f(z) = 4\sigma(z) - 2.
\end{equation}

A more complicated variant named \gls{nsigmoid} was proposed in \cite{Mulindwa2023}; however, it seems that the formula presented in the paper is not as the authors intended and, therefore, we omit this \gls{AF} from the list.

\subsubsection{Shifted and scaled sigmoid (SSS)}
\label{sec:sss}
The \gls[prereset]{SSS} was used in \cite{Arai2018}; it is the \gls{logisticsigmoid} with horizontal scaling and translation defined as
\begin{equation}
    f(z) = \sigma\left(a\left(z-b\right)\right) = \frac{1}{1+\exp\left(-a\left(z-b\right)\right)},
\end{equation}
where $a$ and $b$ are predetermined parameters; \citeauthor{Arai2018} used $a=0.02$ and $b=600$.

\subsubsection{Variant sigmoid function (VSF)}
\label{sec:vsf}
The \gls[prereset]{VSF} is an older parametric variant of the \gls{logisticsigmoid} proposed in \cite{Han1995}. It is defined as
\begin{equation}
    f(z) = a\sigma\left(bz\right) - c = \frac{a}{1+\exp\left(-bz\right)}-c,
\end{equation}
where $a$, $b$, and $c$ are predetermined parameters \cite{Han1995}.

\subsubsection{Scaled hyperbolic tangent}
\label{sec:stanh}
A parametric version called \gls{stanh} was used in \cite{Lecun1998}:
\begin{equation}
    f(z) = a \tanh\left(b\cdot z\right),
\end{equation}
where $a$ and $b$ are fixed hyperparameters that control the scaling of the function. \citeauthor{Lecun1998} proposed using $a=1.7159$ and $b=\frac{2}{3}$

A similar concept was analyzed in \cite{SinghSodhi2014} where \glspl{sigmoid} with bi-modal derivatives were used as \glsxtrlongpl{AF}. An example of such a function is 
\begin{equation}
    f(z) = \frac{1}{2}\left(\frac{1}{1+\exp(-z)} + \frac{1}{1+\exp(-z-b)}\right),
\end{equation}
where $b$ is a hyperparameter \cite{SinghSodhi2014, }; similarly, additional three \glsxtrlongpl{AF} with bi-modal derivates were proposed in \cite{SinghSodhi2014}.

\subsubsection{Arctan}
\label{sec:arctan}
The \gls{arctan} function and its variation were used as \glsxtrlongpl{AF} in \cite{Sivri2022}:
\begin{equation}
    f(z) = \tan^{-1}(z).
\end{equation}
The \gls{arctan} resembles a \gls{logisticsigmoid} activation, however, it covers wider range $\left(-\frac{\pi}{2}, \frac{\pi}{2}\right)$ \cite{Sivri2022}. The \gls{arctan} and several its variation were compared with the \gls{tanh}, \gls{ReLU}, \gls{LReLU}, \gls{logisticsigmoid} activation, and \gls{swish} in \cite{Sivri2022}; the best-performing functions in the presented experiments were the \gls{arctan} and its variation arctanGR \cite{Sivri2022}. Interestingly, the \gls{arctan} was used as an \gls{AF} twenty years earlier in \cite{Kamruzzaman2002}. The arctanGR is a scaled version of the \gls{arctan} and is defined as
\begin{equation}
    f(z) = \frac{\tan^{-1}(z)}{\frac{1+\sqrt{2}}{2}}.
\end{equation}
Other scaling variants such as division by the $\pi$, $\frac{1+\sqrt{5}}{2}$, or the Euler number are presented in \cite{TmerSivri2023}.

\subsubsection{Sigmoid-Algebraic activation function}
\label{sec:sigmoidalgebraic}
The \gls{Sigmoid-Algebraic} is a \gls{sigmoid} variant defined in \cite{Koak2021}. It is defined as
\begin{equation}
 f(z) = \frac{1}{1+\exp\left(-\frac{z\left(1+a|z|\right)}{1+|z|\left(1+a|z|\right)}\right)},
\end{equation}
where $a \geq 0$ is a parameter \cite{Koak2021}.

\subsubsection{Triple-state sigmoid}
\label{sec:tssigmoid}
The \gls{TS-sigmoid} is a cascaded \gls{AF} similar to \glsxtrshort{TS-swish} (see \cref{sec:tsswish}) \cite{Koak2021}; it is defined as
\begin{equation}
f(z) = \frac{1}{1+\exp\left(-z\right)}\left(\frac{1}{1+\exp\left(-z\right)}+\frac{1}{1+\exp\left(-z+a\right)}+\frac{1}{1+\exp\left(-z+b\right)}\right),
\end{equation}
where $a$ and $b$ are fixed parameters \cite{Koak2021}.

\subsubsection{Improved logistic sigmoid}
\label{sec:improved_logistic_sigmoid}
The \gls{improvedlogisticsigmoid} is yet another \gls{sigmoid} based \glsxtrlong{AF} designed to deal with the vanishing gradient problem
\begin{equation}
    f(z)=\begin{cases}
                a(z - b) + \sigma(b), \quad & z \geq b, \\
                \sigma(z), \quad & -b < z < b , \\
                a(z + b) + \sigma(b), \quad & z \leq -b, \\
        \end{cases}
\end{equation}
where $a$ and $b$ are fixed parameters \cite{Qin2019}; $a$ controls the slope and $b$ is a thresholding parameter. The authors recommend a bound on the slope parameter $a$:
\begin{equation}
    a > a_{\mathrm{min}} = \frac{\exp(-b)}{\left(1+\exp(-b)\right)^2}.
\end{equation}
Even though the parameters are fixed during the training of a network, a procedure for preseting them based on the network and data was proposed in \cite{Qin2019}. The output range of the \gls{SiLU} is $(-\infty, \infty)$ \cite{Dubey2022}. The authors \citeauthor{Qin2019} also showed that the \gls{improvedlogisticsigmoid} \gls{AF} has a higher convergence speed than the \gls{logisticsigmoid} \gls{AF} \cite{Qin2019}.

\subsubsection{Combination of the sigmoid and linear activation (SigLin)}
\label{sec:siglin}
A \gls[prereset]{SigLin}\footnote{This abbreviation is used only in this work; \citeauthor{Roodschild2020} did not name the function in \cite{Roodschild2020}.} was used as an \gls{AF} in \cite{Roodschild2020}. The \gls{SigLin} is defined as
\begin{equation}
    f(z) = \sigma\left(z\right)+ az,
\end{equation}
where $\sigma(z)$ is the \gls{logisticsigmoid} \gls{AF} and $a$ is a fixed parameter \cite{Roodschild2020}; however, this \gls{AF} was used only in a modified optimization procedure \cite{Roodschild2020}. \Citeauthor{Roodschild2020} experimented with $a \in \{0, 0.05, 0.1, 0.15\}$ \cite{Roodschild2020}.

\subsubsection{Penalized hyperbolic tangent}
\label{sec:penalized_hyperbolic_tangent}
A \gls{ptanh} the \gls{LReLU} (see \cref{sec:relu}) but uses the \gls{tanh} function instead of the linear function \cite{Xu2016}:
\begin{equation}
    f(z)=\begin{cases}
                \tanh(z), \quad & z \geq 0, \\
                \dfrac{\tanh(z)}{a},  \quad & z < 0, \\
        \end{cases}
\end{equation}
where $a \in (1, \infty)$. This function has similar values near 0 as the \gls{LReLU} with identical parameter $a$ as they both share the same Taylor expansion up to the first order \cite{Xu2016}; however this function saturates to $-\frac{1}{a}$ for $z \rightarrow -\infty$ and to 1 for $z \rightarrow \infty$ \cite{Xu2016}. The \gls{ptanh} \gls{AF} was found to perform consistently well for various \gls{NLP} tasks compared to \gls{ReLU}, \gls{LReLU} and several other \glsxtrlongpl{AF} \cite{Eger2018}.

\subsubsection{Soft-root-sign (SRS)}
\label{sec:srs}
A \gls{SRS} \glsxtrlong{AF} is a parametric, smooth, non-monotonic, and bounded \glsxtrlong{AF} \cite{Li2020SoftRootSign}. It is defined as
\begin{equation}
    f(z) = \frac{z}{\frac{z}{a} + \exp\left(-\frac{z}{b}\right)},
\end{equation}
where $a$ and $b$ are predetermined parameters \cite{Li2020SoftRootSign}; the authors \citeauthor{Li2020SoftRootSign} propose using $a = 2$ and $b = 3$ whereas the parameters are said to be learnable in \cite{Dubey2022}. The output range of \gls{SRS} is $\left[ \frac{ab}{b-a\mathrm{e}}, a\right]$ \cite{Li2020SoftRootSign,Dubey2022}. The performance of the \gls{SRS} was demonstrated using hte CIFAR-10 and CIFAR-100 \cite{Krizhevsky2009} task in comparison with the \gls{ReLU} (see \cref{sec:relu} for the description of the \gls{ReLU} family of AFs), \gls{LReLU}, \gls{PReLU}, \gls{softplus}, \gls{ELU}, \gls{SELU}, and \gls{swish} \cite{Li2020SoftRootSign}.

\subsubsection{Soft clipping (SC)}
\label{sec:sc}
The \gls[prereset]{SC} \cite{Ohn2019,Klimek2020} \gls{AF} is another bounded \gls{AF}; it is approximatelly piecewise linear in the range $z \in (0,1)$ and it is defined as
\begin{equation}
    f(z) = \frac{1}{a}\ln\left(\frac{1+\exp\left(az\right)}{1+\exp\left(a(z-1)\right)}\right),
\end{equation}
where $a$ is a fixed parameter \cite{Klimek2020}.

\subsubsection{Hexpo}
\label{sec:hexpo}
The \gls{hexpo} \glsxtrlong{AF} \cite{Kong2017} was proposed in order to minimize the problem of vanishing gradient \cite{Dubey2022}; it resembles a \gls{tanh} \glsxtrlong{AF} with scaled gradients \cite{Dubey2022}:
\begin{equation}
    f(z)=\begin{cases}
                -a \left(\exp\left(-\frac{z}{b}\right) - 1\right), \quad & z \geq 0, \\
                c \left(\exp\left(-\frac{z}{d}\right) - 1\right),  \quad & z < 0, \\
            \end{cases}
    \end{equation}
where $a$, $b$, $c$, and $d$ are fixed parameters. While the parameters could be trainable in theory, it is not recommended as it would lead to the vanishing gradient problem \cite{Kong2017}. The \gls{hexpo}  functions allow for control over the gradient by tunning the parameters $a$, $b$, $c$, and $d$ and the ratios $\frac{a}{b}$ and $\frac{c}{d}$ --- with increasing the ratios $\frac{a}{b}$ or $\frac{c}{d}$, the rate of gradient decay to zero decreases; increasing only $a$ and $c$ scales the gradient around the origin up \cite{Kong2017}.

\subsubsection{Softsign}
\label{sec:softsign}
A \gls{softsign} \glsxtrlong{AF} is a smooth \glsxtrlong{AF} similar to the \gls{tanh} activation; however, it is less prone to vanishing gradients \cite{Pandey2023}. It is defined as
\begin{equation}
    f(z) = \frac{z}{1+|z|},
\end{equation}
where $|z|$ denotes the absolute value of $z$ \cite{Pandey2023}.

\subsubsection{Smooth step}
\label{sec:smooth_step}
The \gls{smooth_step} is a \gls{sigmoid} \gls{AF}; it is defined as
\begin{equation}
    f(z)=\begin{cases}
                1 \quad & z \geq \frac{a}{2}, \\
                -\frac{2}{a^3}z^3 + \frac{3}{2a}z  + \frac{1}{2}, \quad & -\frac{a}{2} \leq z \leq \frac{a}{2}, \\
                0 \quad & z \leq -\frac{a}{2}, \\
            \end{cases}
\end{equation}
where $a$ is a fixed hyperparameter \cite{Hazimeh2020}.

\subsubsection{Elliott activation function}
\label{sec:elliott}
\Gls{Elliott} \glsxtrlong{AF} is one of the earliest proposed \glsxtrlongpl{AF} to replace to replace the \gls{logisticsigmoid} or \gls{tanh} \glsxtrlongpl{AF} \cite{Elliott1993}; the \gls{Elliott} \gls{AF} is a scaled and translated \gls{softsign} \gls{AF}. It is defined as \cite{Farzad2017,Dubey2022}
\begin{equation}
    f(z) = \frac{0.5z}{1+\left|z\right|} + 0.5.
\end{equation}
The output of the \gls{Elliott} \glsxtrlongpl{AF} is in range $[0,1]$ \cite{Dubey2022, Farzad2017}. The main advantage of the \gls{Elliott} \gls{AF} is that it can be calculated much faster than the \gls{logisticsigmoid} \cite{Burhani2015}.

\subsubsection{Sinc-Sigmoid}
\label{sec:sincsigmoid}
The \gls{SincSigmoid} is a \gls{sigmoid}-based \gls{AF} proposed in \cite{Koak2021}. It is defined as
\begin{equation}
    f(z) = \mathrm{sinc}\left(\sigma\left(z\right)\right),
\end{equation}
where $\mathrm{sinc}\left(x\right)$ is the unnormalized\footnote{\Citeauthor{Koak2021} did not specify whether it is the normalized or unnormalized variant. Still, they provided the derivative of the \gls{SincSigmoid}, which suggests that the unnormalized variant was used.} sinc function \cite{Koak2021}.

\subsubsection{Sigmoid-Gumbel activation function}
\label{sec:sg}
The \gls[prereset]{SG} is a non-adaptive \gls{AF} proposed recently in \cite{Kaytan2022}; it is defined as
\begin{equation}
 f(z) = \frac{1}{1+\exp\left(-z\right)}\exp\left(-\exp\left(-z\right)\right).
\end{equation}

\subsubsection{NewSigmoid}
\label{sec:newsigmoid}
The \gls{NewSigmoid} is a \gls{sigmoid} variant proposed in \cite{Kumar2023Image}. It is defined as
\begin{equation}
    f(z) = \frac{\exp(z)-\exp(-z)}{\sqrt{2\left(\exp(2z)+\exp(-2z)\right)}}.
\end{equation}

\subsubsection{Root2sigmoid}
\label{sec:root2sigmoid}
The \gls{root2sigmoid} is another \gls{sigmoid} variant proposed in \cite{Kumar2023Image}. It is defined\footnote{The author had probably a typo in the definition in the original paper \cite{Kumar2023Image}; we present the formula we think \citeauthor{Kumar2023Image} intended to write --- it resembles the \gls{NewSigmoid} and fits the numerical values given in the paper.} as
\begin{equation}
    f(z) = \frac{\sqrt{2}^z-\sqrt{2}^{-z}}{2\sqrt{2}\sqrt{2\left(\sqrt{2}^{2z}+\sqrt{2}^{-2z}\right)}}.
\end{equation}

\subsubsection{LogLog}
\label{sec:loglog}
The \gls{LogLog} is a simple \gls{AF} proposed in \cite{daSGomes2010}; it is defined as
\begin{equation}
    f(z) = \exp\left(-\exp\left(-z\right)\right).
\end{equation}
The LogLog, \glsxtrshort{cLogLog} (see \cref{sec:cloglog}) were used in \glspl{NN} for forecasting financial time-series in \cite{daSGomes2010}.

\subsubsection{Complementary Log-Log (cLogLog)}
\label{sec:cloglog}
The \gls[prereset]{cLogLog} is another simple \gls{AF} proposed in \cite{daSGomes2010} complementing the \gls{LogLog} (see \cref{sec:loglog}); it is defined as
\begin{equation}
    f(z) = 1-\exp\left(-\exp\left(-z\right)\right).
\end{equation}
The variant called \gls[prereset]{cLogLogm} \cite{daSGomes2010} was also proposed:
\begin{equation}
    f(z) = 1-2\exp\left(-0.7\exp\left(-z\right)\right).
\end{equation}

\subsubsection{SechSig}
\label{sec:sechsig}
The \gls{SechSig} \cite{Kzkurt2023} is another \gls{AF} utilizing the \gls{logisticsigmoid} in its definition; it is defined as
\begin{equation}
    f(z) =\left(z+\sech\left(z\right)\right)\sigma\left(z\right).
\end{equation}
\Citeauthor{Kzkurt2023} also proposed a parametric version which we will call \gls[prereset]{pSechSig}:
\begin{equation}
    f(z) =\left(z+a\cdot\sech\left(z+a\right)\right)\sigma\left(z\right),
\end{equation}
where $a$ is a fixed parameter \cite{Kzkurt2023}.

\subsubsection{TanhSig}
\label{sec:tanhsig}
The \gls{TanhSig} \cite{Kzkurt2023} is an \gls{AF} similar to \gls{SechSig}; it is defined as
\begin{equation}
    f(z) =\left(z+\tanh\left(z\right)\right)\sigma\left(z\right).
\end{equation}
\Citeauthor{Kzkurt2023} also proposed a parametric version which we will call \gls[prereset]{pTanhSig}:
\begin{equation}
    f(z) =\left(z+a\cdot\tanh\left(z+a\right)\right)\sigma\left(z\right),
\end{equation}
where $a$ is a fixed parameter \cite{Kzkurt2023}.

\subsubsection{Multistate activation function (MSAF)}
\label{sec:msaf}
The \gls[prereset]{MSAF} is a \gls{logisticsigmoid} based \gls{AF} proposed in \cite{Cai2015}. The general \gls{MSAF} is defined as
\begin{equation}
    f(z) =a + \sum_{k=1}^N\frac{1}{1+\exp\left(-z+b_k\right)},
\end{equation}
where $a$ and $b_k$, $k=1,\ldots,N$ are fixed parameters; $a \in \mathbb{R}$, $N \in \mathbb{N}^+$, $b_k \in \mathbb{R}^+$, and $b_1 < b_2 < \ldots < b_N$ \cite{Cai2015}. If $a=0$, it is named as $N$-order\footnote{This does not exactly fit into the exemplar \gls{MSAF} of order two presented in \cite{Cai2015}; it is possible that authors intended another constraint $b_1=0$ for such case.} \gls{MSAF}.

There is also a special case called \gls[prereset]{SymMSAF} defined as
\begin{equation}
    f(z) =-1 + \frac{1}{1 + \exp\left(-z\right)} +\frac{1}{1+\exp\left(-z-a\right)},
\end{equation}
where $a$ is required to be significantly smaller than 0 \cite{Cai2015}

\subsubsection{Rootsig and others}
\label{sec:rootsig}
The \gls{rootsig} is one of the activations listed in \cite{Duch1999}. It is defined as
\begin{equation}
    f(z) = \frac{az}{1+\sqrt{1+a^2z^2}},
\end{equation}
where $a$ is a parameter \cite{Duch1999}. This function is called \gls{rootsig} in \cite{daSGomes2010} where the authors list a variant with $a=1$.

There are also several other unnamed sigmoids in \cite{Duch1999}:
\begin{equation}
    f(z) = z\frac{\mathrm{sgn}\left(z\right)z-a}{z^2-a^2},
\end{equation}
\begin{equation}
    f(z) = \frac{az}{1+\left|az\right|},
\end{equation}
and 
\begin{equation}
    f(z) = \frac{az}{\sqrt{1+a^2z^2}}.
\end{equation} 

\subsubsection{Sigmoid and tanh combinations}
\label{sec:sigandtanh}
\Citeauthor{Guevraa2021} proposed several activations mostly combining the \gls{logisticsigmoid}, \gls{tanh}, and linear function in \cite{Guevraa2021}. The general approach is
\begin{equation}
    f(z) = \begin{cases}
        g(z), \quad & z \geq 0,\\
        h(z), \quad & z < 0,\\
    \end{cases},
\end{equation}
where $g(z)$ and $h(z)$ are two different \glspl{AF} \cite{Guevraa2021}. The authors used the following pairs $\left\{g(z),h(z)\right\}$: $\left\{\sigma_2(z), \tanh(z)\right\}$, $\left\{\sigma_2(z), \tanh(z)\right\}$, $\left\{\sigma_2(z), 0\right\}$, $\left\{\tanh(z), 0\right\}$,  $\left\{\sigma_2(z), az\right\}$, and $\left\{\tanh(z), az\right\}$, where $a > 0$ is a fixed parameter and 
\begin{equation}
    \sigma_2(z) = \frac{2}{1+\exp\left(-z\right)} -1.
\end{equation}
\Citeauthor{Guevraa2021} also proposed an \gls{AF} we termed \glsxtrshort{SigLU} (see \cref{sec:siglu}) and nonadaptive variant of \glsxtrshort{PTELU}.

\subsection{Class of sigmoid-weighted linear units}
\label{sec:silu}
The \gls{SiLU} is the most common example of a larger class of sigmoidal units defined as
\begin{equation}
    f(z) = z\cdot s(z),
\end{equation}
where $s(z)$ is any \glslink{sigmoid}{sigmoidal} function; it becomes the \gls{SiLU} if the \gls{logisticsigmoid} function is used. The \gls{SiLU} is thus defined as
\begin{equation}
    f(z) = z\cdot\sigma(z),
\end{equation}
where $\sigma(z)$ is the \gls{logisticsigmoid} \cite{Elfwing2018}. The \gls{SiLU} has the output range of $(-0.5, \infty)$ \cite{Dubey2022} and was first used \cite{Elfwing2018} for reinforcement learning tasks such as SZ-Tetris and Tetris. The \gls{SiLU} was also found to work well for the CIFAR-10/100 \cite{Krizhevsky2009} and ImageNet \cite{Deng2009, Russakovsky2015} tasks in \cite{Ramachandran2017}. The adaptive variant of the \gls{SiLU} is called \gls{swish} (see \cref{sec:swish}) \cite{Ramachandran2017}.

For the purposes of this work, we also consider any squashing functions $s(z)$ and not necessarily only \glspl{sigmoid} --- for example, we classify \gls{rectifiedhyperbolicsecant} (see \cref{sec:resech}) as a member of this class. We also list functions that are closely based on the \gls{SiLU} and its variants.

A similar approach named \gls{WiG} was proposed in \cite{Tanaka2020}, where the \gls{AF} was used only for gating each of the raw inputs:
\begin{equation}
    f(\vec{x})_i = x_i\cdot \sigma(z) = x_i \cdot \sigma(\vec{w}_i^T\vec{x}+b_i),
\end{equation}
where $\vec{x}$ denotes the vector of raw inputs, $\vec{w}_i$ the weights of neuron $i$ and $b_i$ its bias \cite{Tanaka2020}

\subsubsection{Gaussian error linear unit (GELU)}
\label{sec:gelu}
\Gls{GELU} \cite{Hendrycks2016} is an \glsxtrlong{AF} based on the standard Gaussian cumulative distribution function, and it weights inputs by their value rather than gating them as \glspl{ReLU} do \cite{Hendrycks2016}. It is defined as
\begin{equation}
    f(z) = z\cdot\Phi\left(z\right) = z \cdot \frac{1}{2}\left(1+\mathrm{erf}\left(\frac{z}{\sqrt{2}}\right)\right),
\end{equation}
where $\Phi\left(z\right)$ is the standard Gaussian \gls{CDF} and $\mathrm{erf}\left(x\right)$ is the Gauss error function \cite{Hendrycks2016}. It is similar to the \gls{SiLU} but it uses $\Phi\left(z\right)$ instead of the $\sigma\left(z\right)$. However, due to the complicated formula, the \gls{GELU} can be approximated as
\begin{equation}
    f(z) = \frac{1}{2}z\left(1+\tanh\left(\sqrt{\frac{2}{\pi}}\left(z+0.044715z^3\right)\right)\right)
\end{equation}
or
\begin{equation}
    f(z) = z\cdot\sigma\left(1.702z\right),
\end{equation}
if the performance gains are worth the loss of exactness \cite{Hendrycks2016}. The function is similar to \gls{SiLU} (see \cref{sec:silu}), it only uses Gaussian \gls{CDF} $\Phi\left(z\right)$ instead of the logistic distribution \gls{CDF} $\sigma(z)$ \cite{Hendrycks2016}. \gls{GELU} was found to outperform many competitors (e.g., \gls{ReLU}, \gls{ELU}, \gls{SELU}, \gls{CELU}, sigmoid, \gls{tanh}) in \cite{Lee2023GELU}. \citeauthor{Hendrycks2016} also proposed to parameterize the \gls{GELU} by $\mu$ and $\sigma^2$ --- the parameters defining mean and variance of the Gaussian distribution whose \gls{CDF} is used in the \gls{GELU} \cite{Hendrycks2016}, however, only the standard Gaussian distribution was used in experiments in \cite{Hendrycks2016}. Replacing \glspl{ReLU} with GELUs led to better performance in \cite{Kang2022}. More details about \gls{GELU} are available in \cite{Lee2023GELU}.

\subsubsection{Symmetrical Gaussian error linear unit (SGELU)}
\label{sec:sgelu}
A symmetric variant of \gls{GELU} called \gls[prereset]{SGELU} was proposed in \cite{Yu2019Symmetrical}.
It is defined as
\begin{equation}
    f(z) = a\cdot z \cdot \mathrm{erf}\left(\frac{z}{\sqrt{2}}\right),
\end{equation}
where $a$ is a fixed hyperparameter \cite{Yu2019Symmetrical}. The symmetrical nature of the \gls{SGELU} also leads to more symmetrically distributed weights of the \glsxtrlong{NN} compared to \gls{SGELU} \cite{Yu2019Symmetrical}; it is believed that normal distribution of the weights can make the network more rational, accurate, and robust \cite{Yu2019Symmetrical}.

\subsubsection{Cauchy linear unit (CaLU)}
\label{sec:calu}
Another function related to the \gls{GELU} and \gls{SiLU} is the \gls[prereset]{CaLU} \cite{Wu2023TheAdaptive} which uses the \gls{CDF} of the standard Cauchy distribution instead of the Gaussian \gls{CDF} in \gls{GELU} and \gls{logisticsigmoid} in \gls{SiLU}. It is defined as
\begin{equation}
    f(z) = z\cdot\Phi_{\mathrm{Cauchy}}\left(z\right)=z\cdot\left(\frac{\tan^{-1}\left(z\right)}{\pi}+\frac{1}{2}\right),
\end{equation}
where $\Phi_{\mathrm{Cauchy}}\left(z\right)$ is the \gls{CDF} of the standard Cauchy distribution \cite{Wu2023TheAdaptive}.

\subsubsection{Laplace linear unit (LaLU)}
\label{sec:lalu}
Another function related to the \gls{GELU} and \gls{SiLU} is the \gls[prereset]{LaLU} \cite{Wu2023TheAdaptive} which uses the \gls{CDF} of the Laplace distribution; it is defined as
\begin{equation}
    f(z) = z\cdot\Phi_{\mathrm{Laplace}}\left(z\right)=z\cdot\begin{cases}
        1-\frac{1}{2}\exp\left(-z\right), \quad & z \geq 0,\\
        \frac{1}{2}\exp\left(z\right), \quad & z < 0,\\
    \end{cases},
\end{equation}
where $\Phi_{\mathrm{Laplace}}\left(z\right)$ is the \gls{CDF} of the Laplace distribution \cite{Wu2023TheAdaptive}.

\subsubsection{Collapsing linear unit (LaLU)}
\label{sec:colu}
The \gls[prereset]{CoLU} is an \gls{AF} similar to the \gls{SiLU} proposed in \cite{Vagerwal2021}. It is defined as
\begin{equation}
    f(z) = z \cdot \frac{1}{1-z\exp\left(-\left(z+\exp\left(z\right)\right)\right)}.
\end{equation}

\subsubsection{Triple-state swish}
\label{sec:tsswish}
The \gls{TS-swish}\footnote{\Citeauthor{Koak2021} called the function \gls{swish} but it is actually based on the \gls{SiLU}.} is a cascaded \gls{AF} similar to \gls{TS-sigmoid} (see \cref{sec:tssigmoid}) \cite{Koak2021}; it is defined as
\begin{equation}
f(z) = \frac{z}{1+\exp\left(-z\right)}\left(\frac{1}{1+\exp\left(-z\right)}+\frac{1}{1+\exp\left(-z+a\right)}+\frac{1}{1+\exp\left(-z+b\right)}\right),
\end{equation}
where $a$ and $b$ are fixed parameters \cite{Koak2021}.

\subsubsection{Generalized swish}
\label{sec:generalized_swish}
A \gls{SiLU} variant called \gls{generalized_swish}\footnote{Also based on the \gls{SiLU} instead of its adaptive variant \gls{swish}.} was proposed in \cite{Koak2021}. It is defined as
\begin{equation}
    f(z) = z\cdot\sigma\left(\exp(-z)\right).
\end{equation}

\subsubsection{Exponential swish}
\label{sec:exponential_swish}
Another \gls{SiLU} variant called \gls{exponential_swish}\footnote{Again, based on the \gls{SiLU} instead of its adaptive variant \gls{swish}.} was proposed in \cite{Koak2021}. It is defined as
\begin{equation}
    f(z) = \exp\left(-z\right)\sigma\left(z\right).
\end{equation}

\subsubsection{Derivative of sigmoid function}
\label{sec:dsigmoid}
The derivative of \gls{logisticsigmoid} was used as an \gls{AF} in \cite{Koak2021}. \Citeauthor{Koak2021} formulate the \gls{AF} using the following form
\begin{equation}
    f(z) = \exp(-z)\left(\sigma\left(z\right)\right)^2.
\end{equation}

\subsubsection{Gish}
\label{sec:gish}
\Gls{gish} is another \gls{SiLU} variant \cite{Kaytan2023}; the \gls{gish} is defined as
\begin{equation}
    f(z) = z\cdot\ln\left(2-\exp\left(-\exp\left(z\right)\right)\right).
\end{equation}
\Citeauthor{Kaytan2023} found that \gls{gish} outperformed \gls{logisticsigmoid}, \gls{softplus}, \gls{ReLU}, \gls{LReLU}, \gls{ELU}, \gls{swish}, \gls{mish}, \gls{logish}, and \gls{smish} on the MNIST \cite{LiDeng2012} and CIFAR-10 \cite{Krizhevsky2009} datasets \cite{Kaytan2023}.

\subsubsection{Logish}
\label{sec:logish}
\Gls{logish} is yet another \gls{SiLU} variant \cite{Zhu2021Logish}; it is defined as
\begin{equation}
    f(z) = z\cdot\ln\left(1+\sigma\left(z\right)\right).
\end{equation}

\subsubsection{LogLogish}
\label{sec:loglogish}
\Gls{LogLogish} is a \gls{SiLU} variant based on the \gls{LogLog} (see \cref{sec:loglog}) \cite{Wu2023TheAdaptive}; it is defined as
\begin{equation}
    f(z) = z\cdot\left(1-\exp\left(-\exp\left(z\right)\right)\right).
\end{equation}

\subsubsection{ExpExpish}
\label{sec:expexpish}
\Gls{ExpExpish} is a \gls{SiLU} variant \cite{Wu2023TheAdaptive}; it is defined as
\begin{equation}
    f(z) = z\cdot \exp\left(-\exp\left(-z\right)\right).
\end{equation}

\subsubsection{Self arctan}
\label{sec:selfarctan}
The \gls[prereset]{selfarctan} is an \gls{AF} proposed in \cite{TmerSivri2023} whose formula resembles the \gls{SiLU}. The \gls{selfarctan} is defined as
\begin{equation}
    f(z) = z\cdot\tan^{-1}\left(z\right),
\end{equation}
where $\tan^{-1}\left(z\right)$ is the arctangent function \cite{TmerSivri2023}.

\subsubsection{Parametric logish}
\label{sec:plogish}
\Citeauthor{Zhu2021Logish} also proposed a parametric variant of \gls{logish} --- we will call it \gls{pLogish} in this work. It is defined as
\begin{equation}
    f(z_i) = a_iz_i\cdot\ln\left(1+\sigma\left(b_iz_i\right)\right),
\end{equation}
where $a$ and $b$ are fixed parameters \cite{Zhu2021Logish}; \citeauthor{Zhu2021Logish} used $a=1$ and $b=10$ in \cite{Zhu2021Logish}.

\subsubsection{Phish}
\label{sec:phish}
\Gls{phish} is a \gls{SiLU} variant combining \gls{GELU} and \gls{tanh} \cite{Naveen2022}; it is defined as
\begin{equation}
    f(z) = z\cdot \tanh\left(\mathrm{GELU}\left(z\right)\right).
\end{equation}
The \gls{phish} was found to outperform \gls{GELU}, \gls{tanh}, \gls{logisticsigmoid}, and \gls{ReLU}; it performed similarly as the \gls{mish} and \gls{swish} in the experiments in \cite{Naveen2022}.

\subsubsection{Suish}
\label{sec:suish}
The \gls{suish} \cite{Xu2020Comparison} was proposed as an alternative to the \gls{swish} \gls{AF} in \cite{Su2017ABrief}. It is defined as
\begin{equation}
    f(z) = \max\left(z, z\cdot\exp\left(-\left|z\right|\right)\right).
\end{equation}

\subsubsection{Tangent-sigmoid ReLU (TSReLU)}
\label{sec:tsrelu}
The \gls[prereset]{TSReLU} \cite{Mercioni2020Improving} is an \gls{AF} very similar to \gls{phish}, \gls{mish}, and \gls{TanhExp} --- it just uses the \gls{logisticsigmoid} instead of the \gls{GELU} in \gls{phish}, \gls{softplus} in \gls{mish}, and the exponential in \gls{TanhExp}. It is defined as
\begin{equation}
    f(z) = z\cdot \tanh\left(\sigma\left(z\right)\right).
\end{equation}

\subsubsection{Tangent-bipolar-sigmoid ReLU (TBSReLU)}
\label{sec:tbsrelu}
The \gls[prereset]{TBSReLU} is a variant of \gls{TSReLU} proposed in \cite{Mercioni2020Improving}. It is defined as
\begin{equation}
    f(z) = z\cdot \tanh\left(\frac{1-\exp\left(-z\right)}{1+\exp\left(-z\right)}\right).
\end{equation}

\subsubsection{Log-sigmoid}
A logarithm of the \gls{logisticsigmoid} is sometimes used as an \glsxtrlong{AF} \cite{Wong2022}.
It is defined as
\begin{equation}
    f(z) = \ln\left(\sigma(z)\right) = \ln\left(\frac{1}{1+\exp(-z)}\right).
\end{equation}

\subsubsection{Derivative of sigmoid-weighted linear unit (dSiLU)}
\label{sec:dsilu}
The \gls[prereset]{dSiLU} can also be used as an \glsxtrlong{AF} resembling a sigmoid \cite{Elfwing2018}. It is defined as 
\begin{equation}
    f(z) = \sigma(z)\left(1+z\left(1-\sigma(z)\right)\right),
\end{equation}
where $\sigma(z)$ is the \gls{logisticsigmoid} \cite{Elfwing2018}. The \gls{dSiLU} has a maximum value of around 1.1, and the minimum is approximately -0.1 \cite{Elfwing2018}.

\subsubsection{Double sigmoid-weighted linear unit (DoubleSiLU)}
\label{sec:double_silu}
The \gls[prereset]{DoubleSiLU}\footnote{\Citeauthor{Verma2023} termed the unit as DSiLU but that would collide with the \gls{dSiLU} (see \cref{sec:dsilu}) proposed earlier by \citeauthor{Elfwing2018}.} is an \gls{AF} proposed in \cite{Verma2023}. It is defined as
\begin{equation}
    f(z) = z\cdot\frac{1}{1+\exp\left(-z\cdot\frac{1}{1+\exp\left(-z\right)}\right)},
\end{equation}
where $\sigma(z)$ is the \gls{logisticsigmoid} \cite{Verma2023}.

\subsubsection{Modified sigmoid-weighted linear unit (MSiLU)}
\label{sec:msilu}
A \gls[prereset]{MSiLU} is a variant of the \gls{SiLU} that has faster convergence than the \gls{SiLU} \cite{Hayou2018}. It is defined as
\begin{equation}
    f(z) = z\cdot\sigma(z) + \frac{\exp\left(-z^2-1\right)}{4},
\end{equation}
where $\sigma(z)$ is the \gls{logisticsigmoid} \cite{Hayou2018}.

\subsubsection{Hyperbolic tangent sigmoid-weighted linear unit (TSiLU)}
\label{sec:tsilu}
Another \gls{SiLU} variant is the \gls[prereset]{TSiLU} \cite{Verma2023}, which combines the \gls{tanh} and \gls{SiLU}. It is defined\footnote{The formula in \cite{Verma2023} was wrong as it evaluated to $\frac{2x}{0}$, we present the formula we think authors intented.} as
\begin{equation}
 f(z) = \frac{\exp\left(\frac{z}{1+\exp\left(-z\right)}\right)-\exp\left(-\frac{z}{1+\exp\left(-z\right)}\right)}{\exp\left(\frac{z}{1+\exp\left(-z\right)}\right)+\exp\left(\frac{z}{1+\exp\left(-z\right)}\right)}.
\end{equation}

\subsubsection{Arctan sigmoid-weighted linear unit (ASiLU)}
\label{sec:atsilu}
\Gls{ATSiLU} is yet another \gls{SiLU} variant proposed in \cite{Verma2023}; it is defined as 
\begin{equation}
 f(z) = \tan^{-1}\left(z\cdot\frac{1}{1+\exp\left(-z\right)}\right).
\end{equation}

\subsubsection{SwAT}
\label{sec:swat}
\Citeauthor{Verma2023} proposed an \gls{AF} named \gls{SwAT} combining the \gls{SiLU} and arctan in\cite{Verma2023}. This function is defined as 
\begin{equation}
 f(z) = z\cdot\frac{1}{1+\exp\left(-\tan^{-1}|left(z)\right)}.
\end{equation}

\subsubsection{Rectified hyperbolic secant}
\label{sec:resech}
A \gls{rectifiedhyperbolicsecant} \glsxtrlong{AF} was proposed in \cite{SamatinNjikam2016}. This function is totally differentiable, symmetric about the origin, and is approaching zero for inputs going to positive or negative infinity:
\begin{equation}
    f(z) = z\cdot\mathrm{sech}(z),
\end{equation}
where $\mathrm{sech}(z)$ is the hyperbolic secant function \cite{SamatinNjikam2016}.

\subsubsection{Linearly scaled hyperbolic tangent (LiSHT)}
\label{sec:lisht}
A \gls{LiSHT} \glsxtrlong{AF} was proposed in \cite{Roy2023} to address the problem of vanishing gradients and the non-utilization of large negative input values. The \gls{LiSHT}  function is defined as
\begin{equation}
    f(z) = z \cdot \tanh(z).
\end{equation}
The output range of \gls{LiSHT}  function is $[0, \infty]$ \cite{Dubey2022}.The output of \gls{LiSHT}  is close to the \gls{ReLU} (see \cref{sec:relu}) and \gls{swish} for large positive values \cite{Roy2023}; however, unlike the aforementioned \glspl{AF}, the output is symmetric, and, therefore, it behaves identically for large negative values. While the \gls{LiSHT}  is symmetric, the fact that its output is unbounded and non-negative could be considered a disadvantage \cite{Dubey2022}.  The effectiveness of the \gls{LiSHT}  \glsxtrlong{AF} was tested on several different architectures ranging from multilayer perceptron (MLP) and residual \glsxtrlongpl{NN} to \glsxtrshort{LSTM}-based networks and on various tasks --- the Iris dataset, the MNIST \cite{LiDeng2012}, CIFAR-10 and CIFAR-100 \cite{Krizhevsky2009} and the \textit{sentiment140} dataset from Twitter \cite{Go2009, Sahni2017} for sentiment analysis \cite{Roy2023}.

A parametric version of \gls{LiSHT} named \gls{softmodulust} (see \cref{sec:softmodulust}) was proposed in \cite{VallsPrez2023}.

\subsubsection{Mish}
\label{sec:mish}
A popular activation function \gls{mish}\cite{Misra2020} is a combination of the \gls{tanh} and \gls{softplus} \glsxtrlong{AF}; the function resembles \gls{swish} activation (see \cref{sec:swish}). It is defined as
\begin{equation}
    f(z) = z\cdot\tanh\left(\mathrm{softplus}(z)\right) = z\cdot\tanh\left(\ln\left(1+\exp\left(z\right)\right)\right).
\end{equation}
\Gls{mish} was found to outperform \gls{swish}; it performed similarly to $f(z) = z\cdot\ln\left(1+\tanh\left(\exp\left(z\right)\right)\right)$ but this \glsxtrlong{AF} was found to often lead to unstable training \cite{Misra2020}. The \gls{mish} was found to outperform \gls{swish} and \gls{ReLU} for many architectures such as various ResNet architectures \cite{He2016}, Inception v3 \cite{Szegedy2015}, DenseNet-121 \cite{Huang2017}, and others \cite{Misra2020}.  Detailed comparison with other \glsxtrlongpl{AF} was run using the Squeeze Net \cite{Hu2018SqueezeAndExcitation} where it outperformed \gls{swish}, \gls{GELU}, \gls{ReLU}, \gls{ELU}, \gls{LReLU}, \gls{SELU}, \gls{softplus}, \gls{SReLU} , \gls{ISRU}, \gls{tanh}, and \gls{RReLU} \cite{Misra2020}. The \gls{mish} \glsxtrlong{AF} was, for example, used in the YOLOv4 \cite{Bochkovskiy2020} and its variant Scaled-YOLOv4 \cite{Wang2021ScaledYOLOv4}.

\subsubsection{Smish}
\label{sec:smish}
The \gls{smish} \cite{Wang2022Smish} is a variant of the \gls{mish} where the exponential function is replaced by the \gls{logisticsigmoid}. It is, therefore, defined as
\begin{equation}
    f(z) = az\cdot\tanh\left(\ln\left(1+\sigma\left(b z\right)\right)\right),
\end{equation}
where $a$ and $b$ are parameters \cite{Wang2022Smish}; however, \citeauthor{Wang2022Smish} recommend $a=1$ and $b=1$ based on a small parameter search in \cite{Wang2022Smish}.

\subsubsection{TanhExp}
\label{sec:tanhexp}
Similarly as the \gls{mish} is the combination of \gls{tanh} and \gls{softplus}, the \gls{TanhExp} \cite{Liu2021TanhExp} is a combination of \gls{tanh} and the exponential function \cite{AdrianaMercioni2020,Liu2021TanhExp}. It is defined as
\begin{equation}
    f(z) = z\cdot\tanh\left(\exp(z)\right).
\end{equation}

\subsubsection{Serf}
\label{sec:serf}
The \gls{serf} is an \gls{AF} similar to the \gls{mish}; however, it uses the error function instead of the \gls{tanh} \cite{Nag2023}. It is defined as
\begin{equation}
    f(z) = z\operatorname{erf} \left( {\ln \left( {1 + \exp(z)} \right)} \right),
\end{equation}
where $\operatorname{erf}$ is the Gauss error function \cite{Nag2023}. It was found to outperform \gls{mish}, \gls{GELU}, and \gls{ReLU} for various architectures on Multi30K \cite{Elliott2016}, ImageNet \cite{Deng2009, Russakovsky2015}, the CIFAR-10, and CIFAR-100 \cite{Krizhevsky2009} datasets; see \cite{Nag2023} for details.

\subsubsection{Efficient asymmetric nonlinear activation function (EANAF)}
\label{sec:eanaf}
An \glsxtrlong{AF} combining \gls{tanh} and \gls{softplus} called \gls{EANAF} was proposed in \cite{Chai2022}. The function is defined as
\begin{equation}
    f(z) = z\cdot \mathrm{g}\left(\mathrm{h}\left(z\right)\right),
\end{equation}
where $\mathrm{h}(z)$ is the \gls{softplus} function and $\mathrm{g}(z) = \tanh\left(\frac{z}{2}\right)$, which can be simplified to
\begin{equation}
    f(z) = \frac{z\cdot\exp\left(z\right)}{\exp\left(z\right)+2}.
\end{equation}
The \gls{EANAF} is continuously differentiable. The \gls{EANAF} is very similar to \gls{swish} with similar amount of computation but \citeauthor{Chai2022} found that it performs better than \gls{swish} and several other \glsxtrlongpl{AF} in RetinaNet \cite{Lin2020} and YOLOv4 \cite{Wang2021ScaledYOLOv4} architectures on object detection tasks \cite{Chai2022}.

\subsubsection{SinSig}
\label{sec:sinsig}
\Gls{sinsig} \cite{Douge2021} is a self-gated non-monotonic \glsxtrlong{AF} defined as
\begin{equation}
    f(z) = z \cdot \sin\left(\frac{\pi}{2}\sigma\left(z\right)\right),
\end{equation}
where $\sigma(z)$ is the \gls{logisticsigmoid} function \cite{Douge2021}. While \gls{sinsig} is similar to \gls{swish} and \gls{mish}, it outperformed them in experiments in \cite{Douge2021} as the number of layers in a \glsxtrlong{NN} increased. It was also shown that the \gls{sinsig} converges faster. The \gls{sinsig} outperformed \gls{ReLU} and \gls{mish} on several deep architectures including ResNet 20 v2 \cite{He2016Identity}, ResNet 110 v2 \cite{He2016Identity}, SqueezeNet \cite{Iandola2016}, and ShuffleNet \cite{Zhang2018ShuffleNet} among others on the CIFAR-100 task \cite{Krizhevsky2009} in experiments in \cite{Douge2021}.

\subsubsection{Gaussian error linear unit with sigmoid activation function (SiELU)}
\label{sec:sielu}
The \gls[prereset]{SiELU} was proposed in \cite{Ahmed2023}; it is defined as
\begin{equation}
    f(z) = z \sigma\left(2\sqrt{\frac{2}{\pi}}\left(z + 0.044715z^3\right)\right).
\end{equation}

\subsection{Gated linear unit (GLU)}
\label{sec:glu}
A gated activation called \gls{GLU} similar to \glsxtrshort{SiLU} (see \cref{sec:silu}) for use in \glspl{RNN} was proposed in \cite{Dauphin2017}. The \gls{GLU} is defined as 
\begin{equation}
    f(z, z') = z \otimes \sigma(z'),
\end{equation}
where $\otimes$ is the element-wise product and $z$ and $z'$ are two learned linear transformations of input vector $\vec{x}$ \cite{Shazeer2020,Gehring2017}.

\subsubsection{Gated tanh unit (GTU)}
\label{sec:gtu}
A gated activation called \gls{GTU} similar to \glsxtrshort{GLU} (see \cref{sec:glu}) for use in \glspl{RNN} was proposed in \cite{Oord2016Conditional}. The \gls{GTU} is defined as 
\begin{equation}
    f(z, z') = \tanh(z) \otimes \sigma(z'),
\end{equation}
where $\otimes$ is the element-wise product and $z$ and $z'$ are two learned linear transformations of input vector $\vec{x}$ \cite{Shazeer2020}.

\subsubsection{Gated ReLU (ReGLU)}
\label{sec:reglu}
Another \gls{GLU} extension is the \gls[prereset]{ReGLU} \cite{Dauphin2017,Shazeer2020}.  The \gls{ReGLU} is defined as 
\begin{equation}
    f(z, z') = z \otimes \mathrm{ReLU}(z'),
\end{equation}
where $\otimes$ is the element-wise product and $z$ and $z'$ are two learned linear transformations of input vector $\vec{x}$ \cite{Shazeer2020}.

\subsubsection{Gated GELU (GEGLU)}
\label{sec:geglu}
A \gls{GELU}-based \gls{GLU} extension is the \gls[prereset]{GEGLU} \cite{Shazeer2020}; it is defined as 
\begin{equation}
    f(z, z') = z \otimes \mathrm{GELU}(z'),
\end{equation}
where $\otimes$ is the element-wise product and $z$ and $z'$ are two learned linear transformations of input vector $\vec{x}$ \cite{Shazeer2020}.

\subsubsection{Swish GELU (SwiGLU)}
\label{sec:swiglu}
A \gls{swish}-based \gls{GLU} extension is the \gls[prereset]{SwiGLU} \cite{Shazeer2020}; it is defined as 
\begin{equation}
    f(z, z') = z \otimes \mathrm{swish}(z'),
\end{equation}
where $\otimes$ is the element-wise product, $z$ and $z'$ are two learned linear transformations of input vector $\vec{x}$, and $\mathrm{swish}$ is the \gls{swish} with its own trainable parameter \cite{Shazeer2020}.

\subsection{Softmax}
\label{sec:softmax}
The \gls{softmax} is not a usual type of \gls{AF} taking in a single value, but it takes all the output value of the unit $i$ and, also, the output values of other units in order to compute a soft argmax of the values. It is defined as

\begin{equation}
    f(z_j) = \frac{\exp\left(z_j\right)}{\sum_{k=1}^N\exp\left(z_k\right)},
\end{equation}
where $f(z_j)$ is the output of a neuron $j$ in a \gls{softmax} layer consisting of $N$ neurons \cite{Bridle1989,Pearce2021}.

\subsubsection{\texorpdfstring{$\beta$}{Beta}-softmax}
\label{sec:betasoftmax}
The  \gls{betasoftmax} is a \gls{softmax} extension proposed in \cite{Bhuvaneshwari2021}; it is defined as
\begin{equation}
    f(z_j) = \frac{\int\exp\left(bz_j\right)}{\sum_{k=1}^N\int\exp\left(bz_k\right)},
\end{equation}
where $f(z_j)$ is the output of a neuron $j$ in a \gls{softmax} layer consisting of $N$ neurons  and $b$ takes random value from $\mathbb{N}^+$\footnote{No further specification was provided in \cite{Bhuvaneshwari2021}.}\cite{Bhuvaneshwari2021}.

\subsection{Rectified linear function (ReLU)}
\label{sec:relu}
The \glsxtrfull{ReLU} \cite{Nair2010} is widely regarded as the most popular \glsxtrlong{AF} in modern feedforward networks  \cite{Glorot2011, Goodfellow2016, CS231n} due to its simplicity and improved performance \cite{Dubey2022}. It has been observed that \glspl{ReLU}  can significantly expedite the convergence of stochastic gradient descent \cite{Krizhevsky2012}.  Additionally, traditional \glspl{ReLU} are computationally less expensive compared to \glsxtrlongpl{AF} like the logistic or \gls{tanh} functions \cite{CS231n}. \glspl{ReLU} often outperform sigmoidal \glsxtrlongpl{AF} \cite{Glorot2011}. However, a drawback of \glspl{ReLU} is the potential for neurons to become "dead" or "disabled" during training. This means that they may never activate again for any input, resulting in a permanently zero output gradient \cite{CS231n}. This issue can occur after a weight update when a large gradient flows through the unit \cite{CS231n}.  This might happen after a weight update after a large gradient flows through
the unit \cite{CS231n}. However, \glspl{ReLU} often lead to faster convergence than for sigmoid activation, as shown in \cite{Hara2015}. It can also be shown that \glspl{ReLU} and rational function efficiently approximate each other \cite{Telgarsky2017}. The \gls{ReLU} was used as an example of the more general class of piecewise affine \glspl{AF} for \glsxtrlong{NN} verification\footnote{More details are out of the scope of this work, see \cite{Aleksandrov2023} for more details.} using theorem provers in \cite{Aleksandrov2023}.

A \gls{ReLU} is mathematically defined as the maximum of zero and the input value \cite{Glorot2011, Mishkin2017}:
\begin{equation}
    f(z) = \max \left(0, z\right).
\end{equation}

\gls{ReLU} is commonly recommended as the default choice for feedforward networks due to its usually superior performance compared to sigmoidal functions and its computational efficiency \cite{CS231n}; furthermore, it works
comparably to its modifications \cite{Mishkin2017}. Many popular NN models utilize \gls{ReLU} as the \glsxtrlong{AF} of choice, e.g., \cite{Wang2015Effective, Krizhevsky2012}.

Many \gls{ReLU} modification and derivations were proposed \cite{Mishkin2015, Mishkin2017} --- e.g.
\glsxtrfull{LReLU} \cite{Maas2013}, \glsxtrfull{VLReLU} \cite{Graham2014}, parametric ReLU \cite{He2015}, \glsxtrfull{RReLU} \cite{Xu2015} or
S-shaped ReLU \cite{Jin2016}. Smoothed modifications are, for example, exponential linear unit \cite{Clevert2015}
and \gls{softplus} \cite{Glorot2011}. Most of the modifications solve the problem of dying out neurons as
they allow for gradient flows for any input.

\subsubsection{Shifted ReLU}
\label{sec:shifted_relu}
A \gls{shifted_relu} \cite{Clevert2015} is a simple translation of a \gls{ReLU} and is defined as
\begin{equation}
    f(z) = \max \left(-1, z\right).
\end{equation}

\subsubsection{Leaky ReLU (LReLU)}
\label{sec:lrelu}
\Gls[prereset]{LReLU} \cite{Maas2013} is defined as 
\begin{equation}
f(z)=\begin{cases}
            z, \quad & z \geq 0, \\
            \frac{z}{a},  \quad & z < 0, \\
        \end{cases}
\end{equation}
where $a \in (1, \infty)$ is set to large number;\footnote{Depending on the source, researchers use either this form $\frac{z}{a}$ or the inverted form $az$ for the negative inputs.} the recommended setting from \cite{Maas2013} is $a = 100$.

\Gls{LReLU} solves the problem of dying neurons when neurons have permanently zero output gradient in classical \gls{ReLU} by "leaking" the information for $z < 0$ instead of outputting exact zero. Both \gls{ReLU} and \gls{LReLU} can be considered to be a special case of the \gls{maxout} (see \cref{sec:maxout}) \cite{Dubey2022}. A theoretical analysis of the \gls{ReLU} and \gls{LReLU} is available in \cite{Parhi2020}.

\Gls{VLReLU} \cite{Graham2014} is almost identical to the \gls{LReLU} but has much higher slope when the $z$ is negative for faster training \cite{Graham2014} by setting $a_i = 3$. While it can be considered as a special case of \gls{LReLU}, some researchers consider it as a separate case, e.g., \cite{Mishkin2015}.

The so-called \gls{OLReLU} \cite{Nayef2021} propose another reformulation of \gls{LReLU} and calculation of the slope parameter $a$ that is inspired by the \glsxtrshort{RReLU} (see \cref{sec:rrelu}):
\begin{equation}
    f(z)=\begin{cases}
                z, \quad & z \geq 0, \\
                z \cdot \exp(-a),  \quad & z < 0, \\
            \end{cases}
\end{equation}
where 
\begin{equation}
a = \frac{u+l}{u-l},
\end{equation}
where $u$ and $l$ are hyperparameters of the bounds of the \gls{RReLU} \cite{Nayef2021}.

\subsubsection{Randomized leaky ReLU (RReLU)}
\label{sec:rrelu}
\Gls{RReLU} is a \glsxtrlong{LReLU} where the leakiness is stochastic during the training \cite{Xu2015}, i.e.:
\begin{equation}
    f(z_i)=\begin{cases}
                z_i, \quad & z_i \geq 0, \\
                \frac{z}{a_i},  \quad & z_i < 0, \\
            \end{cases}
    \end{equation}
where $a_i$ is a sampled for each epoch and neuron $i$  from the uniform distribution: $a_i \sim U(l, u)$ where $l < u $ and $l, u \in (0, \infty)$ 
\cite{Xu2015}. Similarly as in the dropout approach \cite{Srivastava2014}, an average over all $a_i$ over is taken during inference phase --- the $a_i$ is set to $\frac{l+u}{2}$:
\begin{equation}
    f(z_i)=\begin{cases}
                z_i, \quad & z_i \geq 0, \\
                \frac{z}{\frac{l+u}{2}},  \quad & z_i < 0. \\
            \end{cases}
    \end{equation}
Recommended distribution is $U(3,8)$ for sampling the $a_i$ \cite{Xu2015}.

\subsubsection{Softsign randomized leaky ReLU (S-RReLU)}
\label{sec:srrelu}
The \gls[prereset]{S-RReLU}\footnote{\Citeauthor{Elakkiya2024} used \glsxtrshort{S-RReLU} as a name and not an abbreviation; however, since \glsxtrshort{S-RReLU} is a combination of the \gls{softsign} and \gls{RReLU}, we feel that using it as an abbreviation is appropriate.} is a \gls{RReLU} combined with the \gls{softsign} proposed in \cite{Elakkiya2023, Elakkiya2024}. It is defined as

\begin{equation}
    f(z_i)=\begin{cases}
                \frac{1}{\left(1+z_i\right)^2} + z_i, \quad & z_i \geq 0, \\
                \frac{1}{\left(1+z_i\right)^2} + a_iz_i,  \quad & z_i < 0, \\
            \end{cases}
\end{equation}
where $a_i$ is a sampled for each epoch and neuron $i$  from the uniform distribution: $a_i \sim U(l, u)$ where $l < u $ and $l, u \in (0, \infty)$ \cite{Elakkiya2024}. \Citeauthor{Elakkiya2024} used $l = \frac{1}{8}$ and $u=\frac{1}{3}$ \cite{Elakkiya2024}.

\subsubsection{Sloped ReLU (SlReLU)}
\label{sec:slrelu}
A \gls{SlReLU} \cite{Seo2017} is similar to the \gls{LReLU} --- whereas the \gls{LReLU} parameterizes the slope for negative inputs, the \gls{SlReLU} parameterizes the slope of \gls{ReLU} for positive inputs. It is, therefore, defined as
\begin{equation}
    f(z)=\begin{cases}
                az, \quad & z \geq 0, \\
                0,  \quad & z < 0, \\
            \end{cases}
    \end{equation}
where $a$ is a fixed, predetermined parameter \cite{Seo2017}. \citeauthor{Seo2017} recommended $a \in [1,10]$ based on their experiments in \cite{Seo2017}.

\subsubsection{Noisy ReLU (NReLU)}
\label{sec:nrelu}
A stochastic variant of the \gls{ReLU} called \gls{NReLU} was proposed in \cite{Nair2010}:
\begin{equation}
    f(z) = \max \left(0, z+a\right),
\end{equation}
where $a$ is a stochastic parameter $a \sim \mathrm{N}\left(0, \sigma(z)\right)$, $N\left(0, \sigma^2\right)$ is the Gaussian distribution with zero mean and variance $\sigma^2$ and $\sigma(z)$ is the standard deviation of the inputs $z$. The \gls{NReLU} was designed for use with Restricted Boltzmann machines \cite{Nair2010}. More details about the \gls{NReLU} is available in \cite{Nair2010}.

\subsubsection{SineReLU}
\label{sec:sinerelu}
The \gls{SineReLU} \cite{Gupta2021Parametric, Rodrigues2018} is a \gls{ReLU} based activation that uses trigonometric functions for negative inputs. It is defined as
\begin{equation}
    f(z)=\begin{cases}
                z, \quad & z \geq 0, \\
                a\left(\sin\left(z\right)-\cos\left(z\right)\right),  \quad & z < 0, \\
            \end{cases}
    \end{equation}
where $a$ is a fixed parameter \cite{Gupta2021Parametric,Rodrigues2018}.

\subsubsection{Minsin}
\label{sec:minsin}
The \gls{minsin} is a \gls{ReLU}-based \gls{AF} used in \cite{Xu2020Comparison}. It is defined as
\begin{equation}
    f(z) = \min \left(z, \sin\left(z\right)\right) = \begin{cases}
        \sin(z), \quad & z \geq 0, \\
        z,  \quad & z < 0. \\
    \end{cases}
\end{equation}

\subsubsection{Variational linear unit (VLU)}
\label{sec:vlu}
The \gls[prereset]{VLU} is an \gls{AF} combining the \gls{ReLU} and sine functions proposed in \cite{Gupta2021Parametric}. It is defined as
\begin{equation}
    f(z) = \mathrm{ReLU}\left(z\right) + a \sin\left(b z\right) = \max \left(0, z\right) + a \sin\left(b z\right),
\end{equation}
where $a$ and $b$ are fixed parameters \cite{Gupta2021Parametric}.

\subsubsection{Spatial context-aware activation (SCAA)}
\label{sec:scaa}
The \gls[prereset]{SCAA} is a \gls{ReLU} extension proposed in \cite{Yamamichi2021}. The \gls{ReLU} performs an element-wise max operation on the feature map $\matr{X}$:
\begin{equation}
    \mathbf{\mathrm{ReLU}}\left(\matr{X}\right) = \max\left(\matr{X},\matr{0}\right),
\end{equation}
where $\mathbf{\mathrm{ReLU}}\left(\matr{X}\right)$ is the \gls{ReLU} in the matrix notation and $\matr{0}$ is a matrix of zeroes with the same shape as $\matr{X}$ \cite{Yamamichi2021}. The \gls{SCAA} first applies a depth-wise convolution on $\matr{X}$ to produce spatial context aggregated feature map denoted $\mathbf{f}_{\mathrm{DW}}\left(\matr{X}\right)$ and then proceeds with the elementwise max operation \cite{Yamamichi2021}; the \gls{SCAA} is, therefore, defined as
\begin{equation}
    \mathbf{f}\left(\matr{X}\right) = \max\left(\matr{X},\mathbf{f}_{\mathrm{DW}}\left(\matr{X}\right)\right).
\end{equation}

\subsubsection{Randomly translational ReLU (RT-ReLU)}
\label{sec:rtrelu}
A \gls{RT-ReLU} is a \gls{ReLU} with a randomly added jitter during each iteration of the training process \cite{Cao2018Randomly}. It is defined as
\begin{equation}
    f(z_i)=\begin{cases}
                z_i + a_i, \quad & z_i + a_i \geq 0, \\
                0,  \quad & z_i + a_i < 0, \\
            \end{cases}
\end{equation}
where $a_i$ is stochastic parameter for each neuron $i$ randomly sampled from the Gaussian distribution at each iteration, $a_i \sim \mathrm{N}\left(0, \sigma^2\right)$, where $\sigma^2$ is the variance of the Gaussian distribution. The authors \citeauthor{Cao2018Randomly} set the $\sigma^2 = 0.75^2$ for their experiments \cite{Cao2018Randomly}. The $a_i$ is set to 0 during the test phase \cite{Dubey2022}.

\subsubsection{Natural-Logarithm-ReLU (NLReLU)}
\label{sec:nlrelu}
The \gls{NLReLU} introduces non-linearity to \gls{ReLU} similarly as \gls{ReLTanh} (see \cref{sec:reltanh}) but only for positive part of the \glsxtrlong{AF} \cite{Dubey2022}:
\begin{equation}
    f(z) = \ln\left(a\cdot\max\left(0,z\right) + 1\right),
\end{equation}
where $a$ is a predefined constant \cite{Liu2019NaturalLogarithmRectified}.

\subsubsection{Softplus linear unit (SLU)}
\label{sec:slu}
An \glsxtrlong{AF} \gls{SLU} combining the \gls{ReLU} with the \gls{softplus} \glsxtrlong{AF} was proposed in \cite{Zhao2017}; the function is based around the assumption that zero mean activations improve learning performance \cite{Zhao2017}. The \gls{SLU} is defined as 
\begin{equation}
    f(z)=\begin{cases}
            az, \quad & z \geq 0, \\
            b \ln\left(\exp\left(z\right)+1\right)-c,  \quad & z < 0, \\
        \end{cases}
\end{equation}
where $a_i$, $b_i$, and $c_i$ are predefined parameters; however, to ensure that the function is continuous, differentiable at zero and to avoid vanishing or exploding gradients, its parameters are set to $a=1$, $b=2$, and $c=2\ln\left(2\right)$ \cite{Zhao2017}. The \gls{SLU} is therefore equal to
\begin{equation}
    f(z)=\begin{cases}
            z, \quad & z \geq 0, \\
            2 \ln\frac{\exp\left(z\right)+1}{2},  \quad & z < 0. \\
        \end{cases}
\end{equation}

\subsubsection{Rectified softplus (ReSP)}
\label{sec:resp}
Another \glsxtrlong{AF} combining \gls{ReLU} and \gls{softplus} called \gls{ReSP} \cite{Dubey2022}  was proposed in \cite{Xu2018Novel}. The function is defined as
\begin{equation}
    f(z)=\begin{cases}
            az + \ln(2), \quad & z \geq 0, \\
            \ln\left(1+\exp\left(z\right)\right),  \quad & z < 0, \\
        \end{cases}
\end{equation}
where $a$ is a fixed hyperparameter controlling the slope \cite{Xu2018Novel}. Larger values of $a$ between 1.4 and 2.0 were found to work well \cite{Xu2018Novel}.

\subsubsection{Parametric rectified non-linear unit (PReNU)}
\label{sec:prenu}
A \gls{ReLU} variant called \gls{PReNU} \cite{ElJaafari2020} replaces the linear part of the \gls{ReLU} for positive inputs by a non-linear function similarly to \gls{RePU} (see \cref{sec:repu}). It is defined as
\begin{equation}
    f(z)=\begin{cases}
            z-a\cdot\ln\left(z+1\right), \quad & z \geq 0, \\
            0,  \quad & z < 0, \\
        \end{cases}
\end{equation}
where $a$ is a fixed hyperparameter \cite{ElJaafari2020} --- however, this parameter could be adaptive similarly as in \gls{PReLU} (see \cref{sec:prelu}) that \gls{PReNU} extends since \citeauthor{ElJaafari2020} thought of the \gls{PReLU} as non-adaptive function for some reason \cite{ElJaafari2020}.

\subsubsection{Bounded ReLU (BReLU)}
\label{sec:brelu}
A \gls{BReLU} \cite{Liew2016} is a variant of \gls{ReLU} that limits the output as the unlimited output of the original \gls{ReLU} might lead to an instability \cite{Dubey2022}.  It is defined as
\begin{equation}
    f(z)= \min\left(\max\left(0, z\right), a\right) = \begin{cases}
                0, \quad & z \leq 0, \\
                z, \quad & 0 < z < a, \\
                a, \quad & z > a, \\
            \end{cases}
\end{equation}
where $a$ is a predefined parameter \cite{Liew2016}. The \gls{BReLU} appeared later in the literature under the name \textit{ReLUN} in \cite{Pishchik2023}, where it seems that it was independently proposed.

\subsubsection{Hard sigmoid}
\label{sec:hard_sigmoid}
A \gls{hard_sigmoid} is very similar to \gls{BReLU}; it is a very crude approximation of the \gls{logisticsigmoid} and is commonly defined \cite{Courbariaux2016, Basirat2018} as
\begin{equation}
    f(z)=\max\left(0, \min\left(\frac{z+1}{2}, 1\right)\right).
\end{equation}
Other definitions are sometimes used; e.g. variant from \cite{Avenash2019} is defined as
\begin{equation}
    f(z)=\max\left(0, \min\left(0.2z + 0.5, 1\right)\right).
\end{equation}

While the \gls{hard_sigmoid} is not as commonly used as the \gls{logisticsigmoid}, it can be used, for example, in binarized \glsxtrlong{NN} with stochastic \glsxtrlongpl{AF} \cite{Courbariaux2016} --- the binaryized \glsxtrlongpl{NN} can lead to much faster inference than regular \glsxtrlongpl{NN}, e.g., \citeauthor{Courbariaux2016} reached up to $7\times$ speed up without any loss in classification accuracy \cite{Courbariaux2016} (however, even better speed-ups can be obtained using, for example, \gls{FPGA} implementations as in \cite{Waseem2021}).

\subsubsection{HardTanh}
\label{sec:hard_tanh}
The \gls{hard_tanh} is another piecewise linear function; it is very similar to \gls{hard_sigmoid}, but it approximates the \gls{tanh} instead of the \gls{logisticsigmoid}. It is defined as
\begin{equation}
    \label{eq:hard_tanh}
    f(z)= \begin{cases}
                a, \quad & z < a, \\
                z, \quad & a \leq z \leq b, \\
                b, \quad & z > b, \\
            \end{cases}
\end{equation}
where $a$ and $b$ are fixed parameters \cite{Lupu2023}; \citeauthor{Liu2021Adaptive} used $a=-1$ and $b=11$ in \cite{Liu2021Adaptive}. \Glspl{NN} with \glspl{hard_tanh} are more suitable for linear predictive control than \glspl{NN} with \glspl{ReLU} as they usually require less hidden layers and neurons for representing identical min-max maps \cite{Lupu2023}.

\subsubsection{Shifted HardTanh}
\label{sec:shifted_hardtanh}
\Citeauthor{Kim2021} proposed \gls{hard_tanh} variants with vertical and horizontal shifts in \cite{Kim2021}. The \gls[prereset]{SvHardTanh}\footnote{Both \glsxtrshort{SvHardTanh} and \glsxtrshort{ShHardTanh} are named using the same convention as \glsxtrshortpl{shifted_elu} (see \cref{sec:shifted_elus}) for the purposes of this work.} is defined as
\begin{equation}
    f(z)= \begin{cases}
                -1+a, \quad & z < -1, \\
                z+a, \quad & -1 \leq z \leq 1, \\
                1+a, \quad & z > 1, \\
            \end{cases}
\end{equation}
where $a$ is a fixed parameter \cite{Kim2021}. \Citeauthor{Kim2021} used \gls{hard_tanh} variant with thresholds $-1$ and $1$; a more general variant with parametric thresholds from \cref{eq:hard_tanh} could be defined similarly.

The \gls[prereset]{SvHardTanh} is defined as
\begin{equation}
    f(z)= \begin{cases}
                -1+a, \quad & z < -1, \\
                z+a, \quad & -1 \leq z \leq 1, \\
                1+a, \quad & z > 1, \\

            \end{cases}
\end{equation}
where $a$ is a fixed parameter \cite{Kim2021}. 

The \gls[prereset]{ShHardTanh} is defined as
\begin{equation}
    f(z)= \begin{cases}
                -1, \quad & z < -1-a, \\
                z, \quad & -1-a \leq z \leq 1-a, \\
                1, \quad & z > 1-a, \\
            \end{cases}
\end{equation}
where $a$ is a fixed parameter \cite{Kim2021}.

\Citeauthor{Kim2021} used \gls{hard_tanh} variant with thresholds $-1$ and $1$; more general variants of \glsxtrshort{SvHardTanh} and \glsxtrshort{ShHardTanh} with parametric thresholds from \cref{eq:hard_tanh} could be defined similarly.

\subsubsection{Hard swish}
A linearized variant of the \gls{swish} \gls{AF} (see \cref{sec:swish}) was proposed in \cite{Howard2019}. It is defined as
\begin{equation}
    f(z)= z \cdot \begin{cases}
                0, \quad & z \leq -3, \\
                1, \quad & z \geq 3, \\
                \frac{z}{6}+\frac{1}{2}, \quad & -3 < z < 3. \\
            \end{cases}
\end{equation}
The linearization allows for more efficient computation \cite{Howard2019}.

\subsubsection{Truncated rectified (TRec) activation function}
\label{sec:trec}
The \gls[prereset]{TRec} \gls{AF} is a truncated variant of the \gls{ReLU} \cite{Konda2015}. It resembles onesided variant of the \gls{hardshrink} (see \cref{sec:hardshrink}) --- it is defined as
\begin{equation}
    f(z)= \begin{cases}
        z, \quad & z > a, \\
        0, \quad & z \leq a, \\
    \end{cases}
\end{equation}
where $a$ is a fixed parameter. \Citeauthor{Konda2015} used $a=1$ for most of their experiments \cite{Konda2015}.

\subsubsection{Hardshrink}
\label{sec:hardshrink}
The \gls{hardshrink} \cite{Goroshin2013,Konda2015,Golilarz2017,Singh2023} (named \textit{thresholded linear} \gls{AF} in \cite{Konda2015}\footnote{\citeauthor{Konda2015} proposed it as a novel \gls{AF} but it was already proposed in \cite{Goroshin2013}.}) is very similar to \gls{hard_sigmoid}, \gls{TRec}, and other piecewise linear functions; it is defined as 
\begin{equation}
    f(z)= \begin{cases}
        z, \quad & z > a, \\
        0, \quad & -a \leq z \leq a, \\
        z, \quad & z < -a, \\  
    \end{cases}
\end{equation}
where $a>0$ is a fixed parameter.

\subsubsection{Softshrink}
\label{sec:softshrink}
The \gls{softshrink} is an \gls{AF} similar to the \gls{hardshrink} used in \cite{Sipper2021, Zhang2022Fault}. It is defined as
\begin{equation}
    f(z)= \begin{cases}
        z-a, \quad & z > a, \\
        0, \quad & -a \leq z \leq a, \\
        z+a, \quad & z < -a, \\
    \end{cases}
\end{equation}
where $a>0$ is a fixed thresholding parameter \cite{Zhang2022Fault}.

\subsubsection{Bounded leaky ReLU (BLReLU)}
\label{sec:blrelu}
Similarly as the \gls{BReLU} is a bounded variant of the \gls{ReLU}, the \gls{BLReLU} is a bounded variant of \gls{LReLU} (see \cref{sec:lrelu}) \cite{Liew2016}. It is defined as 
\begin{equation}
    f(z)= \begin{cases}
                az, \quad & z \leq 0, \\
                z, \quad & 0 < z < b, \\
                az+c, \quad & z > b, \\
            \end{cases}
\end{equation}
where $a$ and $b$ are predefined parameters and $c$ is computed such that $b = ab + c$ \cite{Liew2016}, i.e. $c= (1-a)b$. The parameter $a$ controls the leakiness, the parameter $b$ is the threshold of saturation, and $c$ is computed such that the function is continuous.

\subsubsection{V-shaped ReLU (vReLU)}
\label{sec:vrelu}
A V-shaped variant of \gls{ReLU} called \gls{vReLU} is proposed in \cite{Hu2018, Hu2018Symmetric} and tackles the problem of dying neurons that is present with \glspl{ReLU} \cite{Hu2018}. The \gls{vReLU} is identical to the absolute value function and is defined as 
\begin{equation}
    f(z)=\begin{cases}
    z, \quad & z \geq 0, \\
   -z,  \quad & z < 0. \\
\end{cases}
\end{equation}
The output range of \gls{vReLU} is $[0, \infty)$ \cite{Dubey2022}. The \textit{modulus} \glsxtrlong{AF} later proposed in the literature by \citeauthor{VallsPrez2023} in \cite{VallsPrez2023} is identical to the \gls{vReLU}. The absolute value function was used as an \gls{AF} also in \cite{Berngardt2023}.

\subsubsection{Pan function}
\label{sec:pan}
The \gls{pan} function is an \gls{AF} similar to the \gls{vReLU} and \gls{softshrink} \cite{Pan2023Smoothing,Bresson2007}. It is defined as
\begin{equation}
    f(z)=\begin{cases}
    z - a, \quad & z \geq a, \\
    0, \quad & -a < z < a, \\
    -z - a, \quad & z \leq -a, \\
\end{cases}
\end{equation}
where $a$ is a fixed boundary parameter \cite{Pan2023Smoothing}.

\subsubsection{Absolute linear unit (AbsLU)}
\label{sec:abslu}
The \gls[prereset]{AbsLU} \cite{Alkhouly2021} is a \gls{ReLU}-based \gls{AF} similar to the \gls{vReLU}. It is defined as
\begin{equation}
    f(z)=\begin{cases}
    z, \quad & z \geq 0, \\
    a\cdot\left|z\right|,  \quad & z < 0, \\
\end{cases}
\end{equation}
where $a \in [0,1]$ is a fixed hyperparameter \cite{Alkhouly2021}.

\subsubsection{Mirrorer rectified linear unit (mReLU)}
\label{sec:mrelu}
The \gls[prereset]{mReLU} is a bounded \gls{AF} that suppresses the output for unusual inputs \cite{Zhao2016}. It is defined as
\begin{equation}
    f(z)=\min\left(\mathrm{ReLU}\left(1-z\right),\mathrm{ReLU}\left(1+z\right)\right)=\begin{cases}
    1+z, \quad & -1 \leq z \leq 0, \\
    1-z, \quad & 0 < z \leq 1, \\
   0,  \quad & \text{ otherwise }. \\
\end{cases}
\end{equation}

\subsubsection{Leaky single-peaked triangle linear unit (LSPTLU)}
\label{sec:lsptlu}
An \gls{AF} similar to \gls{vReLU}, \gls{AbsLU}, and \gls{tent} activation named \gls[prereset]{LSPTLU} was proposed in \cite{Shan2019}. It is defined as 
\begin{equation}
    f(z)= \begin{cases}
        0.2z, \quad & z < 0, \\
        z, \quad & 0 \leq z \leq a, \\
        2a-z, \quad & a < z \leq 2a, \\
        0, \quad & z \geq 2a, \\
    \end{cases}
\end{equation}
where $a$ is a fixed parameter \cite{Shan2019}. An identical \gls{AF} was proposed under the name \gls[prereset]{LRTLU} in \cite{Desabathula2022}.

\subsubsection{SoftModulusQ}
\label{sec:softmodulusq}
The \gls{softmodulusq} is a quadratic approximation of the \gls{vReLU} proposed in \cite{VallsPrez2023}. The \gls{softmodulusq} is defined as
\begin{equation}
    f(z)=\begin{cases}
    z^2\left(2-\left|z\right|\right), \quad & \left|z\right| \geq 1, \\
    \left|z\right|, \quad & \left|z\right| > 1. \\
\end{cases}
\end{equation}

\subsubsection{SoftModulusT}
\label{sec:softmodulust}
While the \gls{softmodulusq} (see \cref{sec:softmodulusq}) is a quadratic approximation of the \gls{vReLU} (see \cref{sec:vrelu}), the \gls{softmodulust} \cite{VallsPrez2023} is a \gls{tanh} based approximation of the \gls{vReLU}. It is basically a parametric version of the \gls{LiSHT} \glsxtrlong{AF} (see \cref{sec:lisht}):
\begin{equation}
    f(z) = z \cdot \tanh \left(\frac{z}{a}\right),
\end{equation}
where $a$ is a predetermined parameter; the authors \citeauthor{VallsPrez2023} used $a = 0.01$ in their experiments \cite{VallsPrez2023}. When $a=1$, the \gls{softmodulust} becames the \gls{LiSHT} \glsxtrlong{AF}.

\subsubsection{SignReLU}
\label{sec:signrelu}
The combination of \gls{ReLU} and \gls{softsign} resulted in \gls{SignReLU} \cite{Lin2018} that improves the convergence rate and alleviates the vanishing gradient problem \cite{Lin2018}. The \gls{SignReLU} is defined as
\begin{equation}
f(z)=\begin{cases}
            z, \quad & z \geq 0, \\
            a\frac{z}{|z| + 1},  \quad & z < 0, \\
        \end{cases}
\end{equation}
where $a$ is a fixed parameter \cite{Lin2018,Li2024}; the \gls{SignReLU} becomes \gls{ReLU} for $a=0$. The \gls{SignReLU} was independently proposed under the name \gls{DLU} in \cite{Li2022ANewActivation};\footnote{\cite{Li2022ANewActivation} is a preprint of \cite{Li2024}.} this name is sometimes used in the literature --- e.g., \cite{Pan2023Smoothing}.

\subsubsection{Li-ReLU}
\label{sec:lirelu}
\Citeauthor{Elakkiya2024} proposed a combination of a linear function and the \gls{ReLU} in \cite{Elakkiya2024}; they named the function \gls{Li-ReLU}\footnote{Not an abbreviation.} and it is defined as

\begin{equation}
    f(z)=\begin{cases}
                az + z, \quad & z_i \geq 0, \\
                az,  \quad & z_i < 0, \\
            \end{cases}
\end{equation}
where $a_i$ is a fixed parameter \cite{Elakkiya2024}.

\subsubsection{Concatenated ReLU (CReLU)}
\label{sec:crelu}
A \gls{CReLU} is an adaptation of the \gls{ReLU} function proposed based on the observation that filters in \glspl{CNN} in the lower layers form pairs consisting of filters with opposite phase \cite{Shang2016}. The \gls{CReLU} conserves both negative and positive linear responses after convolution by concatenating the output of two \glspl{ReLU} (hence the name) \cite{Shang2016}. The \gls{CReLU} is a function $\mathbb{R} \rightarrow \mathbb{R}^2$ and is defined as \cite{Shang2016}
\begin{equation}
    \vec{f}(z) = \begin{bmatrix}
        \mathrm{ReLU(z)} \\
        \mathrm{ReLU(-z)} \\
        \end{bmatrix},
\end{equation}
with the output range of $[0, \infty)$ for both output elements \cite{Dubey2022}.

\subsubsection{Negative CReLU (NCReLU)}
\label{sec:ncrelu}
A \gls{CReLU} extension named \gls{NCReLU} was proposed in \cite{Zagoruyko2017}; while it is very similar to \gls{CReLU}, it multiplies the second element by $-1$:
\begin{equation}
    \vec{f}(z) = \begin{bmatrix}
        \mathrm{ReLU(z)} \\
        \mathrm{-ReLU(-z)} \\
        \end{bmatrix}.
\end{equation}

Very similar \gls{AF} was proposed concurrently in \cite{Eidnes2017} under the name \gls{BAF}. Unlike the \gls{NCReLU}, it does not produce a vector output but is applied in an alternating manner similar to \glsxtrshort{All-ReLU} (see \cref{sec:allrelu}) but for neurons instead of layers. It is defined for the {$i$-th} neuron as
\begin{equation}
    f(z_i)=\begin{cases}
                g\left(z_i\right), \quad & i \% 2 = 0, \\
                -g\left(-z_i\right), \quad &  i \% 2 = 1, \\
            \end{cases}
\end{equation}
where $g\left(z_i\right)$ is any \gls{ReLU} family \gls{AF} and $\%$ is the modulo operation.

\subsubsection{DualReLU}
\label{sec:dualrelu}
Where \gls{CReLU} \glsxtrlongpl{AF} takes a single value and outputs a vector of two values, the \gls{dualrelu} \cite{Godin2018} takes two values as an input and outputs a single value. The \gls{dualrelu} is a two-dimensional \glsxtrlong{AF} meant as a replacement of the \gls{tanh} \glsxtrlong{AF} for Quasi-Recurrent \glsxtrlongpl{NN} \cite{Godin2018}. It is defined as
\begin{equation}
f(z, z')= \max\left(0, z\right) - \max\left(0, z'\right) = 
\begin{cases}
            0, \quad & z \leq 0 \land z' \leq 0, \\
            z, \quad & z > 0 \land z' \geq 0, \\
            -b, \quad & z \leq 0 \land z' > 0, \\
            a-b, \quad & z > 0 \land z' > 0. \\
        \end{cases}
\end{equation}

\subsubsection{Orthogonal permutation liner unit}
\label{sec:oplu}
The \gls{OPLU} is not applied to a single neuron but always to a pair of neurons \cite{Chernodub2016}. First, the neurons are grouped into pairs of neurons $\{i,j\}$ and the \gls{OPLU} takes two inputs $z_i$ and $z_j$ of neurons $i$ and $j$ and produces the output
\begin{equation}
    f(z_i, z_j)= \max\left(z_i,z_j\right)
\end{equation}
for neuron $i$ and 
\begin{equation}
    f(z_i, z_j)= \min\left(z_i,z_j\right)
\end{equation}
for neuron $j$ \cite{Chernodub2016}.

\subsubsection{Elastic ReLU (EReLU)}
\label{sec:erelu}
Another extension is the \gls{EReLU}, which slightly randomly changes the slope of the positive part of the \gls{ReLU} during the training \cite{Jiang2018}. The \gls{EReLU} is defined as 
\begin{equation}
    f(z_i)=\begin{cases}
                k_i z_i, \quad & z_i \geq 0, \\
                0,  \quad & z_i < 0, \\
            \end{cases}
    \end{equation}
where $k_i$ is a sampled for each epoch and neuron $i$  from the uniform distribution: $a_i \sim U(1-\alpha, 1+\alpha)$ where $\alpha \in (0,1)$ is a parameter controlling the degree of response fluctuations \cite{Jiang2018}. The \gls{EReLU} thus complements the principle of \gls{RReLU}, which randomly changes the leakiness during the training while keeping the positive part fixed, while the \gls{EReLU} changes the positive part and keeps the output constantly zero for negative inputs. The \gls{EReLU} sets the $k_i$ its expected value $\E(k_i)$ which is equal to one --- the \gls{EReLU} becomes the \gls{ReLU} during the test phase \cite{Jiang2018}.

\subsubsection{Power activation functions \& rectified power units (RePU)}
\label{sec:repu}
A \gls{power_activation_function} extending \gls{ReLU} together with a training scheme for better generalization was proposed in \cite{Berradi2018}. This \glsxtrlong{AF} was later independently proposed under the name \gls{RePU} in \cite{Li2019PowerNet}. The \gls{RePU} is defined as
\begin{equation}
    f(z)=\begin{cases}
                z^a, \quad & z \geq 0, \\
                0,  \quad & z < 0, \\
        \end{cases}
\end{equation}
where $a$ is a fixed parameter \cite{Berradi2018,Heeringa2023}. The \gls{RePU} is a generalization of several \glsxtrlongpl{AF} --- it becomes the Heaviside step function for $a=0$ and \gls{ReLU} for $a=1$; the case $a=2$ is called \textit{rectified quadratic unit} (ReQU) in \cite{Li2019PowerNet} and \textit{squared ReLU} in \cite{So2021}; finally, the case $a=3$ is called \textit{rectified cubic unit} (ReCU) \cite{Li2019PowerNet}. The disadvantage of \gls{RePU} is its unbounded and asymmetric nature and that it is prone to vanishing gradient \cite{Dubey2022}. Theoretical analysis of the \gls{RePU} is available in \cite{Heeringa2023}.

However, \citeauthor{Berradi2018} recommends alternating using $a = b$ and $a=\frac{1}{b}$ each epoch; i.e.:
\begin{equation}
    f_1(z)=\begin{cases}
                z^b, \quad & z \geq 0, \\
                0,  \quad & z < 0, \\
        \end{cases}
\end{equation}
and
\begin{equation}
    f_2(z)=\begin{cases}
                z^\frac{1}{b}, \quad & z \geq 0, \\
                0,  \quad & z < 0. \\
        \end{cases}
\end{equation}
Then the \glsxtrlong{AF} $f_1(z)$ is used during odd epochs and $f_2(z)$ during even epochs; their mean is used during the test phase \cite{Berradi2018}. The value $b >1$ was used in the experiments in \cite{Berradi2018} - $b \in \{1.05, 1.1,1.15, 1.20, 1.25\}$.

\subsubsection{Approximate ReLU (AppReLU)}
\label{sec:apprelu}
The \gls[prereset]{AppReLU}\footnote{\Citeauthor{Saha2020} used the abbreviation \glsxtrshort{AReLU} but this is already used for the \glsxtrlong{AReLU} in this work.} \cite{Saha2020, Mediratta2021} is the \gls{RePU} with aditional scaling parameter; it is defined as
\begin{equation}
    f(z)=\begin{cases}
                az^b, \quad & z \geq 0, \\
                0,  \quad & z < 0. \\
        \end{cases}
\end{equation}

\subsubsection{Power linear activation function (PLAF)}
\label{sec:plaf}
The \gls[prereset]{PLAF}\footnote{Originally, \citeauthor{Nasiri2021} named \glsxtrshort{PLAF} as \textit{PowerLinear} \gls{AF}. Also, its variants \glsxtrshort{EPLAF} and \glsxtrshort{OPLAF} were named as \textit{EvenPowLin} and \textit{OddPowLin} in \cite{Nasiri2021}.} is a class of two similar \glspl{AF} proposed in \cite{Nasiri2021}. The first, \gls[prereset]{EPLAF}, is defined as
\begin{equation}
    f(z)=\begin{cases}
                z - \left(1-\frac{1}{d}\right), \quad & z \geq 1, \\
                \frac{1}{d}\left|z\right|^d,  \quad &  -1 \leq z < 1, \\
                - z - \left(1-\frac{1}{d}\right),  \quad & z < -1, \\
        \end{cases}
\end{equation}
where $d$ is a fixed parameter \cite{Nasiri2021}. Similarly, the second \gls{AF} ---  \gls[prereset]{OPLAF} --- is defined as
\begin{equation}
    f(z)=\begin{cases}
                z - \left(1-\frac{1}{d}\right), \quad & z \geq 1, \\
                \frac{1}{d}\left|z\right|^d,  \quad &  0 \leq z < 1, \\
                -\frac{1}{d}\left|z\right|^d,  \quad &  -1 \leq z < 0, \\
                - z - \left(1-\frac{1}{d}\right),  \quad & z < -1, \\
        \end{cases}
\end{equation}
where $d$ is a fixed parameter \cite{Nasiri2021}. \Citeauthor{Nasiri2021} focused on the \gls{EPLAF} in their work \cite{Nasiri2021} and showed that \gls{EPLAF} with $d=2$ performed similarly as the \gls{ReLU} for some of the tasks but it performed significantly better for other tasks; the \gls{OPLAF} was not experimentally validated in \cite{Nasiri2021}.

\subsubsection{Average biased ReLU (ABReLU)}
\label{sec:abrelu}
Similarly as the \gls{RT-ReLU} (see \cref{sec:rtrelu}), the \gls{ABReLU} \cite{Dubey2021} uses horizontal shifting in order to handle negative values \cite{Dubey2022}. It is defined as
\begin{equation}
    f(z_i)=\begin{cases}
                z_i - a_i, \quad & z_i - a_i \geq 0, \\
                0,  \quad & z_i - a_i < 0, \\
            \end{cases}
\end{equation}
where $a_i$ is the average of input activation map to the neuron/filter $i$ \cite{Dubey2021, Dubey2022}, which makes the function data dependent and adjusts the threshold based on the positive and negative data dominance \cite{Dubey2021}. The output range is $[0, \infty)$ \cite{Dubey2022}.

\subsubsection{Delay ReLU (DRLU)}
\label{sec:drlu}
The \gls{DRLU}\footnote{Authors termed the function \gls{DRLU}; however, the usual notation in this work would be \glsxtrshort{DReLU}. Since such notation would collide with the \glsxtrlong{DReLU}, we will use the original notation from \cite{Shan2022} despite the inconsistency.} is a function that also adds a horizontal shift to the \gls{ReLU} \cite{Shan2022}; however, the \gls{DRLU} uses a fixed, predetermined shift whereas \gls{RT-ReLU} uses stochastic shifts (see \cref{sec:rtrelu}) and \gls{ABReLU} computes the shift as the average of input activation map (see \cref{sec:abrelu}). The \gls{DRLU} is defined as
\begin{equation}
    f(z)=\begin{cases}
                z - a, \quad & z - a \geq 0, \\
                0,  \quad & z - a < 0, \\
            \end{cases}
\end{equation}
where $a$ is a fixed, predetermined parameter \cite{Shan2022}. \Citeauthor{Shan2022} also add a constraint $a>0$ \cite{Shan2022} and they used $a \in \{0.06,0.08,0.10\}$ in their experiments \cite{Shan2022}.

\subsubsection{Displaced ReLU (DisReLU)}
\label{sec:disrelu}
Very similar to the \gls{FReLU} (see \cref{sec:frelu}) and \gls{DReLU} (see \cref{sec:drelu}) is the \gls[prereset]{DisReLU}\footnote{\Citeauthor{Macdo2019} originally abbreviated the \glsxtrlong{DisReLU} as \glsxtrshort{DReLU} but that is already taken by \glsxtrlong{DReLU} from \cref{sec:drelu}.} as it also shifts the \gls{ReLU} \cite{Macdo2019}: 
\begin{equation}
    f(z)= \begin{cases}
        z , \quad & z + a \geq 0, \\
        -a,  \quad & z + a < 0, \\
    \end{cases}
\end{equation}
where $a$ is a predefined hyperparameter \cite{Macdo2019,Dubey2022}. A \gls{shifted_relu} (see \cref{sec:shifted_relu}) is a special case of \gls{DisReLU} with $a = 1$ \cite{Macdo2019}. The VGG-19 \cite{Simonyan2014} with \gls{DisReLU}s outperform the \gls{ReLU}, \gls{LReLU}, \gls{PReLU}, and \gls{ELU} \glsxtrlongpl{AF} with a statistically significant difference in performance on the CIFAR-10 and CIFAR-100 datasets \cite{Krizhevsky2009} as shown in \cite{Macdo2019}.

\subsubsection{Modified LReLU}
\label{sec:mlrelu}
Inspired by the \gls{DisReLU} \cite{Macdo2019}, \citeauthor{Yang2022Deep} proposed the \gls[prereset]{MLReLU} in \cite{Yang2022Deep}. The \gls{MLReLU} is a translated \gls{LReLU} and is defined as
\begin{equation}
    f(z)= \begin{cases}
        z , \quad & z + a > 0, \\
        -az,  \quad & z + a \leq 0, \\
    \end{cases}
\end{equation}
where $a$ is a fixed parameter controlling both the slope and the threshold \cite{Yang2022Deep}.

\subsubsection{Flatted-T swish}
\label{sec:fts}
An \glsxtrlong{AF} \gls{FTS} \cite{Chieng2018} combines \gls{ReLU} and the \gls{logisticsigmoid} \glsxtrlong{AF}; it is defined as
\begin{equation}
    f(z) = \mathrm{ReLU}(z) \cdot \sigma(z) + T = \begin{cases}
        \frac{z}{1+\exp(-z)}+T,  \quad & z \geq 0, \\
        T,  \quad & z < 0, \\
    \end{cases}
\end{equation}
where $T$ is a predefined hyperparameter \cite{Chieng2018}, the recommended value is $T=-0.20$ \cite{Chieng2018}. The \gls{FTS} is identical to a shifted \gls{swish} for the positive $z$. The \gls{FTS} was shown to outperform \gls{ReLU}, \gls{LReLU}, \gls{swish}, \gls{ELU}, and \gls{FReLU} \glsxtrlongpl{AF} \cite{Chieng2018}. The special case with $T=0$ was proposed independently under the name of \gls{ReLUSwish} in \cite{Koak2021}.

\subsubsection{Optimal activation function (OAF)}
\label{sec:oaf}
The so-called \gls{OAF} is a combination of \gls{ReLU} and \gls{swish} activations proposed in \cite{Sharma2020ANovel}. It is defined as
\begin{equation}
    f(z) = \mathrm{ReLU}(z) + z\cdot\sigma(z) = \begin{cases}
        z + z\cdot\sigma(z) , \quad & z \geq 0, \\
        z\cdot\sigma(z),  \quad & z < 0. \\
    \end{cases}
\end{equation}

\subsubsection{Exponential linear unit (ELU)}
\label{sec:elu}
An \gls{ELU} is an extension of \gls{LReLU} where the function employs an exponential function for the negative inputs, which speeds up the learning process \cite{Clevert2015}:
\begin{equation}
    f(z)=\begin{cases}
                z, \quad & z \geq 0, \\
                \frac{\exp(z)-1}{a},  \quad & z < 0, \\
          \end{cases}
\end{equation}
where $a$ is a hyperparameter; the authors \citeauthor{Clevert2015} used $a = 1$ in their work \cite{Clevert2015}. The $a$ determines the value to which an \gls{ELU} saturates for inputs going to negative infinity \cite{Clevert2015}. 

\subsubsection{Rectified exponential unit (REU)}
\label{sec:reu}
A \gls{REU} \cite{Ying2019} is an \glsxtrlong{AF} inspired by the \gls{ELU} and \gls{swish} (see \cref{sec:elu, sec:swish}) and is based on the assumption that the success of the \gls{swish} \glsxtrlongpl{AF} is due to the non-monotonic property in the negative quadrant \cite{Ying2019}. The \gls{REU} is defined as
\begin{equation}
    f(z)=\begin{cases}
                z, \quad & z \geq 0, \\
                z\cdot\exp(z),  \quad & z < 0. \\
          \end{cases}
\end{equation}
A parametric version called \gls{PREU} was also proposed in \cite{Ying2019}; see \cref{sec:preu} for details.

\subsubsection{Apical dendrite activation (ADA)}
\label{sec:ada}
A biologically inspired \gls{AF} named \gls[prereset]{ADA} was proposed in \cite{Georgescu2023}. It is similar to the \gls{ELU}, but it applies an exponential function for positive inputs. It is defined as
\begin{equation}
    f(z)=\begin{cases}
            \exp\left(-az+b\right), \quad & z \geq 0, \\
            0,  \quad & z < 0, \\
          \end{cases}
\end{equation}
where $a$ and $b$ are fixed parameters \cite{Georgescu2023}.

\subsubsection{Leaky apical dendrite activation (LADA)}
\label{sec:lada}
As \gls{LReLU} extends the \gls{ReLU}, the \gls[prereset]{LADA} \cite{Georgescu2023} extends the \gls{ADA}.
\begin{equation}
    f(z)=\begin{cases}
            \exp\left(-az+b\right), \quad & z \geq 0, \\
            cz,  \quad & z < 0, \\
          \end{cases}
\end{equation}
where $a$, $b$, and $c \in [0,1]$ are fixed parameters \cite{Georgescu2023}. \Citeauthor{Georgescu2023} used $c=0.01$ in their experiments in \cite{Georgescu2023}.

\subsubsection{Sigmoid linear unit (SigLU)}
\label{sec:siglu}
The \gls[prereset]{SigLU}\footnote{The \gls{AF} is unnamed in the original work \cite{Guevraa2021}.} is an \gls{ELU} alternative that uses a modified \gls{logisticsigmoid} instead of the exponential \cite{Guevraa2021}. It is defined as
\begin{equation}
    f(z)=\begin{cases}
                z, \quad & z \geq 0, \\
                \frac{1-\exp\left(-2z\right)}{1+\exp\left(-2z\right)},  \quad & z < 0. \\
          \end{cases}
\end{equation}

\subsubsection{Swish and ReLU activation (SaRa)}
\label{sec:sara}
The \gls[prereset]{SaRa} is an \gls{AF} combining the \gls{swish} and \gls{ReLU} \glspl{AF} proposed in \cite{Qureshi2022}. It is defined\footnote{The formula in \cite{Qureshi2022} is malformed; we believe that this is the intended case. It is possible that authors intended that the \gls{SaRa} is actually only the part that is defined for the negative inputs in \cref{eq:sara} --- however, we think that it is less likely as that would be only a \gls{swish} (see \cref{sec:swish}) \gls{AF} with some fixed scaling of the output or the \glsxtrshort{AHAF} (see \cref{sec:ahaf}) with fixed parameters.} as
\begin{equation}
    \label{eq:sara}
    f(z)=\begin{cases}
                z, \quad & z \geq 0, \\
                \frac{z}{1+a \cdot\exp\left(-b z\right)},  \quad & z < 0, \\
          \end{cases}
\end{equation}
where $a$ and $b$ are fixed parameters; \citeauthor{Qureshi2022} recommend $a=0.5$ and $b=0.7$ \cite{Qureshi2022}.

\subsection{Maxsig}
\label{sec:maxsig}
The \gls{maxsig} is one of the \glspl{AF} listed in \cite{Xu2020Comparison}. The \gls{maxsig} is similar to the \gls{SigLU} (see \cref{sec:siglu}) and is  defined as
\begin{equation}
    f(z) = \max\left(z, \sigma(z)\right),
\end{equation}
where $\sigma(z)$ is the \gls{logisticsigmoid} \cite{Xu2020Comparison}.

\subsubsection{Tanh linear unit (ThLU)}
\label{sec:thlu}
The \gls[prereset]{ThLU} \cite{Fu2022Research}\footnote{The ref \cite{Fu2022Research} is not the original work with \glspl{ThLU}; it references another work but that uses pure \gls{tanh} as the \glspl{AF}.} is an \gls{AF} combining \gls{tanh} and \gls{ReLU}. It is defined as
\begin{equation}
    f(z)=\begin{cases}
                z, \quad & z \geq 0, \\
                \frac{2}{1+\exp\left(-z\right)} -1,  \quad & z < 0, \\
          \end{cases} = \begin{cases}
            z, \quad & z \geq 0, \\
            \tanh\left(\frac{z}{2}\right),  \quad & z < 0. \\
      \end{cases}
\end{equation}
The \gls{ThLU} is a special case of the \gls{TReLU} with $b_i=\frac{1}{2}$. Similar \gls{AF} was used under the name \gls{maxtanh} in \cite{Xu2020Comparison} --- it just omitted the scaling factor. The \gls{maxtanh} can also be written as $f(z) = \max\left(z, \tanh\left(z\right)\right)$ \cite{Xu2020Comparison}.

\subsubsection{DualELU}
\label{sec:dualelu}
The \gls{dualelu} \cite{Godin2018} is equivalent of \gls{dualrelu} (see \cref{sec:dualrelu}) for ELUs and are defined as
\begin{equation}
f(z, z')= f_{\mathrm{EL}}\left(z\right) - f_{\mathrm{EL}}\left(z'\right),
\end{equation}
where $f_{\mathrm{EL}}\left(z\right)$ is the \gls{ELU} \glsxtrlong{AF} applied to an input $z$.

\subsubsection{Difference ELU (DiffELU)}
\label{sec:diffelu}
An \gls{ELU} variant named \gls[prereset]{DiffELU}\footnote{\Citeauthor{Hu2020Improving} used the abbreviation \gls{DELU} but this name is used for the \gls{AF} proposed by \citeauthor{Pishchik2023} in \cite{Pishchik2023} throughout this work.} was proposed in \cite{Hu2020Improving}. It is defined as
\begin{equation}
f(z)=\begin{cases}
    z, \quad & z \geq 0, \\
    a \left(z \exp\left(z\right) - b \exp\left(bz\right)\right),  \quad & z < 0, \\
\end{cases}
\end{equation}
where $a$ and $b \in (0,1)$ are fixed parameters \cite{Hu2020Improving}. \Citeauthor{Hu2020Improving} also tested setting the parameters to be trainable but that led to worse performance \cite{Hu2020Improving}. The recommended setting is $a=0.3$ and $b=0.1$ \cite{Hu2020Improving}.

\subsubsection{Polynomial linear unit (PolyLU)}
\label{sec:polylu}
The \gls[prereset]{PolyLU} is an \gls{AF} similar to the \gls{ELU} proposed in \cite{Feng2023PolyLU}. It is defined as
\begin{equation}
    f(z)=\begin{cases}
                z, \quad & z \geq 0, \\
                \frac{1}{1-z}-1,  \quad & z < 0. \\
          \end{cases}
\end{equation}
Despite the similarity with the \gls{ELU}, \citeauthor{Feng2023PolyLU} have shown that the \gls{PolyLU} outperformed the \gls{ELU} on the CIFAR-10/100 \cite{Krizhevsky2009} and Dogs vs. Cats \cite{DogsvsCats,Elson2007} datasets \cite{Feng2023PolyLU}. The \gls{PolyLU} was also proposed under the name \gls[prereset]{FPLUS}\footnote{\Citeauthor{Duan2022Activation} used the equivalent definition $f(z) = \left(\mathrm{sgn}\left(z\right)\cdot z + 1\right)^{\mathrm{sgn}(z)}-1$ in \cite{Duan2022Activation}, hence the name.} in \cite{Duan2022Activation}.

\subsubsection{Inverse polynomial linear unit (IpLU)}
\label{sec:iplu}
The \gls[prereset]{IpLU} was proposed in \cite{Alkhouly2021}; it is defined as
\begin{equation}
    f(z)=\begin{cases}
                z, \quad & z \geq 0, \\
                \frac{1}{1+|z|^a},  \quad & z < 0, \\
          \end{cases}
\end{equation}
where $a>0$ is a fixed hyperparameter guaranteeing a small slope for negative inputs \cite{Alkhouly2021}.

\subsubsection{Power linear unit (PoLU)}
\label{sec:polu}
The \gls[prereset]{PoLU} \cite{Li2018Training} is an \gls{AF} similar to the \gls{ELU}. It is defined as
\begin{equation}
    f(z)=\begin{cases}
                z, \quad & z \geq 0, \\
                \left(1-z\right)^{-a}-1,  \quad & z < 0, \\
          \end{cases}
\end{equation}
where $a$ is a fixed parameter \cite{Li2018Training}. \Citeauthor{Li2018Training} used $a \in \{1, 1.5, 2\}$ in their experiments in \cite{Li2018Training}.

\subsubsection{Power function linear unit (PFLU)}
\label{sec:pflu}
The \gls[prereset]{PFLU} is an \gls{AF} proposed in \cite{Zhu2021PFLU}; it is defined as
\begin{equation}
    f(z) = z\cdot\frac{1}{2}\left(1+\frac{z}{\sqrt{1+z^2}}\right).
\end{equation}

\subsubsection{Faster power function linear unit (FPFLU)}
\label{sec:fpflu}
The \gls[prereset]{FPFLU} is an \gls{AF} proposed in \cite{Zhu2021PFLU} that resembles the \gls{IpLU} (see \cref{sec:iplu}) It is defined as
\begin{equation}
    f(z)=\begin{cases}
                z, \quad & z \geq 0, \\
                z + \frac{z^2}{\sqrt{1+z^2}},  \quad & z < 0. \\
          \end{cases}
\end{equation}

\subsubsection{Elastic adaptively parametric compounded unit (EACU)}
\label{sec:eacu}
The \gls[prereset]{EACU} \cite{Zhang2023Elastic} is a stochastic \gls{AF}. It is defined as
\begin{equation}
    f(z_i)=\begin{cases}
                b_iz_i, \quad & z_i \geq 0, \\
                a_iz_i\cdot \tanh\left(\ln\left(1+\exp\left(a_i,z_i\right)\right)\right),  \quad & z_i < 0, \\
          \end{cases}
\end{equation}
where $b_i$ is stochastically sampled during training as
\begin{equation}
    b_i = \begin{cases}
        s_i, \quad & 0.5 < s_i < 1.5, \\
        1,  \text{ otherwise}, \\
  \end{cases}
\end{equation}
\begin{equation}
    s_i \sim mathrm{N}\left(0, 0.01\right),
\end{equation}
and $a_i$ is an adaptive parameter for each neuron or channel $i$ \cite{Zhang2023Elastic}.

\subsubsection{Lipschitz ReLU ({L--ReLU})}
\label{sec:l-relu}
A \gls{L-ReLU} \cite{Basirat2020} is a piecewise linear \glsxtrlong{AF}. The slope of the negative part is selected with respect to a data-dependent Lipschitz constant \cite{Basirat2020}. It builds on a proposed piecewise function that treats the positive $z>0$ and negative values ($z \leq 0$) separately:

\begin{equation}
    f(z) = p(z|z >0) + n(z | z \leq 0),
\end{equation}
where
\begin{equation}
    p(z) = \max\left(\phi(z), 0\right),
\end{equation}
and
\begin{equation}
    n(z) = \min(\mu(z), 0),
\end{equation}
where $\phi(z)$ and $\mu(z)$ can be any function $f: \mathbb{R} \rightarrow \mathbb{R}$ \cite{Basirat2020}. This makes the positive part of the piecewise lay in the first quadrant of the Cartesian coordinate system and the negative part in the third quadrant \cite{Basirat2020}.

\subsubsection{Scaled exponential linear unit (SELU)}
\label{sec:selu}
A \gls{SELU} \cite{Klambauer2017} was proposed in order to make the network self-normalize by automatically converging towards zero mean and unit variance \cite{Dubey2022}. The \gls{ELU} was chosen as the basis for \glspl[prereset]{SNN} because these cannot be derived with \glspl{ReLU}, \gls{sigmoid}, and \gls{tanh} units or even \glspl{LReLU} \cite{Klambauer2017} --- the \glsxtrlong{AF} has to have negative and positive values for controlling the mean, saturation region where derivatives approach zero in order to dampen the variance if it is too large, a slope larger than one in order to increase the variance if it is too small, and a continuous curve to ensure a fixed point where the variance dampening is balanced out by the variance increasing \cite{Klambauer2017}.
The \gls{SELU} is defined as
\begin{equation}
    f(z)=\begin{cases}
                 a z, \quad & z \geq 0, \\
                a b \left(\exp\left(z\right)-1\right),  \quad & z < 0, \\
            \end{cases}
\end{equation}
where $a > 1$ and $b$  are predefined parameters \cite{Dubey2022, Klambauer2017}; the recommended values are $a \approx 1.05078$ and $b \approx 1.6733$ \cite{Klambauer2017}.

\subsubsection{Leaky scaled exponential linear unit (LSELU)}
\label{sec:lselu}
A leaky variant of \gls{SELU} called \gls{LSELU} was proposed in \cite{Chen2021Redefining} and is defined as
\begin{equation}
    f(z)=\begin{cases}
                 a z, \quad & z \geq 0, \\
                a b \left(\exp\left(z\right)-1\right) + acz,  \quad & z < 0, \\
            \end{cases}
\end{equation}
where $a > 1$ and $b$  are predefined parameters of the original \gls{SELU} (see \cref{sec:selu}), and $c$ is a new, predefined parameter controlling the leakiness of the unit \cite{Chen2021Redefining}.

\subsubsection{Scaled exponentially-regularized linear unit (SERLU)}
\label{sec:serlu}
The \gls[prereset]{SERLU} is a modification of the \gls{SELU} proposed in \cite{Zhang2018Effectiveness}; it is defined as
\begin{equation}
    f(z)=\begin{cases}
                 a z, \quad & z \geq 0, \\
                a b z \exp\left(z\right),  \quad & z < 0, \\
            \end{cases}
\end{equation}
where $a > 0$ and $b>0$  are predefined parameters \cite{Zhang2018Effectiveness}. An extension of this approach named \gls{ASERLU} for \gls{BiLSTM} architectures was proposed in \cite{Hermawan2021}.

\subsubsection{Scaled scaled exponential linear unit (sSELU)}
\label{sec:sselu}
Additional scaling of the negative pre-activations was introduced in the \gls{sSELU} \cite{Chen2021Redefining}:
\begin{equation}
    f(z)=\begin{cases}
                 a z, \quad & z \geq 0, \\
                a b \left(\exp\left(cz\right)-1\right),  \quad & z < 0, \\
            \end{cases}
\end{equation}
where $a > 1$ and $b$  are predefined parameters of the original \gls{SELU} (see \cref{sec:selu}), and $c$ is a new, predefined parameter controlling the scaling of the negative inputs to the unit \cite{Chen2021Redefining}.

\subsubsection{RSigELU}
\label{sec:rsigelu}
A parametric \gls{ELU} variant called \gls{rsigelu} \cite{Kiliarslan2021} is defined as
\begin{equation}
    f(z)=\begin{cases}
                 z \left( \frac{1}{1+\exp(-z)}\right) a + z , \quad & 1 < z < \infty, \\
                 z , \quad & 0 \geq z \geq 1, \\
                 a\left(\exp(z)-1\right) , \quad & -\infty < z < 0, \\
            \end{cases}
\end{equation}
where $a$ is a predefined parameter, \citeauthor{Kiliarslan2021} used $0 < a < 1$ in their work \cite{Kiliarslan2021}. For $a = 0$, the \gls{rsigelu} becomes \gls{ReLU} \cite{Kiliarslan2021}. The \gls{rsigelu} was shown to outperform \gls{ReLU}, \gls{LReLU}, \gls{softsign}, \gls{swish}, \gls{ELU}, SEU, \gls{GELU}, LISA, \gls{hexpo}  and \gls{softplus} on the MNIST dataset \cite{LiDeng2012}, Fashion MNIST \cite{Xiao2017} and the IMDB Movie dataset; it still outperformed these \glsxtrlongpl{AF} on the CIFAR-10 dataset \cite{Krizhevsky2009} but it was outperformed by its variant \gls{rsigelud} \cite{Kiliarslan2021}.

\subsubsection{HardSReLUE}
\label{sec:hardsrelue}
Another \gls{AF} proposed by Kili\c{c}arslan is the \gls{HardSReLUE} \cite{Kiliarslan2022ANovel}. \Citeauthor{Kiliarslan2022ANovel} defined the \gls{AF} as
\begin{equation}
    f(z)=\begin{cases}
                 az \left(\max\left(0,\min\left(1, \frac{z+1}{2}\right)\right)\right) + z, \quad & z \geq 0, \\
                a\left(\exp(z)-1\right),  \quad & z < 0, \\
            \end{cases}
\end{equation}
where $a$ is a fixed slope parameter \cite{Kiliarslan2022ANovel}.

\subsubsection{Exponential linear sigmoid squashing (ELiSH)}
\label{sec:elish}
An \glsxtrlong{AF} \gls{ELiSH} \cite{Basirat2018} combines the \gls{swish} (see \cref{sec:swish}) and the \gls{ELU} function \cite{Dubey2022}. It is defined as
\begin{equation}
    f(z)=\begin{cases}
                 \frac{z}{1+\exp(-z)}, \quad & z \geq 0, \\
                \frac{\exp(z)-1}{1+\exp(-z)},  \quad & z < 0. \\
            \end{cases}
\end{equation}

\subsubsection{Hard exponential linear sigmoid squashing (HardELiSH)}
\label{sec:hardelish}
As \gls{ELiSH} (see \cref{sec:elish}) combines \gls{swish} with \gls{ELU} and linear function, the \gls{HardELiSH} combines the \gls{hard_sigmoid} \cite{Courbariaux2016} with \gls{ELU} and linear function \cite{Basirat2018}. It is defined as
\begin{equation}
    f(z)=\begin{cases}
                 z \cdot \max\left(0, \min\left(\frac{z+1}{2}, 1\right)\right), \quad & z \geq 0, \\
                 \left(1+\exp\left(-z\right)\right) \cdot \max\left(0, \min\left(\frac{z+1}{2}, 1\right)\right),  \quad & z < 0. \\
            \end{cases}
\end{equation}

\subsubsection{RSigELUD}
\label{sec:rsigelud}
The \gls{rsigelud} is a double parameter variant of the \gls{rsigelu} (see \cref{sec:rsigelu}) \cite{Kiliarslan2021} that is defined as
\begin{equation}
    f(z)=\begin{cases}
                 z \left( \frac{1}{1+\exp(-z)}\right) a + z , \quad & 1 < z < \infty, \\
                 z , \quad & 0 \leq z \leq 1, \\
                 b\left(\exp(z)-1\right), \quad & -\infty < z < 0, \\
            \end{cases}
\end{equation}
where $a$ and $b$ are predefined parameters, \citeauthor{Kiliarslan2021} used $0 < a < 1$ and $0 < b < 1$ in their work \cite{Kiliarslan2021}. For $a = b = 0$, the \gls{rsigelud} becomes the \gls{ReLU} the same as the \gls{rsigelu}; however, for $a=0$ and positive $b$, the function resembles the vanilla \gls{ELU} \cite{Kiliarslan2021}.

\subsubsection{LS--ReLU}
\label{sec:lsrelu}
The \gls{LSReLU}\footnote{Not an abbreviation.} is a \gls{ReLU}-inspired \gls{AF} proposed in \cite{Wang2020TheInfluence}. It is defined as
\begin{equation}
f(z)=\begin{cases}
    \frac{z}{1+\left|z\right|}, \quad & z \leq 0, \\
    z , \quad & 0 \leq z \leq b, \\
    \log\left(az+1\right)+\left| \log\left(ab + 1\right) -b \right|, \quad & z \geq b, \\
\end{cases}
\end{equation}
where $a$ and $b$ are fixed\footnote{\Citeauthor{Wang2020TheInfluence} do not specify whether the parameters are trainable or fixed.} parameters \cite{Wang2020TheInfluence}.

\subsection{Square-based activation functions}
\label{sec:squarebased}
Several square-based \glsxtrlongpl{AF} were proposed in \cite{Wuraola2021, Wuraola2018, Bilski2016} for better computational efficiency, especially on low-power devices \cite{Wuraola2021}. The approach uses the square function to replace the potentially costly exponential function. These function leads to significantly more efficient computation when there is no hardware implementation of the exponential function \cite{Wuraola2021}.
The efficiency gains can be further improved with a custom hardware operator
\begin{equation}
    f_h(x) = -\left|x\right|\cdot x,
\end{equation}
which can be used for efficient hardware implementation of all of the \glsxtrlongpl{AF} of the square-based family \cite{Wuraola2021}. The usage of the AFs from the family can lead to performance gains of one order of magnitude compared to traditional AFs \cite{Wuraola2021} for both forward and backward passes (depends on the particular \glsxtrlong{AF} and the usage of fixed or floating point representations) \cite{Wuraola2021}.

\subsubsection{SQNL}
\label{sec:sqnl}
A computationally efficient \glsxtrlong{AF} was proposed in \cite{Wuraola2018}; unlike many other sigmoidal functions, it uses the square operator instead of the exponential function in order to achieve better computational efficiency. The derivative of the function is linear, which leads to a less computationally costly computation of the gradient. The function is defined in \cite{Wuraola2021} (the original paper \cite{Wuraola2018} had several mistakes in the definition) as
\begin{equation}
    f(z)=\begin{cases}
            1, \quad & z > 2, \\
            z - \frac{z^2}{4},  \quad & 0 \leq z \leq 2, \\
            z + \frac{z^2}{4},  \quad & -2 \leq z < 0, \\
            -1, \quad & z < -2. \\
        \end{cases}
\end{equation}

The \gls{SQNL}\footnote{\gls{SQNL} is not an abbreviation but rather a name given by \citeauthor{Wuraola2018}.} has bounded range $[-1, 1]$ \cite{Wuraola2021}. The performance of the \gls{SQNL} was verified on several datasets from the {UCI} Machine Learning Repository \cite{Dua2017} and on the MNIST dataset \cite{LiDeng2012}; more details available in \cite{Wuraola2018}.

\subsubsection{Square linear unit (SQLU)}
\label{sec:sqlu}
Similarly as the \gls{SQNL} (see \cref{sec:sqnl}) uses square function to form a sigmoidal function to approximate \gls{tanh}, the \gls{SQLU} \cite{Wuraola2021} uses square function to form a \gls{ELU}-like \glsxtrlong{AF} that is computationally efficient:
\begin{equation}
    f(z)=\begin{cases}
            z, \quad & z > 0, \\
            z + \frac{z^2}{4},  \quad & -2 \leq z \leq 0, \\
            -1, \quad & z < -2. \\
        \end{cases}
\end{equation}
The \gls{SQLU} basically uses the negative part of the \gls{SQNL} and replaces the positive part with a linear function.

\subsubsection{Square swish (squish)}
\label{sec:squish}
Another example of the family of \glsxtrlongpl{AF} based on the square operator is the \gls{squish} \cite{Wuraola2021}, which is an AF inspired by the \gls{swish} and \gls{GELU} (see \cref{sec:gelu}). It uses the square non-linearity in order to achieve good computational efficiency:
\begin{equation}
    f(z)=\begin{cases}
            z + \frac{z^2}{32}, \quad & z > 0, \\
            z + \frac{z^2}{2},  \quad & -2 \leq z \leq 0, \\
            0, \quad & z < -2. \\
        \end{cases}
\end{equation}
While the \gls{squish} was inspired by the \gls{swish} and \gls{GELU} \glsxtrlongpl{AF}, it is an approximation of neither \cite{Wuraola2021}.

\subsubsection{Square REU (SqREU)}
Similarly as \gls{REU} (see \cref{sec:reu}) is a combination of the \gls{ReLU} and \gls{swish} \glsxtrlongpl{AF}, the \textit{square REU} (SqREU) \cite{Wuraola2021} is a combination of \gls{ReLU} and \gls{squish}:
\begin{equation}
    f(z)=\begin{cases}
            z, \quad & z > 0, \\
            z + \frac{z^2}{2},  \quad & -2 \leq z \leq 0, \\
            0, \quad & z < -2. \\
        \end{cases}
\end{equation}

\subsubsection{Square softplus (SqSoftplus)}
\label{sec:sqsoftplus}
A \gls{SqSoftplus} is another square-based computationally efficient replacement of an \glsxtrlong{AF} --- \gls{softplus} \cite{Wuraola2021}:
\begin{equation}
    f(z)=\begin{cases}
            z, \quad & z > \frac{1}{2}, \\
            z + \frac{\left(z+\frac{1}{2}\right)^2}{2},  \quad & -\frac{1}{2} \leq z \leq \frac{1}{2}, \\
            0, \quad & z < \frac{1}{2}. \\
        \end{cases}
\end{equation}

\subsubsection{Square logistic sigmoid (LogSQNL)}
\label{sec:logsqnl}
While the \gls{SQNL} \cite{Wuraola2018} replaces the \gls{tanh} AF, the \gls{LogSQNL} \cite{Wuraola2021} is a square-based replacement for the \gls{logisticsigmoid}:
\begin{equation}
    f(z)=\begin{cases}
            1, \quad & z > 2, \\
            \frac{1}{2}\left(z - \frac{z^2}{4}\right) + \frac{1}{2},  \quad & 0 \leq z \leq 2, \\
            \frac{1}{2}\left(z + \frac{z^2}{4}\right) + \frac{1}{2},  \quad & -2 \leq z < 0, \\
            0, \quad & z < -2. \\
        \end{cases}
\end{equation}

\subsubsection{Square softmax (SQMAX)}
\label{sec:sqmax}
The \gls{SQMAX} is a square-based replacement for the \gls{softmax}, which is exponential-based. It is defined as 
\begin{equation}
    f(z_j) = \frac{\left(z_j + c\right)^2}{\sum_{k=1}^N\left(z_k+c\right)^2},
\end{equation}
where $f(z_j)$ is the output of a neuron $j$ in a \gls{softmax} layer consisting of $N$ neurons and $c = 4$ is a predefined constant \cite{Wuraola2021}.

\subsubsection{Linear quadratic activation}
\label{sec:linq}
Another square-based \gls{AF} called \gls[prereset]{LinQ} was proposed in \cite{Bilski2016}.
\begin{equation}
    f(z)=\begin{cases}
            az + \left(1-2z+z^2\right), \quad & z \geq 2 - 2a, \\
            \frac{1}{4}z\left(4-\left|z\right|\right), \quad & -2 + 2a < z < 2 - 2a, \\
            az - \left(1-2z+z^2\right), \quad & z \leq -2 + 2a, \\
        \end{cases}
\end{equation}
where $a$ is a fixed parameter controlling the slope of the function's linear parts \cite{Bilski2016}.

\subsubsection{Inverse square root linear unit (ISRLU)}
\label{sec:isrlu}
\Gls{ISRLU} \cite{Carlile2017} is an \glsxtrlong{AF} similar to the \gls{ELU} (see \cref{sec:elu}). It has similar properties and a shape as \gls{ELU}; however, it is faster to compute, leading to more efficient training and inference \cite{Carlile2017}. It is defined as
\begin{equation}
    f(z)=\begin{cases}
            z, \quad & z \geq 0, \\
            z\cdot\frac{1}{\sqrt{1+az^2}}, \quad & z < 0, \\
        \end{cases}
\end{equation}
where $a$ is a hyperparameter controlling the value to which the \gls{ISRLU} saturates for negative inputs \cite{Carlile2017}. While the authors state that the hyperparameter $a$ could be trainable for each neuron $i$, only the non-trainable variant was analyzed \cite{Carlile2017}. \citeauthor{Carlile2017} analysed \gls{ISRLU} with $a=1$ and $a=3$ \cite{Carlile2017}.

\subsubsection{Inverse square root unit (ISRU)}
\label{sec:isru}
\Gls{ISRU} \cite{Carlile2017} is an \glsxtrlong{AF} meant to replace sigmoidal \glsxtrlongpl{AF}. It is defined as
\begin{equation}
    f(z) = z\cdot\frac{1}{\sqrt{1+az^2}},
\end{equation}
where $a$ si a fixed hyperparameter controlling the saturation values; the parameter could be trainable similarly as in the \gls{ISRLU} (see \cref{sec:isrlu}) but only the nonadaptive variant was used \cite{Carlile2017}.

\subsubsection{Modified Elliott function (MEF)}
\label{sec:mef}
The \gls[prereset]{MEF} \cite{Burhani2015} is an \gls{AF} inspired by the \gls{Elliott} function (see \cref{sec:elliott}); it can also be considered to be a translated special case of the \gls{ISRU} (see \cref{sec:isru}) with $a=1$. It is defined as
\begin{equation}
    f(z) = z\cdot\frac{1}{\sqrt{1+z^2}} + \frac{1}{2}.
\end{equation}

\subsection{Square-root-based activation function (SQRT)}
\label{sec:sqrt}
A \gls{SQRT} is a monotonically increasing, unbounded \glsxtrlong{AF} proposed in \cite{Yang2018Square} with a similar structure as the earlier proposed \gls{LAF} but with the square root function instead of the natural logarithm used in \gls{LAF}. It is defined as
\begin{equation}
    f(z)=\begin{cases}
        \sqrt{z}, \quad & z \geq 0, \\
        -\sqrt{-z}, \quad & z < 0. \\
    \end{cases}
\end{equation}
The \gls{SQRT} \glsxtrlong{AF} was found to outperform both \gls{tanh} and \gls{ReLU} \glsxtrlongpl{AF} on the CIFAR-10 dataset \cite{Krizhevsky2009} in experiments in \cite{Yang2018Square}.

A parametric variant of the \gls{SQRT} called \gls[prereset]{SSAF} was proposed independently in \cite{Khachumov2023}. It is defined as
\begin{equation}
    f(z)=\begin{cases}
        \sqrt{2az}, \quad & z \geq 0, \\
        -\sqrt{-2az}, \quad & z < 0, \\
    \end{cases}
\end{equation}
where $a$ is a fixed parameter \cite{Khachumov2023}.

\subsection{Bent identity}
\label{sec:bent_identity}
The \gls{bent_identity} \cite{MateLabs2017} is an \gls{AF} approximating the \gls{ReLU}; it can be seen as a fixed variant of the \gls{BLU} (see \cref{sec:blu}) with $a_i=\frac{1}{2}$. It is defined as
\begin{equation}
    f(z) = \frac{\sqrt{z^2+1} - 1}{2}+ z.
\end{equation}

\subsection{Mishra activation function}
\label{sec:mishra}
The \gls{Mishra}\footnote{The \gls{AF} was unnamed in the original papers \cite{Mishra2021ANew, Mishra2021ANonMonotonic}; however, the work \cite{Hurley2023} named it using the name of the original author. We keep the naming in this work.} \gls{AF} is defined as
\begin{equation}
    f(z) = \frac{1}{2}\left(\frac{z}{1+\left|z\right|}\right)^2+\frac{1}{2}\frac{z}{1+\left|z\right|}.
\end{equation}

\subsection{Saha-Bora activation function (SBAF)}
\label{sec:sbaf}
A \gls{SBAF} was proposed in \cite{Saha2018, Saha2020} to be used for the habitability classification of exoplanets. It employs two non-trainable parameters $\alpha$ and $k$, which were set to $k = 0.98$ and  $\alpha = 0.5$,   where authors determined a stable fixed point. It is defined as:
\begin{equation}
    f(z) = \frac{1}{1 + kz^\alpha(1-z)^{(1-\alpha)}}.
\end{equation}

\subsection{Logarithmic activation function}
\label{sec:laf}
The \gls[prereset]{LAF} was proposed in \cite{Bilski2000} (ref. from \cite{Kamruzzaman2002}). According to \cite{Kamruzzaman2002}, it is defined as
\begin{equation}
    f(z)=\begin{cases}
        \ln{z+1}, \quad & z \geq 0, \\
        -\ln{-z+1}, \quad & z < 0. \\
    \end{cases}
\end{equation}
The \gls{LAF} was independently proposed under the name \gls{symlog} in \cite{Hafner2023}.

\subsection{Symexp}
\label{sec:symexp}
The \gls{symexp} \cite{Hafner2023} is an activation function that is inverse of the \gls{LAU}. It is defined as
\begin{equation}
    f(z) = \mathrm{sgn}\left(z\right)\left(\exp\left(\left|z\right|\right)-1\right).
\end{equation}

\subsection{Scaled polynomial constant unit (SPOCU)}
\label{sec:spocu}
The \gls[prereset]{SPOCU} is a polynomial-based \gls{AF} proposed in \cite{Kisek2020,Kisek2020Correction}. It is defined as
\begin{equation}
    f(z) = a h\left(\frac{z}{c}+b\right)-a h\left(b\right),
\end{equation}
where
\begin{equation}
    h(x)=\begin{cases}
        r(d), \quad & x \geq d, \\
        r(x), \quad & 0 \leq x < d, \\
        x, \quad & x < 0, \\
    \end{cases}
\end{equation}
\begin{equation}
    r(x) = x^3\left(x^5-2x^4+2\right),
\end{equation}
and $a>0$, $b \in (0,1)$, $c>0$, and $d \in [1,\infty)$ are fixed parameters satisfying additional conditions listed  in \cite{Kisek2020,Kisek2020Correction}.

\subsection{Polynomial universal activation function (PUAF)}
\label{sec:puaf}
Similarly as the \gls{UAF} (see \cref{sec:uaf}), the \gls[prereset]{PUAF}\footnote{\Citeauthor{Hwang2023} named the function only as the \textit{universal activation function} but this name is already taken by the \gls{UAF} by \citeauthor{Yuen2021} from \cite{Yuen2021}.} is able to approximate popular \glspl{AF} such as the \gls{logisticsigmoid}, \gls{ReLU}, and \gls{swish} \cite{Hwang2023}. It is defined as

\begin{equation}
    f(z)=\begin{cases}
        z^a, \quad & z > c,\\
        z^a \frac{\left(c+z\right)^{b}}{\left(c+z\right)^{b}+\left(c-z\right)^{b}}, \quad & \left|z\right| \leq c,\\
        0, \quad & z < -c,\\
    \end{cases}
\end{equation}

where $a$, $b$ and $c$ are fixed parameters \cite{Hwang2023}. The \gls{PUAF} becomes the \gls{ReLU} with $a=1$, $b=0$, and $c=0$; the \gls{logisticsigmoid} is approximated with $a=0$, $b=5$, and $c=10$; finally, the \gls{swish} is approximated using $a=1$, $b=5$, and $c=10$ \cite{Hwang2023}.

\subsection{Softplus}
\label{sec:softplus}
The \gls{softplus} function was proposed in \cite{Dugas2000} and is defined as
\begin{equation}
    f(z) = \ln\left(\exp\left(z\right) + 1 \right).
\end{equation}
The \gls{softplus} was used as an \glsxtrlong{AF} in \cite{Glorot2011} where it was used alongside with a \gls{ReLU}. The advantage of \gls{softplus} over \gls{ReLU} is that it is smooth and it has a non-zero gradient for negative inputs; thus, it does not suffer from the phenomenon of dying out neurons that is common in networks with \gls{ReLU} activations \cite{HaoZheng2015}. The \gls{softplus} was found to outperform \gls{ReLU} for certain applications and architectures \cite{HaoZheng2015}. A noisy variant was used for spiking \glsxtrlongpl{NN} in \cite{Liu2016}.

\subsection{Parametric softplus (PSoftplus)}
\label{sec:psoftplus}
\Gls{PSoftplus} \cite{Sun2019} is a \gls{softplus} variant that allows for scaling and shifting using two additional parameters. The \gls{PSoftplus} is defined as
\begin{equation}
    f(z) = a\left(\ln\left(\exp\left(z\right) + 1 \right)-b\right),
\end{equation}
where $a$ and $b$ are fixed predetermined hyperparameters \cite{Sun2019}. The creation of the \gls{softplus} was motivated by the assumption that activations with mean outputs close to zero can improve the performance of a \glsxtrlong{NN}; since the output of the \gls{softplus} is always positive, a shift parameter $b$ was introduced to shift the mean output closer to zero \cite{Sun2019}. The slope controlling parameter $a$ is used to adjust the function and the gradient disappearance or overflow during training \cite{Sun2019}. The recommended values are $a=1.5$ and $b=\ln(2)$ \cite{Sun2019}.

\subsubsection{Soft++}
\label{sec:softpp}
Another \gls{softplus} extension \textit{Soft++} is a multiparametric nonsaturating nonlinear \glsxtrlong{AF} proposed in \cite{Ciuparu2020}. It is defined as
\begin{equation}
    f(z) = \ln\left(1+\exp\left(az\right)\right) + \frac{z}{b} - \ln(2),
\end{equation}
where $a$ and $b$ are fixed predetermined hyperparameters \cite{Ciuparu2020}; however, \citeauthor{Ciuparu2020} proposed they could be adaptable in future works. Multiple values of the parameters were used in the experiments in \cite{Ciuparu2020}, but $a=1$ and $b=2$ were found to work well \cite{Ciuparu2020}; nevertheless, a hyperparameter optimization is recommended \cite{Ciuparu2020}.

\subsection{Rand softplus (RSP)}
\label{sec:rsp}
A \gls{softplus} variant \gls{RSP} \cite{Chen2019} introduces a stochastic parameter $a_l$ that is determined by the noise level of the input data \cite{Chen2019}. The \gls{RSP} is defined as
\begin{equation}
    f(z_l) = (1-a_l)\max\left(0,z_l\right) + a_l\cdot\ln\left(1+\exp\left(z_l\right)\right),
\end{equation}
where $a_l$ is adapting to the input noise levels of each layer $l$ --- the exact procedure is described in \cite{Chen2019}.

\subsection{Aranda-Ordaz}
\label{sec:arandaordaz}
The \gls{Aranda-Ordaz} \gls{AF}\cite{daSGomes2013,ARANDAORDAZ1981}  was used in \glspl{NN} in \cite{daSGomes2013}. It is defined as
\begin{equation}
    f(z) = 1-\left(1+a\exp\left(z\right)\right)^{-\frac{1}{a}},
\end{equation}
where $a>0$ is a fixed parameter \cite{daSGomes2013}. \Citeauthor{EssaiAli2022} used $a=2$ in their work \cite{EssaiAli2022}.

\subsection{Bi-firing activation function (bfire)}
\label{sec:bfire}
A \gls{bfire} was proposed in \cite{Li2013} and is defined as
\begin{equation}
    f(z)=\begin{cases}
    z - \frac{a}{2}, \quad & z > a, \\
    \frac{z^2}{2a}, \quad & -a \geq z, \geq a \\
    -z - \frac{a}{2}, \quad & z < -a, \\
\end{cases}
\end{equation}
where $a$ is a predefined smoothing hyperparameter \cite{Li2013}. The \gls{bfire} is basically a smoothed variant of the later proposed \gls{vReLU} (see \cref{sec:vrelu}) as it becomes \gls{vReLU} as $a \rightarrow 0$.

\subsection{Bounded bi-firing activation function (bbfire)}
\label{sec:bbfire}
A bounded variant of the bi-firing (\gls{bfire}) \glsxtrlong{AF} (see \cref{sec:bfire}) called \glsxtrshort{bbfire} was proposed in \cite{Liew2016}; similarly as \gls{BReLU} and \gls{BLReLU} bounds \gls{ReLU} and \gls{LReLU} respectively (see \cref{sec:brelu, sec:blrelu}), the bounded bi-firing function (bbfire) is defined as
\begin{equation}
    f(z)=\begin{cases}
    b, \quad & z < -b - \frac{a}{2}, \\
    -z - \frac{a}{2}, \quad & -b -\frac{a}{2} \geq z < -a, \\
    \frac{z^2}{2a}, \quad & -a \geq z \geq a, \\
    z - \frac{a}{2}, \quad & a < z \geq b + \frac{a}{2}, \\
    b, \quad & z > b + \frac{a}{2}, \\
\end{cases}
\end{equation}
where $a$ and $b$ are predefined hyperparameters \cite{Liew2016}. The is symmetrical about the origin and has a near inverse-bell-shaped activation curve \cite{Liew2016}. While authors of the original \gls{bfire} \cite{Li2013} solved potential numerical instabilities caused by the unboundedness by imposing a small $L_1$ penalty on the hidden activation values \cite{Li2013}, the \glsxtrshort{bbfire} alleviates this problem explicitly without any need for such penalty.

\subsection{Piecewise Mexican-hat activation function (PMAF)}
\label{sec:pmaf}
The \gls[prereset]{PMAF} was used in \cite{Liu2016Multistability}; it is defined as
\begin{equation}
    f(z)=\begin{cases}
    \left(\frac{2}{\sqrt{3}}\pi^{-\frac{1}{4}}\right)\left(1-\left(z+a\right)^2\right)\exp\left(-\frac{\left(z+a\right)^2}{2}\right), \quad & z < 0, \\
    \left(\frac{2}{\sqrt{3}}\pi^{-\frac{1}{4}}\right)\left(1-\left(z-a\right)^2\right)\exp\left(-\frac{\left(z-a\right)^2}{2}\right), \quad & z \geq 0, \\
    \end{cases}
\end{equation}
where $a$ is a fixed parameter --- \citeauthor{Liu2016Multistability} used $a=4$ \cite{Liu2016Multistability}.

\subsection{Piecewise radial basis function (PRBF)}
\label{sec:prbf}
The \gls[prereset]{PRBF} was used in \cite{Liu2016Multistability}; it is defined as
\begin{equation}
    f(z)=\begin{cases}
        \exp\left(-\frac{\left(z-2a\right)^2}{b^2}\right), \quad & z \geq a\\
        \exp\left(-\frac{z^2}{b^2}\right), \quad & -a < z < a\\
        \exp\left(-\frac{\left(z+2a\right)^2}{b^2}\right), \quad & z \leq -a\\
\end{cases}
\end{equation}
where $a$ and $b$ are fixed parameters \cite{Liu2016Multistability} --- \citeauthor{Liu2016Multistability} used $a=3$ and $b=1$ \cite{Liu2016Multistability}.

\subsection{Comb-H-sine}
\label{sec:combhsine}
A \gls{comb-H-sine} is an \glsxtrlong{AF} that was found using an evolutionary approach in \cite{Vijayaprabakaran2022}. It is defined as
\begin{equation}
    f(z) = \sinh\left(az\right) + \sinh^{-1}\left(az\right),
\end{equation}
where $\sinh(x)$ is the hyperbolic sine, $\sinh^{-1}(x)$ is its inverse, and $a$ is a predefined hyperparameter \cite{Vijayaprabakaran2022}. This function was found to outperform \gls{ReLU}, \gls{tanh}, \gls{logisticsigmoid}, and several other \glsxtrlongpl{AF} in LSTM models in \cite{Vijayaprabakaran2022}.

\subsection{Modified arcsinh}
\label{sec:marcsinh}
The \gls[prereset]{m-arcsinh} \gls{AF} was proposed in \cite{Parisi2020} and is defined as
\begin{equation}
    f(z) = \frac{1}{12}\sinh^{-1}\left(z\right)\sqrt{\left|z\right|}.
\end{equation}
Interestingly, the \gls{m-arcsinh} can be used either as an \gls{AF} in a \gls{NN} or as a kernel function in the \gls{SVM} \cite{Parisi2020}.

\subsection{hyper-sinh}
\label{sec:hypersinh}
The \gls{hyper-sinh} is an \gls{AF} that uses the $\sinh$ and cubic functions \cite{Parisi2021hypersinh,Parisi2021}; it is defined as
\begin{equation}
    f(z) = \begin{cases}
        \frac{\sinh\left(z\right)}{3},  \quad & z > 0, \\
        \frac{z^3}{4},  \quad & z \leq 0. \\
    \end{cases}
\end{equation}

\subsection{Arctid}
\label{sec:arctid}
The \gls{arctid} is an \gls{arctan}-based \gls{AF} used in \cite{Xu2020Comparison}; it is defined as
\begin{equation}
    f(z) = \left(\tan^{-1}\left(z\right)\right)^2 - z.
\end{equation}

\subsection{Sine}
\label{sec:sine}
The sine with inputs scaled by $\pi$ was used as an activation in \cite{Hara1994}:
\begin{equation}
    f(z) = \sin\left(\pi z\right).
\end{equation}
It was, for example, used recently with a data-driven determination of a network's biases in \cite{Dudek2021}. Just the sine function without any scaling was used as an activation in \cite{Gashler2014, Zhang2014, Sopena1999, Faroughi2023, Sitzmann2020, Cheng2021}.

Scaled sine with vertical shift was used in \cite{Pan2020}; the used \gls{AF} is defined as
\begin{equation}
    f(z) = 0.5 \sin\left(a z\right)  + 0.3,
\end{equation}
where $a$ is a fixed parameter; $a \in \{0.2, 0.8, 1.2, 1.8, 4\}$ \cite{Pan2020}.

\subsection{Cosine}
\label{sec:cosine}
A cosine activation was used in simulations in \cite{Daskin2018}; it was defined as
\begin{equation}
    f(z) = 1 - \cos\left(z\right).
\end{equation}

\subsection{Cosid}
\label{sec:cosid}
The \gls{cosid} is one of the \glspl{AF} listed in \cite{Xu2020Comparison}. It is defined as
\begin{equation}
    f(z) = \cos\left(z\right) - z.
\end{equation}

\subsection{Sinp}
\label{sec:sinp}
A parametric \gls{AF} similar to the \gls{cosid} was proposed in \cite{Chan2018} under the name \gls{sinp}.\footnote{Technically, the full name used by \citeauthor{Chan2018} is $\mathrm{SinP}[N]$ but we ommited the parameter from the name of the \gls{AF}.} It is defined as
\begin{equation}
    f(z) = \sin\left(z\right) - az,
\end{equation}
where $a$ is a fixed parameter \cite{Chan2018}. \Citeauthor{Chan2018} used $a \in \{1, 1.5, 2\}$ \cite{Chan2018}.

\subsection{Growing cosine unit (GCU)}
\label{sec:gcu}
Another cosine-based \gls{AF} is the \gls[prereset]{GCU} proposed in \cite{Noel2021Growing}. It is defined as
\begin{equation}
    f(z) = z \cos\left(z\right).
\end{equation}
Empirical evaluation of the performance of \gls{GCU} compared to \gls{ReLU}, \gls{PReLU}, and \gls{mish} is available in \cite{Sharma2022}; its brief evaluation with respect to the generation of \glsxtrshortpl{NFT} is available in \cite{Sharma2022NFT}.

\subsection{Amplifying sine unit (ASU)}
\label{sec:asu}
The \gls[prereset]{ASU} is the sine equivalent of the \gls{GCU} \cite{Rahman2023Amplifying,Rahman2023ASUCNN}
\begin{equation}
    f(z) = z \sin\left(z\right).
\end{equation}

\subsection{Sinc}
\label{sec:sinc}
The \gls{sinc} is an older \gls{AF} proposed in \cite{Efe2008}. It is defined as
\begin{equation}
    f(z)=\begin{cases}
                \frac{\sin\left(\pi z\right)}{\pi z}, \quad & z \neq 0, \\
                1,  \quad & z = 1. \\
            \end{cases}
\end{equation}

A shifted variant was proposed under the name \gls[prereset]{SSU} in \cite{Gustineli2022}. It is defined as
\begin{equation}
    f(z) = \pi \mathrm{sinc}\left(z-\pi\right).
\end{equation}

\subsection{Decaying sine unit (DSU)}
\label{sec:dsu}
The \gls[prereset]{DSU} is a \gls{sinc} based \gls{AF} proposed in \cite{Gustineli2022}. It is defined as
\begin{equation}
    f(z) = \frac{\pi}{2}\left(\mathrm{sinc}\left(z-\pi\right)-\mathrm{sinc}\left(z+\pi\right)\right).
\end{equation}

\subsection{Hyperbolic cosine linearized squashing function (HcLSH)}
\label{sec:hclsh}
The \gls[prereset]{HcLSH} is an \gls{AF} proposed in \cite{AbdelNabi2023}; it is defined as
\begin{equation}
    f(z)=\begin{cases}
                \ln\left(\cosh\left(z\right) + z\cdot\cosh\left(\frac{z}{2}\right)\right), \quad & z \geq 0, \\
                \ln\left(\cosh\left(z\right)\right)+z,  \quad & z < 0. \\
            \end{cases}
\end{equation}

\subsection{Polyexp}
\label{sec:polyexp}
The \gls{polyexp} is an \gls{AF} combining quadratic function and an exponential function \cite{Efe2008};\footnote{The \cite{Hara1994} is referenced as the origin of \gls{polyexp} in \cite{Efe2008} but we have not seen the definition there.} it is defined as
\begin{equation}
    f(z) = \left(az^2+bz +c\right)\exp\left(-dz^2\right),
\end{equation}
where $a$, $b$, $c$, and $d$ are fixed parameters \cite{Efe2008}.

\subsection{Exponential}
\label{sec:exponential}
The exponential was used as an \gls{AF} in \cite{Pan2020}. The \gls{AF} was defined as 
\begin{equation}
    f(z) = \exp\left(-z\right).
\end{equation}

\subsection{E-Tanh}
\label{sec:etanh}
An \gls{AF} named \gls{E-Tanh} combining the exponential and \gls{tanh} functions was proposed in \cite{Kalaiselvi2022}. It is defined as
\begin{equation}
    f(z) = a \cdot \exp\left(z\right)\tanh\left(z\right),
\end{equation}
where $a$ is a fixed scaling parameter \cite{Kalaiselvi2022,AfshariNia2023}.

\subsubsection{Evolved combination of tanh and ReLU}
\label{sec:evolved_tanh_ReLU}
The combination of \gls{tanh} and \gls{ReLU} was found using neuroevolution in \cite{Vijayaprabakaran2021} --- while \citeauthor{Vijayaprabakaran2021} also mentioned other \glspl{AF}, this combination led to the best performance on the HAR dataset using the \gls{LSTM} units. The best-performing recurrent \gls{AF} was
\begin{equation}
    f(z) = a \left(\tanh\left(z^2\right) + \mathrm{ReLU}\left(z\right)\right)
\end{equation}
and the regular \gls{AF} was
\begin{equation}
    f(z) = \max\left(\tanh\left(\log\left(z\right)\right), \mathrm{ReLU}\left(z\right)\right).
\end{equation}
See \cite{Vijayaprabakaran2021} for evaluation details and for other top \glspl{AF}.

\subsection{Wave}
\label{sec:wave}
The \gls{wave} is an \gls{AF} combining quadratic function and an exponential function \cite{Efe2008};\footnote{The \cite{Hara1994} is referenced as the origin of \gls{wave} in \cite{Efe2008} but we have not seen the definition there.} similarly as the \gls{polyexp} but only with a single parameter; it is defined as
\begin{equation}
    f(z) = \left(1-z^2\right)\exp\left(-az^2\right),
\end{equation}
where $a$ is a fixed parameter \cite{Efe2008}.

\subsection{Non-monotonic cubic unit (NCU)}
\label{sec:ncu}
A simple \gls{AF} based on a third-degree polynomial was proposed in \cite{Gustineli2022}. It is named \gls[prereset]{NCU} and is defined as
\begin{equation}
    f(z) = z - z^3.
\end{equation}

\subsection{Triple}
\label{sec:triple}
Another \gls{AF} based on a third-degree polynomial called \gls{triple} was proposed in \cite{Chen2022Target}. It is defined as 
\begin{equation}
    f(z) = a\cdot z^3,
\end{equation}
where $a$ is a fixed parameter \cite{Chen2022Target}. \Citeauthor{Chen2022Target} tested values of $a \in \{0.1, 0.5, 1, 2\}$ and observed that $a=1$ reaches the best results \cite{Chen2022Target}.

\subsection{Shifted quadratic unit (SQU)}
\label{sec:squ}
The \gls[prereset]{SQU} \cite{Gustineli2022} is a simple non-monotonic \gls{AF} defined as
\begin{equation}
    f(z) = z^2 + z.
\end{equation}

\subsection{Knowledge discovery activation function (KDAC)}
\label{sec:kdac}
\Citeauthor{Wang2022Why} proposed a special \gls{AF} for knowledge discovery in \cite{Wang2022Why}. This function named \gls[prereset]{KDAC} has two adaptive parameters $a>0$ and $b>0$ and one fixed parameter $c$. It is defined\footnote{The original code by \citeauthor{Wang2022Why} is available at \url{https://github.com/pyy-copyto/KDAC/blob/main/KDAC.py}.} as
\begin{equation}
    f(z) = p \cdot \left(1-\mathrm{h}_\mathrm{max}\left(p , r\right)\right)+r\cdot \mathrm{h}\left(p , r\right) + k\mathrm{h}_\mathrm{max}\left(p , r\right)\left(1-\mathrm{h}_\mathrm{max}\left(p , r\right)\right),
\end{equation}
where 
\begin{equation}
    \mathrm{h}_\mathrm{max}(x,y) = \mathrm{clip}\left(\frac{1}{2}-\frac{1}{2}\frac{x-y}{c}\right),
\end{equation}
\begin{equation}
    \mathrm{clip}(x)=\begin{cases}
                0,  \quad & x \leq 0, \\
                x, \quad & 0 < x < 1, \\
                1, \quad & x \geq 0, \\
            \end{cases}
\end{equation}
\begin{equation}
    p = az,
\end{equation}
\begin{equation}
    q = \mathrm{h}_\mathrm{min}\left(bz , s\right),
\end{equation}
\begin{equation}
    r = \begin{cases} 
        p,  \quad & z > 0, \\
            bz \cdot \left(1-q\right)+s\cdot \mathrm{h}\left(q , s\right) +  kq\left(1-q\right), \quad & z \leq 0, \\
    \end{cases}
\end{equation}
\begin{equation}
    s = \tanh(z),
\end{equation}
and
\begin{equation}
    \mathrm{h}_\mathrm{min}(x,y) = \mathrm{clip}\left(\frac{1}{2}+\frac{1}{2}\frac{x-y}{c}\right).
\end{equation}
\Citeauthor{Wang2022Why} used fixed $c = 0.01$ \cite{Wang2022Why}.

\subsection{K-winner-takes-all activation function (k-WTA)}
\label{sec:kwta}
The \gls{k-WTA} \gls{AF} was used to improve adversarial robustness in \cite{Xiao2020}. It is defined as
\begin{equation}
    f(\vec{z})_j = \begin{cases}
        z_j,  \quad & z_j \in \left\{ k \text{ largest elements of $\vec{z}$}\right\}, \\
        0, \quad & \text{ otherwise}, \\
    \end{cases}
\end{equation}
where $f\left(\vec{z}\right): \mathbb{R}^N \rightarrow \mathbb{R}^N$ is the \gls{k-WTA} \gls{AF} and $f(\vec{z})_j$ its $j$-th element, $\vec{z}$ is the input to the \gls{AF}, and $k$ a fixed parameter \cite{Xiao2020}.

\subsection{Volatility-based activation function (VBAF)}
\label{sec:vbaf}
The \gls[prereset]{VBAF}\footnote{\Citeauthor{Kayim2022} named the function originally only \textit{volatility \glsxtrlong{AF}}.} is an \gls{AF} with multiple intputs proposed in \cite{Kayim2022}. It is meant for time-series forecasting and was used in a \gls{LSTM} \gls{NN} in \cite{Kayim2022}. It is defined as
\begin{equation}
    f(z_1, \ldots, z_n) = \sqrt{\frac{\sum_{j=1}^n\left(\mean{z}-z_j\right)}{n}},
\end{equation}
where
\begin{equation}
    \mean{z} = \frac{\sum_{j=1}^nz_j}{n},
\end{equation}
$n$ is the number of time-series samples in the given period \cite{Kayim2022, Xin2023}. Unfortunately, no more details about the application of the \gls{VBAF} were provided in \cite{Kayim2022}; thus, it remains unclear whether the \gls{VBAF} was applied only directly to the inputs, or it was used on intermediary representations of a \gls{NN}.

\subsection{Chaotic activation functions}
\label{sec:caf}
The \glspl[prereset]{CAF} listed in this work are \glspl{AF} that use a recursive definition to produce a chaotic behavior. 

\subsubsection{Hybrid chaotic activation function}
\label{sec:hcaf}
The \gls[prereset]{HCAF} is a multi-output type of \gls{AF} proposed in \cite{Reid2021}. Neuron $i$ in layer $l$ takes an input $z_i^l$, applies the \gls{logisticsigmoid} \gls{AF} and then maps the outputs using logistic map function to individual outputs going to the neurons in layer $l+1$ \cite{Reid2021}. Therefore, a single neuron in layer $l$ emits a different activation value to each neuron in the layer $l+1$ \cite{Reid2021}.

For a neuron $i$, the \gls{HCAF} first applies the \gls{logisticsigmoid} function to produce activation $a_i$
\begin{equation}
    a_i = f(z_i) = \sigma\left(z_i\right),
\end{equation}
then the first value going to the neuron $1$ in the following layer is calculated as
\begin{equation}
    c_{i,1} = r a_i \left(1-a_i\right),
\end{equation}
 and the output values going to the other neurons in the following layer are calculated recursively as
 \begin{equation}
    c_{i,j} = r c_{i,{j-1}} \left(1-c_{i,{j-1}}\right),
\end{equation}
where $j$ is the number of a neuron in a following layer and $r$ represents an excitatory rate in a neuron \cite{Reid2021}. \Citeauthor{Reid2021} used $r=4$ as this value produces a chaotic behavior of the logistic map \cite{Reid2021}; generally, values below 0 or above 4 lead to the output to become unbounded, values between 0 and 1 lead to convergence toward the zero, values between 1 and 3 lead to convergence to a fixed number, values between 3 and 3.5 lead to a periodic solution and only values between 3.5 and 4 produce chaotic behavior \cite{Reid2021}.

\subsubsection{Fusion of chaotic activation function (FCAF)}
\label{sec:fcaf}
Similarly as \gls{HCAF}, also the \gls[prereset]{FCAF} \cite{Kabir2012} uses a recursive definition for computing the output of a neuron. The \gls{FCAF} is defined\footnote{\Citeauthor{Kabir2012} did not explicitly defined what the index $i$ denotes but most likely it denotes the $i$-th neuron in a given layer.} for hidden units as\footnote{The formula given in \cite{Kabir2012} probably missed a minus sign after the parameter $a$.}
\begin{equation}
    f(z_{i+1}) = r z_i\left(1-z_i\right) + z_i + a-\frac{b}{2\pi}\sin\left(2\pi z_i\right)
\end{equation}

and for the output units as
\begin{equation}
    f(z_{i+1}) = r z_i\left(1-z_i\right) + z_i + a-\frac{b}{2\pi}\sin\left(2\pi z_i\right) +\exp\left(-cz_i^2\right) +d,
\end{equation}
where $r$, $a$, $b$, $c$, and $d$ are fixed parameters \cite{Kabir2012}; the suitable values for the parameter $r$ are discussed in \cref{sec:hcaf} where an equivalent parameter is used.

\subsubsection{Cascade chaotic activation function (CCAF)}
\label{sec:ccaf}
The \gls[prereset]{CCAF} was introduced in \cite{Abbasi2022} and is recursively defined for the neuron $i+1$ in a given layer using the preceding neuron $i$ from the same layer as 
\begin{equation}
    f(z_{i+1}) = a\cdot\sin\left(\pi\cdot b \cdot\sin\left(\pi z_i\right)\right),
\end{equation}
where $a$ and $b$ are two fixed parameter from the interval $[0,1]$ \cite{Abbasi2022}.

\section{Adaptive activation functions}
\label{sec:adaptive_activation_functions}

The \glsxtrlong{AF} introduces non-linearities to \glsxtrlongpl{NN} and is crucial for \glslink{NN}{network}'s performance \cite{Goodfellow2016}. Even though it might be suboptimal, the same \glsxtrlong{AF} is usually used for the whole \glslink{NN}{network} or at least for all neurons in a single layer.
Over the last few decades, there have been several attempts to use \glsxtrlongpl{AF} that might differ across neurons (e.g., \cite{Wu1997, Wu2001, Flennerhag2018, Vecci1998, He2015}). The \glsxtrlongpl{AAF} --- i.e., functions that have a trainable parameter that changes their shape --- have been receiving more attention recently (e.g., \cite{He2015, Dushkoff2016, Hou2017, Scardapane2018}) and might become a new standard in the field. One of the first descriptions of the general \gls{AAF} approach is available in \cite{Wu1997} where \citeauthor{Wu1997} described an \gls{AF}\footnote{The authors used the name \gls[prereset]{TAF} rather than the \gls[prereset]{AAF} that is used throughout this work.} that has one or more trainable parameters that are trained together with the rest of the network's weights \cite{Wu1997}. The simplest forms just add a parameter to a particular \glsxtrlong{NN} that controls one of its properties (e.g., slope), while the more complex ones allow for the learning of a large number of \glsxtrlongpl{AF} (e.g., adaptive spline \glsxtrlongpl{AF} in \cite{Vecci1998}).

\subsection{Transformative adaptive activation function (TAAF)}
The \gls[prereset]{TAAF} \cite{Kunc2021, Kunc2020} is a family of \glspl{AF} that adds four adaptive parameters for scaling and translation of any given \gls{AF} --- as such, the \glspl{TAAF} represent a simple framework with a small set of additional parameters that generalizes a lot of \glspl{AAF} that are listed in this work. While there are even more general approaches such as the \gls{ABU} (see \cref{sec:abu}) that allows for a combination of several different activation functions, the \glspl{TAAF} are conceptually simpler and add only four additional parameters. The \gls{TAAF} is defined as
\begin{equation}
    \label{eq:taaf}
    g(f,z_i) = \alpha_i \cdot f(\beta_i\cdot z_i + \gamma_i) +\delta_i,
\end{equation}
where $z_i$ input to the \gls{AF}, $\alpha_i$, $\beta_i$, $\gamma_i$, and $\delta_i$, are adaptive parameters for each neuron $i$ \cite{Kunc2021}. Therefore, the output of a neuron with \gls{TAAF} with inputs $x_i$ is:
\begin{equation}
   \alpha_i \cdot f\left(\beta_i\cdot \sum_{i=1}^nw_ix_i + \gamma_i\right) +\delta_i,
\end{equation}
where $x_i$ are individual inputs, $w_i$ are its weights, and $n$ is the number of incoming connections \cite{Kunc2021}. If there is no unit $x_i$, then the parameter $\gamma$ is equivalent to the bias term of the neuron \cite{Kunc2021}. It was shown in \cite{Kunc2021} that each of the four adaptive parameters statistically significantly improve the performance of the \gls{AF} --- this is not surprising as many of the \glspl{AF} presented in this work use subset of these parameters. For example, equivalent of $\alpha$ is used in the \gls{PPReLU} (see \cref{sec:pprelu}), equivalent of $\beta$ in the \gls{swish} (see \cref{sec:swish}), and equivalent of $\gamma$ and $\delta$ in the \gls{FReLU} (see \cref{sec:frelu}).

\subsection{The ReLU-based family of adaptive functions}
The are numerous \gls{ReLU} extensions that are adaptive \cite{Dubey2022}. Some of the adaptive activations have a non-adaptive counterpart --- e.g.,  \gls{PReLU} (see \cref{sec:prelu}), which is basically a \gls{LReLU} with an adaptive parameter of leakiness.

\subsubsection{Parametric rectified linear unit (PReLU)}
\label{sec:prelu}
However, \glspl{AAF} might be very useful even in the simplest form with a single added parameter --- an \gls{AAF} called \gls{PReLU} was used to obtain a state-of-the-art result on the ImageNet Classification in 2015, the first surpassing human-level performance \cite{He2015}. The \gls{PReLU} generalize the \gls{ReLU} by adding a parameter that controls the slope of the \glsxtrlong{AF} for negative inputs (the \gls{ReLU} is constant at zero for negative inputs) that is learned with other weights \cite{Zhang2018Multiple}: 
\begin{equation}
    f(z_i)=\begin{cases}
                z_i, \quad & z_i \geq 0, \\
                \frac{z_i}{a_i},  \quad & z_i < 0, \\
            \end{cases}
\end{equation}
where $a_i$ is an optimized parameter for each neuron/filter $i$.
The \gls{LReLU} \cite{Maas2013} is essentially a \gls{PReLU} but with the parameter $a_i$ fixed and not trainable (see \cref{sec:lrelu} for LReLU details). \Glspl{PReLU} are better than \glspl{ReLU} for verification-friendly \glspl{NN} \cite{Leofante2023}.

\subsubsection{Positive parametric rectified linear unit (PReLU\textsuperscript{+})}
\label{sec:pprelu}
The \gls[prereset]{PPReLU} is an adaptive variant of the \gls{SlReLU} (see \cref{sec:slrelu}) proposed in \cite{Dai2022}; it is also a special case of, for example, \gls{DPReLU}, \gls{dual_line}, and \gls{PiLU}. It is defined as
\begin{equation}
    f(z_i)=\begin{cases}
                a_iz_i, \quad & z_i \geq 0, \\
                0,  \quad & z_i < 0, \\
            \end{cases}
    \end{equation}
where $a_i$ is a trainable parameter \cite{Dai2022}.

\subsubsection{Margin Relu}
\label{sec:marrelu}
The \gls[prereset]{MarReLU}\footnote{\Citeauthor{Heo2019} abbreviated it as MReLU, but this abbreviation is used for the \glsxtrlong{mReLU} (see \cref{sec:mrelu}) in this work.} is an adaptive variant of the \gls{shifted_relu} where the shift $a_i$ is determined as the channel-by-channel expectation value of the negative response \cite{Heo2019}. It is defined as
\begin{equation}
    f(z_i)=\max\left(z, a_i\right) = \begin{cases}
                z, \quad & z_i - a_i\geq 0, \\
                a_i,  \quad & z_i - a_i < 0. \\
            \end{cases}
    \end{equation}

\subsubsection{Funnel parametric rectified linear unit (FunPReLU)}
\label{sec:funprelu}
The \gls[prereset]{FunReLU}\footnote{\Citeauthor{Ma2020Funnel} originally named the unit \glsxtrshort{FReLU} but its abbreviation would collide with the \glsxtrlong{FReLU}.} and \gls[prereset]{FunPReLU} are 2D \glspl{AF} proposed in \cite{Ma2020Funnel}. The \gls{FunReLU} and \gls{FunPReLU} introduce a spatial context into the \gls{AF} by comparing the input to a funnel condition instead of the zero that is used as the threshold in \gls{ReLU} and \gls{PReLU} \cite{Ma2020Funnel}.
The \gls{FunReLU} is defined as 
\begin{equation}
    f(z_{c,m,n}) = \max\left(z_{c,m,n}, \mathrm{t}\left(z_{c,m,n}\right)\right),
\end{equation}
where $z_{c,m,n}$ is the input on the $c$-th channel at the 2D spatial position $m,n$ and $\mathrm{t}\left(z_{c,m,n}\right)$ is the spatial context from a $3 \times 3$ window\footnote{Other sizes were also tested in \cite{Ma2020Funnel} but \citeauthor{Ma2020Funnel} found $3 \times 3$  to work the best.}
\begin{equation}
    \mathrm{t}\left(z_{c,m,n}\right) = \sum_{m-1\leq h \leq m +1, n-1\leq w \leq n +1} z_{c,h,w} \cdot p_{c,h,w}
\end{equation}
and $p_{c,h,w}$ denotes the coefficients on this window \cite{Ma2020Funnel}. The \gls{FunPReLU} is defined similarly. The \gls{FunReLU} was, for example, used in \cite{Gao2020SelfSupervised, Bogoi2022}.

\subsubsection{React-PReLU (RPReLU)}
\label{sec:rprelu}
The \gls[prereset]{RPReLU} is an adaptive variant of the \gls{PReLU} with vertical and horizontal shifts; it is defined as
\begin{equation}
    f(z_i)=\begin{cases}
                z_i-a_c+b_c, \quad & z_i \geq a_c, \\
                c_c\left(z_i-a_c\right)+b_c,  \quad & z_i < a_c, \\
            \end{cases}
    \end{equation}
where $a_c$, $b_c$, and $c_c$ are trainable parameters for each channel $c$ and $z_i$ denotes the input to the neuron $i$ in the channel $c$ \cite{Liu2020ReActNet}; $a_c$ controls the horizontal shift, $b_c$ controls the vertical shift, and $c_c$ is the slope parameter for negative inputs as in the original \gls{PReLU}.

\subsubsection{Smooth activation unit (SAU)}
\label{sec:sau}
The \gls[prereset]{SAU} is a smoothed variant of the \gls{PReLU}\footnote{The principle could be, however, applied to other \glspl{AF}.} using the convolution operation with the Gaussian function \cite{Biswas2022SAU}. It is defined as
\begin{equation}
 f(z_i) = \left(\mathrm{PReLU}_{a_i} * \phi_{b_i}\right)(z_i),
\end{equation}
where $*$ is the convolution operation, $\mathrm{PReLU}_{a_i}$ is the \gls{PReLU}\footnote{\Citeauthor{Biswas2022SAU} used the \gls{LReLU} in the definition of the \gls{SAU} but since they consider the parameter $a_i$ trainable, we stick to the usage of \gls{PReLU} in the defintion.} parametrized by $a_i$\footnote{To conform to the used definition of the \gls{PReLU} unit, we will use the slope scaling by $\frac{1}{a_i}$ even though authors originally used the $a_i$ for slope scaling of negative inputs.} and $\phi_{b_i}(x)$ is the Gaussian function parameterized by $b_i$ inversely controlling the deviation of the function \cite{Biswas2022SAU}. The resulting \gls{AF} is then
\begin{equation}
    f(z_i) = \frac{1}{2b_i}\sqrt{\frac{2}{\pi}}\exp\left(-\frac{b_i^2z_i^2}{2}\right)+\frac{1+\frac{1}{a_i}}{2}z_i+\frac{1-\frac{1}{a_i}}{2}z_i\cdot\mathrm{erf}\left(\frac{b_iz_i}{\sqrt{2}}\right),
\end{equation}
where $a_i$ and $b_i$ are either fixed or trainable parameters \cite{Biswas2022SAU}.

\subsubsection{Smooth maximum unit (SMU)}
\label{sec:smu}
The \gls[prereset]{SMU} \cite{Biswas2022Smooth} is an \gls{AAF} that uses a smooth approximation of the absolute value function. The \gls{SMU} is defined as
\begin{equation}
    f(z_i) = \frac{\left(1+a_i\right)z_i + \left(1-a_i\right)z_i\operatorname{erf}\left(b_i\left(1-a_i\right)z_i\right)}{2}
\end{equation}
where $a_i$ and $b_i$ are learnable parameters \cite{Biswas2022Smooth}. This smooth approximation of the absolute value function using the Gaussian error function could be used to create a whole class of \glspl{AF} similarly as in \cref{sec:msrf}.

\subsubsection{Leaky Learnable ReLU (LeLeLU)}
\label{sec:lelelu}
An adaptive \gls{LReLU} variant named \gls[prereset]{LeLeLU} was proposed in \cite{Maniatopoulos2021}. It is a \gls{LReLU} with learnable scaling parameter:
\begin{equation}
    f(z_i)=\begin{cases}
                a_iz_i, \quad & z_i \geq 0, \\
                0.01a_iz_i,  \quad & z_i < 0, \\
            \end{cases}
\end{equation}
where $a_i$ is a trainable parameter for each neuron $i$ \cite{Maniatopoulos2021}.

\subsubsection{Parametric rectified exponential unit (PREU)}
\label{sec:preu}
Similarly as \gls{PReLU} extends the \gls{ReLU} (see \cref{sec:prelu}), the \gls{PREU} extends the \gls{swish} and \gls{ELU} inspired \gls{REU} \cite{Ying2019}. It is defined as
\begin{equation}
    f(z_i)=\begin{cases}
                a_i z_i, \quad & z_i \geq 0, \\
                a_i z_i\cdot\exp(b_iz_i),  \quad & z_i < 0, \\
            \end{cases}
\end{equation}
where $a_i$ and $b_i$ are trainable parameters for each neuron/filter $i$ \cite{Ying2019}. The advantage of \gls{PREU} is that it uses the negative information near zero \cite{Dubey2022} --- unlike the \gls{ReLU}.

\subsubsection{Randomly translational PReLU (RT-PReLU)}
\label{sec:rtprelu}
A \gls{RT-PReLU} an equivalent extension to \gls{PReLU} as is {\gls{RT-ReLU}} to \gls{ReLU} (see \cref{sec:rtrelu}) \cite{Cao2018Randomly}. It is defined as
\begin{equation}
    f(z_i)=\begin{cases}
                z_i, \quad & z_i + b_i \geq 0, \\
                \frac{z_i}{a_i},  \quad & z_i + b_i < 0, \\
            \end{cases}
\end{equation}
where $a_i$ is a trainable parameter and $b_i$ is a stochastic parameter for each neuron $i$ randomly sampled from the Gaussian distribution at each iteration, $b_i \sim \mathrm{N}\left(0, \sigma^2\right)$, where $\sigma^2$ is the variance of the Gaussian distribution. The offset $b_i$ is set to zero during the test phase \cite{Dubey2022}. The authors \citeauthor{Cao2018Randomly} set the $\sigma^2 = 0.75^2$ for their experiments \cite{Cao2018Randomly}. It is also possible to have the parameter $b_i$ sampled for each neuron $i$, but the $a_i$ is shared by all neurons in a channel $c$ \cite{Kim2020}.

\subsubsection{Probabilistic activation (ProbAct)}
\label{sec:probact}
A probabilistic class of \glsxtrlongpl{AF} \gls{ProbAct} that adds a random noise to any \glsxtrlong{AF} \cite{Shridhar2019}.
It is defined as 
\begin{equation}
    f(z) = g(z) + \sigma e,
\end{equation}
where $g(z)$ is any function (either fixed or trainable) defining the mean of the probabilistic activation, $e\sim \mathrm{N}(0,1)$ is a random value sampled from a standard normal distribution, and $\sigma$ is either fixed or learnable parameter controlling the range of the perturbation \cite{Shridhar2019}. $\sigma$ can be either a global learnable parameter or different for each neuron $i$ \cite{Shridhar2019}. The \gls{ProbAct} used in \cite{Shridhar2019} is a \gls{ReLU} based \gls{ProbAct} defined as
\begin{equation}
    f(z) = \max(0, z) + \sigma e,
\end{equation}
which resembles \gls{NReLU} (see \cref{sec:nrelu}) that adds random noise for output values for the positive inputs $z$ \cite{Nair2010}. A similar concept for sigmoid and \gls{tanh} activation was used in \cite{Gulcehre2016} (see \cref{sec:noisy_af}).

The chaotic injections presented in \cite{Reid2022} represent a similar approach; however, the injections are not stochastic but rather defined using the chaos theory. Furthermore, \citeauthor{Reid2022} discuss several approaches for injections of a  chaotic value $s_n$ into a \gls{ReLU}: $\mathrm{ReLU}\left(z+s_n\right)$,  $\mathrm{ReLU}\left(z\cdot s_n\right)$,  $\mathrm{ReLU}\left(z+z\cdot s_n\right)$,  $\mathrm{ReLU}\left(z\right) + s_n$,  $\mathrm{ReLU}\left(z\right)\cdot s_n$,  $\mathrm{ReLU}\left(z\right) + zs_n$, and  $\mathrm{ReLU}\left(z\right)+\mathrm{ReLU}\left(z\right)\cdot s_n$ \cite{Reid2022}.

\subsubsection{Adaptive offset activation function (AOAF)}
\label{sec:aoaf}
Another \gls{ReLU} variant with adaptive shift termed \gls{AOAF} was defined in \cite{Jiang2022}. The \gls{AOAF} introduces two hyperparameters and one data-dependent adaptive parameter; it is defined as
\begin{equation}
     f(z_i) = \max\left(0, z_i - ba_i\right) + ca_i,
\end{equation}
where $b$ and $c$ are predefined, fixed parameters and $a_i$ is the mean value of the inputs of neuron $i$ \cite{Jiang2022}. The recommended values for the parameters $b$ and $c$ are $b = c = 0.17$ as it yielded the best image classification accuracy in the experiments \cite{Jiang2022}.

\subsubsection{Dynamic leaky ReLU (DLReLU)}
\label{sec:dlrelu}
An error based \gls{DLReLU} was proposed in \cite{Hu2019} (under the name of Dynamic ReLU --- DReLU --- but this naming collides with \gls{DReLU} established in \cite{Si2018, Chen2020Dynamic}; see \cref{sec:drelu} and \cref{sec:dyrelu}). The \gls{DLReLU} is a \gls{LReLU} where the leakiness depends on the test error from the previous epoch \cite{Dubey2022}
\begin{equation}
    f(z)=\begin{cases}
                z, \quad & z \geq 0, \\
                ab_tz,  \quad & z < 0, \\
            \end{cases}
    \end{equation}
where $a \in (0,1)$ is a predefined parameter controlling the leakiness similarly as in \gls{LReLU} (see \cref{sec:lrelu}) and $b_t$ is a dynamic parameter computed for current training epoch $t$ as the test erroch from the previous epoch $t-1$, i.e., $b_t = \mathrm{MSE}_{t-1}$ \cite{Hu2019}.

A version {exp--\gls{DLReLU}} was proposed to deal with deeper networks with more than seven hidden layers in order to avoid too large error values causing the training to fail \cite{Hu2019}:
\begin{equation}
    f(z)=\begin{cases}
                z, \quad & z \geq 0, \\
                ac_tz,  \quad & z < 0, \\
            \end{cases}
    \end{equation}
where $c_t = \exp(-b_t) = \exp\left(\mathrm{MSE}_{t-1}\right)$ \cite{Hu2019}.

The advantage of \gls{DLReLU} and {exp--\gls{DLReLU}} is that the changes in leakiness are big at the beginning of the training due to higher test error and diminish towards the end \cite{Hu2019}. A similar effect could be obtained by a schedule of the leakiness parameter in the \gls{LReLU}.

\subsubsection{Dynamic ReLU (DReLU)}
\label{sec:drelu}
Similar approach to the \gls{ABReLU} is presented by the \gls{DReLU} \cite{Si2018} (not to be confused with identically named activations in \cite{Hu2019, Chen2020Dynamic}),

\begin{equation}
    f(z_i)=\begin{cases}
                z_i, \quad & z_i - a_i \geq 0, \\
                a_i,  \quad & z_i - a_i < 0, \\
            \end{cases}
\end{equation}
where $a_i$ is a threshold value that is computed as the midpoint of the range of input values for each batch; e.g., if the values range from -4 to 8, then $a_i = \frac{-4+8}{2} = 2$ \cite{Si2018}. The \gls{DReLU} can be considered to be a variant of \gls{DisReLU} (see \cref{sec:disrelu}) with data-dependent determination of the shifting point.

\subsubsection{Flexible ReLU (FReLU)}
\label{sec:frelu}
The \gls{FReLU} is a \gls{ReLU} extension with zero-like property and the ability to capture negative information \cite{Qiu2018FReLU}. The zero-like property is the ability to push activation means closer to zero \cite{Qiu2018FReLU} as this might speed up learning \cite{Clevert2015}. The \gls{FReLU} builds on the ability to shift the AF
\begin{equation}
    f(z_i)=\mathrm{ReLU}\left(z_i + a_i\right) + b_i,
\end{equation}
where $a_i$ and $b_i$ would be optimized parameters \cite{Qiu2018FReLU}. However, since the parameter $a_i$ can be learned by the bias term of the neuron to whose output is the \glsxtrlong{AF} applied, the authors \citeauthor{Qiu2018FReLU} formulate the \gls{FReLU} as
\begin{equation}
    f(z_i)=\mathrm{ReLU}\left(z_i\right) + b_i = \begin{cases}
        z_i + b_i, \quad & z_i \geq 0, \\
        b_i,  \quad & z_i < 0, \\
    \end{cases},
\end{equation}
where $b_i$ is a trainable parameter \cite{Qiu2018FReLU}.

\subsubsection{Adaptive shifted ReLU (ShiLU)}
\label{sec:shilu}
An \gls{ShiLU} \cite{Pishchik2023} is another adaptive variant of the \gls{ReLU} activation; it is a variant that adds a trainable slope and a vertical shift:
\begin{equation}
    f(z_i) = a_i\mathrm{ReLU}\left(z_i\right) + b_i = a_i\cdot\max \left(0, z\right) + b_i,
\end{equation}
where $a_i$ and $b_i$ are trainable parameters for each neuron $i$ \cite{Pishchik2023}. They used the name \textit{shifted ReLU}, but that name is already taken by the non-adaptive \gls{shifted_relu}; hence, the full name is \glsxtrlong{ShiLU} throughout this work to avoid confusion.

\subsubsection{StarReLU}
\label{sec:starrelu}
The \gls{StarReLU} \cite{Yu2024} is an adaptive version of the \gls{RePU} of power 2 using a similar approach as the \gls{ShiLU}; it is defined as

\begin{equation}
    f(z_i) = a_i \left(\mathrm{ReLU}\left(z\right)\right)^2 + b_i,
\end{equation}
where $a_i$ and $b_i$ are trainable parameters for each neuron $i$ \cite{Yu2024}. If the parameters are not used in an adaptive manner, \citeauthor{Yu2024} recommend setting $a_i = 0.8944$ and $b_i = -0.4472$ \cite{Yu2024}.

\subsubsection{Adaptive HardTanh}
\label{sec:adaptive_hardtanh}
An adaptive variant of \gls{hard_tanh} was used in \cite{Liu2021Adaptive}; it is defined as
\begin{equation}
    f(z) = \mathrm{HardTanh}\left(a_t\left(z-b\right)\right),
\end{equation}
where $a_t$ is a scale factor for each epoch $t$ such that $1 \leq a_1 \leq a_2 \leq \ldots \leq a_t \leq \ldots \leq a_T$, $T$ is the total number of training epochs and $b$ is an adaptive parameter trained using \gls{BP} with other parameters of the \gls{NN} \cite{Liu2021Adaptive}. The parameters $a_t$ are set such that the function starts in a similar shape as a regular \gls{hard_tanh} (see \cref{sec:hard_tanh}) and gradually approaches the sign function \cite{Liu2021Adaptive}. This allows for training a network that will gradually become a binary \gls{NN} where each activation is the sign function which can be used for speeding the inference \cite{Liu2021Adaptive}.

\subsubsection{Attention-based ReLU (AReLU)}
\label{sec:arelu}
\Gls{AReLU} \cite{Chen2020AReLU} is a adaptive \gls{ReLU} variant that uses ELSA --- element-wise attention mechanism proposed in \cite{Chen2020AReLU}. It is defined as
\begin{equation}
    f(z_l)=\begin{cases}

        \left(1+\sigma(b_l)\right)z_l, \quad & z_l \geq 0, \\
        \mathrm{C}(a_l)z_l,  \quad & z_l < 0, \\
    \end{cases},
\end{equation}
where $a_l$ and $b_l$ are learnable parameters for each layer $l$, $\sigma(x)$ is the \gls{logisticsigmoid} function, $C(x)$ is a function that clips the input into $[0.01, 0.99]$ \cite{Chen2020AReLU}. The derivative of $C(a_l)$ is handled by just not using the \gls{BP} when $a_l < 0.01$ or $a_l > 0.99$ \cite{Chen2020AReLU}. While \citeauthor{Chen2020AReLU} observe that the parameters $a_l$ and $b_l$ are insensitive to the initialization, they recommend initializing $a_l = 0.9$ and $b_l = 2.0$ as a larger initial value of $b_l$ can speed up the convergence \cite{Chen2020AReLU}. The \gls{AReLU} was found to outperform \gls{CELU}, \gls{ELU}, \gls{GELU}, LReLU, Maxout, Relu, \gls{RReLU}, \gls{SELU}, sigmoid, \gls{softplus}, \gls{swish}, \gls{tanh}, \gls{APLU}, \gls{PAU}, \gls{PReLU}, and \gls{SLAF} in experiments with various learning rates in \cite{Chen2020AReLU}. The performance of \gls{AReLU} was validated under different settings in \cite{Kang2021, Gupta2021, Kang2022}.

\subsubsection{Dual parametric ReLU (DPReLU) and Dual Line activation function}
\label{sec:dprelu}

A \gls{DPReLU} \cite{Balaji2020} extends the concept of \gls{PReLU} even further:
\begin{equation}
    f(z_i)=\begin{cases}
            a_i z_i, \quad & z_i \geq 0, \\
            b_i z_i,  \quad & z_i < 0, \\
        \end{cases}
\end{equation}
where $a_i$ and $b_i$ are trainable parameters for each neuron $i$; these are initialized the same as \gls{PReLU} --- $a_i = 1$, $b_i = 0.01$ \cite{Balaji2020}. The \gls{DPReLU} was also later proposed independently in \cite{Varshney2021} under the name \textit{fully parametric ReLU} and the abbreviation \gls{FReLU} (which is already used in the literature for the \glsxtrlong{FReLU}; see \cref{sec:frelu}).

\subsubsection{Dual Line}
\label{sec:dual_line}
The \gls{DPReLU} was further extended into a \gls{dual_line} \glsxtrlong{AF} that adds a shift parameter

\begin{equation}
    f(z_i)=\begin{cases}
            a_i z_i + m_i, \quad & z_i \geq 0, \\
            b_i z_i + m_i,  \quad & z_i < 0, \\
        \end{cases}
\end{equation}
where $a_i$ and $b_i$ are trainable parameters for each neuron $i$ the same as in \gls{DPReLU} and $m_i$ is an additional trainable shift parameter for each neuron or filter $i$; $m_i$ was initialized to $m_i=-0.22$ \cite{Balaji2020}.

\subsubsection{Piecewise linear unit (PiLU)}
\label{sec:pilu}
An \gls{AF} very similar to the \gls{dual_line} is the \gls[prereset]{PiLU} proposed in \cite{Inturrisi2021}; it just extends the \gls{dual_line} concept by adding horizontal shifts. It is defined as
\begin{equation}
    f(z_i)=\begin{cases}
            a_i z_i + c_i(1-a_i), \quad & z_i \geq c_i, \\
            b_i z_i + c_i(1-b_i),  \quad & z_i < c_i, \\
        \end{cases}
\end{equation}
where $a_i$, $b_i$, and $c_i$ are adaptive parameters for each neuron $i$ \cite{Inturrisi2021}. The \gls{PiLU} geneneralizes, for example, \gls{ReLU}, \gls{LReLU}, \gls{PReLU}, \gls{SlReLU}, \gls{DPReLU}, and \gls{dual_line}.

\subsubsection{Dual parametric family of activation functions}
\label{sec:dpaf}
The \gls{DPReLU} approach (see \cref{sec:dprelu}) can be extended to a general concept transforming any \glsxtrlong{AF} $g(z)$:
\begin{equation}
    f(z_i)=\begin{cases}
            a_i g(z_i) + m_i, \quad & z_i \geq 0, \\
            g(z_i) + m_i,  \quad & z_i < 0, \\
        \end{cases}
\end{equation}
where $g(z_i)$ is any \glsxtrlong{AF} and $a_i$ and $m_i$ are trainable parameters for each neuron $i$ \cite{Balaji2020}. The functions of this family are called \glspl{DPAF} throughout this text.

\subsubsection{Fully parameterized activation function (FPAF)}
\label{sec:fpaf}
Similar approach to \gls{DPAF} (see \cref{sec:dpaf}) was proposed under the name of \gls{FPAF} in \cite{Varshney2021}; the \gls{FPAF} is defined as
\begin{equation}
    f(z_i)=\begin{cases}
            a_i g_1(z_i), \quad & z_i \geq 0, \\
            b_i g_2(z_i),  \quad & z_i < 0, \\
        \end{cases}
\end{equation}
where $a_i$ and $b_i$ are trainable parameters for each neuron $i$ and $g_1(z_i)$ and $g_2(z_i)$ can be any function \cite{Varshney2021}.
The \gls{FPAF}, in contrast to the family of \glsxtrshortpl{DPAF}, has no trainable shift but allows for learnable slopes for both parts of the piecewise function.

\subsubsection{Elastic PReLU (EPReLU)}
\label{sec:eprelu}
The same as \gls{EReLU} extends the concept of \gls{ReLU} (see \cref{sec:relu}) \cite{Jiang2018}, the \gls[prereset]{EPReLU} extends the \gls{PReLU} --- it adds a varying coefficient to the positive part of the \gls{PReLU} \cite{Jiang2018}:
\begin{equation}
    f(z_i)=\begin{cases}
                k_i z_i, \quad & z_i \geq 0, \\
                \frac{z_i}{a_i},  \quad & z_i < 0, \\
            \end{cases}
\end{equation}
where $a_i$ is the optimized parameter, $k_i$ is a sampled for each epoch and neuron $i$  from the uniform distribution: $a_i \sim U(1-\alpha, 1+\alpha)$ where $\alpha \in (0,1)$ \cite{Jiang2018}. A modified training procedure for \gls{EPReLU} is also proposed --- the neuron weights and the trainable parameter $a_i$ are updated with $k_i = 1$ in odd epochs, while in even epochs, the $a_i$ is kept fixed, the parameter $k_i$ is sampled from the uniform distribution, and only the neuron weights are updated \cite{Jiang2018}. It was shown that the \gls{EPReLU} leads to improved performance over the \gls{ReLU}, \gls{PReLU}, \gls{EReLU}, \gls{APLU}, \gls{NIN}, and \gls{maxout} networks on several datasets \cite{Jiang2018}.

\subsubsection{Paired ReLU}
\label{sec:paired_relu}
A \gls{paired_relu} \cite{Tang2018} is a concept similar to \gls{CReLU} (see \cref{sec:crelu}), but it introduces four trainable parameters. It is defined as
\begin{equation}
    \vec{f}(z) = \begin{bmatrix}
        \max\left(a_iz_i-b_i, 0\right) \\
        \max\left(c_iz_i-d_i, 0\right) \\
        \end{bmatrix},
\end{equation}
where $a_i$, $b_i$, $c_i$, and $d_i$ are trainable parameters for each neuron $i$ \cite{Tang2018,Dubey2022}. The parameters $a_i$ and $c_i$ are scale parameters and $b_i$ and $d_i$ are trainable thresholds; the inital values of scale parameters are $a_i = 0.5$ and $c_i = -0.5$ \cite{Tang2018}.

\subsubsection{Tent}
\label{sec:tent}
The \gls{tent} is a \gls{ReLU}-based \gls{AF} proposed in \cite{Rozsa2019}; it is defined as
\begin{equation}
    f(z_i) = \max\left(0, a_i-|z_i|\right),
\end{equation}
where $a_i$ is a trainable parameter \cite{Rozsa2019}. \Citeauthor{Rozsa2019} recommend using batch normalization and initializing $a_i=1$ \cite{Rozsa2019}. Also, having a weight decay on the parameter $a_i$ during training proved beneficial for certain tasks \cite{Rozsa2019}.

\subsubsection{Hat}
\label{sec:hat}
The \gls{hat} \cite{Wang2022CNNs} is an \gls{AAF} very similar to the \gls{tent} \gls{AF} --- the only difference is that the \gls{tent} \gls{AF} is centered around zero while the \gls{hat} is positive only for positive inputs. The \gls{hat} \gls{AF} is defined as
\begin{equation}
    f(z_i)=\begin{cases}
            0,  \quad & z_i < 0, \\
            z_i, \quad & 0 \leq z_i \leq \frac{a_i}{2}, \\
            z_i, \quad & \frac{a_i}{2} \leq z_i \leq a_i , \\
            0,  \quad & z_i > a_i, \\
        \end{cases}
\end{equation}
where $a_i$ is can be either fixed or trainable parameter \cite{Wang2022CNNs}. \Citeauthor{Wang2022CNNs} used $a_i=2$ for the fixed variant in \cite{Wang2022CNNs}; this value was also used in \cite{Hong2022OnTheActivation}.

\subsubsection{ReLU memristor-like activation function (RMAF)}
\label{sec:rmaf}
The \gls{RMAF} is an \glsxtrlong{AF} similar to the \gls{swish} \gls{AF} (see \cref{sec:swish}) and it also attempts to leverage the negative values \cite{Yu2020}. It is defined as
\begin{equation}
    f(z_i)=\left[b\frac{1}{\left(0.25\left(1+\exp(-z_i)\right)+0.75\right)^c}\right]a_iz_i,
\end{equation}
where $a_i$ is a trainable parameter initialized $a_i=1$ for each neuron $i$ or it is a fixed hyperparameter and $b$ and $c$ are fixed hyperparameters \cite{Yu2020}.

\subsubsection{Parametric tanh linear unit (PTELU)}
\label{sec:ptelu}
A \gls{PTELU} \cite{Gupta2017} is an adaptive function that, for positive inputs, behaves just as a \gls{ReLU}; however, the negative part is parameterized \gls{tanh} function \cite{Dubey2022}. It can also be seen as an extension of the \gls{PReLU} (see \cref{sec:prelu}). It is an adaptive variant of \gls{ThLU} (see \cref{sec:thlu}).
It is defined as
\begin{equation}
    f(z_i)=\begin{cases}
                 z_i, \quad & z_i \geq 0, \\
                a_i \tanh\left(b_i z_i\right),  \quad & z_i < 0, \\
            \end{cases}
\end{equation}
where $a_i$ and $b_i$ are trainable parameters for each neuron $i$; $a_i \geq0$ and $b_i \geq 0$ \cite{Gupta2017}. It has output range of $[-a_i, \infty)$ \cite{Dubey2022}. The parameter $a_i$ controls the saturation value, and the parameter $b_i$ controls the convergence rate \cite{Gupta2017}. While the AF resembles an adaptive extension of an \gls{ELU} \glsxtrlongpl{AF}, the author \citeauthor{Gupta2017} decided to use \gls{tanh} function for the negative inputs because it gives a higher gradient for small negative inputs and saturates earlier than $\exp(z) - 1$ and thus the noise-robust deactivation state earlier and faster \cite{Gupta2017}. The nonadaptive variant of \gls{PTELU} with $a_i=1$ and $b_i=1$ was proposed in \cite{Guevraa2021}.

\subsubsection{Tangent linear unit (TaLU)}
\label{sec:talu}
The \gls[prereset]{TaLU} \cite{Mercioni2021Developing} is an \gls{AF} similar to the \gls{PTELU}. The \gls{TaLU} is defined as
\begin{equation}
    f(z_i)=\begin{cases}
                 z_i, \quad & z_i \geq 0, \\
                \tanh\left(z_i\right)  \quad &  a_i < z_i < 0, \\
                \tanh\left(a_i\right)  \quad &  z_i \leq a_i, \\
            \end{cases}
\end{equation}
where $a_i < 0$ is either a learnable\footnote{The variant with the adaptive parameter was named \textit{\gls{TaLU} learnable} by the authors.} or fixed parameter \cite{Mercioni2021Developing}.

\subsubsection{PTaLU}
\label{sec:ptalu}
The \gls{PTaLU}\footnote{Not an abbreviation but a name given by \citeauthor{Mercioni2021Developing} in \cite{Mercioni2021Developing}.} is a variant of \gls{TaLU} with another learnable parameter \cite{Mercioni2021Developing}. It is defined as
\begin{equation}
    f(z_i)=\begin{cases}
                 z_i, \quad & z_i \geq b_i, \\
                \tanh\left(z_i\right)  \quad &  a_i < z_i < b_i, \\
                \tanh\left(a_i\right)  \quad &  z_i \leq a_i, \\
            \end{cases}
\end{equation}
where $a_i$ and $b_i$ are trainable parameters \cite{Mercioni2021Developing}. \Citeauthor{Mercioni2021Developing} used initial values $a_i = -0.75$ and $b_i=1$ in \cite{Mercioni2021Developing}.

\subsubsection{TanhLU}
\label{sec:tanhlu}
The \gls{tanhLU}\footnote{Not an abbreviation but a name given by \citeauthor{Shen2022} in \cite{Shen2022}.} is a parametric combination of the \gls{tanh} and a linear function proposed in \cite{Shen2022}. It is defined as

\begin{equation}
    f(z_i) = a_i \cdot \tanh\left(b_iz_i\right) + c_iz_i,
\end{equation}
where $a_i$, $b_i$, and $c_i$ are trainable parameters for each neuron $i$ \cite{Shen2022}.

\subsubsection{TeLU}
\label{sec:telu}
Despite the similar name, the \gls[prereset]{TeLU}\footnote{\cite{AdrianaMercioni2020} used the \glsxtrshort{TeLU} as the name and not as an abbreviation; nevertheless, the long name \glsxtrlong{TeLU} fits the usual naming convention and, therefore, it is used in this work.} \cite{AdrianaMercioni2020} is quite different from the \gls{PTELU}. The \gls{TeLU} is closely related to the \gls{mish} and \gls{TanhExp} activations, but, unlike these two \glspl{AF}, it also has an additional adaptive parameter. It is defined as
\begin{equation}
    f(z_i) = z_i\cdot\tanh\left(\mathrm{ELU}\left(a_iz_i\right)\right),
\end{equation}
where $a_i$ is either learnable or fixed scaling parameter \cite{AdrianaMercioni2020}.

\subsubsection{Tanh based ReLU (TReLU)}
\label{sec:trelu}
A \gls{TReLU} was proposed in \cite{Zhang2019Research}; however, it is is only a special case of previously proposed \gls{PTELU} (see \cref{sec:ptelu}). It is defined as
\begin{equation}
    f(z_i)=\begin{cases}
                 z_i, \quad & z_i \geq 0, \\
                \tanh\left(b_i z_i\right),  \quad & z_i < 0, \\
            \end{cases}
\end{equation}
where $b_i$ is a trainable parameter for each neuron $i$\cite{Zhang2019Research}. This function is identical to the \gls{PTELU} with its parameter $a_i$ fixed to $a_i=1$. Another special case of \gls{PTELU} was proposed in \cite{Nakhua2023} also under the name of \gls{TReLU} --- this time, the parameter $a_i$ becomes a predefined fixed parameter, and $b_i$ becomes fixed to $b_i=1$. This function is denoted \gls{TReLU2} in this work and is defined as
\begin{equation}
    f(z)=\begin{cases}
                 z, \quad & z \geq 0, \\
                a\tanh\left(z\right),  \quad & z < 0, \\
            \end{cases}
\end{equation}
where $a$ is fixed\footnote{While \citeauthor{Nakhua2023} used the parameter fixed during their experiments, they also speculated that making it learnable might improve the performance.} parameter \cite{Nakhua2023}.

\subsubsection{Rectified linear tanh (ReLTanh)}
\label{sec:reltanh}
A \gls{ReLTanh} is a piecewise adaptive \glsxtrlong{AF} that improves traditional \gls{tanh} \glsxtrlong{AF} \cite{Wang2019ReLTanh} --- it replaces the positive and negative saturated regions of the \gls{tanh} \glsxtrlongpl{AF} with straight lines whose slopes are identical to the slope of the \gls{tanh} at the thresholds \cite{Wang2019ReLTanh}. It is defined as
\begin{equation}
    f(z_i)=\begin{cases}
                 \tanh'(a_i)(z_i-a_i) + \tanh(a_i), \quad & z_i \leq a_i, \\
                 \tanh(z_i), \quad & a_i < z_i < b_i, \\
                 \tanh'(b_i)(z_i-b_i) + \tanh(b_i), \quad & z_i \geq b_i, \\
            \end{cases}
\end{equation}
where $\tanh'(x)$  is the derivative of the \gls{tanh} function
\begin{equation}
    \tanh'(x) = \frac{4}{\left(\exp(x) + \exp(-x)\right)^2},
\end{equation}
and $a_i \in [a_{\text{low}}, a_{\text{high}}]$ and $b_i \in [b_{\text{low}}, b_{\text{high}}]$ are two trainable parameters that may be defined for each neuron $i$ but are rather recommended to be shared by a whole layer $l$ \cite{Wang2019ReLTanh} in order to decrease computational burden. The limits $a_{\text{low}}$, $a_{\text{high}}$, $b_{\text{low}}$, and $b_{\text{high}}$ for the parameters are to constraint the learnable range and are predefined hyperparameters. \citeauthor{Wang2019ReLTanh} used $a_{\text{low}} = -\infty$, $a_{\text{high}}=-1.5$, $b_{\text{low}}=0$, and $b_{\text{high}}=0.5$ in their work \cite{Wang2019ReLTanh}. The initial values were set to $a_i = -1.5$ and $b_i = 0$ for all layers (the parameters were shared layer-wise) in order to speed up the training process in early stages by the larger gradients \cite{Wang2019ReLTanh}.

\subsubsection{Bendable linear unit (BLU)}
\label{sec:blu}
A \gls{BLU} \cite{Godfrey2019} is an adaptive function that allows for any interpolation between the identity function and a rectifier \cite{Godfrey2019, Dubey2022}. It is defined as
\begin{equation}
    f(z_i) = a_i \left(\sqrt{z_i^2+1} - 1\right) + z_i,
\end{equation}
where $a_i \in [-1, 1]$ is a trainable parameter for each neuron or filter \cite{Godfrey2019}. One of the main advantages of the \gls{BLU} is that it can model an identity function; the identity function is useful because its gradient cannot vanish or explode, and it also allows for a layer to be "skipped" \cite{Godfrey2019} --- it is one of the reasons of why ResNets \cite{He2016} became so popular \cite{Godfrey2019} as it is rather hard to learn an identity transformation using conventional \glsxtrlong{NN} and the architecture with skip connections allows for easy learning of the identity mapping \cite{He2016}. Unless the magnitude $|a_i|$ is exactly 1, the derivative of \gls{BLU} is non-zero for both positive and negative inputs (similarly to \gls{LReLU} and in contrast to vanilla \gls{ReLU} and \gls{ELU}) \cite{Godfrey2019}. \gls{BLU} has a slope higher than 1 for positive inputs for $a_i$ approaching 1 (or for negative inputs for $a_i$ approaching -1) \cite{Godfrey2019}; this property helps to avoid vanishing gradient problems \cite{Godfrey2019, Klambauer2017}. Another useful benefit is that \gls{BLU} are $C^\infty$ continuous \cite{Godfrey2019}, which can be theoretically exploited for speeding up the optimization \cite{Godfrey2019}, e.g., \cite{Zhang2013, Yang2013, Bergou2022, Mahdavi2013}. It was also shown that smooth \glsxtrlongpl{AF} provide better signal propagation \cite{Hayou2019}.

\subsubsection{Rectified BLU (ReBLU)}
\label{sec:reblu}
A variant of the \gls{BLU} (see \cref{sec:blu}) was proposed in \cite{Dai2022} under the name \gls[prereset]{ReBLU}. It is defined as
\begin{equation}
    f(z_i)=\begin{cases}
        \mathrm{BLU}\left(z_i\right), \quad & z_i > 0, \\
       0,  \quad & z_i \leq 0, \\
   \end{cases}=\begin{cases}
    a_i \left(\sqrt{z_i^2+1} - 1\right) + z_i, \quad & z_i > 0, \\
   0,  \quad & z_i \leq 0, \\
\end{cases}
\end{equation},
where $a_i$ is a trainable parameter \cite{Dai2022}.

\subsubsection{DELU}
\label{sec:delu}
The \gls{DELU}\footnote{\gls{DELU} is not an abbreviation but rather a name given by \citeauthor{Pishchik2023}.} \glsxtrlong{AF} \cite{Pishchik2023} is a \gls{ReLU} variation that utilizes the \gls{SiLU} function (see \cref{sec:silu}). It is defined as
\begin{equation}
    f(z_i)=\begin{cases}
                (a_i + 0.5)z_i + \left|\exp\left(-z_i\right)-1\right|, \quad & z_i \geq 0, \\
                z_i\sigma\left(z_i\right),  \quad & z_i < 0, \\
            \end{cases}
\end{equation}
where $a_i$ is a trainable parameter for each neuron $i$ and $\sigma(z_i)$ is the \gls{logisticsigmoid} function \cite{Pishchik2023}.

\subsubsection{Soft clipping mish}
\label{sec:scmish}
A \gls{ReLU} variant called \gls{SC-mish} was proposed in \cite{Mercioni2021}. It adds soft clipping to the positive inputs using the \gls{mish} \gls{AF}; it is defined as
\begin{equation}
    f(z_i) = \max\left(0, z_i\cdot \tanh\left(\mathrm{softplus}(a_iz_i)\right)\right),
\end{equation}
where $a_i$ is a fixed parameter \cite{Mercioni2021}; \citeauthor{Mercioni2021} used $a_i=1$. It also has a variant where the parameter $a_i$ is trainable. Such a variant is called \gls{SCL-mish}. When using the \gls{SCL-mish}, \citeauthor{Mercioni2021} initalized the parameter $a_i=0.25$ \cite{Mercioni2021}.

\subsubsection{Soft clipping swish}
\label{sec:scswish}
Yet another \gls{AF} proposed by \citeauthor{Mercioni2022} is the \gls[prereset]{SC-swish} \cite{Mercioni2022,Mercioni2022Weather,Mercioni2021SoftClipping}. This function is very similar to \gls{SC-mish} and is defined as
\begin{equation}
    f(z) = \max\left(0, z\cdot \sigma\left(z\right)\right),
\end{equation}
where $\sigma(z)$ is the \gls{logisticsigmoid} \cite{Mercioni2022}.

\subsubsection{Parametric swish (p-swish)}
\label{sec:pswish}
The \gls[prereset]{p-swish} \cite{Mercioni2020PSwish} is another \gls{AF} proposed by \citeauthor{Mercioni2020PSwish}. It is defined as
\begin{equation}
    f(z_i)=\begin{cases}
                a_iz_i\sigma\left(b_iz_i\right), \quad & z_i \leq c_i, \\
                z_i,  \quad & z_i > c_i, \\
            \end{cases}
\end{equation}
where $a_i$, $b_i$, and $c_i$ are either trainable or fixed parameters (or combination thereof) \cite{Mercioni2020PSwish}. The parameters were initialized to $a_i=1$, $b_i=1$ and $c_i=0$ in experiments in \cite{Mercioni2020PSwish}. An \gls{AF} named \textit{R\_S} similar to the \gls{p-swish} was independently proposed in \cite{Shi2022Image}; it is equivalent to a \gls{p-swish} with fixed $a_i = 1$.

\subsubsection{Parametric exponential linear unit (PELU)}
\label{sec:pelu}
Similarly as \gls{PReLU} extends the concept of \gls{ReLU}, the \gls{PELU} \cite{Trottier2016} extends the concept of \gls{ELU} (see \cref{sec:elu}). 
The \gls{PELU} builds on a parameterization that separately controls the saturation point, the decay, and the slope:
\begin{equation}
    f(z_i)=\begin{cases}
                c_i z_i, \quad & z_i \geq 0, \\
                a_i \left(\exp\left(\frac{z_i}{b_i}\right)-1\right),  \quad & z_i < 0, \\
            \end{cases}
\end{equation}
where $a_i$, $b_i$, and $c_i$ are trainable parameter for each neuron $i$ \cite{Trottier2016}. However, the \gls{PELU} introduces only two new parameters --- $a_i$ controlling the saturation point, and $b_i$ controlling the decay --- to control the shape of the \glsxtrlong{AF}; the slope is  not controlled separately through another parameter as it could lead to non-differentiability at $z_i = 0$; therefore the slope is set such that the derivatives on both sides of zero are equal which leads to $c_i = \frac{a_i}{b_i}$ \cite{Trottier2016} and therefore the \gls{PELU} is defined as
\begin{equation}
    f(z_i)=\begin{cases}
                \frac{a_i}{b_i} z_i, \quad & z_i \geq 0, \\
                a_i \left(\exp\left(\frac{z_i}{b_i}\right)-1\right),  \quad & z_i < 0. \\
            \end{cases}
\end{equation}

The \gls{PELU} combined with mixing different \glsxtrlongpl{AF} which use an adaptive linear combination or hierarchical gated combination of \glsxtrlong{AF} was shown to perform well \cite{Qian2018} --- see \cref{sec:adaptive_combination}. 

\subsubsection{Extended exponential linear unit (EDELU)}
\label{sec:edelu}
An adaptive function called \gls{EDELU}\footnote{Authors called the function \textit{extendeD \gls{ELU}} resulting in an abbreviation \gls{DELU} but that name is already taken by an \gls{AF} proposed a few months earlier in \cite{Pishchik2023}.} was proposed in \cite{Catalba2023}. This function is the same as the \gls{PELU}, but it omits the vertical scaling for positive inputs, adds a parameter controlling the threshold, and uses inverse definitions of the parameters present in the \gls{PELU}. It is defined as
\begin{equation}
    f(z_i)=\begin{cases}
                z_i, \quad & z_i \geq c_i, \\
                \frac{\left(\exp\left(a_iz_i\right)-1\right)}{b_i},  \quad & z_i < c_i, \\
            \end{cases}
\end{equation}
where $a_i\geq0$\footnote{The \gls{PELU} equivalent would be $\frac{1}{b_i}$.} and $b_i\geq0$\footnote{The \gls{PELU} equivalent would be $\frac{1}{a_i}$.} are trainable \gls{PELU} parameters and $c_i \geq 0$ is the novel parameter for controlling the threshold that has to satisfy the relationship
\begin{equation}
b_ic_i = \exp\left(a_ic_i\right)-1;
\end{equation} 
while the $c_i = 0$ is always a solution of the equation, there are other solutions for $b_i > a_i > 0$ \cite{Catalba2023}.

\subsubsection{Adaptive combination of PELU and PReLU}
\label{sec:adaptive_combination}
Two different \glsxtrlongpl{AF} can be mixed together, as shown in \cite{Qian2018}. One such example of mixed \glsxtrlong{AF} is
\begin{equation}
    f(z_i) = a_i\cdot \mathrm{LReLU}(z_i) + \left(1-a_i\right)\mathrm{ELU}(z_i),
\end{equation}
where $a_i$ is a combination coefficient that might be learned from the data \cite{Qian2018}.
Another mixing approach was shown for combining \gls{PReLU} and \gls{PELU}:
\begin{equation}
    f(z_i) = \sigma(a_iz_i) \mathrm{PReLU}(z_i) +  \left(1-\sigma\left(a_iz_i\right)\right) \mathrm{PELU}(z_i),
\end{equation}
where $\sigma(x)$ is the \gls{logisticsigmoid} \cite{Qian2018}. \citeauthor{Qian2018} also proposed other mixing schemes such as hierarchical activation, winner-take-all selection whose performance were shown on the MNIST \cite{LiDeng2012}, CIFAR-10 and CIFAR-100 \cite{Krizhevsky2009} datasets; see \cite{Qian2018} for details.

\subsubsection{Fast exponential linear unit (FELU)}
\label{sec:felu}
An \gls{ELU} variant called \gls{FELU} aiming at efficient training and network inference was proposed in \cite{Qiumei2019} --- it is inspired by fast approximation of the exponential function proposed in \cite{Schraudolph1999} to replace the exponential in the \gls{ELU}:
\begin{equation}
    f(z_i)=\begin{cases}
                z_i, \quad & z_i \geq 0, \\
                a_i \left(2^{\frac{z_i}{\ln(2)}} - 1\right),  \quad & z_i < 0, \\
            \end{cases}
\end{equation}
where $a_i$ is a trainable parameter controlling the soft saturation region \cite{Qiumei2019}.

\subsubsection{P+FELU}
\label{sec:pfelu}
\Citeauthor{Adem2022PFELU} proposed variant of the \gls{FELU} function named \gls{PFELU}; this variant has an added parameter and is defined as
\begin{equation}
    f(z_i)=\begin{cases}
                z_i+b, \quad & z_i \geq 0, \\
                a_i \left(2^{\frac{z_i}{\ln(2)}} - 1\right) + b,  \quad & z_i < 0, \\
            \end{cases}
\end{equation}
where $a_i$ is a trainable parameter same as the original \gls{FELU} and $b$ is the added trainable parameter \cite{Adem2022PFELU}.

\subsubsection{Multiple parametric exponential linear unit (MPELU)}
\label{sec:mpelu}
A \gls{PELU} extension, \glsxtrfull{MPELU} \cite{Li2018Improving}, uses two trainable parameters to allow for a combination of a \gls{ReLU} and \gls{ELU} \cite{Dubey2022}. The \gls{MPELU} is defined as
\begin{equation}
    f(z_i)=\begin{cases}
                z_i, \quad & z_i \geq 0, \\
                a_i \left(\exp\left(b_iz_i\right)-1\right),  \quad & z_i < 0, \\
            \end{cases}
\end{equation}
where $a_i$ and $b_i$ are trainable parameters for each neuron $i$ \cite{Li2018Improving, Dubey2022}. The \gls{ReLU}, certain parameterizations of \gls{PELU}, and \gls{ELU} are special cases of the \gls{MPELU} \cite{Li2018Improving}. A special method for weight initialization of neurons with \gls{MPELU} units was also proposed; depending on a particular initialization, the method can become the initialization for \gls{ELU} networks or for \gls{ReLU} networks \cite{Li2018Improving}. The MSRA\footnote{The initialization method was unnamed in the original paper \cite{He2015} but was later named \textit{Microsoft Research Asia} (MSRA) filler \cite{Pham2019}.} filler approach \cite{He2015} can be considered as a special case of the \gls{MPELU} initialization \cite{Li2018Improving}. The MPLU initialization is a similar approach to LSUV initialization \cite{Mishkin2015}, but unlike the LSUV initialization, it provides an analytic solution for \gls{ELU} and \gls{MPELU} and therefore it has lower computational costs \cite{Li2018Improving}. It was also shown that the \gls{MPELU} works better with batch normalization compared to the vanilla \gls{ELU} \cite{Li2018Improving}. The performance of the \gls{MPELU} was empirically shown on the CIFAR-10 and CIFAR-100 datasets \cite{Krizhevsky2009} using multiple \glsxtrlong{NN} architectures \cite{Li2018Improving}, e.g. nine-layer deep \gls{NIN} \cite{Lin2013} or even a ResNet with 1001 layers \cite{He2016}.

\subsubsection{P-E2-ReLU}
\label{sec:pe2relu}
The \gls{AAF} named \gls{P-E2-ReLU} is combining two \glspl{ELU} and a \gls{ReLU} using two adaptive parameters \cite{Jie2021}. It is defined as 
\begin{equation}
    \label{eq:pe2relu}
    f(z_i) = a_i\cdot\mathrm{ReLU}\left(z_i\right) + b_i\cdot\mathrm{ELU}\left(z_i\right) + \left(1-a_i-b_i\right)\cdot\left(-\mathrm{ELU}\left(-z_i\right)\right),
\end{equation}
where $a_i$ and $b_i$ are trainable parameters for each neuron $i$ \cite{Jie2021}. The parameters were initialized to $a_i=0.4$ and $b_i=0.3$ in experiments in \cite{Jie2021}.
\Citeauthor{Jie2021} mentioned that other combinations could be considered and called this family \gls{P-E2-X}. One such combination is denoted \gls{P-E2-Id} and is defined as
\begin{equation}
    f(z_i) = a_iz_i + \left(1-a_i\right)\cdot\left(mathrm{ELU}\left(z_i\right)-\mathrm{ELU}\left(-z_i\right)\right),
\end{equation}
and another is named \gls{P-E2-ReLU-1}
\begin{equation}
    f(z_i) = a_i\cdot\mathrm{ReLU}\left(z_i\right) + \left(1-a_i\right)\cdot\left(mathrm{ELU}\left(z_i\right)-\mathrm{ELU}\left(-z_i\right)\right),
\end{equation}
whera $a_i$ is a trainable parameter in both \glspl{AAF} \cite{Jie2021}. The parameter was initialized to $a_i=0.5$ in experiments in \cite{Jie2021}.

\subsubsection{Soft exponential}
\label{sec:soft_exponential}
The \gls{soft_exponential} \glsxtrlong{AF} is an \glsxtrlong{AAF} that is able to interpolate between logarithmic, linear, and exponential functions \cite{Godfrey2015}. It is defined as
\begin{equation}
    f(z_i)=\begin{cases}
                \frac{\exp\left(z_i\right)-1}{a_i} + a_i,  \quad & a_i > 0, \\
                z_i,  \quad & a_i = 0, \\
                -\frac{\ln\left(1-a_i\left(z_i+a_i\right)\right)}{a_i}, \quad & a_i < 0, \\
            \end{cases}
\end{equation} 
where $a_i$ is a trainable parameter \cite{Godfrey2015}. The \gls{soft_exponential} \glsxtrlongpl{AF} is continuously differentiable with respect to $z_i$ and also with respect to $a_i$ \cite{Godfrey2015}; furthermore, for any constant $a_i$, the function is monotonic \cite{Godfrey2015}. When $a_i = -1$, the function becomes $f(z_i) = \ln(z_i)$, while for $a_i=0$ it becomes linear function $f(z_i) = z_i$ and for $a_i = 1$, it is the exponential function $f(z_i) = \exp(z_i)$ \cite{Godfrey2015}.

\subsubsection{Continuously differentiable ELU (CELU)}
\label{sec:celu}
A \gls{CELU} was proposed in \cite{Barron2017}; \gls{CELU} is very similar to the original parameterization of \gls{ELU} \cite{Dubey2022} but reformulated such that the derivative at $z_i = 0$ is 1 for all values of $a_i$ \cite{Barron2017}. \gls{CELU} is defined as
\begin{equation}
    f(z_i)=\begin{cases}
                z_i, \quad & z_i \geq 0, \\
                a_i \left(\exp\left(\frac{z_i}{a_i}\right)-1\right),  \quad & z_i < 0, \\
            \end{cases}
\end{equation}
where $a_i$ is a learnable parameter for each neuron $i$. Its main advantages are that its derivative with respect to $z_i$ is bounded and that it contains both the linear transfer function and \gls{ReLU} \cite{Barron2017}.

\subsubsection{Erf-based ReLU (ErfReLU)}
\label{sec:erfrelu}
The \gls[prereset]{ErfReLU} \cite{Rajanand2023} is an \gls{AAF} similar to the \gls{ELU}. It is defined as
\begin{equation}
    f(z_i)=\begin{cases}
                z_i, \quad & z_i \geq 0, \\
                a_i \mathrm{erf}\left(z_i\right),  \quad & z_i < 0, \\
            \end{cases}
\end{equation}
where $a_i$ is a learnable parameter for each neuron $i$ and $\mathrm{erf}\left(z_i\right)$ is the Gauss error function \cite{Rajanand2023}.

\subsubsection{Parametric scaled exponential linear unit (PSELU)}
\label{sec:pselu}
A \gls{PSELU} \cite{Pratama2020} is basically a \gls{SELU} (see \cref{sec:selu}) where the parameters $a$ and $b$ controlling the behavior are trainable. It is defined as 
\begin{equation}
    f(z_i)=\begin{cases}
                 a_i z_i, \quad & z_i \geq 0, \\
                a_i b_i \left(\exp\left(z_i\right)-1\right),  \quad & z_i < 0, \\
            \end{cases}
\end{equation}
where $a_i$ and $b_i$  are trainable parameters for each neuron $i$.

\subsubsection{Leaky parametric scaled exponential linear unit (LPSELU)}
\label{sec:lpselu}

A \gls{LPSELU} \cite{Pratama2020} is a leaky extension of the \gls{PSELU} (see \cref{sec:pselu}) to avoid small gradients hindering the learning process \cite{Pratama2020}:
\begin{equation}
    f(z_i)=\begin{cases}
                 a_i z_i, \quad & z_i \geq 0, \\
                a_i b_i \left(\exp\left(z_i\right)-1\right) + c_iz_i,  \quad & z_i < 0, \\
            \end{cases}
\end{equation}
where $a_i$ and $b_i$  are trainable parameters for each neuron $i$ and $c_i$ is either a predefined constant or a trainable parameter \cite{Pratama2020}.

\subsubsection{Leaky parametric scaled exponential linear unit with reposition parameter (LPSELU\_RP)}
\label{sec:lpselurp}
The \gls{LPSELU} can be extended by a reposition parameter similarly as \gls{FReLU} extends \gls{ReLU} (see \cref{sec:frelu}) \cite{Pratama2020}; such function is called \gls{LPSELURP} and is defined as
\begin{equation}
    f(z_i)=\begin{cases}
                 a_i z_i + m_i, \quad & z_i \geq 0, \\
                a_i b_i \left(\exp\left(z_i\right)-1\right) + c_iz_i + m_i,  \quad & z_i < 0, \\
            \end{cases}
\end{equation}
where $a_i$ and $b_i$  are trainable parameters for each neuron $i$, and $c_i$ is either a predefined constant or a trainable parameter the same as for \gls{LPSELU} (see \cref{sec:lpselu}) and $m_i$ is a trainable reposition parameter \cite{Pratama2020}. It was empirically observed that the shift parameter $m_i$ converges to a small negative value, which supports the hypothesis that the negative output of \glsxtrlongpl{AF} is important \cite{Pratama2020}.

\subsubsection{Shifted ELU family}
\label{sec:shifted_elus}
A family of several \glsxtrlongpl{AF}, \gls{shifted_elu}, was proposed in \cite{Grelsson2018}; functions in this family have either vertical or horizontal shift of an \gls{ELU} \glsxtrlong{AF} that can be either constant or trainable. An \gls{ELU} with fixed horizontal shift is ShELU, with fixed vertical \gls{SvELU} and \gls{PELU} (see \cref{sec:pelu}) with trainable horizontal shift is PShELU \cite{Grelsson2018}.
The ShELU is defined as
\begin{equation}
    f(z)=\begin{cases}
                z + b, \quad & z + b \geq 0, \\
                a\left(\exp\left(z+b\right)-1\right),  \quad & z + b < 0, \\
            \end{cases}
\end{equation}
where $a$ is a fixed parameter similarly as in the vanilla \gls{ELU} and $b$ is novel, preset parameter controlling the horizontal shift \cite{Grelsson2018}. The \gls{SvELU} is defined similarly:
\begin{equation}
    f(z)=\begin{cases}
                z + b, \quad & z \geq 0, \\
                a\left(\exp\left(z\right)-1\right)+b,  \quad & z < 0, \\
            \end{cases}
\end{equation}
where $a$ is a fixed parameter similarly as in the vanilla \gls{ELU} and $b$ is novel, preset parameter controlling the vertical shift \cite{Grelsson2018}. \citeauthor{Grelsson2018} define also a variant of \gls{PELU} with horizontal shift called PSheLU:
\begin{equation}
    f(z_i)=\begin{cases}
                 \frac{a_i}{b_i} \left(z_i+c_i\right), \quad & z_i + c_i \geq 0, \\
                a_i\left(\exp\left(\frac{z_i+c_i}{b_i}\right)-1\right),  \quad & z_i + c_i < 0, \\
            \end{cases}
\end{equation}
where $a_i$ and $b_i$ are trainable parameters of the original \gls{PELU}, and $c_i$ is a novel trainable parameter controlling the horizontal shift for each neuron $i$ \cite{Grelsson2018}. For some reason, \citeauthor{Grelsson2018} did not propose a \gls{PELU} with vertical shift (PSvELU), but it could be defined in a similar manner
\begin{equation}
\label{eq:psvelu}
    f(z_i)=\begin{cases}
                 \frac{a_i}{b_i}z_i+c_i, \quad & z_i\geq 0, \\
                a_i\left(\exp\left(\frac{z_i}{b_i}\right)-1\right)+c_i,  \quad & z_i < 0, \\
            \end{cases}
\end{equation}
where $a_i$, $b_i$ are trainable parameters of the original \gls{PELU} and $c_i$ is a novel trainable parameter controlling the vertical shift.
Note that the shifted \glsxtrlongpl{AF} with horizontal shifts are equivalent to non-shifted variants with biases that are individual for each neuron and not shared in the same tiling pattern as the convolutional kernel \cite{Grelsson2018}.

\subsubsection{Tunable swish (T-swish)}
\label{sec:tswish}
The \gls[prereset]{T-swish} proposed in \cite{Javid2022} is an \gls{AAF} combining the \gls{ELU}, \gls{E-swish} (see \cref{sec:eswish}) and \gls{swish} (see \cref{sec:swish}) as it has trainable parameters for both horizontal and vertical scaling for negative inputs. It is defined as
\begin{equation}
    f(z_i) = \begin{cases}
        z_i, \quad & z_i \geq c_i, \\
        a_iz_i\cdot\sigma(b_iz_i), \quad & z_i < c_i, \\
    \end{cases}
\end{equation}
where $a_i$, $b_i$, and $c_i$ are either fixed or trainable parameters for each neuron $i$ \cite{Javid2022}.

\subsubsection{Rectified parametric sigmoid unit (RePSU)}
\label{sec:repsu}
The \gls[prereset]{RePSU} is an \gls{AAF} proposed in \cite{Atto2023}; it consists of a linear combination of two components --- \gls{RePSKU} and \gls{RePSHU}. It is defined as

\begin{equation}
    f(z_i)= a_i \mathrm{RePSKU}_{b_i,c_i,d_i,e_i}\left(z_i\right) + \left(1-a_i\right)\mathrm{RePSHU}{b_i,c_i,d_i,e_i}\left(z_i\right),
\end{equation}
where
\begin{equation}
    \mathrm{RePSKU}_{b_i,c_i,d_i,e_i}\left(z_i\right)=\begin{cases}
                 \frac{z_i - b_i}{1+\exp\left(-\mathrm{sgn}\left(z_i-c_i\right)\left(\frac{\left|z_i-c_i\right|}{d_i}\right)^{e_i}\right)}, \quad & z_i \geq b_i, \\
                0,  \quad & z_i < b_i, \\
            \end{cases}
\end{equation}
\begin{equation}
    \mathrm{RePSHU}_{b_i,c_i,d_i,e_i}\left(z_i\right)=\begin{cases}
                 2z_i - \mathrm{RePSKU}_{b_i,c_i,d_i,e_i}\left(z_i\right) \quad & z_i \geq b_i, \\
                0,  \quad & z_i < b_i, \\
            \end{cases}
\end{equation}
and $a_i$, $b_i$, $c_i$, $d_i$, and $e_i$ are parameters (common for both $\mathrm{RePSKU}_{b_i,c_i,d_i,e_i}\left(z_i\right)$ and $\mathrm{RePSHU}{b_i,c_i,d_i,e_i}\left(z_i\right)$) \cite{Atto2023}.  The \gls{RePSU} is a generalization of the \gls{SSBS} function \cite{Atto2008} used for image denoising \cite{Atto2023}.

\subsubsection{Parametric deformable exponential linear unit (PDELU)}
\label{sec:pdelu}
An \glsxtrlong{AAF} \gls{PDELU} \cite{Cheng2020} is based on the premise that shifting the mean value of the output closer to zero speeds up the learning \cite{Cheng2020}. The \gls{PDELU} is defined as
\begin{equation}
    f(z_i)=\begin{cases}
                z_i, \quad & z_i\geq 0, \\
                a_i\left(\left[ 1+\left(1-b\right)z_i \right] ^\frac{1}{1-b} - 1\right),  \quad & z_i < 0, \\
            \end{cases}
\end{equation}
where $a_i$ is a trainable parameter for each neuron $i$ and $b$ is a fixed hyperparameter controlling the degree of deformation \cite{Cheng2020}. The \citeauthor{Cheng2020} recommend setting $b=0.9$ \cite{Cheng2020}. The authors found that the MSRA initialization method \cite{He2015} is consistent with \gls{PDELU} \cite{Cheng2020}.
The performance of \gls{PDELU} was empirically shown on the CIFAR-10 and CIFAR-100 datasets \cite{Krizhevsky2009} and on the ImageNet dataset \cite{Russakovsky2015} where it outperformed \gls{ReLU}, \gls{APLU}, \gls{LReLU}, \gls{PReLU}, \gls{SReLU}, \gls{ELU}, \gls{MPELU} (and other) \glsxtrlongpl{AF} \cite{Cheng2020}.

\subsubsection{Elastic exponential linear unit (EELU)}
\label{sec:eelu}
An adaptive variant of the \gls{ELU} function that has a stochastic component was proposed in \cite{Kim2020} --- the \gls{EELU}. The \gls{EELU} combines \gls{EReLU} (see \cref{sec:erelu}) and \gls{MPELU} (see \cref{sec:mpelu}) and is defined as
\begin{equation}
    f(z_i^c)=\begin{cases}
                k_i^c z_i^c, \quad & z_i^c \geq 0, \\
                a^c \left(\exp\left(b^cz_i^c\right)-1\right),  \quad & z_i^c < 0, \\
            \end{cases}
\end{equation}
where $a^c$ and $b^c$ are trainable parameters shared among all neurons of a channel $c$ and $k_i^c$ is a randomly sampled noise parameter for each neuron $i$ in channel $c$ during the training stage \cite{Kim2020} and set to 1 during the testing stage. The $k_i^c$ is sampled coefficient from Gaussian distribution with a random standard deviation that is truncated from 0 to 2; $k_i^c$ is therefore sampled as
\begin{equation}
    k_i^c = \max\left(0, \min\left(s_i^c, 2\right)\right)
\end{equation}
where
\begin{equation}
    s_i^c \sim \mathrm{N}\left(1, \sigma^2\right),
\end{equation}
\begin{equation}
    \sigma \sim \mathrm{U}(0, \epsilon), \epsilon \in (0, 1],
\end{equation}
where $\mathrm{N}\left(1, \sigma^2\right)$ is Gaussian distribution with mean 1 and variance $\sigma^2$, $U$ denotes the uniform distribution \cite{Kim2020}. The $\epsilon$ is a hyperparameter; the authors recommend smaller values, e.g., 0.1 or 0.2 \cite{Kim2020}.

The training algorithm is also modified and works in two steps --- first, the \gls{EELU} parameter and the weights are updated with fixed $k_i^c = 1$, and then weights are updated with random $k_i^c$ and fixed \gls{EELU} parameters \cite{Kim2020}. The authors also recommend using the \gls{MPELU} initialization \cite{Li2018Improving} method \cite{Kim2020}.

\subsubsection{Parametric first power linear unit with sign (PFPLUS)}
\label{sec:pfplus}
The \gls[prereset]{PFPLUS} is an \gls{AAF} proposed in \cite{Duan2022Activation,Duan2022Feature}. It is defined as
\begin{equation}
    f(z_i) = a_iz_i\cdot\left(1-b_iz_i\right)^{\mathrm{H}(z_i)-1},
\end{equation}
where $\mathrm{H}(z_i)$ is Heaviside step function (see \cref{sec:binaryaf})
\begin{equation}
    \mathrm{H}(z_i)=\begin{cases}
                1, \quad &  z_i \geq 0, \\
                0, \quad & z_i < 0. \\
        \end{cases}
\end{equation}

and $a_i > 0$ and $b_i > 0$ are trainable parameters for each neuron $i$ \cite{Duan2022Activation}. For example, the \gls{PFPLUS} is similar to the \gls{ReLU} when $a_i=0.2$ and $b_i=10$ and similar to a linear mapping when $a_i=5$ and $b_i=0.1$ \cite{Duan2022Activation}.

\subsubsection{Parametric variational linear unit (PVLU)}
\label{sec:pvlu}
The \gls[prereset]{PVLU} is an adaptive variant of the \gls{VLU} proposed \cite{Gupta2021Parametric}. It is defined as
\begin{equation}
    f(z_i) = \mathrm{ReLU}\left(z_i\right) + a_i \sin\left(b_i z_i\right) = \max \left(0, z_i\right) + a_i \sin\left(b_i z_i\right),
\end{equation}
where $a_i$ and $b_i$ are trainable parameters \cite{Gupta2021Parametric}.

\subsection{Sigmoid-based adaptive functions}
\label{sec:sigmoid_afs}
Many different \glsxtrlongpl{AAF} based on the sigmoid family were proposed in the literature \cite{Dubey2022}, one of the earliest examples is a \gls{logisticsigmoid} \glsxtrlong{AF} with shape autotuning \cite{Yamada1992}. The function proposed by \citeauthor{Yamada1992} uses a single parameter controlling both the amplitude and the slope of the \glsxtrlong{AF} \cite{Yamada1992, Mishra2017}. The proposed adaptive function is defined as
\begin{equation}
    \label{eq:yamada1992}
    f(z) = 2\frac{1-\exp\left(-az\right)}{a\left(1+\exp\left(-az\right)\right)},
\end{equation}
where $a \in (0, \infty)$ is a learnable parameter \cite{Mishra2017}.

\subsubsection{Generalized hyperbolic tangent}
\label{sec:generalized_hyperbolic_tangent}
The \gls{generalized hyperbolic tangent} \cite{Chen1996} introduces two trainable parameters that control the scale of the \glsxtrlong{AF}:
\begin{equation}
    f(z_i) = \frac{a_i\left(1-\exp(-b_iz_i)\right)}{1+\exp(-b_iz_i)},
\end{equation}
where $a_i$ and $b_i$ are trainable parameters for each neuron $i$ \cite{Yamada1992}. A non-adaptive version with fixed parameters was used for document recognition in \cite{Lecun1998} in order to improve convergence toward the end of the learning session \cite{Lecun1998} (see \cref{sec:stanh}).

\subsubsection{Trainable amplitude}
\label{sec:trainable_amplitude}
A more general approach was introduced in \cite{Trentin2001}, which used networks with a trainable amplitude of \glsxtrlongpl{AF}; the same approach was later used for \glsxtrlongpl{RNN} \cite{Goh2003}. The class of adaptive functions with a trainable amplitude is defined as 
\begin{equation}
    \label{eq:trainable_amplitude}
    f(z_i) = a_i g(z_i) + b_i,
\end{equation}
where $a_i$ and $b_i$ are trainable parameters for each neuron $i$. The $a_i$ determines the trainable amplitude and the $b_i$ trainable offset. These parameters can be either different for each neuron or may be shared by a whole layer or even a whole network \cite{Trentin2001}.

\subsubsection{Adaptive slope sigmoidal function (ASSF)}
\label{sec:assf}
A \gls{ASSF} based on the work of \citeauthor{Yamada1992,Yamada1992Remarks} was used in \cite{Nawi2009,Sharma2011}.  It is defined as
\begin{equation}
    \label{eq:assf}
    f(z) = \sigma\left(a\cdot z\right),
\end{equation}
where $\sigma$ is the \gls{logisticsigmoid} and $a$ is a global trainable parameter \cite{Sharma2011}. The \gls{ASSF} was also rediscovered by \citeauthor{Mercioni2019} in \cite{Mercioni2019}.

\subsubsection{Slope varying activation function (SVAF)}
\label{sec:svaf}
A \gls{SVAF} was proposed in \cite{Bai2009}
\begin{equation}
    \label{eq:svaf}
    f(z) = \tanh\left(a\cdot z\right),
\end{equation}
where $a$ is a global trainable parameter. The slope varying activation function was proposed together with a \gls{BP} modification that has two different learning rates \cite{Bai2009}. The slope varying activation function was implemented as a modification of the \gls{BP} algorithm rather; a different example of modification of the \gls{BP} algorithm resulting in an \glsxtrlong{AAF} is presented in \cite{ChienChengYu2002}.

\subsubsection{TanhSoft}
\label{sec:tanhsoft}
The \gls{TanhSoft} is a family of \glspl{AAF} proposed in \cite{Biswas2020TanhSoft} that combine the \gls{softplus} and \gls{tanh} that contains three notable cases ---  {\gls{TanhSoft}-1}, {\gls{TanhSoft}-2}, and {\gls{TanhSoft}-3} \cite{Biswas2020TanhSoft,Biswas2021TanhSoft}.

The general \gls{TanhSoft} is defined as
\begin{equation}
    f(z_i) = \tanh\left(a_iz_i+b_i\exp\left(c_iz_i\right)\right)\ln\left(d_i+\exp\left(z_i\right)\right),
\end{equation}
where $a_i$, $b_i$, $c_i$, and $d_i$ are either trainable or fixed parameters \cite{Biswas2020TanhSoft}; $a_i \in (-\infty,1]$, $b_i \in [0,\infty)$, $c_i \in (0,\infty)$, and $d_i \in [0,1]$ \cite{Biswas2020TanhSoft}.

The first \gls{AF}, named \gls{TanhSoft}-1, is defined as
\begin{equation}
    f(z_i) = \tanh\left(a_iz_i\right)\ln\left(1+\exp\left(z_i\right)\right),
\end{equation}
where $a_i$ is a trainable parameter \cite{Biswas2020TanhSoft,Biswas2021TanhSoft}; it can be obtained from the general \gls{TanhSoft} by setting $b_i=0$ and $d_i=1$ \cite{Biswas2020TanhSoft}. The second \gls{AF} from \cite{Biswas2021TanhSoft}, \gls{TanhSoft}-2, is defined as
\begin{equation}
    f(z_i) = z_i \tanh\left(b_i\exp\left(c_iz_i\right)\right),
\end{equation}
where $b_i$ and $c_i$ are trainable parameters \cite{Biswas2020TanhSoft, Biswas2021TanhSoft}. The {\gls{TanhSoft}-2} can be obtained from the general \gls{TanhSoft} by setting $a_i=0$ and $d_i=0$ \cite{Biswas2020TanhSoft}. The last \gls{AF} from \cite{Biswas2021TanhSoft}, \gls{TanhSoft}-3, is defined as
\begin{equation}
    f(z_i) = \ln\left(1+\exp\left(z_i\right)\tanh\left(a_iz_i\right)\right),
\end{equation}
where $a_i$ is a trainable parameter \cite{Biswas2021TanhSoft}. It can be obtained from the general \gls{TanhSoft} by setting $b_i=0$ and $d_i=1$.

\subsubsection{Parametric sigmoid (psigmoid)}
\label{sec:psigmoid}
An adaptive variant of \gls{logisticsigmoid} named \gls[prereset]{psigmoid}\footnote{Not to be confused with \glsxtrfull{PSF} from \cref{sec:psf}.} was proposed in \cite{Ying2021,Ozbay2010}.\footnote{It seems that this \gls{AAF} was first proposed in 2010 in \cite{Ozbay2010} and then independently in 2021 in \cite{Ying2021}.} Similarly as in \gls{generalized hyperbolic tangent}, it introduces two scaling parameters to a \gls{logisticsigmoid}:
\begin{equation}
    \label{eq:psigmoid}
    f(z_i) = a_i\sigma\left(b\cdot z_i\right),
\end{equation}
where $a_i$ is a trainable parameter for each neuron or channel $i$ and $b$ is a global trainable parameter \cite{Ying2021}.

\subsubsection{Parametric sigmoid function (PSF)}
\label{sec:psf}
A \gls{PSF} is a continuous, differentiable, and bounded function proposed in \cite{Chandra2004, Singh2003}\footnote{\cite{Singh2003} contains the definition equivalent to $f(z) = \mathrm{PSF}\left(\frac{z}{2}\right)$.} and is defined as
\begin{equation}
    \mathrm{PSF}(z) = \frac{1}{\left(1+\exp(-z)\right)^m},
\end{equation}
where $m$ is a global trainable parameter \cite{Dubey2022,Chandra2004Case}. The parameter $m$ controls the slope of the sigmoid and the position of the maximum derivative; the envelope of the relevant derivatives for different values of $m$ is also a sigmoid function \cite{Chandra2004}. The larger values of $m$ improve the gradient flow \cite{Dubey2022}. The \gls{PSF} is only one instance of a larger class of \glsxtrlongpl{AF} proposed in \cite{Chandra2003}.

\subsubsection{Slope and threshold adaptive activation function with tanh function (STAC-tanh)}
\label{sec:stactanh}
The \gls[prereset]{STAC-tanh} was proposed in \cite{Zhang2021ANovel}. It is basically a $\tanh$ based equivalent of the \gls{improvedlogisticsigmoid} with adaptive parameters. It is defined as
\begin{equation}
    f(z_i) = \begin{cases}
        \tanh{-a_i} + b_i\left(z_i+a_i\right),  \quad & z_i < -a_i, \\
        \tanh{z_i},  \quad & -a_i \leq z_i \leq a_i, \\
        \tanh{a_i} + b_i\left(z_i-a_i\right),  \quad & z_i > a_i, \\
    \end{cases}
\end{equation}
where $a_i$ and $b_i$ are trainable parameters \cite{Zhang2021ANovel}.

\subsubsection{Generalized Riccati activation (GRA)}
\label{sec:gra}
The \gls[prereset]{GRA} is an adaptive variant of a \gls{sigmoid} \gls{AF} proposed in \cite{Protonotarios2022}. It is defined as
\begin{equation}
    f(z_i) = 1 - \frac{a_i}{a_i+\left(1+b_iz_i\right)^{c_i}},
\end{equation}

where $a_i$, $b_i$, and $c_i$ are adaptive parameters --- $b_i > 0$ and $c_i > 0$ \cite{Protonotarios2022}.

\subsection{Adaptive sigmoid-weighted linear units}
\label{sec:adaptive_silus}
There are several \glspl{AF} that are based on the \gls{SiLU} but have an adaptive parameter; the most common example is the \gls{swish} \gls{AF}, but there are also other popular functions based on the same principle.

\subsubsection{Swish}
\label{sec:swish}
A \gls{swish} \glsxtrlong{AF} \cite{Ramachandran2017} is an adaptive variant of the \gls{SiLU} \cite{Elfwing2018} (see \cref{sec:silu}); it is also the member of the \glsxtrshort{LAAF} class (see \cref{sec:laaf}):
\begin{equation}
    f(z_i) = z_i\cdot\sigma(a_iz_i),
\end{equation}
where $\sigma(z)$ is the \gls{logisticsigmoid}, $a_i$ is either a fixed hyperparameter or a trainable parameter \cite{Ramachandran2017}. The \gls{swish} has an output range of $(-\infty, \infty)$ \cite{Dubey2022}. The parameter $a_i$ controls the amount of non-linearity the \gls{swish} activation has \cite{Dubey2022}. 
The \gls{swish} might also be considered a member of the family of \glspl{ACON} \cite{Ma2021Activate}; it is then named \gls{ACON-A}. The \gls[prereset]{PSiLU} is another name for the \gls{swish} activation used in \cite{Dai2022}.

\subsubsection{Adaptive hybrid activation function (AHAF)}
\label{sec:ahaf}
A \gls{swish} variant with vertical scaling was proposed in \cite{Bodyanskiy2022} under the name \gls[prereset]{AHAF}. It is defined as
\begin{equation}
    f(z_i) = a_iz_i\cdot\sigma(b_iz_i),
\end{equation}
where $a_i$ and $b_i$ are trainable parameters \cite{Bodyanskiy2022}.

\subsubsection{Parametric shifted SiLU (PSSiLU)}
\label{sec:pssilu}
The \gls[prereset]{PSSiLU} is a \gls{swish} based \gls{AAF} proposed in \cite{Dai2022}. It is defined as
\begin{equation}
    f(z_i) = \frac{z_i\cdot\left(\sigma(a_iz_i)-b_i\right)}{1-b_i},
\end{equation}
where $a_i$ and $b_i$ are trainable parameters \cite{Dai2022}.

\subsubsection{E-swish}
\label{sec:eswish}
\gls{E-swish} \cite{Alcaide2018} is an \gls{AAF} inspired by the \gls{swish} \cite{Ramachandran2017} \glsxtrlong{AF} (see \cref{sec:swish}); the \gls{E-swish} has a scaling parameter that allows for vertical scaling of the \glsxtrlong{AF} \cite{Alcaide2018}. The name of the \glsxtrlong{AF} is not chosen well as the \gls{E-swish} is rather extending the \gls{SiLU} (see \cref{sec:silu}) and not \gls{swish} which is its adaptive variant.\footnote{Calling the \gls{SiLU} as \gls{swish} is quite common in the literature, e.g., \gls{exponential_swish}, \gls{generalized_swish}, and \glsxtrshort{TS-swish}.} The function is defined as
\begin{equation}
    f(z) = az\cdot\sigma(z),
\end{equation}
where $\sigma(z)$ is the \gls{logisticsigmoid}  and $a$ is a preset parameter \cite{Alcaide2018} --- however, the parameter $a$ is considered to be trainable in review \cite{Dubey2022}. \citeauthor{Alcaide2018} recommends setting $a \in [1,2]$ to avoid exploding gradients that are hypothesized to more likely occur for higher values of $a$ \cite{Alcaide2018}. The \gls{E-swish} was found to outperform the \gls{SiLU} (called \gls{swish} in the paper) on the the MNIST \cite{LiDeng2012}, CIFAR-10 and CIFAR-100 \cite{Krizhevsky2009} datasets using the Wide ResNet (WRN) \cite{Zagoruyko2016} architecture \cite{Alcaide2018}.

\subsubsection{ACON-B}
\label{sec:aconb}
The \gls{ACON} family conists of \gls{swish} \gls{AF} and several extensions; one is named \gls{ACON-B} and is defined as
\begin{equation}
    f(z_i) = \left(1-b_i\right)z_i\cdot\sigma\left(a_i\left(1-b_i\right)z_i\right)+b_iz_i,
\end{equation}
where $a_i$ and $b_i$ are trainable parameters \cite{Ma2021Activate}. The $b_i$ is initalized to 0.25 and $a_i$ to 1.\footnote{There is no initial value for $a_i$ in \gls{ACON-B} mentioned explicitly in \cite{Ma2021Activate}; however, there is one for its extension \gls{ACON-C}.}

\subsubsection{ACON-C}
\label{sec:aconc}
The \gls{ACON-C} is another member of the \gls{ACON} family from \cite{Ma2021Activate}. It is defined as
\begin{equation}
    f(z_i) = \left(c_i-b_i\right)z_i\cdot\sigma\left(a_i\left(c_i-b_i\right)z_i\right)+b_iz_i,
\end{equation}
where $a_i$, $b_i$, and $c_i$ are trainable parameters \cite{Ma2021Activate,Zhang2023Traffic}. \Citeauthor{Ma2021Activate} used initial values $a_i=1$, $b_i=0$, and $c_i=1$ in \cite{Ma2021Activate}.

\Citeauthor{Ma2021Activate} also proposed a general extension to the \gls{ACON} family named \gls{MetaACON} which uses a small \gls{NN} to determine the value of the parameter $a_i$; they used the variant \gls{ACON-C} for the experiments with \gls{MetaACON} resulting in \gls{MetaACON-C}\footnote{The implementation of \gls{MetaACON-C} and other \glspl{AF} from the \gls{ACON} family is available at \url{https://github.com/nmaac/acon}.} \cite{Ma2021Activate}. The \gls{MetaACON} was used to improve YOLOv7 \cite{Wang2023YOLOv7} in \cite{Ye2023}. \Citeauthor{Kan2022} extented the \gls{ACON} \glspl{AF} into an \gls{AF} they named {CBAC}\footnote{No further description is provided in \cite{Kan2022}.} \cite{Kan2022}. The \glspl{ACON} were used, for example, in \cite{Ma2021Activate,Ye2023,Zhang2023Traffic,Tu2022,Xi2022,Li2022HARNUNet,Wu2022AFast,Niu2022,Zhang2022ALightweight,Cao2022Smaller,Jia2023AADHYOLOv5,Xu2023Unmanned,He2023Method, Qin2023,Chen2023LODNU,Zhao2023Lightweight}. The \gls{1Dmeta-ACON} is a \gls{MetaACON} extension proposed in \cite{Liu2023WindTurbine}.

\subsubsection{Parameterized self-circulating gating unit (PSGU)}
\label{sec:psgu}
The \gls[prereset]{PSGU} \cite{Li2021PSGU} is related to the \gls{LiSHT} and \gls{GTU} \glsxtrlongpl{AF} as it is basically a \gls{LiSHT} with gated input with learnable scaling parameter. It is defined as
\begin{equation}
    f(z_i) = z_i \cdot \tanh\left(a_i\sigma\left(z_i\right)\right),
\end{equation}
where $a_i$ is a learnable parameter and $\sigma(z)$ is the \gls{logisticsigmoid} function \cite{Li2021PSGU}. \Citeauthor{Li2021PSGU} also propose a novel initialization method for \glspl{NN} with the \gls{PSGU} \gls{AF} and show that it is more suitable for the use with \gls{PSGU} than other common methods \cite{Li2021PSGU}. The \gls{PSGU} is shown to outperform \gls{ReLU}, \gls{mish}, \gls{swish}, \gls{PATS} and \gls{GELU} using various \gls{NIN} and ResNet architectures \cite{Li2021PSGU}. The \gls{PSGU} was also proposed in \cite{Mercioni2020Improving} under the name \gls[prereset]{TSReLUl} as the adaptive variant of \gls{TSReLU}. \Citeauthor{Mercioni2020Improving} used $a_i = 0.5$ as the initial value \cite{Mercioni2020Improving}.

\subsubsection{Tangent-bipolar-sigmoid ReLU learnable (TBSReLUl)}
\label{sec:tbsrelul}
Similarly as \gls{TSReLUl} is an adaptive variant of \gls{TSReLU}, the \gls[prereset]{TBSReLUl} \cite{Mercioni2020Improving} is an adaptive variant of \gls{TBSReLU}. This variant is defined as
\begin{equation}
    f(z) = z_i\cdot \tanh\left(a_i\frac{1-\exp\left(-z_i\right)}{1+\exp\left(-z_i\right)}\right).
\end{equation}
where $a_i$ is a trainable parameter \cite{Mercioni2020Improving}. \Citeauthor{Mercioni2020Improving} used $a_i = 0.5$ as the initial value \cite{Mercioni2020Improving}.

\subsubsection{PATS}
\label{sec:pats}
The \gls{AF} named \gls{PATS}\footnote{Not an abbreviation.} \cite{Zheng2020} is very similar to \gls{PSGU}, but it uses arctan and a random scaling parameter instead of the \gls{tanh} and the adaptive parameter in \gls{PSGU}. It is defined as
\begin{equation}
    f(z_i) = z_i \tan^{-1}\left(a_i \pi \sigma\left(z_i\right)\right),
\end{equation}
where $\sigma(z)$ is the \gls{logisticsigmoid} function and
\begin{equation}
    a_i \sim \mathrm{U}\left(l,u\right),
\end{equation}
is sampled during training\footnote{Unfortunately, the author did not specify what happens during the test phase in \cite{Zheng2020}, one can only assume that the expected value is used.} from the uniform distribution with bounds $l$ and $u$ such that $0 < l < u < 1$ \cite{Zheng2020}. The authors experimented with fixed, deterministic values of $a_i \in \{\frac{1}{4}, \frac{1}{2}, \frac{5}{8}, \frac{3}{4}\}$ --- the value $\frac{5}{8}$ led to lowest test error on the CIFAR-10 \cite{Krizhevsky2009}; they also deemed that suitable values for $l$ and $u$ are $\frac{1}{2}$ and ${3}{4}$ respectively \cite{Zheng2020}. However, only fixed variant with $a_i = \frac{5}{8}$ was used in the follow-up works such as \cite{Lin2023LowArea}.

\subsubsection{Adaptive quadratic linear unit (AQuLU)}
\label{sec:aqulu}
The \gls[prereset]{AQuLU} is an adaptive \gls{SiLU} variant proposed in \cite{Wu2023TheAdaptive}; it is defined as
\begin{equation}
    f(z_i) = \begin{cases}
        z_i,  \quad & z_i \geq \frac{1-b_i}{a_i}, \\
        a_iz_i^2+b_iz_i,  \quad & -\frac{b_i}{a_i} \geq z_i < \frac{1-b_i}{a_i}, \\
        0,  \quad & z_i < -\frac{b_i}{a_i}, \\
    \end{cases}
\end{equation}
where $a_i$ and $b_i$ are trainable parameters for each neuron $i$ \cite{Wu2023TheAdaptive}.

\subsubsection{Sinu-sigmoidal linear unit (SinLU)}
\label{sec:sinlu}
Another adaptive \gls{SiLU} variant is the \gls[prereset]{SinLU}, which adds an adaptive term using the sine function to the linear part of the \gls{SiLU} \cite{Paul2022}. The \gls{SinLU} is defined as
\begin{equation}
    f(z_i) = \left(z_i + a_i \sin\left(b_i z_i\right)\right)\cdot\sigma(z_i),
\end{equation}
where $\sigma(z_i)$ is the \gls{logisticsigmoid} function and $a_i$ and $b_i$ are trainable parameters for each neuron $i$ \cite{Paul2022}.

\subsubsection{ErfAct}
\label{sec:erfact}
An \gls{AAF} based on the Gauss error function was proposed in \cite{Biswas2022}. The \gls{AAF} is named \gls{ErfAct} and is defined as
\begin{equation}
    f(z_i) = z_i\cdot\mathrm{erf}\left(a_i\exp\left(b_iz_i\right)\right),
\end{equation}
where $a_i$ and $b_i$ are trainable parameters for each  neuron $i$ and $\mathrm{erf}(x)$ is the Gauss error function \cite{Biswas2022}.

\subsubsection{Parametric serf (pserf)}
\label{sec:pserf}
An adaptive version of the \gls{serf} \gls{AF} named \gls[prereset]{pserf} was proposed in \cite{Biswas2022}. It is defined as
\begin{equation}
    f(z_i) = z_i\cdot\mathrm{erf}\left(a_i\ln\left(1+\exp\left(b_iz_i\right)\right)\right),
\end{equation}
where $a_i$ and $b_i$ are trainable parameters for each  neuron $i$ and $\mathrm{erf}(x)$ is the Gauss error function \cite{Biswas2022}.

\subsubsection{Swim}
\label{sec:swim}
The \gls{swim} is an adaptive variant of the \gls{PFLU} (see \cref{sec:pflu}) independently proposed in \cite{Abdool2023}. It is defined as
\begin{equation}
    f(z_i) = z_i\cdot\frac{1}{2}\left(1+\frac{a_iz_i}{\sqrt{1+{z_i}^2}}\right),
\end{equation}
where $a_i$ is either fixed or trainable parameter for each neuron $i$ \cite{Abdool2023}. \Citeauthor{Abdool2023} used fixed $a_i=0.5$ in their experiments in \cite{Abdool2023}.

\subsection{Tuned softmax (tsoftmax)}
\label{sec:tsoftmax}
A \gls{softmax} (see \cref{sec:softmax}) variant named \gls[prereset]{tsoftmax} was proposed in \cite{Bhuvaneshwari2021}; it is defined as

\begin{equation}
    f(z_j) = \frac{\int\exp\left(z_j\right)}{\sum_{k=1}^N\int\exp\left(z_k\right)} + c,
\end{equation}
where $f(z_j)$ is the output of a neuron $j$ in a \gls{softmax} layer consisting of $N$ neurons and $c$ is an adaptive parameter \cite{Bhuvaneshwari2021}.

\subsection{Generalized Lehmer softmax (glsoftmax)}
\label{sec:glsoftmax}
The \gls[prereset]{glsoftmax} is a \gls{softmax} variant proposed in \cite{Terziyan2022}.
It is defined as
\begin{equation}
    f(z_j) = \frac{\exp\left(\mathrm{LNORM}\left(z_j\right)\right)}{\sum_{k=1}^N\exp\left(\mathrm{LNORM}\left(z_k\right)\right)},
\end{equation}
where $\mathrm{LNORM}\left(z_j\right)$ is a generalized Lehmer-based Z-score-like normalization with four trainable parameters $a_i$, $b_i$, $c_i$, and $d_i$ defined in \cite{Terziyan2022}:
\begin{equation}
\mathrm{LNORM}\left(z_i\right) = \frac{z_i - M_{a_i,b_i}}{\mathrm{GLM}_{c_i,d_i}\left(\vec{z}-M_{a_i,b_i}\right)},
\end{equation}
\begin{equation}
    M_{a_i,b_i}=\mathrm{GLM}_{a_i,b_i}\left(\vec{z}\right),
\end{equation}
\begin{equation}
    \mathrm{GLM}_{\alpha,\beta}\left(\vec{x}\right) = \frac{\ln\left(\frac{\sum_{k=1}^N\alpha^{(\beta+1)x_k}}{\sum_{k=1}^N\alpha^{\beta x_k}}\right)}{\ln\left(\alpha\right)},
\end{equation}
$\vec{x}$ is a vector of elements $x_k$, $k=1,\ldots,N$ and $\vec{z}-M_{a_i,b_i}$ represents a vector with elements $z_k - M_{a_i,b_i}$, $k=1,\ldots,N$ \cite{Terziyan2022}.

\subsection{Generalized power softmax (gpsoftmax)}
\label{sec:gpsoftmax}
The \gls[prereset]{gpsoftmax} is another \gls{softmax} variant proposed in \cite{Terziyan2022}.
It is defined as
\begin{equation}
    f(z_j) = \frac{\exp\left(\mathrm{PNORM}\left(z_j\right)\right)}{\sum_{k=1}^N\exp\left(\mathrm{PNORM}\left(z_k\right)\right)},
\end{equation}
where $\mathrm{PNORM}\left(z_j\right)$ is a generalized power-based Z-score-like normalization with four trainable parameters $a_i$, $b_i$, $c_i$, and $d_i$ defined in \cite{Terziyan2022}:
\begin{equation}
\mathrm{PNORM}\left(z_i\right) = \frac{z_i - M_{a_i,b_i}}{\mathrm{GPM}_{c_i,d_i}\left(\vec{z}-M_{a_i,b_i}\right)},
\end{equation}
\begin{equation}
    M_{a_i,b_i}=\mathrm{GPM}_{a_i,b_i}\left(\vec{z}\right),
\end{equation}
\begin{equation}
    \mathrm{GPM}_{\alpha,\beta}\left(\vec{x}\right) = \frac{\ln\left(\sum_{k=1}^N \alpha^{\beta x_k}\right) - \ln\left(N\right)}{\beta \ln\left(\alpha\right)},
\end{equation}
$\vec{x}$ is a vector of elements $x_k$, $k=1,\ldots,N$ and $\vec{z}-M_{a_i,b_i}$ represents a vector with elements $z_k - M_{a_i,b_i}$, $k=1,\ldots,N$ \cite{Terziyan2022}.

\subsection{Adaptive radial basis function (ARBF)}
\label{sec:arbf}
The \gls[prereset]{ARBF} was used in \cite{Jiang2021Multilayer}. It is defined as
\begin{equation}
    f(z_i) \exp\left(-\frac{\left(z_i-a_i\right)^2}{2b_i^2}\right),
\end{equation}
where $a_i$ and $b_i$ are adaptive parameters for each neuron $i$ \cite{Jiang2021Multilayer}. The parameter $a_i$ controls the center while the parameter $b_i$ controls the width \cite{Jiang2021Multilayer}.

\subsection{Parametric Gaussian error linear unit (PGELU)}
\label{sec:pgelu}
The \gls{AAF} named \gls[prereset]{PGELU} was proposed in \cite{Duan2024} as the result of noise injection. It is an \gls{GELU} (see \cref{sec:gelu}) adaptive variant defined as
\begin{equation}
    f(z_i) = z\cdot\Phi\left(\frac{z}{a}\right),
\end{equation}
where $\Phi\left(z\right)$ is the standard Gaussian \gls{CDF} and $a$ is a global learnable parameter representing the \gls{RMS} noise \cite{Duan2024}.

\subsection{Parametric flatted-T swish (PFTS)}
\label{sec:pfts}
A \gls{PFTS} \cite{Chieng2020} is an adaptive extension of the \gls{FTS} (see \cref{sec:fts}); \gls{PFTS} is identical to \gls{FTS} except for that the parameter $T$ is adaptive --- i.e.:
\begin{equation}
    f(z_i) = \mathrm{ReLU}(z_i) \cdot \sigma(z_i) + T_i = \begin{cases}
        \frac{z_i}{1+\exp(-z_i)}+T_i,  \quad & z_i \geq 0, \\
        T_i,  \quad & z_i < 0, \\
    \end{cases}
\end{equation}
where $T_i$ is a trainable parameter for each neuron $i$ \cite{Chieng2020}; the parameter $T_i$ is initialized to the value -0.20 \cite{Chieng2020}.

\subsection{Parametric flatten-p mish (PFPM)}
\label{sec:pfpm}
The \gls[prereset]{PFPM} is an \gls{AAF} proposed in \cite{Mondal2022ANovel}; it is defined as
\begin{equation}
    f(z_i) = \begin{cases}
        z_i \tanh\left(\ln\left(1+\exp\left(z_i\right)\right)\right)+p_i,  \quad & z_i \geq 0, \\
        p_i,  \quad & z_i < 0, \\
    \end{cases}
\end{equation}
where $p_i$ is a trainable parameter \cite{Mondal2022ANovel}.

\subsection{Gaussian error unit (GEU)}
\label{sec:geu}
The \gls{AAF} named \gls[prereset]{GEU} was proposed in \cite{Duan2024} as the result of noise injection. It is defined as
\begin{equation}
    f(z_i) = \Phi\left(\frac{z}{a}\right),
\end{equation}
where $\Phi\left(z\right)$ is the standard Gaussian \gls{CDF} and $a$ is a global learnable parameter representing the \gls{RMS} noise \cite{Duan2024}. The \gls{GEU} multiplied by $z$ becomes the \gls{PGELU} (see \cref{sec:pgelu}).

\subsection{Scaled-gamma-tanh activation function (SGT)}
\label{sec:sgt}
The \gls[prereset]{SGT} \gls{AF} is a piecewise polynomial function proposed in \cite{Khagi2022}. It is defined as
\begin{equation}
    f(z_i)=\begin{cases}
        a z_i^{b_i}, \quad & z_i \geq 0, \\
        c z_i^{d_i}, \quad & z_i < 0, \\
    \end{cases}
\end{equation}
where $a$ and $c$ are fixed, predefined parameters and $b_i$ and $c_i$ are trainable parameters for each neuron or filter $i$ \cite{Khagi2022}.

\subsection{RSign}
\label{sec:rsign}
An adaptive variant of the sign function was used in \cite{Liu2020ReActNet}. It is called \gls{RSign} and is defined as
\begin{equation}
    f(z_i) = \begin{cases}
        1,  \quad & z_i \geq a_c, \\
        -1,  \quad & z_i < a_c, \\
    \end{cases}
\end{equation}
where $a_c$ is an adaptive threshold for each channel \cite{Liu2020ReActNet}. An extension was used in \cite{Ding2022IENet}, where \citeauthor{Ding2022IENet} used multiple \gls{RSign} functions for each channel.

\subsection{P-SIG-RAMP}
\label{sec:psigramp}
An \gls{AAF} combining the \gls{logisticsigmoid} and \gls{ReLU} was proposed in \cite{Jie2021} under the name \gls{P-SIG-RAMP}. The \gls{P-SIG-RAMP} is defined as
\begin{equation}
    f(z_i) = a_i \sigma\left(z_i\right) + \left(1-a_i\right)\cdot\begin{cases}
        1,  \quad & z_i \geq \frac{1}{2b_i}, \\
        b_iz_i+\frac{1}{2},  \quad & -\frac{1}{2b_i} < z_i < \frac{1}{2b_i}, \\
        0,  \quad & z_i \leq -\frac{1}{2b_i}, \\
    \end{cases}
\end{equation}
where $a_i \in [0,1]$ and $b_i$ are trainable parameters \cite{Jie2021}.

\subsection{Locally adaptive activation function (LAAF)}
\label{sec:laaf}
A general class of slope varying functions called \gls[prereset]{LAAF} was proposed in \cite{Jagtap2020,Jagtap2020Locally}:
\begin{equation}
    f(z_i) = g(a_i\cdot z_i),
\end{equation}
where $a_i$ is a trainable parameter for each neuron $i$ and $g$ is any activation function; \citeauthor{Jagtap2020} used \gls{logisticsigmoid}, \gls{tanh}, \gls{ReLU}, and \gls{LReLU} as $g$ in their \glspl{LAAF} in \cite{Jagtap2020}.
The corresponding activations are thus given by
\begin{equation}
    f(z_i) = \sigma(a_i z_i) = \frac{1}{1 + \exp\left(-a_i z_i\right)},
\end{equation}
\begin{equation}
    f(z_i) = \tanh(a_i z_i) = \frac{\exp\left(a_i z_i\right) - \exp\left(-a_i z_i\right)}{\exp\left(a_i z_i\right) + \exp\left(-a_i z_i\right)},
\end{equation}

\begin{equation}
    f(z_i) = \mathrm{ReLU}(a_iz_i) = \max\left(0,a_i z_i\right),
\end{equation}
and 
\begin{equation}
    f(z_i) = \mathrm{LReLU}(a_iz_i) = \max\left(0,a_i z_i\right) - b \max\left(0,-a_i z_i\right),
\end{equation}
where $b$ is the \gls{LReLU} leakiness parameter \cite{Jagtap2020}. To accelerate the convergence, \citeauthor{Jagtap2020} add additional fixed parameter to the expression:
\begin{equation}
    f(z_i) = g(na_iz_i),
\end{equation}
where $n>1$ is a fixed parameter \cite{Jagtap2020}.
It was found that this additional parameter improves both the convergence rate and the solution accuracy \cite{Jagtap2020}.

\subsubsection{Adaptive slope hyperbolic tangent}
\label{sec:astanh}
A \gls{tanh} \glsxtrlong{AF} with adaptive slope was used in an \gls{MLP} architecture in \cite{Kapoor2021}. The used \glsxtrlong{AF} is defined as
\begin{equation}
    f(z_i) = \tanh\left(a_iz_i\right),
\end{equation}
where $a_i$ is a trainable parameter for each neuron $i$.

\subsubsection{Parametric scaled hyperbolic tangent (PSTanh)}
\label{sec:pstanh}
A parametric \glsxtrlong{AF} similar to the \gls{swish} but based on the \gls{tanh} function instead of the \gls{logisticsigmoid} called \gls{PSTanh} was proposed in \cite{Adu2021}. It is defined as
\begin{equation}
    f(z_i) = z_i \cdot a_i\left(1+\tanh\left(b_iz_i\right)\right),
\end{equation}
where $a_i$ and $b_i$ are trainable parameters for each neuron $i$ \cite{Adu2021}. The function is also very similar to the \gls{PTELU} (see \cref{sec:ptelu}) as for $z_i > 0$ and $a_i \approx 1$, the output is close to $z_i$ \cite{Adu2021} (the exact distance depends on the parameters $a_i$ and $b_i$).

\subsubsection{Scaled sine-hyperbolic function (SSinH)}
\label{sec:ssinh}
An \gls{AF} similar to \gls{PSTanh} is the \gls[prereset]{SSinH} \cite{Husain2021}; it is defined as
\begin{equation}
    f(z_i) = a_i \sinh\left(b_iz_i\right),
\end{equation}
where $a_i$ and $b_i$ are trainable scaling parameters and $\sinh$ is the hyperbolic sine \cite{Husain2021}.

\subsubsection{Scaled exponential function (SExp)}
\label{sec:sexp}
\Citeauthor{Husain2021} also proposed \gls[prereset]{SExp} along with the \gls{SSinH} in \cite{Husain2021}. It is defined as
\begin{equation}
    f(z_i) = a_i \left(\exp\left(b_iz_i\right)-1\right),
\end{equation}
where $a_i$ and $b_i$ are trainable scaling parameters and $\sinh$ is the hyperbolic sine \cite{Husain2021}.

\subsubsection{Logmoid activation unit (LAU)}
\label{sec:lau}
A learnable \gls{LAU} was proposed in \cite{Zhou2023Lau, Zhou2023TwoParameter}; which utilise two learnable parameters $a_l$ and $b_l$ for each network layer $l$
\begin{equation}
    f(z_{i,l}) = z \ln\left(1+a_l\sigma\left(b_l\cdot z_{i,l}\right)\right),
\end{equation}
where $z_{i,l}$ is the output of the neuron $i$ in layer $l$ without the \glsxtrlong{AF} and $\sigma$ is the \gls{logisticsigmoid} \cite{Zhou2023TwoParameter}. The author used initial values of the parameters $a_l=b_l=1$ for each network's layer $l$ and trained these parameters together with the rest of the network's weights \cite{Zhou2023TwoParameter}.

\subsubsection{Cosinu-sigmoidal linear unit (CosLU)}
\label{sec:coslu}
The \gls{CosLU} is an \glsxtrlong{AAF} proposed in \cite{Pishchik2023} that is based on the \gls{logisticsigmoid}. It is defined as
\begin{equation}
    f(z_i) = \left(z + a_i \cos\left(b_iz_i\right)\right)\sigma\left(z_i\right),
\end{equation}
where $a_i$ and $b_i$ are trainable parameters for neuron $i$ and $\sigma(z_i)$ is the \gls{logisticsigmoid} function \cite{Pishchik2023}. The cosine amplitude is controlled by the parameter $a_i$, whereas its frequency is controlled by the parameter $b_i$.

\subsubsection{Adaptive Gumbel (AGumb)}
\label{sec:AGumb}
An \glsxtrlong{AF} \gls{AGumb} is based approach of viewing \glsxtrlongpl{AF} as a combination of unbounded and bounded components where the bounded component is based upon a cumulative distribution function of a continuous distribution \cite{Farhadi2019}. While the \gls{logisticsigmoid} activation is a \gls{CDF} of the symmetric logistic distribution, the \gls{AGumb} is based on the Gumbel distribution \cite{Farhadi2019}. It is defined as
\begin{equation}
    f(z_i) = 1 - \left(1+ a_i \cdot \exp\left(z_i\right)\right)^{-\frac{1}{a_i}},
\end{equation}
where $a_i \in \mathbb{R}^+$ is trainable parameter for each neuron $i$ \cite{Farhadi2019}.

\subsection{Shape autotuning adaptive activation function (SAAAF)}
\label{sec:saaaf}
The \gls[prereset]{SAAAF}\footnote{\Citeauthor{Zhou2021Shape} named the function as \textit{shape autotuning activation function} but the resulting abbreviation \glsxtrshort{SAAF} is already taken by \glsxtrlong{SAAF} (see \cref{sec:aplu}). Since the proposed function is an \gls{AAF}, we term it as such to avoid the abbreviation collision.} is an \gls{AAF} proposed in \cite{Zhou2021Shape}. It is defined as
\begin{equation}
    f(z_i) = \frac{z_i}{\frac{z_i}{a_i}+\exp\left(-frac{z_i}{b_i}\right)},
\end{equation}
where $a_i \geq 0$ and $b_i \geq 0$ are trainable parameters for neuron $i$ and $0 < \frac{b_i}{a_i} < \mathrm{e}$ \cite{Zhou2021Shape}.

\subsection{Noisy activation functions}
\label{sec:noisy_af}
Stochastic variants of saturing \glsxtrlongpl{AF} such as the \gls{logisticsigmoid} or \glsxtrlong{tanh} were proposed in \cite{Gulcehre2016} where an additional noise is injected to the \glsxtrlong{AF} when it operates in the saturation regimes \cite{Gulcehre2016}. The noisy \glsxtrlong{AF} is defined as
\begin{equation}
    \label{eq:noisy_af}
    f(z_i) = a \mathrm{h}(z_i) + \left(1-a\right)\mathrm{u}(z_i) -\mathrm{sgn}(z_i)\mathrm{sgn}(1-a)c\bigg(\sigma\Big(p_i\left(\mathrm{h}(z_i)-\mathrm{u}(z_i)\right)\Big)\bigg)^2\epsilon,
\end{equation}
where $\mathrm{h}(z_i)$ is any saturating \glsxtrlong{AF} such as hard-tanh or hard-sigmoid, $\mathrm{u}(z_i)$ is its linearization using first-order Taylor expansion around zero, $c$ is a hyperparameter changing the scale of the standard deviation of the noise, $p_i$ is a trainable parameter adjusting the magnitude of the noise for each neuron $i$, $a$ is a hyperparameter influencing the mean of the added term, and $\sigma(x)$ is the \gls{logisticsigmoid} function \cite{Gulcehre2016}. $\epsilon$ is the added noise; it is defined as $\epsilon = |\xi|$ if the noise term $\xi$ is sampled from half-normal distribution and as $\epsilon = \xi$ if the noise term $\xi$ is sampled from normal distribution with mean 0 and variance 1 \cite{Gulcehre2016}.

\Citeauthor{Gulcehre2016} also experimented with adding noise to the input of the \glsxtrlong{AF}, resulting in an \glsxtrlong{AF} defined as
\begin{equation}
    f(z_i) = \mathrm{h}\left(z_i+s(z_i)\epsilon\right),
\end{equation}
where $s(z_i)$ is either fixed parameter $s(z_i) = b$ or it is a trainable term 
\begin{equation}
    s(z_i)  = c\left(\sigma\left(p_i\left(\mathrm{h}(z_i)-\mathrm{u}(z_i)\right)\right)\right)^2,
\end{equation}
where the meaning of $c$, $\sigma$, $p_i$, $\mathrm{h}(z_i)$, and $\mathrm{u}(z_i)$ is same as in \cref{eq:noisy_af} \cite{Gulcehre2016}.

A similar concept in \gls{ReLU} settings is the \gls{ProbAct} \glsxtrlong{AF} (see \cref{sec:probact}).

\subsection{Fractional adaptive activation functions}
\label{sec:faaf}
\Glspl[prereset]{FAAF} were proposed in \cite{Zamora2021, Zamora2022, Job2022, Ramadevi2024, Zamora2019} as a generalization of several activation functions using the fractional calculus (see \cite{Ortigueira2011} for a general introduction to the fractional calculus). Generally, for any activation function $f(z)$, its generalization $g(z)$ using fractional derivatives is defined as the $a-th$ fractional derivative of $f$:
\begin{equation}
    g(z) = D^{a} f(z),
\end{equation}
where $a$ can be a learnable\footnote{\cite{Job2022} did not specified whether the parameter is trainable but \cite{Zamora2021} explicitly uses a trainable $a$.} parameter \cite{Zamora2021}. The \glspl{FAAF} proposed in \cite{Job2022}  were further evaluated in \cite{Ramadevi2024}.

\subsubsection{Fractional ReLU}
\label{sec:fracrelu}
The \gls[prereset]{FracReLU} is defined as
\begin{equation}
    f(z_i) = \frac{z_i^{1-a_i}}{\Gamma\left(2-a_i\right)},
\end{equation}
where $\Gamma\left(x\right)$ is the Gamma function and $a_i$ is a trainable parameter \cite{Zamora2021}. The \gls{FracReLU} was later independently proposed in \cite{Job2022} under the name FReLU (but this abbreviation is already taken by \glsxtrlong{FReLU}). 

\subsubsection{Fractional softplus}
\label{sec:fracsoftplus}
The \gls[prereset]{FracSoftplus} is using the \gls{softplus} function to generalize sigmoid-like functions through fractional derivatives \cite{Zamora2021}. It is defined as
\begin{equation}
    f(z_i)=D^{a_i} \ln \left(1+\exp(z_i)\right),
\end{equation}
which is then computed as
\begin{equation}
    f(z_i)=\lim_{h \rightarrow 0} \frac{1}{h^{a_i}} \sum_{n=0}^{\infty}(-1)^{n} \frac{\Gamma(a_i+1) \ln \left(1+\exp(z_i-n h)\right)}{\Gamma(n+1) \Gamma(1-n+a_i)},
\end{equation}
where $a_i$ is a trainable parameter \cite{Zamora2021}. Particularly interesting cases are when $a_i=0$ as it is the \gls{softplus} function, $a_i=1$ \gls{logisticsigmoid}, and $a_i=2$ which leads to a bell-like shape \cite{Zamora2021}.

\subsubsection{Fractional hyperbolic tangent}
\label{sec:fractanh}
The \gls[prereset]{FracTanh} is another fractional generalization proposed in \cite{Zamora2021}; it is defined as 
\begin{equation}
    f(z_i)=D^{a_i} \tanh \left(z_i\right),
\end{equation}
which is then computed as
\begin{equation}
    f(z_i)=\lim_{h \rightarrow 0} \frac{1}{h^{a_i}} \sum_{n=0}^{\infty}(-1)^{n} \frac{\Gamma(a_i+1) \tanh\left(z_i-n \cdot h\right)}{\Gamma(n+1) \Gamma(1-n+a_i)},
\end{equation}
where $a_i$ is a trainable parameter \cite{Zamora2021}. The function becomes the \gls{tanh} for $a_i=0$ and the quadratic hyperbolic secant function for $a_i=1$.

\subsubsection{Fractional adaptive linear unit}
\label{sec:falu}
The \gls[prereset]{FALU} \cite{Zamora2022} is yet another \gls{AAF} based on fractional calculus\footnote{The \gls{FALU} was published in \cite{Zamora2022} without any links to \cite{Zamora2021} even though it was proposed by the same first author and it uses the same principles.} It can be seen as the fractional generalization using the $a_i-th$ fractional derivative of the \gls{swish} function:
\begin{equation}
    f(z_i)=D^{a_i} z_i \sigma\left(b_iz_i\right),
\end{equation}
where $a_i$ and $b_i$ are trainable parameters and $\sigma$ is the \gls{logisticsigmoid} function \cite{Zamora2022}.
The fractional derivative is then calculated as
\begin{equation}
    f(z_i)=\lim_{h \rightarrow 0} \frac{1}{h^{a_i}} \sum_{n=0}^{\infty}(-1)^{n} \frac{\Gamma(a_i+1) z_i \sigma\left(b_iz_i\right)}{\Gamma(n+1) \Gamma(1-n+a_i)}.
\end{equation}
However, as this calculation is not practical, \citeauthor{Zamora2022} use following approximation for $a_i \in \left[0,2\right]$ and $b_i \in \left[1,10\right]$:

\begin{equation}
    f(z_i)\approx\begin{cases}
                g(z_i,b_i) + a_i\sigma(b_iz_i)\left(1-g(z_i,b_i)\right), \quad & a_i \in [0,1],\\
                g(z_i,b_i) + a_i\sigma(b_iz_i)\left(1-2h(z_i,b_i)\right), \quad & a_i \in (1,2],\\
            \end{cases}
\end{equation}
where
\begin{equation}
    g(z_i,b_i)=z_i \sigma\left(b_iz_i\right),
\end{equation}
\begin{equation}
    h(z_i,b_i)=g(z_i,b_i) + \sigma(z_i)\left(1-g(z_i,b_i)\right),
\end{equation}
and $a_i$ and $b_i$ are the two previously mentioned trainable parameters \cite{Zamora2022}. The \gls{FALU} was shown to outperform \gls{ReLU}, \gls{GELU}, \gls{ELU}, \gls{SELU}, and \gls{KAF} on the MNIST \cite{LiDeng2012}, CIFAR-10 \cite{Krizhevsky2009}, ImageNet \cite{Deng2009, Russakovsky2015}, and Fashion MNIST \cite{Xiao2017} datasets for several tested architectures \cite{Zamora2022}.

\subsubsection{Fractional leaky ReLU (FracLReLU)}
\label{sec:fraclrelu}
The \gls[prereset]{FracLReLU} is the fractional variant of the \gls{LReLU} (see \cref{sec:lrelu}) proposed in \cite{Job2022}. It is defined using fractional calculus as
\begin{equation}
    f(z_i)=\begin{cases}
        D^{a_i}z_i, \quad & z_i \geq 0,\\
        D^{a_i}0.1z_i, \quad & z_i < 0,\\
    \end{cases}
\end{equation}
where $a_i \in (0,1)$ is a fixed parameter \cite{Job2022}. The fractional derivative is then calculated as
\begin{equation}
    f(z_i)=\begin{cases}
        \frac{1}{\Gamma\left(2-a_i\right)}z_i^{1-a_i}, \quad & z_i \geq 0,\\
        \frac{b}{\Gamma\left(2-a_i\right)}z_i^{1-a_i}, \quad & z_i < 0.\\
    \end{cases}
\end{equation}

\subsubsection{Fractional parametric ReLU (FracPReLU)}
\label{sec:fracprelu}
The \gls[prereset]{FracPReLU} is the fractional variant of the \gls{PReLU} (see \cref{sec:prelu}) proposed in \cite{Job2022}. It is defined using fractional calculus as
\begin{equation}
    f(z_i)=\begin{cases}
        D^{a_i}z_i, \quad & z_i \geq 0,\\
        D^{a_i}b_iz_i, \quad & z_i < 0,\\
    \end{cases}
\end{equation}
where $a_i \in (0,1)$ is a fixed parameter and $b_i$ is a trainable parameter \cite{Job2022}. The fractional derivative is then calculated as
\begin{equation}
    f(z_i)=\begin{cases}
        \frac{1}{\Gamma\left(2-a_i\right)}z_i^{1-a_i}, \quad & z_i \geq 0,\\
        \frac{b_i}{\Gamma\left(2-a_i\right)}z_i^{1-a_i}, \quad & z_i < 0.\\
    \end{cases}
\end{equation}

\subsubsection{Fractional ELU (FracELU)}
\label{sec:fracelu}
The \gls[prereset]{FracELU} is the fractional variant of the \gls{ELU} (see \cref{sec:elu}) proposed in \cite{Job2022}. It is defined using fractional calculus as
\begin{equation}
    f(z_i)=\begin{cases}
        D^{a_i}z_i, \quad & z_i \geq 0,\\
        D^{a_i}b\left(\exp(z_i-1)\right), \quad & z_i < 0,\\
    \end{cases}
\end{equation}
where $a_i \in (0,1)$  and $b$ are fixed parameters \cite{Job2022}. The fractional derivative is then calculated as
\begin{equation}
    f(z_i)=\begin{cases}
        \frac{1}{\Gamma\left(2-a_i\right)}z_i^{1-a_i}, \quad & z_i \geq 0,\\
        b\sum_{k=0}^{\infty}\left(\frac{1}{k!}\cdot\frac{\Gamma\left(k+1\right)}{\Gamma\left(k+1-a_i\right)}z_i^{k-a_i}\right)-b\frac{1}{\Gamma\left(1-a_i\right)}z_i^{-a_i}, \quad & z_i < 0.\\
    \end{cases}
\end{equation}

\subsubsection{Fractional SiLU (FracSiLU)}
\label{sec:fracsilu}
The \gls[prereset]{FracSiLU} is the fractional variant of the \gls{SiLU} (see \cref{sec:silu}) proposed in \cite{Job2022}. It is defined using fractional calculus as

While \citeauthor{Job2022} intended the \gls[prereset]{FracSiLU} to be the fractional variant of the \gls{SiLU} (see \cref{sec:silu}), they used a wrong definition of the \gls{SiLU}. Here we present both the {FracSiLU} from the \cite{Job2022} and the \gls{FracSiLU} that fit the definition of \gls{SiLU} --- the definition from \cite{Job2022} will be denoted as \gls{FracSiLU1} whereas the variant we derived as \gls{FracSiLU2}.
The \citeauthor{Job2022} used this definition\footnote{\citeauthor{Job2022} referenced \cite{Nwankpa2018,Qian2018} for their definition of \gls{SiLU}; however, the \cite{Nwankpa2018} contains the \gls{SiLU} definition from \cref{sec:silu} and \cite{Qian2018} does not mention \gls{SiLU} at all.} of \gls{SiLU}:
\begin{equation}
    f(z_i)=\begin{cases}
        D^{a_i}z_i, \quad & z_i \geq 0,\\
        D^{a_i}z_i\sigma\left(z_i\right), \quad & z_i < 0.\\
    \end{cases}
\end{equation}
Then the \gls{FracSiLU1} is defined as
\begin{equation}
    f(z_i)=\begin{cases}
        D^{a_i}z_i, \quad & z_i \geq 0,\\
        D^{a_i}z_i\sigma\left(z_i\right), \quad & z_i < 0,\\
    \end{cases}
\end{equation}
where $\sigma\left(z_i\right)$ is the \gls{logisticsigmoid} \cite{Job2022}. The fractional derivative is then calculated as
\begin{equation}
    f(z_i)=\begin{cases}
        \frac{1}{\Gamma\left(2-a_i\right)}z_i^{1-a_i}, \quad & z_i \geq 0,\\
        \sum_{k=0}^{\infty}\left(\left(-1\right)^k+\frac{\left(2^{k+1}-1\right)B_{k+1}\Gamma\left(k+2\right)}{\Gamma\left(k+2-a_i\right)\left(k+1\right)!}\right)z_i^{k+1-a_i}, \quad & z_i < 0,\\
    \end{cases}
\end{equation}
where $B_{n}$ is n-th Bernoulli's number \cite{Job2022}.

When using the \gls{SiLU} definition from \cref{sec:silu}, the \gls{FracSiLU2} is then defined as
\begin{equation}
    f(z_i)=D^{a_i}z_i\sigma\left(z_i\right).
\end{equation}
Since \citeauthor{Job2022} made no assumption about the sign of $z_i$, the fractional derivative of \gls{FracSiLU2} is computed as
\begin{equation}
    f(z_i)= sum_{k=0}^{\infty}\left(\left(-1\right)^k+\frac{\left(2^{k+1}-1\right)B_{k+1}\Gamma\left(k+2\right)}{\Gamma\left(k+2-a_i\right)\left(k+1\right)!}\right)z_i^{k+1-a_i},
\end{equation}
where $B_{n}$ is n-th Bernoulli's number.

\subsubsection{Fractional GELU (FracGELU)}
\label{sec:fracgelu}
Similarly as for \gls{FracSiLU}, \citeauthor{Job2022} intended the \gls[prereset]{FracGELU} to be the fractional variant of the \gls{GELU} (see \cref{sec:gelu}), but they used a wrong definition of the \gls{GELU}. Here we present both the {FracGELU} from the \cite{Job2022} and the \gls{FracGELU} that fit the definition of \gls{GELU} --- the definition from \cite{Job2022} will be denoted as \gls{FracGELU1} whereas the variant we derived as \gls{FracGELU2}.
The \citeauthor{Job2022} used this definition\footnote{\citeauthor{Job2022} referenced \cite{Nwankpa2018,Qian2018} for their definition of \gls{SiLU}; however, neither \cite{Nwankpa2018} nor \cite{Qian2018} contains a definition of \gls{GELU}.} of \gls{GELU}:
\begin{equation}
    f(z)=\begin{cases}
       z, \quad & z \geq 0,\\
       z\cdot\Phi\left(z\right), \quad & z < 0,\\
    \end{cases}
\end{equation}
where $\Phi\left(z\right)$ is the standard Gaussian \gls{CDF} \cite{Job2022}. Then the \gls{FracGELU1} is defined as
\begin{equation}
    f(z_i)=\begin{cases}
        D^{a_i}z_i, \quad & z_i \geq 0,\\
        D^{a_i}z_i\cdot\Phi\left(z_i\right), \quad & z_i < 0.\\
    \end{cases}
\end{equation}

The fractional derivative of \gls{FracGELU1} is then calculated as
\begin{equation}
    f(z_i)=\begin{cases}
        \frac{1}{\Gamma\left(2-a_i\right)}z_i^{1-a_i}, \quad & z_i \geq 0,\\
        0.5\frac{z_i^{1-a_i}}{\Gamma\left(2-a_i\right)}-\frac{1}{\sqrt{2\pi}}\sum_{k=0}^{\infty}\frac{1}{k!}\left(-\frac{1}{2}\right)^k\frac{z_i^{2\left(k+1\right)-a_i}}{2k+1}\frac{\Gamma\left(2k+3\right)}{\Gamma\left(2k+3-a_i\right)}, \quad & z_i < 0.\\
    \end{cases}
\end{equation}

When using the \gls{GELU} definition from \cref{sec:gelu}, the \gls{FracGELU2} is then defined as
\begin{equation}
    f(z_i)=D^{a_i}z_i\cdot\Phi\left(z_i\right).
\end{equation}
Since \citeauthor{Job2022} made no assumption about the sign of $z_i$, the fractional derivative of \gls{FracGELU2} is computed as
\begin{equation}
    f(z_i)= 0.5\frac{z_i^{1-a_i}}{\Gamma\left(2-a_i\right)}-\frac{1}{\sqrt{2\pi}}\sum_{k=0}^{\infty}\frac{1}{k!}\left(-\frac{1}{2}\right)^k\frac{z_i^{2\left(k+1\right)-a_i}}{2k+1}\frac{\Gamma\left(2k+3\right)}{\Gamma\left(2k+3-a_i\right)}.
\end{equation}

\subsection{Scaled softsign}
\label{sec:scaled_softsign}
An \glsxtrlong{AF} called \gls{scaled_softsign} \cite{Pishchik2023} is an adaptive variant of the \gls{softsign} activation (see \cref{sec:softsign}) with variable amplitude. It is defined as
\begin{equation}
    f(z_i) = \frac{a_iz_i}{b_i+\left|z_i\right|},
\end{equation}
where $a_i$ and $b_i$ are trainable parameters for each neuron $i$ \cite{Pishchik2023}. The parameter $a_i$ controls the range of the output while the parameter $b_i$ controls the rate of transition between signs \cite{Pishchik2023}.

\subsection{Parameterized softplus (s\textsubscript{+}2L)}
\label{sec:parameterized_softplus}
\Gls{parameterized_softplus} is an adaptive variant of a \gls{softplus} \glsxtrlong{AF} that allows for vertical shifts \cite{Vargas2023}. It is defined as
\begin{equation}
    f(z_i) = \ln\left(1+\exp(z_i)\right) - a_i,
\end{equation}
where $a_i \in [0,1]$ is a trainable parameter for each neuron $i$ \cite{Vargas2023}. \citeauthor{Vargas2023} also proposed a non-adaptive variant with fixed $a_i$ that is denoted as $s_{+}2$ \cite{Vargas2023}.

\subsection{Universal activation function (UAF)}
\label{sec:uaf}
The so-called \gls[prereset]{UAF} is a \gls{softplus} based \gls{AAF} proposed in \cite{Yuen2021}. It is defined as
\begin{equation}
    f(z_i) = \ln\left(1+\exp\left(a_i\left(z_i+b_i\right)+c_iz^2_i\right)\right) - \ln\left(1+\exp\left(d_i\left(z_i-b_i\right)\right)\right)+e_i,
\end{equation}
where $a_i$, $b_i$, $c_i$, $d_i$, and $e_i$ are trainable parameter for each neuron $i$ \cite{Yuen2021}. For example, the \gls{UAF} is able to well approximate the step function, \gls{logisticsigmoid}, \gls{tanh}, \gls{ReLU}, \gls{LReLU}, and Gaussian function \cite{Yuen2021}.

\subsection{Learnable extended activation function (LEAF)}
\label{sec:leaf}
The \gls[prereset]{LEAF} is an \gls{AAF} proposed in \cite{Bodyanskiy2023} that is able to replace several existing \glspl{AF}. It is defined as
\begin{equation}
    f(z_i) = \left(a_iz_i + b_i\right)\sigma\left(c_iz_i\right)+d_i,
\end{equation}
where $\sigma(x)$ is the \gls{logisticsigmoid} and $a_i$, $b_i$, $c_i$, and $d_i$ are trainable parameters for each neuron $i$ \cite{Bodyanskiy2023}. The \cref{tab:leaf_params} contains a list of \glspl{AF} that are equivalent to a particular \gls{LEAF} parameterization.

\begin{table}[h]
    \centering
    \begin{tabular}{ r| c c c c}
    equiv. \gls{AF} & $a_i$ & $b_i$ & $c_i$ & $d_i$\\ \hline
    \gls{ReLU} & 1 & 0 & $+\infty$ & 0 \\
    \gls{SiLU} & 1 & 0 & 1 & 0 \\
    \gls{tanh} & 0 & 2 & 2 & -1 \\
    \gls{logisticsigmoid} & 0 & 0 & 1 & 0 \\
    \gls{swish} & 1 & 0 & $a_i$ & 0 \\
    \gls{AHAF} & $a_i$ & 0 & $b_i$ & 0 \\
    \hline
    \end{tabular}
    \caption[AF equivalent to LEAF parameterizations]{\textbf{AF equivalent to LEAF parameterizations} \\ The list of \glspl{AF} that have an equivalent \gls{LEAF} parameterization.}
    \label{tab:leaf_params}
\end{table}

\subsection{Generalized ReLU (GReLU)}
\label{sec:grelu}
Theb \gls[prereset]{GReLU} is an \gls{AF} based on the \gls{UAF} (see \cref{sec:uaf}) \cite{Terziyan2022}. It is defined as
\begin{equation}
    f(z_i) = \frac{1}{b_i}\log_{a_i}\left(1+a_i^{b_iz_i}\right) = \frac{\ln\left(1+a_i^{b_iz_i}\right)}{b_i\ln\left(a_i\right)},
\end{equation}
where $a_i$ and $b_i$ are trainable parameters \cite{Terziyan2022}.

\subsection{Multiquadratic activation function (MAF)}
\label{sec:maf}
The \gls[prereset]{MAF} was used in \cite{GuangBinHuang2012,Siouda2022}. It is defined as
\begin{equation}
    f(z_i) = \sqrt{\left|\left|z_i-a_i \right|\right|^2 + b_i^2},
\end{equation}
where $a_i$ and $b_i$ are trainable parameters \cite{Siouda2022} $a_i$ is the slope coefficient and $b_i$ is the bias coefficient \cite{Siouda2022}.

\subsection{EIS activation functions}
\label{sec:eis}
The \gls{EIS}\footnote{The \gls{EIS} is a name given by \citeauthor{Biswas2021}; it is not an abbreviation.} is a family of \glspl{AAF} proposed in \cite{Biswas2020} with three notable examples {\gls{EIS}-1}, {\gls{EIS}-2}, and {\gls{EIS}-3} \cite{Biswas2020,Biswas2021}.

The general \gls{EIS} is defined as
\begin{equation}
    f(z_i) = \frac{z_i\left(\ln\left(1+\exp\left(z_i\right)\right)\right)^{a_i}}{\sqrt{b_i+c_iz_i^2}+d_i\exp\left(-e_iz_i\right)},
\end{equation}
where $a_i$, $b_i$, $c_i$, $d_i$, and $e_i$ are either trainable parameters or fixed hyperparameters; $a_i \in [0,1]$, $b_i \in [0,\infty)$, $c_i \in [0,\infty)$, $d_i \in [0,\infty)$, $e_i \in [0,\infty)$ and $b_i$, $c_i$, and $d_i$ cannot be equal to zero at the same time \cite{Biswas2020}.

The {\gls{EIS}-1} is defined as
\begin{equation}
    f(z_i) = \frac{z_i\ln\left(1+\exp\left(z_i\right)\right)}{z_i+d_i\exp\left(-e_iz_i\right)},
\end{equation}
where $d_i$ and $e_i$ are trainable parameters \cite{Biswas2020,Biswas2021}. It can be obtained from the general \gls{EIS} by setting $a_i = 1$, $b_i=0$, $c_i=1$ \cite{Biswas2020}.

The {\gls{EIS}-2} is defined as
\begin{equation}
    f(z_i) = \frac{z_i\ln\left(1+\exp\left(z_i\right)\right)}{\sqrt{b_i+c_iz_i^2}},
\end{equation}
where $b_i$ is a trainable parameter \cite{Biswas2020}. It can be obtained from the general \gls{EIS} by setting $a_i = 1$, $d_i=0$ \cite{Biswas2020}; however, the {\gls{EIS}-2} from \cite{Biswas2021} also fixes $c_i=1$.

And finally, the {\gls{EIS}-3} is defined as
\begin{equation}
    f(z_i) = \frac{z_i}{1+d_i\exp\left(-e_iz_i\right)},
\end{equation}
where $d_i$ and $e_i$ are trainable parameters \cite{Biswas2020,Biswas2021}; it can be obtained from the general \gls{EIS} by setting $a_i = 0$, $b_i=1$, $c_i=0$ \cite{Biswas2020}.

The \gls{EIS} family contains the \gls{softplus}, \gls{swish}, and \gls{ISRU} as special cases \cite{Biswas2020}.

\subsubsection{Linear combination of parameterized softplus and ELU (ELUs\textsubscript{+}2L)}
A linear combination of \gls{parameterized_softplus} and \gls{ELU} ($\mathrm{ELUs}_{+}2\mathrm{L}$) \cite{Vargas2023} is an \glsxtrlong{AAF} combining ELUs and \gls{parameterized_softplus} \glsxtrlongpl{AF}. It is defined as
\begin{equation}
    f(z_i) = b_i \mathrm{ELU}(z_i) + (1-b_i)\mathrm{s}_{+}2\mathrm{L}(z_i),
\end{equation}
where $b_i$ is a trainable parameter for each neuron $i$, $\mathrm{ELU}(z_i)$ is the \gls{ELU} \glsxtrlong{AF} and $\mathrm{s}_{+}2\mathrm{L}(z_i)$ is the parameterized \gls{softplus} \glsxtrlong{AF} \cite{Vargas2023}. The variant with non-adaptive \glsxtrlong{parameterized_softplus} is denoted as $\mathrm{ELUs}_{+}2$ \cite{Vargas2023}.

\subsection{Global-local neuron (GLN)}
\label{sec:gln}
The \gls[prereset]{GLN} is an \gls{AAF} that is a convex combination of two \glspl{AF} proposed in \cite{Ferreira2021}. It is defined as
\begin{equation}
    f(z_l) = \sigma\left(a\right)\cdot\mathrm{global}(z_l) + \left(1-\sigma\left(a\right)\right)\cdot\mathrm{local}(z_l) - b,
\end{equation}
where $a_l$ and $b_l$ are trainable weights for each layer $l$ and $\mathrm{global}(z_l)$ and $\mathrm{local}(z_l)$ are \glspl{AF} capable of identifying the global and local characteristics respectively \cite{Ferreira2021,DeMedeirosDelgado2022}; the authors used $\mathrm{global}(z_l)\sin\left(z_l\right)$ and $\mathrm{local}(z_l)=\tanh(z_l)$ in \cite{Ferreira2021,DeMedeirosDelgado2022}.

\subsection{Neuron-adaptive activation function}
\label{sec:naf}
A similar approach to trainable amplitude and \gls{generalized hyperbolic tangent} is the so-called \gls{NAF} \cite{ShuxiangXu2000, ShuxiangXu2005, Xu2007}, which comprises of a linear combination of two \glsxtrlongpl{AF} with scalable amplitude:
\begin{equation}
    f(z) = a \exp\left(-b \cdot \left(z\right)^2\right) + \frac{c}{1+\exp\left(-d \cdot z\right)},
\end{equation}
where $a$, $b$, $c$, and $d$ are trainable parameters that are shared by the whole network \cite{ShuxiangXu2000}. The \gls{NAF} was shown to perform superiorly on a few regression tasks \cite{ShuxiangXu2000}.

\subsubsection{Scaled logistic sigmoid}
\label{sec:scaled_logistic_sigmoid}
A scaling variant of \gls{logisticsigmoid} called \gls{scaledlogisticsigmoid} was proposed in \cite{Tezel2007}. The function is defined as
\begin{equation}
    f(z_i) = \frac{a_i}{1+\exp\left(-b_iz_i\right)},
\end{equation}
where $a_i$ and $b_i$ are trainable parameters for each neuron $i$ \cite{Tezel2007}. Note that this activation is identical to the second part of the previously proposed \gls{NAF} (see \cref{sec:naf}).

A variant combining \gls{SLS-SS} was also used in \cite{Tezel2007}; it has four trainable parameters and is defined as
\begin{equation}
    f(z_i) = a_i\cdot \sin\left(b_iz_i\right) + \frac{c_i}{1+\exp\left(-d_iz_i\right)},
\end{equation}
where $a_i$, $b_i$, $c_i$, and $d_i$ are trainable parameters \cite{Tezel2007}. This \glsxtrlong{AF} is a special case of another variant of \gls{NAF} \cite{Xu2002}:
\begin{equation}
    f(z_i) = a_i\cdot \sin\left(b_iz_i\right) + c_i \exp\left(-d_i \cdot \left(z\right)^2\right)+  \frac{e_i}{1+\exp\left(-f_iz_i\right)},
\end{equation}
where $a_i$, $b_i$, $c_i$, $d_i$, $e_i$ and $f_i$ are trainable parameters \cite{Xu2002}.

\subsection{Adaptive piece-wise linear unit (APLU)}
\label{sec:aplu}
\glsreset{APLU}Another generalization of \gls{ReLU} is the \gls[prereset]{APLU}, which uses the sum of hinge-shaped functions as the \glsxtrlong{AF} \cite{Agostinelli2014}. An approach extending \gls{APLU} is \gls{SAAF} with piece-wise polynomial form and was specifically designed for regression and allows for bias--variance trade-off using a regularization term \cite{Hou2017}.

\gls{APLU} is defined as
\begin{equation}
    f(z_i) = \max\left(0, z_i\right) + \sum_{s=1}^S a_i^s \max\left(0, -z_i + b_i^s\right),
\end{equation}
where $S$ is the number of hinges, $i$ is the number of neurons, and $a_i^s$, $b_i^s$, $s \in 1, \ldots, S$ are trainable parameters per unit \cite{Jin2016}. However, the optimizer might choose very large values of $a_i^s$ and balance them by very small weights, which could lead to numerical instabilities; therefore, an $L^2$ penalty is added to the parameters $a_i^s$, $b_i^s$ scaled by 0.001 \cite{Agostinelli2014}. Another adaptive piecewise linear function was proposed in \cite{Aziznejad2020}, where a weighted combination of \glspl{ReLU} with additional parameters was used.

\subsection{Simple piecewise linear and adaptive function with symmetric hinges (SPLASH)}
\label{sec:splash}
The \gls[prereset]{SPLASH} \cite{Tavakoli2021} is an approach similar to the \gls{APLU}. It is defined as
\begin{equation}
    f(z_l) = \sum_{s=1}^{\frac{S+1}{2}}a^{+}_{l,s}\max\left(0,z-b_{l,s}\right) + \sum_{s=1}^{\frac{S+1}{2}}a^{-}_{l,s}\max\left(0,-z-b_{l,s}\right),
\end{equation}
where $S$ is an odd number, $b_{l,s}$ and $-b_{l,s}$ are hinge parameters and $a^{+}_{l,s}$ and $a^{-}_{l,s}$ are scaling parameters for each layer $l$ \cite{Tavakoli2021} ; these max functions form $S+1$ continuous line segments with hinges at $b_{l,s}$ and $-b_{l,s}$ \cite{Tavakoli2021}. While \citeauthor{Tavakoli2021} tried different values for $S$, they found that using $S=7$ usually works well \cite{Tavakoli2021}.

\subsection{Multi-bias activation (MBA)}
\label{sec:mba}
An approach similar to \gls{APLU} and \gls{paired_relu} (see \cref{sec:paired_relu}) termed \gls{MBA} \cite{Li2016} uses the same activation but with multiple biases, which allows to learn more complex activations; in this it resembles \gls{paired_relu} as one input map leads to several output maps with activation with different biases \cite{Li2016}. The weights that will be given to the output maps in the next layer are similar to the weights in the \gls{APLU}; however, the \gls{MBA} is able to provide cross-channel information due to multiple outputs for each activation \cite{Li2016}.
The \gls{MBA} is defined as
\begin{equation}
    \vec{f}(z_i) = \begin{bmatrix}
        g(zi+b_{i, 1}) \\
        g(zi+b_{i, 2}) \\
        \ldots \\
        g(zi+b_{i, k}) \\
        \ldots \\
        g(zi+b_{i, K}) \\
        \end{bmatrix},
\end{equation}
where $b_{i, k}$, $k=1, 2, \ldots, K$ are trainable biases and $g(x)$ is any non-linear \glsxtrlong{AF} \cite{Li2016}; \citeauthor{Li2016} used \gls{ReLU} as the \glsxtrlong{AF} $g(x)$ \cite{Li2016}.

\subsection{Mexican ReLU (MeLU)}
\label{sec:melu}
A \gls{MeLU} is an \glsxtrlong{AF} with a similar approach as the \gls{APLU}, but it does not need any $L^2$ penalty \cite{Maguolo2021}. The \gls{MeLU} is defined as
\begin{equation}
    \label{eq:melu_base}
    f(z_i) = \mathrm{PReLU}(z_i) + \sum_{j}^{k-1} a_{i,j} \phi_{b_jc_j}\left(z_i\right),
\end{equation}
where $a_{i,j}$ are trainable parameters for each neuron/filter $i$, and $k$ is the total number of trainable parameters ($k-1$ for the sum and one for the \gls{PReLU}), $b_j$ and $c_j$ are fixed constants that are chosen recursively (more details in \cite{Maguolo2021}); $\phi_{b_jc_j}\left(z_i\right)$ is defined as
\begin{equation}
    \phi_{b_jc_j}\left(z_i\right) = \max\left(c_j-\left|z_i-b_j\right|,0\right).
\end{equation}
\citeauthor{Maguolo2021} used $k=4$ and $k=8$ for their experiments; the trainable parameters $a_{i,j}$ were all initialized to zero which helps the training at the early stages by exploiting the properties of the \gls{ReLU} (e.g., the \gls{MeLU} is convex for many iterations at the beginning)\cite{Maguolo2021}. The advantage of the \gls{MeLU} over the \gls{APLU} is that it needs only half of the parameters while retaining the same representation power when the parameters are jointly optimized with the network's weights and biases \cite{Maguolo2021}.

\subsubsection{Modified Mexican ReLU (MMeLU)}
\label{sec:mmelu}
The \gls[prereset]{MMeLU} is an \gls{MeLU} inspired \gls{AF} proposed in \cite{Fakhfakh2023}. It is defined as
\begin{equation}
    f(z_i) = a_i \cdot \max\left(b_i-\left|z_i-c_i\right|,0\right) + \left(1-a_i\right) \mathrm{ReLU}\left(z_i\right),
\end{equation}
where $a_i$, $b_i$, and $c_i$ are adaptive parameters estimated using Bayesian procedure outlined in \cite{Fakhfakh2023};  $a_i \in [0,1]$, $b_i \in \mathbb{R}^+$ in , and $c_i \in \mathbb{R}$ \cite{Fakhfakh2023}.

\subsubsection{Gaussian ReLU (GaLU)}
\label{sec:galu}
The \gls[prereset]{GaLU} is a \gls{MeLU}-inspired \gls{AAF} proposed in \cite{Nanni2020}. It uses the same basic form as \gls{MeLU} has in \cref{eq:melu_base} but it uses following $\phi_{b_jc_j}\left(z_i\right)$:
\begin{equation}
    \phi_{b_jc_j}\left(z_i\right) = \max\left(c_j-\left|z_i-b_j\right|,0\right) + \min\left(\left|z-b_j - 2c_j\right|-c_j,0\right),
\end{equation}
where $b_j$ and $c_j$ are similar parameters is in the original \gls{MeLU}; more details about the parameters is available in \cite{Nanni2020}.

\subsubsection{Hard-Swish}
\label{sec:hardswish}
The \gls{hardswish} is an adaptive variant of a scaled \gls{hard_sigmoid} activation \cite{Avenash2019}. It is defined as
\begin{equation}
    f(z_i)=2z_i\cdot\max\left(0, \min\left(0.2b_iz_i + 0.5, 1\right)\right),
\end{equation}
where $b_i$ is either trainable or fixed parameter \cite{Avenash2019}. For $b_i \rightarrow \infty$, the \gls{hardswish} approaches the \gls{ReLU} \cite{Avenash2019}. The \gls{hardswish} outperformed the \gls{logisticsigmoid}, \gls{tanh}, \gls{ReLU}, \gls{LReLU}, and \gls{swish} on the MNIST dataset \cite{LiDeng2012} in \cite{Avenash2019}. The ResNet \cite{He2016}, \gls{WRN} \cite{Zagoruyko2016}, and DenseNet \cite{Huang2017} with  gls{hardswish} outperformed their variants with \gls{ReLU} and \gls{swish} on the CIFAR-10 \cite{Krizhevsky2009} dataset \cite{Avenash2019}.

\subsection{S-shaped rectified linear activation unit (SReLU)}
\label{sec:srelu}
\glsreset{SReLU}
A \gls[prereset]{SReLU} \cite{Jin2016} consists of three piecewise linear functions that are controlled by four trainable parameters that are learned jointly with the whole network. The \gls{SReLU} is able to learn both convex and non-convex functions; in particular, it is able to learn both \gls{ReLU} and also sigmoidlike functions. It is similar to \gls{APLU} (see \cref{sec:aplu}), but \gls{APLU} approximates non-convex functions, and it requires the rightmost linear function to have a unit slope and bias of zero \cite{Jin2016}. \gls{SReLU} is defined as
\begin{equation}
    f(z_i)=\begin{cases}
                t_i^r + a_i^r(z_i-t_i^r), \quad & z_i \geq t_i^r, \\
                z_i, \quad & t_i^r > z_i > t_i^l, \\
                t_i^l + a_i^l(z_i-t_i^l), \quad & z_i \leq t_i^l, \\
            \end{cases}
\end{equation}
where $t_i^r$, $t_i^l$, $a_i^r$, and $a_i^l$ are trainable parameters for each neuron $i$ (or channel $i$ in case of \glsxtrlongpl{CNN}) \cite{Jin2016}. The parameters $t_i^r$ and $t_i^l$ determine thresholds of an interval outside which the slope of the linear parts is controlled by parameters $a_i^r$ and $a_i^l$, respectively. The authors \citeauthor{Jin2016} show that the \gls{SReLU} outperformed the \gls{ReLU}, \gls{LReLU}, \gls{PReLU}, \gls{APLU}, \gls{maxout} and plain \gls{NIN} on several visual tasks. The authors also recommend to initialize the parameters of \gls{SReLU} to $t_i \in \mathbb{R}$, $a_i^r := 1$, $t_i^l := 0$, and $a_i^l \in (0,1)$ which degenerates the \gls{SReLU} into a \gls{LReLU} and then keep these parameter fixed during several initial training epochs \cite{Jin2016}. The \gls{SReLU} can be seen as a more general concept to the later proposed \gls{PLU} (see \cref{sec:plu}) and to the \gls{BLReLU} \cite{Liew2016} (see \cref{sec:blrelu}).

\subsubsection{N-activation}
\label{sec:nactivation}
The \gls{N-activation} is activation very similar to a special case of \gls{SReLU}\footnote{It would be a special case of \gls{SReLU} if the the thresholds were directly trainable and not determined using the $\min$ and $\max$ functions.} proposed in \cite{Prach2023}. The \gls{N-activation} with trainable parameters $a_i$, and $b_i$ is defined as
\begin{equation}
    f(z_i)=\begin{cases}
                z_i-2t_{i,\min}, \quad & z_i < t_{i,\min}, \\
                -z_i, \quad & t_{i,\min} \leq z_i \leq t_{i,\max}, \\
                z_i-2t_{i,\max}, \quad & z_i >t_{i,\max}, \\
            \end{cases}
\end{equation}
where
\begin{equation}
    t_{i,\min} = \min\left(a_i, b_i\right)
\end{equation}
and
\begin{equation}
    t_{i,\max} = \max\left(a_i, b_i\right).
\end{equation}

\subsubsection{ALiSA}
\label{sec:alisa}
A special case of \gls{SReLU} was later proposed under the name \gls{ALiSA} in \cite{Bawa2019}; it can be obtained by setting $t_i^r \coloneqq 1$ and $t_i^l \coloneqq = 0$:
\begin{equation}
    f(z_i)=\begin{cases}
                a_i^rz_i-a_i^r+1, \quad & z_i \geq 1, \\
                z_i, \quad & t_i^r > z_i > t_i^l, \\
                a_i^lz_i, \quad & z_i \leq 0, \\
            \end{cases}
\end{equation}
where $a_i^r$ and $a_i^l$ are adaptive parameters \cite{Bawa2019}. Its nonadaptive variant is called simply \gls{LiSA} and has parameters $a_i^r$ and $a_i^l$ fixed \cite{Bawa2019}.

\subsection{Alternated left ReLU (All-ReLU)}
\label{sec:allrelu}
The \gls{All-ReLU} was proposed in \cite{Curci2021} for usage in sparse \glsxtrlongpl{NN}. It is inspired by the \gls{SReLU} \cite{Curci2021}. It is defined as
\begin{equation}
    f(z_i)=\begin{cases}
                -a z_i, \quad & z_i \leq 0 \text{ and } l \% 2 = 0, \\
                a z_i, \quad & z_i \leq 0 \text{ and } l \% 2 = 1, \\
                z_i, \quad & z_i > 0, \\
            \end{cases}
\end{equation}
where $a$ is a fixed parameter controlling the slope for negative inputs, $l$ is the number of layers, and $\%$ is the modulo operation \cite{Curci2021}.

\subsection{Piecewise linear unit (PLU)}
\label{sec:plu}
A \gls{PLU} \cite{Nicolae2018} resembles two earlier proposed \glsxtrlongpl{AF} --- the \gls{SReLU} (see \cref{sec:srelu}) and  \glsxtrlong{APLU} (see \cref{sec:aplu}); it can be even seen as a special case of the \gls{SReLU}.
\begin{equation}
    f(z_i) = \max\left(a_i\left(z_i+b\right)-b, \min\left(a_i\left(z_i-b\right)+b, z_i\right)\right),
\end{equation}
where $a_i$ is either a trainable parameter or a predefined constant \cite{Dubey2022} and $b$ is a predefined constant \cite{Nicolae2018}; a variant with $a = 0.1$ and $b=1$ was shown in \cite{Nicolae2018}.
The advantage of the \gls{PLU} compared to the \gls{SReLU} is that it produces an invertible function (which is not always the case for the more general \gls{SReLU}) \cite{Nicolae2018}.

\subsection{Adaptive linear unit (AdaLU)}
\label{sec:adalu}
The \gls[prereset]{AdaLU} \cite{Mo2022} is yet another piecewise linear \gls{AAF}. It is defined as
\begin{equation}
    f(z_i)=\begin{cases}
                c_i\left(z_i-a_i\right)+b_i, \quad & (z_i - a_i) > 0 \text{ and } c_i\left(z_i-a_i\right) > e_i, \\
                d_i\left(z_i-a_i\right)+b_i, \quad & (z_i - a_i) \leq 0 \text{ and } d_i\left(z_i-a_i\right) > e_i,\\
                e_i + b_i, \quad & \text{otherwise}, \\
            \end{cases}
\end{equation}
where $a_i$, $b_i$, $c_i$, $d_i$, and $e_i$ are trainable parameters for each neuron $i$ \cite{Mo2022}. The parameters $a_i$ and $b_i$ control the offsets; $c_i$ and $d_i$ control the slope of each linear part, and $e_i$ is the saturation value \cite{Mo2022}.

\subsection{Trapezoid-shaped activation function (TSAF)}
\label{sec:tsaf}
The \gls[prereset]{TSAF} \cite{Mao2022Approximating} (ref. from \cite{Pan2023Smoothing}) is an \gls{AF} consisting of four \glspl{ReLU}. It is defined as
\begin{multline}
 f(z_i) = \frac{1}{c_i}\big(\mathrm{ReLU}\left(z_i-a_i+c_i\right)+\mathrm{ReLU}\left(z_i-a_i\right)+\mathrm{ReLU}\left(z_i+b_i-c_i\right)-\\\mathrm{ReLU}\left(z_i-b_i\right)\big),
\end{multline}
where $a_i$, $b_i$, and $c_i$ are parameters\footnote{\Citeauthor{Pan2023Smoothing} do not state whether they are used in trainable or fixed form.} such that $a_i < b_i$ and $c_i \in (0,1]$ \cite{Pan2023Smoothing}.

\subsection{Adaptive Richard's curve weighted activation (ARiA)}
\label{sec:aria}
Another function motivated by the \gls{swish} \glsxtrlong{AF} is the \gls{ARiA} \cite{Patwardhan2018}, which replaces the \gls{logisticsigmoid} in the \gls{swish} by Richard's curve \cite{Dubey2022}. Richard's curve \cite{Richards1959} is a generalization of the \gls{logisticsigmoid} that is controlled by several hyperparameters.
The Richard's curve is defined as \cite{Patwardhan2018}:
\begin{equation}
    \label{eq:richards_curve}
    \sigma_R(x) = A + \frac{K - A}{\left(C+Q\cdot\exp\left(-Bx\right)\right)^\frac{1}{\upsilon}},
\end{equation}
where $A$ is the lower asymptote, $K$ is the upper asymptote, $C$ is a constant (typically equal to 1 \cite{Patwardhan2018}), $\upsilon>0$ controls the direction of growth and $B$ is the exponential growth rate, $Q$ controls the initial value of the function. The \gls{ARiA} is defined as
\begin{equation}
    f(z) = z \cdot \sigma_R(x),
\end{equation}
where $\sigma_R(x)$ is the Richard's curve from \cref{eq:richards_curve} \cite{Patwardhan2018}. As such, the \gls{ARiA} has five hyperparameters controlling its behavior. To reduce the number of the hyperparameters, \gls{ARiA2} was also proposed \cite{Patwardhan2018} that is defined by only two hyperparameters $a$ and $b$
\begin{equation}
    f(z) = z \cdot \left(1 + \exp\left(-bz\right)\right)^{-a}.
\end{equation}
The \gls{swish} \glsxtrlong{AF} is a special case of \gls{ARiA} with $A=1$, $K=0$, $B=1$, $\upsilon=1$, $C=1$, and $Q=a_i$, where $a_i$ is the parameter of the \gls{swish} \glsxtrlong{AF} (see \cref{sec:swish} for details) \cite{Patwardhan2018}. The \gls{ARiA2} is a special case of \gls{ARiA} with $K=0$, $B=1$, $\upsilon=1$, $C=\frac{1}{a}$, and $Q=b$, where $a$ and $b$ are the \gls{ARiA2} hyperparameters \cite{Patwardhan2018}. \citeauthor{Patwardhan2018} reached best accuracy on the MNIST \cite{LiDeng2012} dataset with a custom \gls{CNN} using \gls{ARiA2} with $a=1.5$ and $b=2$; the best parameters for the DenseNet \cite{Huang2017} were $a=1.75$ and $b=1$ \cite{Patwardhan2018}. While the parameters were fixed in the experiments in \cite{Patwardhan2018}, they can also be trainable as is in the special case of the \gls{swish} \glsxtrlong{AF}.

\subsection{Modified Weibull function}
\label{sec:mwf}
A \gls[prereset]{MWF}  is an Weibull-function-based \gls{AF} proposed in \cite{Husain2021}. It is defined as
\begin{equation}
    f(z_i) = \left(\frac{z_i}{a_i}\right)^{b_i-1}\exp\left(-\left(\frac{z_i}{c_i}\right)^{d_i}\right),
\end{equation}
where $a_i$, $b_i$, $c_i$, and $d_i$ are trainable parameters \cite{Husain2021}. The parameter $b_i$ determines the location of the peak of the \gls{AF} \cite{Husain2021}. The polynomial term dominates for small input values while the exponential starts to dominate with larger values which reduces the output value as the input value further increases \cite{Husain2021}.

\subsection{Sincos}
\label{sec:sincos}
The \gls{sincos} is another older \gls{AF} proposed in \cite{Efe2008}. It is defined as
\begin{equation}
    f(z)= a\cdot \sin\left(bz\right)+c\cdot\cos\left(dz\right),
\end{equation}
where $a$, $b$, $c$, and $d$ are adaptive parameters \cite{Efe2008}.

\subsection{Combination of sine and logistic sigmoid (CSS)}
\label{sec:css}
The \gls[prereset]{CSS}\footnote{The function was unnamed in \cite{Ozbay2010}; we used this abbreviation to distinguish it from \gls{sinsig}.} is an \gls{AAF} proposed in \cite{Ozbay2010}. It is defined as
\begin{equation}
    f(z)= a\cdot \sin\left(bz\right)+c\cdot\sigma\left(dz\right),
\end{equation}
where $a$, $b$, $c$, and $d$ are adaptive parameters \cite{Ozbay2010}.

\subsection{Catalytic activation function (CatAF)}
\label{sec:cataf}
The \gls[prereset]{CatAF} is an \gls{AAF} that uses sinusoidal mixing of any \gls{AF} and the identity to produce the final activation \cite{Sarkar2022}. It is defined as
\begin{equation}
    f(z_i) = z_i \sin\left(a_i\right) + g(z_i)\cos{a_i},
\end{equation}
where $a_i$ is a trainable parameter and $g(z_i)$ is any \gls{AF} such as the \gls{ReLU} \cite{Sarkar2022}.

\subsection{Expcos}
\label{sec:expcos}
An \gls{AAF} combining an exponential function with the cosine was proposed in \cite{Ozbay2010}. It is called \gls{expcos} in this work\footnote{The function was originally unnamed in \cite{Ozbay2010}.} and is defined as
\begin{equation}
    f(z)= \exp\left(-az^2\right)\cdot \cos\left(bz\right),
\end{equation}
where $a$ and $b$ are adaptive parameters \cite{Ozbay2010}.

\subsection{Multi-bin trainable linear unit (MTLU)}
\label{sec:mtlu}
The \gls{MTLU} can be seen as a conceptual extension of the \gls{SReLU} (see \cref{sec:srelu}) into more than three segments \cite{Dubey2022}:
\begin{equation}
    f(z_i)=\begin{cases}
                a_{i,0} z + b_{i, 0}, \quad & z_i \geq c_{i, 0}, \\
                a_{i,1} z + b_{i, 1}, \quad & c_{i, 0} < z_i \geq c_{i, 1}, \\
                \ldots \\
                a_{i,k} z + b_{i, k}, \quad & c_{i, k-1}  <z_i \geq c_{i, k}, \\
                \ldots \\
                a_{i,K} z + b_{i, K}, \quad & c_{i, K-1} < z_i , \\
            \end{cases}
\end{equation}
where $a_{i,0}, \ldots, a_{i, K}$, and $b_{i,0}, \ldots, b_{i, K}$ are trainable parameters for each neuron/filter and $K$ and $c_{i,0}, \ldots, c_{i, K-1}$ are predefined  hyperparameters \cite{Gu2019}. The authors used unifromly distributed anchors $c_{i,0}, \ldots, c_{i, K-1}$ \cite{Gu2019}. The main disadvantage besides the higher number of additional parameters is the higher number of non-differentiable points \cite{Dubey2022}. The \gls{MTLU} was also named \gls{CPN}\textsubscript{m} in \cite{Gao2023}. The \gls{CPN}\textsubscript{mc} is a \gls{MTLU} variant with continuity constraint proposed in \cite{Gao2023}.

An \gls{AF} with the same form as the \gls{MTLU} with only minor differences was proposed in \cite{Zhou2021Learning, Zhu2023PWLU} under the name \gls[prereset]{PWLU}; \citeauthor{Zhu2023PWLU} also proposed its 2D extension in \cite{Zhu2023PWLU}. Unlike the \gls{MTLU}, it uses a uniformly spaced demarcation points $c_{i, k}$ \cite{Zhou2021Learning}. Another \gls{PWLU} variant named \gls[prereset]{N-PWLU} allows for learnable intervals on which the function is piecewise linear, and also it leverages cumulative definition for efficient learning \cite{Zhu2022}. Multistability analysis of such piecewise linear \glspl{AF} is analyzed in \cite{Zhang2021Multistability}. An analysis of a number of regions of piecewise linear \glspl{NN} is available in \cite{Goujon2024}.

\subsection{Continuous piecewise nonlinear activation function {CPN}}
\label{sec:cpn}
A variant of the \gls{MTLU} named \gls{CPN} where the $c_{i, k}$ was used in \cite{Gao2023}. It is defined as
\begin{equation}
    f(z_i)=\begin{cases}
                a_{i,0} z + b_{i, 0} + c_{i, 0}g(z_i), \quad & z_i \geq d_{i, 0}, \\
                a_{i,1} z + b_{i, 1} + c_{i, 1}g(z_i), \quad & d_{i, 0} < z_i \geq d_{i, 1}, \\
                \ldots \\
                a_{i,k} z + b_{i, k} + c_{i, k}g(z_i), \quad & d_{i, k-1}  <z_i \geq d_{i, k}, \\
                \ldots \\
                a_{i,K} z + b_{i, K} + c_{i, K}g(z_i), \quad & d_{i, K-1} < z_i , \\
            \end{cases}
\end{equation}

where $a_{i,0}, \ldots, a_{i, K}$, $b_{i,0}, \ldots, b_{i, K}$, $c_{i,0}, \ldots, d_{i, K-1}$ are trainable parameters for each neuron/filter, $g(z_i)$ is a non-linear function such as the \gls{logisticsigmoid} and $K$ and $c_{i,0}, \ldots, d_{i, K-1}$ are predefined  hyperparameters \cite{Gao2023}.

\Citeauthor{Gao2023} also proposed a variant named \gls{CPN}\textsubscript{nl}, which introduces a non-linear term for each small interval and does not enforce the uniform division of the activation space \cite{Gao2023}. It is defined  as

\begin{equation}
    f(z_i) = \max\left\{p_{i, 0}(z), p_{1, 0}(z), \ldots, p_{i, k}(z), \ldots, p_{i, K}(z)\right\},
\end{equation}
where
\begin{equation}
    p_k(z) = a_{i, k}z_i + b_{i, k}\mathrm{SiLU}(z_i) + c_i,  \quad k = 0,1,\ldots, K,
\end{equation}
where $K$ is the number of functions and $a_{i, k}$, $b_{i, k}$, and $c_{i, k}$ are learnable coefficients for $k = 0,1,\ldots, K$ \cite{Gao2023}.

\subsection{Look-up table unit (LuTU)}
\label{sec:lutu}
A piecewise \glsxtrlong{AF} \gls{LuTU} \cite{Wang2018LookUp, Piazza1993, Fiori2002} is a learnable \glsxtrlong{AF} that consists of several points defining the function; the values between the points are obtained using either linear interpolation or smoothing with single period cosine mask function \cite{Wang2018LookUp}. Similar \glsxtrlong{AAF} using linear interpolation was used in \cite{MiaoKang2005, Kang2007}. A look-up table of anchor points $\{a_{i,j}, b_{i,j}\}, j=0, 1, \ldots, n$ that are uniformly spaced with step $s$, $a_{i,j} = a_{0} + s\cdot j$, controls the shape of the \glsxtrlongpl{AF}. The step $s$, anchor points $a_0$, and $n$ are predetermined hyperparameters, and therefore $a_{i,j}$ are predetermined values for which the output values $b_{i,j}$ are learnable parameters.
Linear-interpolation-based function is defined as
\begin{equation}
    f(z_i) = \frac{1}{s}\left(b_{i,j}\left(a_{i,j+1}-z_i\right) + b_{i,j+1}\left(z_i-a_{i,j}\right)\right), \quad a_{i,j} \geq z \geq a_{i,j+1},
\end{equation}
where $a_{i,j}$ are hyperparameters defined by the step $s$ and initial point $a_0$ shared for all points and $b_{i,j}$ are trainable parameters for each neuron $i$ \cite{Wang2018LookUp}. \citeauthor{Wang2018LookUp} used $a_0=-12$, step $s=0.1$ and $n=240$ to cover the interval $[-12,12]$ using 241 anchor poiints for each neuron.
Therefore, for any input value between $a_{i,j}$ and $a_{i, j+1}$, the output is linearly interpolated from $b_{i,j}$ and $b_{i,j+1}$ \cite{Wang2018LookUp}. However, such a definition might lead to unstable gradients \cite{Wang2018LookUp}; therefore, a variant of \gls{LuTU} with cosine smoothing was also proposed.
The smoothing function is defined as
\begin{equation}
    r(x, \tau)=\begin{cases}
                \frac{1}{2\tau}\left(1+\cos\left(\frac{\pi}{\tau}x\right)\right), \quad & -\tau \geq x \geq \tau, \\
                0, \quad & \text{otherwise}, \\
            \end{cases}
\end{equation}
where $\tau$ is a hyperparameter controlling the period ($2\tau$) of the cosine function \cite{Wang2018LookUp}.
The smoothed variant of the \gls{LuTU} is then defined as
\begin{equation}
    \label{eq:lutu_smoothed}
    f(z_i)= \sum_{j=0}^n y_j r\left(z_i-a_{i,j}, ts\right),
\end{equation}
where $t$ is an integer defining the ration between $\tau$ and $s$ \cite{Wang2018LookUp}. The formula in \cref{eq:lutu_smoothed} can be further simplified as it is not necessary to sum over all $j \in \{0,1,\ldots,n\}$ as the smoothing function has a truncated input domain, more details in \cite{Wang2018LookUp}.

\subsection{Maxout unit}
\label{sec:maxout}
\Gls{maxout} returns the maximum of multiple linear functions per each unit $i$ \cite{Goodfellow2013}:
\begin{equation}
    f(z_i) = \max_{k \in \{1, \ldots, K\}} w_i^kz_i + b_i^k
\end{equation}
where $K$ is the number of linear functions. The \gls{maxout} can also be used directly on inputs of the neuron as shown in \cite{Goodfellow2013} (by replacing $w_i^kz_i$ with $\vec{x_i}^T\vec{w}_{i}^k$ where $\vec{x_i} \in \mathbb{R}^d$ is the vector of individual inputs to a neuron $i$ and $w \in \mathbb{R}^{d}$ are trainable weights \cite{Goodfellow2013}) but the equation presented here uses only the hidden state for simplicity. The advantage of \gls{maxout} is that it is a universal approximator of a convex function \cite{Goodfellow2013, Jin2016}; however, it cannot learn non-convex functions \cite{Jin2016} and introduces a high number of additional parameters per neuron \cite{Jin2016, Goodfellow2013}. While some works show that \gls{maxout} perform superiorly \cite{Goodfellow2013, Hanif2020}, other experiments show that \gls{ReLU}, which is a special case of maxout, performs better \cite{Castaneda2019}. Furthermore, since the \gls{maxout} is more complex than regular \gls{ReLU}, the training is relatively slower \cite{Castaneda2019}.

Empirical comparison of the \gls{maxout} with \gls{ReLU}, \gls{LReLU}, \gls{SELU} and \gls{tanh} is available in \cite{Castaneda2019}; with \gls{ReLU}, \gls{tanh}, sigmoid and \gls{VLReLU} in \cite{Mishkin2015}.

\subsection{Adaptive blending unit (ABU)}
\label{sec:abu}
An approach mixing several \glsxtrlongpl{AF} was described in \cite{Stfeld2020} where \gls{ABU} was introduced. The \gls{ABU} is a weighted sum of several predefined activations \cite{Stfeld2020}. It is defined as
\begin{equation}
    \label{eq:abu_basic}
    f(z_l) = \sum_{j=0}^n a_{j,l} g_j(z_l),
\end{equation}
where $g_j(z_l)$ is an \glsxtrlong{AF} from a pool of $n$ \glsxtrlongpl{AF} and $a_{j,l}$ is a weighting parameter that is trained for each layer $l$ and \glsxtrlong{AF} $g_j(z_l)$. The \gls{ABU} was first proposed as a special case of a general framework called \gls{TAF} already in 1997 \cite{Wu1997}. The blending weights $a_{j,l}$ are initialized to $\frac{1}{n}$ but are then trained alongside the weights of the NN \cite{Stfeld2020}.
\citeauthor{Stfeld2020} used \gls{tanh}, \gls{ELU}, \gls{ReLU}, \gls{swish}, and the identity as the pool of \glsxtrlongpl{AF} $g_j$ but they admit that no exhaustive search was performed to select this set and that there might be other pools that perform better \cite{Stfeld2020}. This approach was also used in \cite{Pishchik2023} where \gls{ReLU}, \gls{logisticsigmoid}, \gls{tanh}, and \gls{softsign} \glsxtrlongpl{AF} were used.

However, similar approach was already proposed in \cite{Wang2018LookUp} where \citeauthor{Wang2018LookUp} inspired by the \gls{MoGU} (see \cref{sec:mogu}) generalized the concept to mixing several different \glsxtrlongpl{AF}
\begin{equation}
    \label{eq:abu_bias}
    f(z_i) = \sum_{j=0}^n a_{i,j} g_j(z_i - b_{i,j}),
\end{equation}
where $g_j(z_i)$ is an \glsxtrlong{AF} from a pool of $n$ \glsxtrlongpl{AF}, $a_{i,j}$ is a trainable weighting parameter of the function $g_j(z_i)$ and $b_{i, j}$ is a trainable parameter controlling the vertical shift of the function $g_j(z_i)$ for each neuron $i$ \cite{Wang2018LookUp}. Furthermore, if $g_j(z_i)$ already contains a way for controlling its scale or shift, the parameters $a_{i,j}$ and $b_{i, j}$ can be discarded \cite{Wang2018LookUp}. This approach is identical to the \gls{ABU} from \cite{Stfeld2020} if $b_{i, j} = 0$ and the parameters are shared by all neurons in the same layer and not learned for each neuron separately.

A very similar approach was proposed in \cite{Manessi2018}, where \citeauthor{Manessi2018} use a linear combination of \glsxtrlongpl{AF} from a selected pool as the final \glsxtrlong{AF}. The difference from the \gls{ABU} is that the weights are constrained such that they sum up to 1 \cite{Manessi2018, Stfeld2020}. \citeauthor{Manessi2018} uses analyses the linear combination of identity, \gls{ReLU}, and \gls{tanh} \glsxtrlongpl{AF} \cite{Manessi2018}. \citeauthor{Stfeld2020} analyzed the performance of unconstrained ABUs and ABUs with various constraints such as $\sum_{j=0}^n a_{j,l} = 1$, $\sum_{j=0}^n \left|a_{j,l}\right| = 1$, and two approaches enforcing $\sum_{j=0}^n a_{j,l} = 1$ and $a_{j,l} > 0$ --- clipping of negative values $a_{j,l}$ before normalization and \gls{softmax} normalization \cite{Stfeld2020}. It was found that the unconstrained \gls{ABU} works the best on average on the selected tasks; however, some of the constrained variants performed better than the unconstrained \gls{ABU} for particular tasks \cite{Stfeld2020}.

Another variant of \gls{ABU} (called by \citeauthor{Klabjan2019} \textit{activation ensemble}) was proposed in \cite{Klabjan2019} --- the final activation is a weighted sum of \glsxtrlongpl{AF}; the weighting coefficients has to sum-up to 1 (similarly to \cite{Manessi2018}). However, unlike in the work \cite{Manessi2018}, the individual \glsxtrlongpl{AF} are scaled before the weighting to the interval $[0,1]$ using min--max scaling \cite{Klabjan2019}:
\begin{equation}
    h_j(z) = \frac{g_j(z)-\min_k\left(g_j\left(z_k\right)\right)}{\max_k\left(g_j\left(z_k\right)\right)-\min_k\left(g_j\left(z_k\right)\right)+\epsilon},
\end{equation}
where $g_j$ are individual \glsxtrlongpl{AF}, $\epsilon$ is a small number and $k$ goes through all training samples in a minibatch.The final output is
\begin{equation}
    f(z_i) = \sum_{j=0}^n a_{j,i} h_j(z_i),
\end{equation}
where $a_{j,i}$ is a weight for each neuron $i$ and \glsxtrlong{AF} $j$, $n$ is the total number of the individual \glsxtrlongpl{AF} in the \gls{ABU}; the weights $a_{j,i} \in [0,1]$ are constrained such that
\begin{equation}
    \sum_{j=0}^n a_{j,i} = 1.
\end{equation}

\subsubsection{Trainable compound activation function (TCA)}
\label{sec:tca}
The \gls[prereset]{TCA} \cite{Baggenstoss2022} is an \gls{AAF} similar to the \gls{ABU} \cite{Stfeld2020} and especially to its variant with the bias (see \cref{eq:abu_bias}) \cite{Wang2018LookUp}; however, unlike the form from \cref{eq:abu_bias} it uses horizontal scaling instead of the vertical. It was defined in \cite{Baggenstoss2022} as
\begin{equation}
    f(z_i) = \frac{1}{k}\sum_{j=1}^k f_j\left(\exp\left(a_{i,j}\right)z_i + b_{i,j}\right),
\end{equation}
where $k$ is the number of mixed functions and $a_{i,j}$ and $b_{i,j}$, $j=1, \ldots, k$, are scaling and translation trainable parameters for each neuron $i$ and function $j$ \cite{Baggenstoss2022}. The \gls{TCA} was found to improve the performance of \glspl{RBM} and \glspl{DBN} \cite{Baggenstoss2022}.

Later, \citeauthor{Baggenstoss2023} introduced a \gls{TCA} also with vertical scaling parameters in \cite{Baggenstoss2023}. This slightly different variant is denoted as \gls[prereset]{TCAv2} throughout this work. \Gls{TCAv2} is defined as
\begin{equation}
    f(z_i) = \frac{\sum_{j=1}^k \exp\left(a_{i,j}\right) f_j\left(\exp\left(b_{i,j}\right)z_i + c_{i,j}\right)}{\sum_{j=1}^k \exp\left(a_{i,j}\right)},
\end{equation}
where $k$ is the number of mixed functions and $a_{i,j}$, $b_{i,j}$ and $c_{i,j}$, $j=1, \ldots, k$, are scaling and translation trainable parameters for each neuron $i$ and function $j$ \cite{Baggenstoss2023}.

\subsubsection{Average of a pool of activation functions (APAF)}
\label{sec:apaf}
An \gls{APAF} was used in \cite{Liao2020}; the output is defined as
\begin{equation}
    f(z_i) =\frac{\sum_{j=0}^n a_{j,i} h_j(z_i)}{\sum_{j=0}^n a_{j,i} }.
\end{equation}
\citeauthor{Liao2020} used the \gls{ReLU}, \gls{logisticsigmoid}, \gls{tanh}, and the linear functions as the candidate functions in the pool \cite{Liao2020}. This approach was also used in \cite{Pishchik2023}.

\subsubsection{Gating adaptive blending unit (GABU)}
\label{sec:gabu}
Yet another approach previously proposed employs a gated linear combination of \glsxtrlongpl{AF} for each neuron \cite{Dushkoff2016} --- the variant is called \gls{GABU} throughout this work. This allows each neuron to choose which \glsxtrlong{AF} (from an existing pool) it may use to minimize the error \cite{Dushkoff2016}. A similar method uses just binary indicators instead of the gates \cite{Ismail2013}.
The gating variant of \gls{ABU} from \cite{Dushkoff2016} is defined as
\begin{equation}
    f(z_i) = \sum_{j=0}^n \sigma\left(a_{j,i}\right) g_j(z_i),
\end{equation}
where $\sigma\left(a_{j,i}\right)$ is the \gls{logisticsigmoid} function acting as  gating function and $a_{j,i}$ is a trainable parameter controlling the weight of the \glsxtrlong{AF} $g_j$ for each neuron $i$.

\subsubsection{Deep Kronecker neural networks}
\label{sec:dknn}
\glsreset{DKNN}
The concept of ABUs was further generalized in the framework of \glspl{DKNN} \cite{Jagtap2022Deep}, which provides an efficient way of constructing wide networks with \glsxtrlongpl{AAF} while keeping the number of parameters low \cite{Jagtap2022Deep}. \Glspl{DKNN} are equivalent to the \glsxtrlongpl{FFNN} with an \glsxtrlong{AAF} $f$ defined as
\begin{equation}
    \label{eq:dknn}
    f(z_l) = \sum_{j=0}^n a_{l,j} g_j(b_{l,j} z_l),
\end{equation}
where $z_l$ is a preactivation of a neuron from a layer $l$, $a_{l,j}$ and $b_{l,j}$ are either trainable or fixed parameters and $g_j$, $j=1,\ldots,n$ are fixed \glsxtrlongpl{AF} \cite{Jagtap2022Deep}.

\subsubsection{Rowdy activation functions}
\textit{Rowdy activation functions} are a general class of \glsxtrlongpl{AF} that is a special case of \glspl{DKNN} (see \cref{sec:dknn}). A rowdy activation function is a \glspl{DKNN} with any \glsxtrlong{AF} (e.g., \gls{ReLU}) that is the function $g_0$ from \cref{eq:dknn} and $n$ other functions that are defined as
\begin{equation}
    g_j(z_l) = c \cdot \sin\left(jcz_l\right),
\end{equation}
or
\begin{equation}
    g_j(z_l) = c \cdot \cos\left(jcz_l\right),
\end{equation}
where $c\geq 1$ is a fixed scaling factor and $j=1,\ldots,n$ \cite{Jagtap2022Deep}. The rowdy activation functions introduce highly fluctuating, non-monotonic terms that remove saturation regions from the output of each layer in the network \cite{Jagtap2022Deep} similarly as does the stochastic noise in \cite{Gulcehre2016}.

\subsubsection{Self-learnable activation function (SLAF)}
\label{sec:slaf}
The \gls{SLAF} \cite{Goyal2019} can be considered to be a special case of the \gls{ABU} where the function $g_j(z_i)$ are increasing powers of $z_i$:
\begin{equation}
    f(z_i) = \sum_{j=0}^{k-1} a_{i,j}z_i^j,
\end{equation}
where $a_{i,j}$ are learnable parameters for each neuron $i$ and $k$ is a hyperparameter defining the number of elements in the polynomial expression \cite{Goyal2019, Dubey2022}. However, since the gradient is proportional to $z_i$ and its powers, \citeauthor{Goyal2019} used mean-variance normalization over the training sample to avoid exploding or vanishing gradients \cite{Goyal2019}. A similar concept was analyzed in \cite{Piazza1992}, where it was applied to the output neuron only. A similar approach was used independently in \cite{Chen2020ATwostage}, where authors used the equivalent of \gls{SLAF} with $k=6$. A quadratic variant (i.e., \gls{SLAF} with $k=2$) was used in \cite{Fonseca2021}.

\subsubsection{Chebyshev polynomial-based activation function (ChPAF)}
\label{sec:chpaf}
A \gls{ChPAF} was proposed in \cite{Deepthi2023}. The function is defined as
\begin{equation}
    f(z) = \sum_{j=0}^k a_j C_j(z),
\end{equation}
where $a_j$, $j=0,\ldots,k$ are learnable parameters shared by a whole network, $k$ is a fixed hyperparameter denoting the maximum order of used Chebyshev polynomials, and $C_j(z)$ is a Chebyshev polynomial of order $j$ defined as
\begin{equation}
    C_{j+1}(z) = 2zC_j(z)-C_{j-1}(z)
\end{equation}
with starting values $C_0(z) = 1$ and $C_1(z)=z$ \cite{Deepthi2023}. \citeauthor{Deepthi2023} used polynomials of a maximum order of 3 in their experiments \cite{Deepthi2023}. The Chebyshev activation function was found to outperform several \glsxtrlongpl{AF} including \gls{ReLU}, \gls{ELU}, \gls{mish} and \gls{swish} while retaining fast convergence using the CIFAR-10 dataset \cite{Krizhevsky2009} as shown in experiments \cite{Deepthi2023}.

\subsubsection{Legendre polynomial-based activation function (LPAF)}
\label{sec:lpaf}
A \gls{LPAF} was used for the study of approximations of several non-linearities in \cite{Venkatappareddy2021}. The activation is a linear combination of Legendre polynomials and is defined as 
\begin{equation}
    f(z) = \sum_{j=0}^k a_j G_j(z),
\end{equation}
where $a_j$, $j=0,\ldots,k$ are learnable parameters shared by a whole network, $k$ is a fixed hyperparameter denoting the maximum order of used Legendre polynomials, and $G_j(z)$ is a Legendre polynomial of order $j$ defined as
\begin{equation}
    G_{j+1}(z) = \frac{2k+1}{k+1}zG_k(z) - \frac{k}{k+1}G_{k-1}(z_i),
\end{equation}
with starting values $G_0(z) = 1$ and $G_1(z)=z$ \cite{Venkatappareddy2021}. The \gls{LPAF} was found to outperform \gls{ELU}, \gls{ReLU}, \gls{LReLU}, and \gls{softplus} on the MNIST \cite{LiDeng2012} and Fashion MNIST \cite{Xiao2017} datasets \cite{Venkatappareddy2021}.

\subsubsection{Hermite polynomial-based activation function (HPAF)}
\label{sec:hpaf}
The \gls[prereset]{HPAF} \cite{Lokhande2020} is an \gls{AAF} similar to \gls{ChPAF} and \gls{LPAF} but it used the Hermite polynomials instead. It is defined as
\begin{equation}
    f(z) = \sum_{j=0}^k a_j H_j(z),
\end{equation}
where $a_j$ is a trainable parameter and $H_j(z)$ is the Hermite polynomial
\begin{equation}
    H_j(z) = \left(-1\right)^j\exp\left(z^2\right)\dv[j]{z}\left(\exp\left(-z^2\right)\right), j > 0
\end{equation}
and
\begin{equation}
    H_0(z) = 1.
\end{equation}

\subsubsection{Mixture of Gaussian unit (MoGU)}
\label{sec:mogu}
The \gls[prereset]{MoGU} was proposed in \cite{Wang2018LookUp} as a byproduct of analysis of the behavior of the \gls{LuTU} unit (see \cref{sec:lutu}) as the shape of learned activation units with the cosine smoothing mostly composed of a few peaks and valleys \cite{Wang2018LookUp}. The \gls{MoGU} is defined as
\begin{equation}
    f(z_i) = \sum_{j=0}^n \frac{a_{i,j}}{\sqrt{2\pi \sigma_{i,j}^2}} \exp\left(-\frac{\left(z_i-\mu_{i,j}\right)^2}{2\sigma_{i,j}^2}\right),
\end{equation}
where $a_{i,j}$, $\sigma_{i,j}$, and $\mu_{i,j}$ are trainable parameters for each neuron $i$ and Gaussian $j$ from the mixture \cite{Wang2018LookUp}. The parameter $a_{i,j}$ controls the scale, $\sigma_{i,j}$ controls the standard deviation, and $\mu_{i,j}$ controls the mean of the Gaussian $j$ for neuron $i$ \cite{Wang2018LookUp}.

\subsubsection{Fourier series activation}
\label{sec:fsa}
The \gls[prereset]{FSA}  was proposed in \cite{Liao2020}. It is defined as
\begin{equation}
    f(z_i)= a_i + \sum_{j=1}^r \left(b_{i,j} \cos\left(jd_iz_i\right)+c_{i,j}\sin\left(jd_iz_i\right)\right),
\end{equation}
where $a_i$, $b_{i,j}$, $c_{i,j}$, $d_i$ are trainable parameters for each neuron $i$, and $r$ is a fixed hyperparameter denoting the rank of the Fourier series \cite{Liao2020}; \citeauthor{Liao2020} used $r=5$ throughout his experiments.

\subsection{Padé activation unit (PAU)}
\label{sec:pau}
\Glspl[prereset]{PAU} \cite{Molina2020} are adaptive activations based on the Padé approximant \cite{Brezinski2002,Brezinski1996}. The \gls{PAU} is defined as
\begin{equation}
    \label{eq:pau}
    f(z) = \frac{\sum_{j=0}^m a_{j}z^j}{1 + \sum_{k=1}^n b_{k}z^k},
\end{equation}
where $m$ and $n$ are hyperparameters denoting the order of the polynomials and $a_{j}$, $j=0,\ldots,m$ and $b_{k}$, $k=1,\ldots,n$ are trainable parameters that are globaly shared by all units \cite{Molina2020}. While the Padé approximation could be used to approximate particular \glsxtrlong{AF}, the parameters $a_{j}$ and $b_{k}$ are optimized freely with other weights of the \glsxtrlong{NN} \cite{Molina2020}. This \gls{PAU} variant was for reinforcement learning in \cite{Delfosse2021} where \citeauthor{Delfosse2021} observed that rational functions might replace some of the residual blocks in ResNets. To avoid numerical instabilities, a \textit{safe PAU} ensures that the polynomial in the denominator cannot be zero \cite{Molina2020}; it is defined as
\begin{equation}
    \label{eq:safe_pau}
    f(z) = \frac{\sum_{j=0}^m a_{j}z^j}{1 + \left|\sum_{k=1}^n b_{k}z^k\right|}.
\end{equation}
The hyperparameters were set to $m=5$ and $n=4$ in experiments in \cite{Molina2020}. The notion of using rational functions in activations was further analyzed in \cite{Boulle2020} where authors used \glsxtrlong{AF} equivalent to \cref{eq:pau} with distinct parameters for each layer to learn \textit{rational neural networks}; the safe variant of \gls{PAU} (\cref{eq:safe_pau}) was not used as it results in non-smooth \glsxtrlong{AF} and expensive calculation of gradient during training \cite{Boulle2020}. \citeauthor{Boulle2020} used low degrees $m=3$ and $n=2$ in their work \cite{Boulle2020}; this is in contrast to \cite{Chen2018Rational} where rational functions of higher orders were used in a \glsxtrlongpl{GNN}.

\subsection{Randomized Padé activation unit (RPAU)}
\label{sec:rpau}
The \gls{PAU} can be extended similarly as \gls{RReLU} extends \gls{ReLU}, resulting in \gls{RPAU} \cite{Molina2020}. Let $\vec{C} = \left\{ a_{0}, \ldots, a_{m}, b_{0}, \ldots, b_{n}\right\}$ be coefficients of \gls{PAU} activation (see \cref{sec:pau}). Then an additive noise is introduced into each coefficient $c_{j} \in \vec{C}$ during training for every input $z_{k}$ such that $c_{j,k} = c_{j} + z_{j,k}$, where $z_{j,k} \sim \mathrm{U}(l_{j}, u_{j})$, $l_{j} = (1-a)c_{j}$ and $u_{j} = (1+a)c_{j}$ \cite{Molina2020}. This results in \gls{RPAU}:
\begin{equation}
    f(z_{k}) = \frac{c_{0,k}+c_{1,k}z_{k}+c_{2,k}z_{k}^2 +\ldots+c_{m,k}z_{k}^m}{1 + \left|c_{m+1,k}z_{k}+c_{m+2,k}z_{k}^2 +\ldots+c_{m+n,k}z_{k}^n\right|},
\end{equation}
where $z_{k}$ is output of a unit for training input $k$ \cite{Molina2020}.

\subsection{Enhanced rational activation (ERA)}
\label{sec:era}
The \gls{ERA} \cite{Trimmel2022} function is very similar to the original \gls{PAU} (see \cref{sec:pau}); however, \citeauthor{Trimmel2022} note similarly as \citeauthor{Boulle2020} that the safe version of \gls{PAU} is costly to compute whereas the original \gls{PAU} has undefined values on poles (values of $z$ where the denominator in \gls{PAU} is equal to zero). To avoid both the poles and the use of absolute value, a modified rational function without the poles is used \cite{Trimmel2022}. The \gls{ERA} is defined as
\begin{equation}
    \label{eq:era}
    f(z) = \frac{P(z)}{Q_C(z)} = \frac{\sum_{j=0}^m a_{j}z^j}{\epsilon + \Pi_{k=1}^n\left(\left(z-c_k\right)^2+d_k^2\right)},
\end{equation}
where $a_j$, $j=0, \ldots, m$, $c_k$, and $d_k$, $k=1, \ldots, n$ are trainable parameters for each layer and $\epsilon > 0$ is a small number helping to avoid numerical instabilities when $d_k$ are small \cite{Trimmel2022}. In practice, \citeauthor{Trimmel2022} used $\epsilon = 10^{-6}$ \cite{Trimmel2022}. The \gls{ERA} in \cref{eq:era} can be rewritten using partial fractions, which reduces the number of operations and, therefore, leads to more efficient computation \cite{Trimmel2022}. \citeauthor{Trimmel2022} used $m=5$ and $n=4$ for their experiments.

\subsection{Orthogonal Padé activation unit (OPAU)}
\label{sec:opau}
The \gls[prereset]{OPAU} is an extension of the \gls{PAU} proposed in \cite{Biswas2021OrthogonalPade}. It is defined as
\begin{equation}
    \label{eq:opau}
    f(z) = \frac{\sum_{j=0}^m a_{j}r_i(z)}{1 + \sum_{k=1}^n b_{k}r_k(x)},
\end{equation}
where $a_j$, $j=0,\ldots,m$ and $b_k$, $k=1,\ldots,n$ are trainable weights, $m$ and $n$ are fixed parameters, and $r_j(z)$ belongs to a set of orthogonal polynomials \cite{Biswas2021OrthogonalPade}. The \textit{sage \gls{OPAU}} is defined\footnote{Using notation as described in the original article by \citeauthor{Biswas2021OrthogonalPade} \cite{Biswas2021OrthogonalPade}.} as 
\begin{equation}
    \label{eq:safe_opau}
    f(z) = \frac{\sum_{j=0}^m a_{j}r_i(z)}{1 + \sum_{k=1}^n \left|b_{k}\right|\left|r_k(x)\right|},
\end{equation}
with identical parameters as the \gls{OPAU} from \cref{eq:opau} \cite{Biswas2021OrthogonalPade}. \Citeauthor{Biswas2021OrthogonalPade} used six bases for orthogonal polynomials --- Chebyshev polynomials (two variants), Hermite polynomials (also two variants), Laguerre, and Legendre polynomials --- as shown in \cref{tab:opau_bases}.

\begin{table}[h!]
    \centering
    \begin{tabular}{ p{4cm} p{7cm}}
    polynomial & definition\\ \hline
    Chebyshev (first kind) & $r_0(z)=1$, $r_1(z)=z$, $r_{n+1}(z)=2zr_n(z)-r_{n-1}(z)$ \\
    Chebyshev (second kind) & $r_0(z)=1$, $r_1(z)=2z$, $r_{n+1}(z)=2zr_n(z)-r_{n-1}(z)$ \\
    Laguerre & $r_0(z)=1$, $r_1(z)=1-z$, $r_{n+1}(z)=\frac{\left(2n+1-z\right)r_n\left(z\right)-nr_{n-1}\left(z\right)}{n+1}$ \\
    Legendre & $r_n(z) = \sum_{k=0}^{\left[\frac{n}{2}\right]}\left(-1\right)^k\frac{\left(2n-2k\right)!}{2^nk!\left(n-2k\right)!(n-k)}z^{n-2k}$ \\
    Probabilist's Hermite & $r_n(z) = \left(-1\right)^n\exp\left(\frac{z^2}{2}\right)\dv[n]{z}\left(\exp\left(-\frac{z^2}{2}\right)\right)$ \\
    Physicist's Hermite & $r_n(z) = \left(-1\right)^n\exp\left(z^2\right)\dv[n]{z}\left(\exp\left(-z^2\right)\right)$ \\
    \hline
    \end{tabular}
    \caption[Polynomial bases used in OPAU]{\textbf{Polynomial bases used in \glsxtrshort{OPAU}} \\ List of polynomial bases used in the \glsxtrshort{OPAU} taken \cite{Biswas2021OrthogonalPade}. The Chebyshev polynomials and the Laguerre polynomial have recurrent definitions; whereas the Legendre and the Hermite polynomials are defined by a single expression.}
    \label{tab:opau_bases}
\end{table}

\subsection{Spline interpolating activation functions}
\label{sec:saf}
More complex approaches include \glspl{SAF} \cite{Vecci1998,Guarnieri1999,Solazzi2000,Scardapane2018,Campolucci1996, Bohra2020, Lane1990, Vaicaitis2022, Kumar2014, Mayer2001, Mayer2002, Solazzi2000Neural,Neumayer2023,Ducotterd2022,Aziznejad2019}, which facilitate the training of a wide variety of \glsxtrlongpl{AF} using interpolation. One common example is the cubic spline interpolation that was used in \cite{Scardapane2018}. The SAFs are controlled by a vector $\vec{q} \in \mathbb{R}^k$ of internal parameters called \textit{knots}, which are a sampling of the AF over $k$ representative points \cite{Scardapane2018}. The output is computed using a spline interpolation using the closest knot and its $p$ rightmost neighbors; $p=3$ results in cubic interpolation \cite{Scardapane2018}. Spline-based \glsxtrlongpl{AF} were also used in the ExSpliNet --- an interpretable approach combining \glsxtrlongpl{NN} and ensembles of probabilistic trees \cite{Fakhoury2022}. A set of fixed but highly redundant knots for spline interpolation was used in \cite{Bohra2020}, where the authors then relied on the sparsifying effect of $L_1$ regularization to nullify the coefficients that are not needed \cite{Bohra2020}. Spline flexible \glsxtrlongpl{AF} were used for sound synthesis in \cite{Uncini2002}. The usage of splines led to the creation of b-spline-based \glsxtrlongpl{NN}, e.g., \cite{Kuzuya2022}.

Similar to the \gls{SAF} is the \gls{PPAF} \cite{LpezRubio2019} that is also defined by a number of points where the function switches from one polynomial to another \cite{LpezRubio2019}. \citeauthor{LpezRubio2019} used zeroth-order, first-order, and third-order polynomials for the piecewise function \cite{LpezRubio2019}; for example, zeroth-order \gls{PPAF} uses step function and is defined as
\begin{equation}
    f(z_i) = \begin{cases}
        0, \quad & z_i < q_{i, 1}, \\
        \frac{k}{m}, \quad q_{i, k} \leq & z_i < q_{i, k+1}, \\
        1, \quad & z_i \geq q_{i, m}, \\
    \end{cases}
\end{equation}
where $m-1$ is the number of controlling points $q_{i,k}$ of a neuron $i$, $k \in \{1, 2, \ldots, m-1\}$ \cite{LpezRubio2019}. The position of control points is determined using the learning procedure outlined in \cite{LpezRubio2019}.

If there are no constrains and the \gls{AF} is limited to linear splines, the \gls{AF} can be also defined using one hidden layer with \glspl{ReLU} \cite{Neumayer2023}:
\begin{equation}
    f(z_i) = \sum_{k=1}^K a_{i,k} \mathrm{ReLU}\left(b_{i,k}z_i + c_{i,k}\right),
\end{equation}
where $K\in \mathbb{N}$ and $a_{i,k}$, $b_{i,k}$, and $c_{i,k}$ are trainable parameters \cite{Neumayer2023}.

\subsection{Truncated Gaussian unit (TruG)}
\label{sec:trug}
A \gls{TruG} \cite{Su2017} is a unit in a probabilistic framework that is able to well approximate sigmoid, \gls{tanh}, and \gls{ReLU}. It is controlled by truncation points $\xi_1$ and $\xi_2$ and under the probabilistic framework described in \cite{Su2017} is defined as 
\begin{equation}
    \E(h|z, \xi_1, \xi_2) = z + \sigma \frac{\phi\left(\frac{\xi_1-z}{\sigma}\right)-\phi\left(\frac{\xi_2-z}{\sigma}\right)}{\Phi\left(\frac{\xi_1-z}{\sigma}\right)-\Phi\left(\frac{\xi_2-z}{\sigma}\right)},
\end{equation}
where $\phi(x)$ is the probability density function (PDF) of a univariate Gaussian distribution with mean $z$ and variance $\sigma^2$ and $\Phi(x)$ its \gls{CDF} \cite{Su2017}. The truncation points can be either selected manually or tuned with the rest of the weights \cite{Su2017}.

\subsection{Mollified square root function (MSRF) family}
\label{sec:msrf}
\Citeauthor{Pan2023Smoothing} used a smoothing approach on piecewise linear \glspl{AF} to create a whole new family of \glspl{AF} in \cite{Pan2023Smoothing}. The approach is based on the \gls[prereset]{MSRF} method. This smoothing approach was first used in \cite{Bresson2007} and then in the \gls{SquarePlus} \gls{AF} in \cite{Barron2021}, which inspired \citeauthor{Pan2023Smoothing} in the creation of the \gls{MSRF} family of \glspl{AF}.

For example, the absolute value $|x|$ is not differentiable at $x=0$, but it can be regularized by mollification as
\begin{equation}
    \left|x\right|_\epsilon = \sqrt{x^2+\epsilon},
\end{equation}
where $\epsilon$ is a small positive parameter and $\lim_{\epsilon\rightarrow0^+}\left|x\right|_\epsilon = \left|x\right|$ \cite{Pan2023Smoothing}.

\subsubsection{SquarePlus}
\label{sec:squareplus}
The \gls{SquarePlus} \cite{Barron2021} is the first \gls{AF} that used the mollification procedure described in \cref{sec:msrf} above. It is defined as
\begin{equation}
    f(z) = \frac{1}{2}\left(z + \left|z\right|_\epsilon\right) = \frac{1}{2}\left(z + \sqrt{z^2+\epsilon}\right).
\end{equation}
The \gls{SquarePlus} is very similar to the \gls{softplus} (see \cref{sec:softplus}) for $\epsilon = 4\left(\ln\left(2\right)\right)^2$ and they produce identical outputs at $z=0$ \cite{Pan2023Smoothing}.

\subsubsection{StepPlus}
\label{sec:stepplus}
As the \gls{SquarePlus} approximates the \gls{ReLU}, the \gls{StepPlus} approximates the step function (see \cref{sec:binaryaf}) similarly as the \gls{logisticsigmoid} does \cite{Pan2023Smoothing}. It is defined as
\begin{equation}
    f(z) = \frac{1}{2}\left(1+\frac{z}{\left|z\right|_\epsilon}\right).
\end{equation}

The sign function is smoothed into the \gls{BipolarPlus} \gls{AF} \cite{Pan2023Smoothing}
\begin{equation}
    f(z) = \frac{z}{\left|z\right|_\epsilon}.
\end{equation}

\subsubsection{LReLUPlus}
\label{sec:lreluplus}
A smoothed variant of the \gls{LReLU} called \gls{LReLUPlus} is defined as
\begin{equation}
    f(z_i) = \frac{1}{2}\left(z_i+a_iz_i+\left|\left(1-a_i\right)z_i\right|_\epsilon\right),
\end{equation}
where $\left|x\right|_\epsilon$ is the \gls{MSRF} procedure \cite{Pan2023Smoothing} described in \cref{sec:msrf} and $a_i$ is a fixed or trainable parameter. 

A function equivalent to the \gls{LReLUPlus} was independently proposed in \cite{Biswas2022Smooth} under the name \gls{SMU1}. The only difference was that \citeauthor{Biswas2022Smooth} used parameter $\mu$ that is the square root of $\epsilon$ from \cref{sec:msrf}: $\epsilon=\mu^2$.

\subsubsection{vReLUPlus}
\label{sec:vreluplus}
The \gls{vReLUPlus} \cite{Pan2023Smoothing} is a \gls{MSRF} smoothed variant of the \gls{vReLU} (see \cref{sec:vrelu}); it is defined as
\begin{equation}
    f(z) = \left|z\right|_\epsilon.
\end{equation}

\subsubsection{SoftshrinkPlus}
\label{sec:softshrinkplus}
The smoothed variant of the \gls{softshrink} (see \cref{sec:softshrink}) is named \gls{SoftshrinkPlus}\footnote{\Citeauthor{Pan2023Smoothing} named the function STFPlus originally \cite{Pan2023Smoothing}.} \cite{Pan2023Smoothing} and is defined as
\begin{equation}
    f(z) = z + \frac{1}{2}\left(\sqrt{\left(z-a\right)^2 + \epsilon}- \sqrt{\left(z+a\right)^2 + \epsilon}\right),
\end{equation}
where $a$ is a fixed parameter similar to the original \gls{softshrink}'s thresholding parameter \cite{Pan2023Smoothing}.

\subsubsection{PanPlus}
\label{sec:panplus}
The \gls{MSRF} procedure can be also used to smooth the \gls{pan} \gls{AF} (see \cref{sec:pan}) \cite{Pan2023Smoothing}; the resulting \gls{PanPlus} \cite{Pan2023Smoothing} is defined as 
\begin{equation}
    f(z) = -a + \frac{1}{2}\left(\sqrt{\left(z-a\right)^2 + \epsilon} + \sqrt{\left(z+a\right)^2 + \epsilon}\right),
\end{equation}
where $a$ is a fixed thresholding parameter of the \gls{pan} function \cite{Pan2023Smoothing}.

\subsubsection{BReLUPlus}
\label{sec:breluplus}
The \gls{BReLUPlus} \cite{Pan2023Smoothing} is a \gls{MSRF} smoothed variant of the \gls{BReLU} (see \cref{sec:brelu}) defined as
\begin{equation}
    f(z) = \frac{1}{2}\left(1+ \left|z\right|_\epsilon - \left|z-1\right|_\epsilon \right).
\end{equation}

\subsubsection{SReLUPlus}
\label{sec:sreluplus}
Another smoothed \gls{AF} is the \gls{SReLUPlus} which is the smoothed variant of the \gls{SReLU} (see \cref{sec:srelu}) \cite{Pan2023Smoothing}; it is defined as
\begin{equation}
    f(z_i) = a_iz_i + \frac{1}{2}\left(a_i-1\right)\left(\left|z_i-t_i\right|_\epsilon - \left|z_i+t_i\right|_\epsilon\right),
\end{equation}
where $a_i$ has similar role as in the original \gls{SReLU} and $t_i$ is a parameter for symmetric variant of \gls{SReLU} with $t_i = t_i^r = t_i^l$ \cite{Pan2023Smoothing}.

\subsubsection{HardTanhPlus}
\label{sec:hardtanhplus}
Similarly, the smoothed variant of the \gls{hard_tanh} (see \cref{sec:hard_tanh}) named \gls{HardTanhPlus} \cite{Pan2023Smoothing} is defined as
\begin{equation}
    f(z) = \frac{1}{2}\left(\left|z+1\right|_\epsilon - \left|z-1\right|_\epsilon\right).
\end{equation}

\subsubsection{HardshrinkPlus}
\label{sec:hardshrinkplus}
The smoothed variant of the \gls{hardshrink} (see \cref{sec:hardshrink}) is named \gls{HardshrinkPlus}\footnote{\Citeauthor{Pan2023Smoothing} named the function HTFPlus originally \cite{Pan2023Smoothing}.} \cite{Pan2023Smoothing}; it is defined as
\begin{equation}
    f(z) = z \left(1 + \frac{1}{2}\left(\frac{z-a}{\sqrt{\left(z-a\right)^2+\epsilon}}-\frac{z+a}{\sqrt{\left(z+a\right)^2+\epsilon}}\right)\right),
\end{equation}
where $a$ is a fixed parameter with a similar function as in the \gls{hardshrink} \cite{Pan2023Smoothing}.

\subsubsection{MeLUPlus}
\label{sec:meluplus}
\Citeauthor{Pan2023Smoothing} also provided a smoothed variant of the \gls{MeLU} (see \cref{sec:melu}); however, the formula written in \cite{Pan2023Smoothing} is not the \gls{MeLU} \gls{AF} but rather its single component $\phi_{b_jc_j}\left(z_i\right)$. Nevertheless, the full smoothed \gls{MeLUPlus} can be obtained easily as the combination of the \gls{LReLUPlus} and the smoothed $\phi_{b_jc_j}^\text{Plus}\left(z_i\right)$ defined as
\begin{equation}
    \phi_{b_jc_j}\text{Plus}\left(z_i\right) = \frac{1}{2} \left(c_j - \left|z_i-b_j\right|_\epsilon + \sqrt{\left(c_j - \left|z_i-b_j\right|_\epsilon\right)^2+\epsilon}\right),
\end{equation}
where $b_j$ and $c_j$ are the same parameters as in the \gls{MeLU}.

\subsubsection{TSAFPlus}
\label{sec:tsafplus}
The smoothed variant of the \gls{TSAF} (see \cref{sec:tsaf}) named \gls{TSAFPlus} \cite{Pan2023Smoothing} is defined as
\begin{equation}
    f(z_i) = \frac{1}{c_i}\left(\left|z_i-a_i+c_i\right|_\epsilon+\left|z_i-a_i\right|+\left|z_i+b_i-c_i\right|-\left|z_i-b_i\right|\right),
\end{equation}
where $a_i$, $b_i$, and $c_i$ have a similar role as in the original \gls{TSAF} \cite{Pan2023Smoothing}.

\subsubsection{ELUPlus}
\label{sec:eluplus}
Even the \gls{ELU} (see \cref{sec:elu}) can be mollified into a "smoothed" variant named \gls{ELUPlus} \cite{Pan2023Smoothing}.  The smoothed variant is defined as
\begin{equation}
    f(z) = \frac{1}{2}\left(z+\left|z\right|_\epsilon\right)+\frac{1}{2}\left(\frac{\exp\left(z\right)-1}{a}+\left|\frac{\exp\left(z\right)-1}{a}\right|_\epsilon\right),
\end{equation}
where $a$ is a fixed parameter\footnote{\Citeauthor{Pan2023Smoothing} used variant with inverse parameter $\frac{1}{a}$; we have used the same parameter variant as in the original \gls{ELU}.} with a similar function as in the \gls{ELU} \cite{Pan2023Smoothing}.

\subsubsection{SwishPlus}
\label{sec:swishplus}
The mollified variant of the \gls{swish} (see \cref{sec:swish}) named \gls{SwishPlus} \cite{Pan2023Smoothing} is defined using the smoothed step function instead of the \gls{logisticsigmoid}; it is, therefore, defined as
\begin{equation}
    f(z) = z\cdot\mathrm{StepPlus}\left(z\right)=\frac{1}{2}\left(z+\frac{z^2}{\left|z\right|_\epsilon}\right).
\end{equation}

\subsubsection{MishPlus}
\label{sec:mishplus}
The mollified variant of the \gls{swish} (see \cref{sec:mish}) named \gls{MishPlus} \cite{Pan2023Smoothing} is defined using the \gls{BipolarPlus} and \gls{SquarePlus} as

\begin{equation}
    f(z) = z\cdot\mathrm{BipolarPlus}\left(\mathrm{BipolarPlus}\left(z\right)\right).
\end{equation}

\subsubsection{LogishPlus}
\label{sec:logishplus}
The mollified variant of the \gls{logish} (see \cref{sec:logish}) named \gls{LogishPlus} \cite{Pan2023Smoothing} is defined as

\begin{equation}
    f(z) = z\cdot \ln\left(1+\mathrm{StepPlus}\left(z\right)\right).
\end{equation}

\subsubsection{SoftsignPlus}
\label{sec:softsignplus}
The mollified variant of the \gls{softsign} (see \cref{sec:softsign}) named \gls{SoftsignPlus} \cite{Pan2023Smoothing} is defined as

\begin{equation}
    f(z) = \frac{z}{1+\left|z\right|_\epsilon}.
\end{equation}

\subsubsection{SignReLUPlus}
\label{sec:signreluplus}
\Citeauthor{Pan2023Smoothing} provide a mollified version for an approximation of the \gls{SignReLU}\footnote{\Citeauthor{Pan2023Smoothing} call it \gls{DLU} throughout their work \cite{Pan2023Smoothing}.} (see \cref{sec:signrelu}) in \cite{Pan2023Smoothing}. They approximate the \gls{SignReLU} as
\begin{equation}
    \mathrm{SignReLU}(z) = \frac{1}{2}\left(z+\left|z\right|\right) + \frac{z-\left|z\right|}{2\left|1-z\right|_\epsilon}.
\end{equation}
Using the approximation, they then define the \gls{SignReLUPlus} as
\begin{equation}
    \mathrm{SignReLU}(z) = \frac{1}{2}\left(z+\left|z\right|_\epsilon\right) + \frac{z-\left|z\right|_\epsilon}{2\left|1-z\right|_\epsilon}.
\end{equation}

\subsection{Complex approaches}
\label{sec:complex_approaches}
The \gls[prereset]{NIN} \cite{Lin2013}, which uses a micro \glsxtrlong{NN} as an \glsxtrlong{AAF}, represents a different approach. A combination of the \gls{NIN} and \glspl{maxout} called \gls{MIN} was shown to have good performance in \cite{Chang2015}. A similar approach is the \gls[prereset]{WHE} layer \cite{Wang2020Wide}, which is a sparselly connected layer with several \glsxtrlongpl{AF} that is used in place of a traditional \glsxtrlong{AF} \cite{Wang2020Wide}.

\Glsxtrlongpl{AAF} called \gls{NPF} that are learned nonparametrically were proposed in \cite{Eisenach2017} where a Fourier series basis expansion is used for nonparametric estimation. Only one \gls{NPF} is learned per filter in \glspl{CNN} while different activation is learned in each neuron of a fully connected layer \cite{Eisenach2017}; the learning is in two stages where the network is first learned with \glspl{ReLU} in the convolution layers and \gls{NPF} in all others and only then the network is learned with all \glsxtrlongpl{AF} being the \gls{NPF} \cite{Eisenach2017}.

Yet another approach is learning \glsxtrlongpl{AF} using hypernetworks \cite{Vercellino2017} resting in \textit{hyperactivations}. The hyperactivation consists of two parts --- a shallow \glsxtrlong{FFNN} called activation network and a hypernetwork, which is a type of \glsxtrlong{NN} that produces weights for another network \cite{Vercellino2017}. The hypernetwork is used for the normalization of the activation network. A single hyperactivation is learned for each layer in the \glsxtrlong{NN} \cite{Vercellino2017}. A \gls{NN} with a combination of more \glsxtrlongpl{AF} was used in \cite{Zhang2021ASelfAdaptive}.

The \glsxtrlong{AAF} might also be trained in a semi--supervised manner \cite{Castelli2012, Castelli2012b, Castelli2014}.


\subsubsection{Variable activation function (VAF)}
\label{sec:vaf}
Similarly to \gls{NIN}, the \gls{VAF} subnetwork approach uses simple \glsxtrlongpl{AF} to produce more complex behavior \cite{Apicella2019}; the activation is replaced by a small subnetwork with one hidden layer with $k$ neurons and only one input and one output neuron \cite{Apicella2019}. Specifically, \gls{VAF} is defined as
\begin{equation}
    f(z_l) = \sum_{j=1}^k a_{l,j}\mathrm{g}\left(b_{l,j}z_l+c_{l, j}\right) + a_{l,0},
\end{equation}
where $a_{l,0}$, $a_{l,j}$, $b_{l,j}$, and $c_{l,j}$, $j=1, \ldots, k$, are trainable parameters for each layer $l$ and $\mathrm{g}(x)$ is an \glsxtrlong{AF} such as \gls{tanh} or \gls{ReLU} that were used in experiments with \gls{VAF} in \cite{Apicella2019}. The same concept of using subnetwork to learn the \glsxtrlong{AF} was also proposed under the name of \textit{activation function unit} (AFU) in \cite{Minhas2019}.

\subsubsection{Flexible activation bag (FAB)}
\label{sec:fab}
The \gls{FAB} \cite{Klopries2023} is an approach similar to \gls{NIN} and \gls{VAF} as it uses a subnetwork to learn the \gls{AF} for each layer $l$ using a pool of $K$ activations $f_k\left(z_l,\vec{a_{l,k}}\right)$. It uses a shallow network with double head with \gls{ReLU} activation in the first layer; then there are two separate heads\cite{Klopries2023}. The first head predicts the parameters $\vec{a_{l,k}}$  of the individual \glspl{AF} $f_k\left(z_l,\vec{a_{l,k}}\right)$ in the bag squashed by a \gls{sigmoid} \gls{AF}, then the parameters are mapped into a valid range of each of the parameters \cite{Klopries2023}. The second head is a selective head for selecting an appropriate \gls{AF} by producing a score $s_{l,k}$ --- it can be either discrete or continuous resulting in soft or hard selection \cite{Klopries2023}.  \Citeauthor{Klopries2023} used five selection methods --- all of the functions are used ($s_{l,k}=1$), hard selection, soft selection using \glspl{logisticsigmoid}, \gls{softmax} selection, and Gumber-Softmax \cite{Jang2017} selection \cite{Klopries2023}. The bag of activations used in the \gls{FAB} consists of a constant function, linear function, exponential function, step function, \gls{ReLU}, step function, sine function, \gls{tanh}, \gls{logisticsigmoid}, and Gaussian function (see \cite{Klopries2023} for exact definitions with the adaptive parameters). Then the output of \gls{FAB} is assembled as
\begin{equation}
    f(z_l) = \sum_{k=1}^K s_{l,k}f_k\left(z_l,\vec{a_{l,k}}\right),
\end{equation}
where $s_{l,k}$ are the selection scores of $f_k$; the $\vec{a_{l,k}}$ consists of parameters of the function $f_k$ (the used \glspl{AF} have from one to three parameters) and the $f_k$ are the individual \glspl{AF} from the bag of $K$ functions \cite{Klopries2023}.

\subsubsection{Dynamic parameter ReLU (DY--ReLU)}
\label{sec:dyrelu}
The \gls{DY--ReLU} (proposed under the name of dynamic \gls{ReLU} in \cite{Chen2020Dynamic} but that collides with previously proposed \glspl{DReLU} in \cite{Si2018, Hu2019}) is an \glsxtrlong{AF} whose parameters are input dependent \cite{Chen2020Dynamic}. The concept of \gls{DY--ReLU} is similar to the \gls{WHE} in \cite{Wang2020Wide} and hyperactivations in \cite{Vercellino2017} as the \gls{DY--ReLU} is an example of a hyperactivation. The dynamic \glsxtrlong{AF} has two components --- hyperfunction that computes parameters for the \glsxtrlong{AF} and the \glsxtrlong{AF} itself \cite{Chen2020Dynamic}. The \gls{DY--ReLU} piecewise linear function is computed as the maximum of multiple linear functions \cite{Dubey2022}. It is defined as
\begin{equation}
    f(z_i) = \max_{1\leq k \leq K} \left(a_{i,k}z_i+b_{i,k}\right),
\end{equation}
where $K$ is a hyperparameter and $a_{i,k}$ and $b_{i,k}$ are coefficients determined by the hyper function $\vec{\theta}(\vec{z})$ using all inputs $z_i$ \cite{Chen2020Dynamic}. The hyperfunction $\vec{\theta}(\vec{z})$ is a light-weight \glsxtrlong{NN} \cite{Chen2020Dynamic}. The parameters generated by the hyperfunction $\vec{\theta}(\vec{z})$ can be different for each filter $i$, or they can be shared in the whole layer \cite{Chen2020Dynamic}. The \gls{DY--ReLU} can be considered as a dynamic and efficient variant of Maxout (see \cref{sec:maxout}) \cite{Chen2020Dynamic}.

\subsubsection{Random NNs with trainable activation functions}
\label{sec:rand_nn_trainable_afs}
A very different approach based on \glsxtrlongpl{AAF} is presented in \cite{Erturul2018} where a \glsxtrlong{NN} with random weights is initialized, and the weights are not trained, but the \glsxtrlongpl{AF} are trained instead. The \glsxtrlongpl{AF} in \cite{Erturul2018} are polynomial \glsxtrlongpl{AF} and are trained separately for each hidden neuron with random weights first; only then the weights of the output layer are estimated \cite{Erturul2018}. \citeauthor{Erturul2018} used five different adaptive variants of \glsxtrlongpl{AF}:
\begin{equation}
    \label{eq:Erturul2018_af1}
    f(z_i) = \frac{1}{1+\exp\left(-a_iz_i - b_i\right)},
\end{equation}
\begin{equation}
    \label{eq:Erturul2018_af2}
    f(z_i) = \sin\left(a_iz_i + b_i\right),
\end{equation}
\begin{equation}
    \label{eq:Erturul2018_af3}
    f(z_i) = \exp\left(-a_i||z_i-b_i||\right),
\end{equation}
\begin{equation}
    \label{eq:Erturul2018_af4}
    f(z_i) = \begin{cases}
       1, \quad & a_iz_i+b_i \leq 0, \\
       0, \quad & \text{otherwise},
    \end{cases}
\end{equation}
and
\begin{equation}
    \label{eq:Erturul2018_af5}
    f(z_i) = \sqrt{||z_i-a_i||^2+b_i^2},
\end{equation}
where $a_i$ and $b_i$ are trainable parameters \cite{Erturul2018}.

\subsubsection{Kernel activation function (KAF)}
\label{sec:kaf}
A \gls[prereset]{KAF} \cite{Scardapane2019} is a non-parametric function that uses kernel expansion together with a dictionary to make the activation flexible \cite{Dubey2022}. The \gls{KAF} uses a weighted sum of kernel terms:
\begin{equation}
    f(z_i) = \sum_{j=1}^D a_{i,j}\kappa\left(z_i, d_j\right),
\end{equation}
where $D$ is a fixed hyperparameter, $a_{i,j}$ are mixing coefficients and $d_j$, $j=1,\ldots,D$ are called \textit{dictionary elements} and $\kappa(z_i,d_j): \mathbb{R}\times\mathbb{R} \rightarrow \mathbb{R}$ is 1D kernel function --- \citeauthor{Scardapane2019} consider only $a_{i,j}$ trainable and the dictionary elements $d_j$ are uniformly spaced around zero \cite{Scardapane2019}. This has the advantage that the resulting model is linear in its parameters and, therefore, can efficiently optimized \cite{Scardapane2019}.

The kernel function $\kappa(z,d_j)$ used in \cite{Scardapane2019} is the 1D Gaussian kernel defined as
\begin{equation}
    \kappa(z,d_j) = \exp\left(-\gamma\left(z-d_j\right)^2\right),
\end{equation}
where $\gamma \in \mathbb{R}$ is a fixed parameter called the \textit{kernel bandwith} \cite{Scardapane2019}. \citeauthor{Scardapane2019} recommend setting the kernel bandwidth to 
\begin{equation}
    \gamma = \frac{1}{6\Delta^2},
\end{equation}
where $\Delta$ is the distance betwen the grid points as adapting $\gamma$ through back-propagation did not yield any gain in accuracy \cite{Scardapane2019}. The mixing coefficients $a_{i,j}$ can be initialized either randomly from a normal distribution --- this provided good diversity for the optimization process \cite{Scardapane2019} --- or using kernel ridge regression to approximate an \glsxtrlong{AF} of choice \cite{Scardapane2019}.
\citeauthor{Scardapane2019} also proposed 2D-\gls{KAF} that works over all possible pairs of incoming values and uses 2D Gaussian kernel \cite{Scardapane2019}.

An extension of the \gls{KAF} approach was presented in \cite{Kiliarslan2022} where \glsxtrlong{AF} used was the sum of the \gls{KAF} and \gls{rsigelu} (see \cref{sec:rsigelu}) or \gls{KAF} and \gls{rsigelud} (see \cref{sec:rsigelud}). Kernel methods are becoming more common in deep learning --- e.g., fully kernected layers are replacing fully connected layers with a kernel-based approach in \cite{Zhang2023Fully}.

\subsection{SAVE-inspired activation functions}
\label{sec:save}
\Citeauthor{Brad2023} produced several \gls{AF} that are, supposedly, motivated by human behavior in \cite{Brad2023}. These \glspl{AF} were created using the \glsxtrshort{SAVE} method \cite{Brad2022} and are mostly variations of the \glspl{AF} listed above. For completeness' sake, a list of these \glspl{AF} is included in our work in \cref{tab:save_inspired} also with the real-life motivations listed in \cite{Brad2023} --- however, no deeper analysis or objective evaluation of these \glspl{AF} was not provided in \cite{Brad2023}.
\begin{landscape}
\begin{table}[h!]
    \centering
    \begin{tabular}{ p{7cm} p{3cm} p{7cm}}
    formula & parameters & principle from \cite{Brad2023}\\ \hline
    $a \sin\left(bz + c\right)$ & $a$, $b$, $c$ & activation of resonance \\
    $a \tanh\left(bz + c\right)$ & $a$, $b$, $c$ & activation of resonance \\
    $z + a$ & $a$ & introduction of neutral elements \\
    $a \left(\mathrm{ReLU}\left(z\right)\right)^b$ & $a$, $b$ & action against the wolf-pack spirit \\
    $ a \exp\left(bz\right)$ & $a$, $b$ & activation of centrifugal forces \\
    $a_1\mathrm{ReLU}\left(f_1(z)\right) + \ldots + a_n\mathrm{ReLU}\left(f_n(z)\right)$ & $a_1,\ldots,a_n$, $f_1(z),\ldots,f_n(z)$ & application of multi-level connections \\
    $\mathrm{ReLU}\left(z-a\right)$ & $a$ & application of asymmetry \\
    $a_1\mathrm{ReLU}\left(z\right) + \ldots + a_n\mathrm{ReLU}\left(z\right)$ & $a_1,\ldots,a_n$ & harmonization of individual goals with collective goal \\
    $a z^{1-b}$ & $a$ , $b$ & transformation for value-added \\
    $a\left(1-\exp\left(-bz\right)\right)$ & $a$ , $b$ & transformation for value-added \\
    $\mathrm{ReLU}\left(f(z)g(z)\right)$ & $f(z)$, $g(z)$ & application of prisoner paradox \\
    $\mathrm{ReLU}\left(\frac{\exp\left(z\right)}{\frac{1}{1+\exp\left(-az\right)}}\right)$ & $a$ & application of prisoner paradox \\
    $\mathrm{ReLU}\left(\frac{az+b}{cz+d}\right)$ & $a$, $b$, $c$, $d$ & application of shipwrecked paradox \\
    \hline
    \end{tabular}
    \caption[SAVE-inspired activations]{\textbf{SAVE-inspired activations} \\ The list of \glsxtrshort{SAVE}-inspired \glspl{AF} from \cite{Brad2023}.}
    \label{tab:save_inspired}
\end{table}
\end{landscape}

\FloatBarrier 

\section{Conclusion}
\label{sec:conclusion}
This paper provides an extensive survey of 400 \glsxtrlong{NN} \glsxtrlongpl{AF}. Despite all its scope, it has some limitations and focuses on the largest family of real-valued \glsxtrlongpl{AF} categorized into two main classes: fixed \glsxtrlongpl{AF} and adaptive \glsxtrlongpl{AF}. The fixed \glsxtrlongpl{AF} that we also refer to as classical are predetermined mathematical functions that apply the same transformation to all inputs regardless of their values. Each neuron within a layer typically applies the same activation function to its inputs. Examples include \gls{logisticsigmoid}, \glsxtrlong{tanh}, and \gls{ReLU} \cite{Nair2010} functions. On the other hand, \glsxtrlongpl{AAF} learn their parameters based on the input data and may thus change their shape. This feature allows for more flexibility and leads to faster convergence during training and improved performance \cite{Jagtap2020}. Examples of \glsxtrlongpl{AAF} include \gls{PReLU} \cite{He2015}, \gls{PELU} \cite{Trottier2016}, \gls{swish} \cite{Ramachandran2017}, and \gls{TAAF} \cite{Kunc2021, Kunc2020}.

The profound impact of \glsxtrlongpl{AF} on network performance is undeniable, and the absence of a consolidated resource often results in redundant proposals and wasteful reinvention. By offering this comprehensive compilation, we aim to prevent the unnecessary duplication of established \glsxtrlongpl{AF}. While recognizing the limitations of our work in not conducting extensive benchmarks or in-depth analyses, we believe this exhaustive list will be a valuable reference for researchers. Even though this list will never be complete due to ongoing proposals of new \glsxtrlongpl{AF}, we believe it establishes a solid foundation for future research.

\defbibheading{bibintoc}[References]{%
  \phantomsection
  \pdfbookmark[0]{#1}{#1}
  \section*{#1}%
}
\AtNextBibliography{\scriptsize}
\printbibliography[heading=bibintoc]

@article{Chen2016b,
  doi = {10.1093/bioinformatics/btw074},
  url = {https://doi.org/10.1093/bioinformatics/btw074},
  year = {2016},
  month = feb,
  publisher = {Oxford University Press ({OUP})},
  volume = {32},
  number = {12},
  pages = {1832--1839},
  author = {Yifei Chen and Yi Li and Rajiv Narayan and Aravind Subramanian and Xiaohui Xie},
  title = {Gene expression inference with deep learning},
  journal = {Bioinformatics}
}

@inproceedings{Nair2010, 
    Publisher = {Omnipress}, 
    Title = {Rectified Linear Units Improve Restricted Boltzmann Machines}, 
    Url = {http://www.cs.toronto.edu/~fritz/absps/reluICML.pdf}, 
    Booktitle = {Proceedings of the 27th International Conference on Machine Learning (ICML-10)}, 
    Author = {Vinod Nair and Geoffrey E. Hinton}, 
    Editor = {Johannes Fürnkranz and Thorsten Joachims}, 
    Year = {2010}, 
    Pages = {807-814} 
   }

@article{Srivastava2014,
 author = {Srivastava, Nitish and others},
 title = {Dropout: A Simple Way to Prevent Neural Networks from Overfitting},
 journal = {J. Mach. Learn. Res.},
 issue_date = {January 2014},
 volume = {15},
 number = {1},
 month = {1},
 year = {2014},
 issn = {1532-4435},
 pages = {1929--1958},
 numpages = {30},
 url = {https://www.cs.toronto.edu/~hinton/absps/JMLRdropout.pdf},
 acmid = {2670313},
 publisher = {JMLR.org},
}

@techreport{Go2009,
 author = {Alec Go and Richa Bhayani and Lei Huang},
 title = {Twitter Sentiment Classification using Distant Supervision},
 year = {2009},
 url = {https://www-cs.stanford.edu/people/alecmgo/papers/TwitterDistantSupervision09.pdf},
 publisher = {Stanford University},
}

@inproceedings{Sahni2017,
  doi = {10.1109/comsnets.2017.7945451},
  url = {https://doi.org/10.1109/comsnets.2017.7945451},
  year = {2017},
  month = jan,
  publisher = {{IEEE}},
  author = {Tapan Sahni and Chinmay Chandak and Naveen Reddy Chedeti and Manish Singh},
  title = {Efficient Twitter sentiment classification using subjective distant supervision},
  booktitle = {2017 9th International Conference on Communication Systems and Networks ({COMSNETS})}
}

@article{Kunc2020,
        doi = {10.1186/s12864-020-06821-6},
        url = {https://doi.org/10.1186/s12864-020-06821-6},
        year = {2020},
        month = dec,
        publisher = {Springer Science and Business Media {LLC}},
        volume = {21},
        number = {S5},
        author = {Vladim{\'{\i}}r Kunc and Ji{\v{r}}{\'{\i}} Kl{\'{e}}ma},
        title = {On tower and checkerboard neural network architectures for gene expression inference},
        journal = {{BMC} Genomics}
    }

@article{Kunc2021,
    doi = {10.1371/journal.pone.0243915},
    url = {https://doi.org/10.1371/journal.pone.0243915},
    year = {2021},
    month = jan,
    publisher = {Public Library of Science ({PLoS})},
    volume = {16},
    number = {1},
    pages = {e0243915},
    author = {Vladim{\'{\i}}r Kunc and Ji{\v{r}}{\'{\i}} Kl{\'{e}}ma},
    editor = {Holger Fr\"{o}hlich},
    title = {On transformative adaptive activation functions in neural networks for gene expression inference},
    journal = {{PLOS} {ONE}}
}

@InProceedings{Glorot2010,
  title = 	 {Understanding the difficulty of training deep feedforward neural networks},
  author = 	 {Xavier Glorot and Yoshua Bengio},
  booktitle = 	 {Proceedings of the Thirteenth International Conference on Artificial Intelligence and Statistics},
  pages = 	 {249--256},
  year = 	 {2010},
  editor = 	 {Yee Whye Teh and Mike Titterington},
  volume = 	 {9},
  series = 	 {Proceedings of Machine Learning Research},
  address = 	 {Chia Laguna Resort, Sardinia, Italy},
  month = 	 {13--15 May},
  publisher = 	 {PMLR},
  url = 	 {http://proceedings.mlr.press/v9/glorot10a.html}
}

@article{Flennerhag2018,
  author    = {Sebastian Flennerhag and
               others},
  title     = {Breaking the Activation Function Bottleneck through Adaptive Parameterization},
  journal   = {CoRR},
  volume    = {abs/1805.08574},
  year      = {2018},
  url       = {http://arxiv.org/abs/1805.08574},
  archivePrefix = {arXiv},
  eprint    = {1805.08574},
  bibsource = {dblp computer science bibliography, https://dblp.org}
}

@article{Vecci1998,
  author    = {Lorenzo Vecci and
              others},
  title     = {Learning and Approximation Capabilities of Adaptive Spline Activation Function Neural Networks},
  journal   = {Neural Networks},
  volume    = {11},
  number    = {2},
  pages     = {259--270},
  year      = {1998},
  url       = {https://doi.org/10.1016/S0893-6080(97)00118-4},
  doi       = {10.1016/S0893-6080(97)00118-4},
  bibsource = {dblp computer science bibliography, https://dblp.org}
}

@inproceedings{Maas2013,
  title={Rectifier nonlinearities improve neural network acoustic models},
  author={Maas, Andrew L. and Hannun, Awni Y. and Ng, Andrew Y.},
  year = {2013},
  booktitle = {Proceedings of the International Conference on Machine Learning ({ICML})},
  url = {https://ai.stanford.edu/~amaas/papers/relu_hybrid_icml2013_final.pdf}, 
  
}

@misc{Agostinelli2014,
  doi = {10.48550/ARXIV.1412.6830},
  url = {https://arxiv.org/abs/1412.6830},
  author = {Agostinelli,  Forest and Hoffman,  Matthew and Sadowski,  Peter and Baldi,  Pierre},
  title = {Learning Activation Functions to Improve Deep Neural Networks},
  publisher = {arXiv},
  year = {2014},
  copyright = {arXiv.org perpetual,  non-exclusive license}
}

@article{Lin2013,
  author    = {Min Lin and
               others},
  title     = {Network In Network},
  journal   = {CoRR},
  volume    = {abs/1312.4400},
  year      = {2013},
  url       = {http://arxiv.org/abs/1312.4400},
  archivePrefix = {arXiv},
  eprint    = {1312.4400},
  bibsource = {dblp computer science bibliography, https://dblp.org}
}

@article{Chen1996,
  doi = {10.1016/0893-6080(96)00006-8},
  url = {https://doi.org/10.1016/0893-6080(96)00006-8},
  year  = {1996},
  month = {6},
  publisher = {Elsevier {BV}},
  volume = {9},
  number = {4},
  pages = {627--641},
  author = {Chyi-Tsong Chen and Wei-Der Chang},
  title = {A feedforward neural network with function shape autotuning},
  journal = {Neural Networks}
}

@article{Trentin2001,
  doi = {10.1016/s0893-6080(01)00028-4},
  url = {https://doi.org/10.1016/s0893-6080(01)00028-4},
  year  = {2001},
  month = {5},
  publisher = {Elsevier {BV}},
  volume = {14},
  number = {4-5},
  pages = {471--493},
  author = {Edmondo Trentin},
  title = {Networks with trainable amplitude of activation functions},
  journal = {Neural Networks}
}

@article{Goh2003,
  doi = {10.1016/s0893-6080(03)00139-4},
  url = {https://doi.org/10.1016/s0893-6080(03)00139-4},
  year  = {2003},
  month = {10},
  publisher = {Elsevier {BV}},
  volume = {16},
  number = {8},
  pages = {1095--1100},
  author = {Su Lee Goh and Danilo P. Mandic},
  title = {Recurrent neural networks with trainable amplitude of activation functions},
  journal = {Neural Networks}
}

@article{Castelli2014,
  doi = {10.1016/j.patrec.2013.06.013},
  url = {https://doi.org/10.1016/j.patrec.2013.06.013},
  year  = {2014},
  month = {2},
  publisher = {Elsevier {BV}},
  volume = {37},
  pages = {178--191},
  author = {Ilaria Castelli and Edmondo Trentin},
  title = {Combination of supervised and unsupervised learning for training the activation functions of neural networks},
  journal = {Pattern Recognition Letters}
}

@incollection{Castelli2012,
  doi = {10.1007/978-3-642-28258-4_7},
  url = {https://doi.org/10.1007/978-3-642-28258-4_7},
  year  = {2012},
  publisher = {Springer Berlin Heidelberg},
  pages = {62--71},
  author = {Ilaria Castelli and Edmondo Trentin},
  title = {Semi-unsupervised Weighted Maximum-Likelihood Estimation of Joint Densities for the Co-training of Adaptive Activation Functions},
  booktitle = {Lecture Notes in Computer Science}
}

@incollection{Castelli2012b,
  doi = {10.1007/978-3-642-28258-4_6},
  url = {https://doi.org/10.1007/978-3-642-28258-4_6},
  year  = {2012},
  publisher = {Springer Berlin Heidelberg},
  pages = {52--61},
  author = {Ilaria Castelli and Edmondo Trentin},
  title = {Supervised and Unsupervised Co-training of Adaptive Activation Functions in Neural Nets},
  booktitle = {Lecture Notes in Computer Science}
}

@article{Yamada1992,
  doi = {10.1109/72.143373},
  url = {https://doi.org/10.1109/72.143373},
  year  = {1992},
  month = {7},
  publisher = {Institute of Electrical and Electronics Engineers ({IEEE})},
  volume = {3},
  number = {4},
  pages = {595--601},
  author = {T. Yamada and T. Yabuta},
  title = {Neural network controller using autotuning method for nonlinear functions},
  journal = {{IEEE} Transactions on Neural Networks}
}

@inproceedings{Yamada1992Remarks,
  title = {Remarks on a neural network controller which uses an auto-tuning method for nonlinear functions},
  url = {http://dx.doi.org/10.1109/IJCNN.1992.226893},
  DOI = {10.1109/ijcnn.1992.226893},
  booktitle = {[Proceedings 1992] IJCNN International Joint Conference on Neural Networks},
  publisher = {IEEE},
  author = {Yamada,  T. and Yabuta,  T.},
  year = {1992},
}

@incollection{Scardapane2018,
  doi = {10.1007/978-3-319-95098-3_7},
  url = {https://doi.org/10.1007/978-3-319-95098-3_7},
  year  = {2018},
  month = {7},
  publisher = {Springer International Publishing},
  pages = {73--83},
  author = {Simone Scardapane and others},
  title = {Learning Activation Functions from Data Using Cubic Spline Interpolation},
  booktitle = {Neural Advances in Processing Nonlinear Dynamic Signals}
}

@article{Guarnieri1999,
  doi = {10.1109/72.761726},
  url = {https://doi.org/10.1109/72.761726},
  year  = {1999},
  month = {5},
  publisher = {Institute of Electrical and Electronics Engineers ({IEEE})},
  volume = {10},
  number = {3},
  pages = {672--683},
  author = {S. Guarnieri and others},
  title = {Multilayer feedforward networks with adaptive spline activation function},
  journal = {{IEEE} Transactions on Neural Networks}
}

@inproceedings{Solazzi2000,
  doi = {10.1109/ijcnn.2000.861352},
  url = {https://doi.org/10.1109/ijcnn.2000.861352},
  year  = {2000},
  publisher = {{IEEE}},
  author = {M. Solazzi and A. Uncini},
  title = {Artificial neural networks with adaptive multidimensional spline activation functions},
  booktitle = {Proceedings of the {IEEE}-{INNS}-{ENNS} International Joint Conference on Neural Networks. {IJCNN} 2000. Neural Computing: New Challenges and Perspectives for the New Millennium}
}

@inproceedings{Piazza1992,
    author = {F. Piazza and others},
    title = {Artificial Neural Networks With Adaptive Polynomial Activation Function},
    year = {1992},
    booktitle = {Proceedings of the International Joint Conference on Neural Networks.{IJCNN}.}
}

@InProceedings{Hou2017,
  title = 	 {{ConvNets with Smooth Adaptive Activation Functions for Regression}},
  author = 	 {Le Hou and others},
  booktitle = 	 {Proceedings of the 20th International Conference on Artificial Intelligence and Statistics},
  pages = 	 {430--439},
  year = 	 {2017},
  editor = 	 {Aarti Singh and Jerry Zhu},
  volume = 	 {54},
  series = 	 {Proceedings of Machine Learning Research},
  address = 	 {Fort Lauderdale, FL, USA},
  month = 	 {20--22 Apr},
  publisher = 	 {PMLR},
  url = 	 {http://proceedings.mlr.press/v54/hou17a.html}
}

@article{Dushkoff2016,
  doi = {10.2352/issn.2470-1173.2016.19.coimg-149},
  url = {https://doi.org/10.2352/issn.2470-1173.2016.19.coimg-149},
  year  = {2016},
  month = {2},
  publisher = {Society for Imaging Science {\&} Technology},
  volume = {2016},
  number = {19},
  pages = {1--5},
  author = {Michael Dushkoff and Raymond Ptucha},
  title = {Adaptive Activation Functions for Deep Networks},
  journal = {Electronic Imaging}
}

@article{Qian2018,
  doi = {10.1016/j.neucom.2017.06.070},
  url = {https://doi.org/10.1016/j.neucom.2017.06.070},
  year  = {2018},
  month = {1},
  publisher = {Elsevier {BV}},
  volume = {272},
  pages = {204--212},
  author = {Sheng Qian and others},
  title = {Adaptive activation functions in convolutional neural networks},
  journal = {Neurocomputing}
}

@inproceedings{ShuxiangXu2000,
  doi = {10.1109/ijcnn.2000.861351},
  url = {https://doi.org/10.1109/ijcnn.2000.861351},
  year  = {2000},
  publisher = {{IEEE}},
  author = {Shuxiang Xu and  Ming Zhang},
  title = {Justification of a neuron-adaptive activation function},
  booktitle = {Proceedings of the {IEEE}-{INNS}-{ENNS} International Joint Conference on Neural Networks. {IJCNN} 2000. Neural Computing: New Challenges and Perspectives for the New Millennium}
}

@inproceedings{ShuxiangXu2005,
  doi = {10.1109/icita.2005.109},
  url = {https://doi.org/10.1109/icita.2005.109},
  publisher = {{IEEE}},
  year = {2005},
  author = {Shuxiang Xu and  Ming Zhang},
  title = {Data Mining {\textemdash} An Adaptive Neural Network Model for Financial Analysis},
  booktitle = {Third International Conference on Information Technology and Applications ({ICITA}{\textquotesingle}05)}
}

@incollection{Xu2007,
  doi = {10.1007/978-3-540-72383-7_147},
  url = {https://doi.org/10.1007/978-3-540-72383-7_147},
  year = {2007},
  publisher = {Springer Berlin Heidelberg},
  pages = {1265--1273},
  author = {Shuxiang Xu and Ming Zhang},
  title = {A New Adaptive Neural Network Model for Financial Data Mining},
  booktitle = {Advances in Neural Networks {\textendash} {ISNN} 2007}
}

@incollection{Xu2002,
  doi = {10.1007/3-540-36187-1_31},
  url = {https://doi.org/10.1007/3-540-36187-1_31},
  year = {2002},
  publisher = {Springer Berlin Heidelberg},
  pages = {356--362},
  author = {Shuxiang Xu and Ming Zhang},
  title = {An Adaptive Activation Function for Higher Order Neural Networks},
  booktitle = {Lecture Notes in Computer Science}
}

@article{Ismail2013,
  doi = {10.1016/j.engappai.2012.12.011},
  url = {https://doi.org/10.1016/j.engappai.2012.12.011},
  year  = {2013},
  month = {5},
  publisher = {Elsevier {BV}},
  volume = {26},
  number = {5-6},
  pages = {1540--1549},
  author = {A. Ismail and others},
  title = {Predictions of bridge scour: Application of a feed-forward neural network with an adaptive activation function},
  journal = {Engineering Applications of Artificial Intelligence}
}

@article{Bai2009,
  doi = {10.1016/j.chaos.2007.07.033},
  url = {https://doi.org/10.1016/j.chaos.2007.07.033},
  year  = {2009},
  month = {4},
  publisher = {Elsevier {BV}},
  volume = {40},
  number = {1},
  pages = {69--77},
  author = {Yanping Bai and others},
  title = {The performance of the backpropagation algorithm with varying slope of the activation function},
  journal = {Chaos,  Solitons {\&} Fractals}
}

@inproceedings{ChienChengYu2002,
  doi = {10.1109/tencon.2002.1181357},
  url = {https://doi.org/10.1109/tencon.2002.1181357},
  publisher = {{IEEE}},
  year = {2002},
  author = {Chien-Cheng Yu and  others},
  title = {An adaptive activation function for multilayer feedforward neural networks},
  booktitle = {2002 {IEEE} Region 10 Conference on Computers,  Communications,  Control and Power Engineering. {TENCOM} {\textquotesingle}02. Proceedings.}
}

@article{Simonyan2014,
  author    = {Karen Simonyan and
               Andrew Zisserman},
  title     = {Very Deep Convolutional Networks for Large-Scale Image Recognition},
  journal   = {CoRR},
  volume    = {abs/1409.1556},
  year      = {2014},
  url       = {http://arxiv.org/abs/1409.1556},
  bibsource = {dblp computer science bibliography, http://dblp.org}
}

@mastersthesis{Krizhevsky2009,
  author       = {Alex Krizhevsky}, 
  title        = {Learning multiple layers of features from tiny images},
  school       = {Department of Computer Science, University of Toronto},
  year         = 2009,
  %address      = {The address of the publisher},
  %month        = 8,
  %note         = {An optional note}
}

@online{DogsvsCats,
    author    = "Kaggle",
    title     = "Dogs vs. Cats",
    url       = "https://www.kaggle.com/c/dogs-vs-cats/data",
    year = 2013,
    urldate = {2024-01-12},
}

@inproceedings{Elson2007,
author = {Jeremy Elson and John R. Douceur and Jon Howell and Jared Saul},
title = {Asirra: A CAPTCHA that Exploits Interest-Aligned Manual Image Categorization},
  series = {CCS07},
  url = {http://dx.doi.org/10.1145/1315245.1315291},
  DOI = {10.1145/1315245.1315291},
  booktitle = {Proceedings of the 14th ACM conference on Computer and communications security},
  publisher = {ACM},
  year = {2007},
  month = oct,
  collection = {CCS07}
}

@article{Subramanian2017,
  doi = {10.1016/j.cell.2017.10.049},
  url = {https://doi.org/10.1016/j.cell.2017.10.049},
  year = {2017},
  month = nov,
  publisher = {Elsevier {BV}},
  volume = {171},
  number = {6},
  pages = {1437--1452.e17},
  author = {Aravind Subramanian and Rajiv Narayan and Steven M. Corsello and David D. Peck and Ted E. Natoli and Xiaodong Lu and Joshua Gould and John F. Davis and Andrew A. Tubelli and Jacob K. Asiedu and David L. Lahr and Jodi E. Hirschman and Zihan Liu and Melanie Donahue and Bina Julian and Mariya Khan and David Wadden and Ian C. Smith and Daniel Lam and Arthur Liberzon and Courtney Toder and Mukta Bagul and Marek Orzechowski and Oana M. Enache and Federica Piccioni and Sarah A. Johnson and Nicholas J. Lyons and Alice H. Berger and Alykhan F. Shamji and Angela N. Brooks and Anita Vrcic and Corey Flynn and Jacqueline Rosains and David Y. Takeda and Roger Hu and Desiree Davison and Justin Lamb and Kristin Ardlie and Larson Hogstrom and Peyton Greenside and Nathanael S. Gray and Paul A. Clemons and Serena Silver and Xiaoyun Wu and Wen-Ning Zhao and Willis Read-Button and Xiaohua Wu and Stephen J. Haggarty and Lucienne V. Ronco and Jesse S. Boehm and Stuart L. Schreiber and John G. Doench and Joshua A. Bittker and David E. Root and Bang Wong and Todd R. Golub},
  title = {A Next Generation Connectivity Map: L1000 Platform and the First 1, 000, 000 Profiles},
  journal = {Cell}
}

@inproceedings{Szegedy2015,
  doi = {10.1109/cvpr.2015.7298594},
  url = {https://doi.org/10.1109/cvpr.2015.7298594},
  year  = {2015},
  month = {jun},
  publisher = {{IEEE}},
  author = {Christian Szegedy and  Wei Liu and  Yangqing Jia and Pierre Sermanet and Scott Reed and Dragomir Anguelov and Dumitru Erhan and Vincent Vanhoucke and Andrew Rabinovich},
  title = {Going deeper with convolutions},
  booktitle = {2015 {IEEE} Conference on Computer Vision and Pattern Recognition ({CVPR})}
}

@inproceedings{He2016,
  doi = {10.1109/cvpr.2016.90},
  url = {https://doi.org/10.1109/cvpr.2016.90},
  year  = {2016},
  month = {6},
  publisher = {{IEEE}},
  author = {Kaiming He and Xiangyu Zhang and Shaoqing Ren and Jian Sun},
  title = {Deep Residual Learning for Image Recognition},
  booktitle = {2016 {IEEE} Conference on Computer Vision and Pattern Recognition ({CVPR})}
}

@incollection{He2016Identity,
  doi = {10.1007/978-3-319-46493-0_38},
  url = {https://doi.org/10.1007/978-3-319-46493-0_38},
  year = {2016},
  publisher = {Springer International Publishing},
  pages = {630--645},
  author = {Kaiming He and Xiangyu Zhang and Shaoqing Ren and Jian Sun},
  title = {Identity Mappings in Deep Residual Networks},
  booktitle = {Computer Vision {\textendash} {ECCV} 2016}
}

@inproceedings{Huang2017,
  doi = {10.1109/cvpr.2017.243},
  url = {https://doi.org/10.1109/cvpr.2017.243},
  year  = {2017},
  month = {7},
  publisher = {{IEEE}},
  author = {Gao Huang and Zhuang Liu and Laurens van der Maaten and Kilian Q. Weinberger},
  title = {Densely Connected Convolutional Networks},
  booktitle = {2017 {IEEE} Conference on Computer Vision and Pattern Recognition ({CVPR})}
}

@misc{Dua2017 ,
author = {Dua, Dheeru and Karra Taniskidou, Efi},
year = {2017},
title = {{UCI} Machine Learning Repository},
url = {http://archive.ics.uci.edu/ml},
institution = {University of California, Irvine, School of Information and Computer Sciences},
}

@article{Chen2019,
  doi = {10.1162/neco_a_01192},
  url = {https://doi.org/10.1162/neco_a_01192},
  year = {2019},
  month = jun,
  publisher = {{MIT} Press - Journals},
  volume = {31},
  number = {6},
  pages = {1215--1233},
  author = {Yunhua Chen and Yingchao Mai and Jinsheng Xiao and Ling Zhang},
  title = {Improving the Antinoise Ability of {DNNs} via a Bio-Inspired Noise Adaptive Activation Function Rand Softplus},
  journal = {Neural Computation}
}

@incollection{Liu2016,
  doi = {10.1007/978-3-319-46681-1_49},
  url = {https://doi.org/10.1007/978-3-319-46681-1_49},
  year = {2016},
  publisher = {Springer International Publishing},
  pages = {405--412},
  author = {Qian Liu and Steve Furber},
  title = {Noisy Softplus: A Biology Inspired Activation Function},
  booktitle = {Neural Information Processing}
}

@article{Jagtap2020,
  doi = {10.1016/j.jcp.2019.109136},
  url = {https://doi.org/10.1016/j.jcp.2019.109136},
  year = {2020},
  month = mar,
  publisher = {Elsevier {BV}},
  volume = {404},
  pages = {109136},
  author = {Ameya D. Jagtap and Kenji Kawaguchi and George Em Karniadakis},
  title = {Adaptive activation functions accelerate convergence in deep and physics-informed neural networks},
  journal = {Journal of Computational Physics}
}

@article{Eetemadi2018,
  doi = {10.1093/bioinformatics/bty945},
  url = {https://doi.org/10.1093/bioinformatics/bty945},
  year = {2018},
  month = nov,
  publisher = {Oxford University Press ({OUP})},
  author = {Ameen Eetemadi and Ilias Tagkopoulos},
  editor = {Bonnie Berger},
  title = {Genetic Neural Networks: an artificial neural network architecture for capturing gene expression relationships},
  journal = {Bioinformatics}
}

@inproceedings{Zagoruyko2016,
  doi = {10.5244/c.30.87},
  url = {https://doi.org/10.5244/c.30.87},
  year = {2016},
  publisher = {British Machine Vision Association},
  author = {Sergey Zagoruyko and Nikos Komodakis},
  title = {Wide Residual Networks},
  booktitle = {Procedings of the British Machine Vision Conference 2016}
}

@article{Jagtap2020Locally,
  doi = {10.1098/rspa.2020.0334},
  url = {https://doi.org/10.1098/rspa.2020.0334},
  year = {2020},
  month = jul,
  publisher = {The Royal Society},
  volume = {476},
  number = {2239},
  pages = {20200334},
  author = {Ameya D. Jagtap and Kenji Kawaguchi and George Em Karniadakis},
  title = {Locally adaptive activation functions with slope recovery for deep and physics-informed neural networks},
  journal = {Proceedings of the Royal Society A: Mathematical,  Physical and Engineering Sciences}
}

@article{Lecun1998,
	doi = {10.1109/5.726791},
	url = {https://doi.org/10.1109/5.726791},
	year = 1998,
	publisher = {Institute of Electrical and Electronics Engineers ({IEEE})},
	volume = {86},
	number = {11},
	pages = {2278--2324},
	author = {Y. Lecun and L. Bottou and Y. Bengio and P. Haffner},
	title = {Gradient-based learning applied to document recognition},
	journal = {Proceedings of the {IEEE}}
}

@inproceedings{Kingma2014,
  author       = {Diederik P. Kingma and
                  Max Welling},
  editor       = {Yoshua Bengio and
                  Yann LeCun},
  title        = {Auto-Encoding Variational Bayes},
  booktitle    = {2nd International Conference on Learning Representations, {ICLR} 2014,
                  Banff, AB, Canada, April 14-16, 2014, Conference Track Proceedings},
  year         = {2014},
  url          = {http://arxiv.org/abs/1312.6114},
  bibsource    = {dblp computer science bibliography, https://dblp.org}
}

@article{Russakovsky2015,
  doi = {10.1007/s11263-015-0816-y},
  url = {https://doi.org/10.1007/s11263-015-0816-y},
  year = {2015},
  month = apr,
  publisher = {Springer Science and Business Media {LLC}},
  volume = {115},
  number = {3},
  pages = {211--252},
  author = {Olga Russakovsky and Jia Deng and Hao Su and Jonathan Krause and Sanjeev Satheesh and Sean Ma and Zhiheng Huang and Andrej Karpathy and Aditya Khosla and Michael Bernstein and Alexander C. Berg and Li Fei-Fei},
  title = {{ImageNet} Large Scale Visual Recognition Challenge},
  journal = {International Journal of Computer Vision}
}

@inproceedings{Deng2009,
  doi = {10.1109/cvpr.2009.5206848},
  url = {https://doi.org/10.1109/cvpr.2009.5206848},
  year = {2009},
  month = jun,
  publisher = {{IEEE}},
  author = {Jia Deng and Wei Dong and Richard Socher and Li-Jia Li and  Kai Li and  Li Fei-Fei},
  title = {{ImageNet}: A large-scale hierarchical image database},
  booktitle = {2009 {IEEE} Conference on Computer Vision and Pattern Recognition}
}

@article{LiDeng2012,
  doi = {10.1109/msp.2012.2211477},
  url = {https://doi.org/10.1109/msp.2012.2211477},
  year = {2012},
  month = nov,
  publisher = {Institute of Electrical and Electronics Engineers ({IEEE})},
  volume = {29},
  number = {6},
  pages = {141--142},
  author = {Li Deng},
  title = {The {MNIST} Database of Handwritten Digit Images for Machine Learning Research [Best of the Web]},
  journal = {{IEEE} Signal Processing Magazine}
}

@misc{Xiao2017,
  doi = {10.48550/ARXIV.1708.07747},
  url = {https://arxiv.org/abs/1708.07747},
  author = {Xiao,  Han and Rasul,  Kashif and Vollgraf,  Roland},
  title = {Fashion-MNIST: a Novel Image Dataset for Benchmarking Machine Learning Algorithms},
  publisher = {arXiv},
  year = {2017},
  copyright = {arXiv.org perpetual,  non-exclusive license}
}

@inproceedings{Krizhevsky2012,
 author = {Krizhevsky, Alex and Sutskever, Ilya and Hinton, Geoffrey E},
 booktitle = {Advances in Neural Information Processing Systems},
 editor = {F. Pereira and C.J. Burges and L. Bottou and K.Q. Weinberger},
 pages = {},
 publisher = {Curran Associates, Inc.},
 title = {ImageNet Classification with Deep Convolutional Neural Networks},
 url = {https://proceedings.neurips.cc/paper_files/paper/2012/file/c399862d3b9d6b76c8436e924a68c45b-Paper.pdf},
 volume = {25},
 year = {2012}
}

@inproceedings{Zhang2018Efficient,
 author = {Zhang, Huan and Weng, Tsui-Wei and Chen, Pin-Yu and Hsieh, Cho-Jui and Daniel, Luca},
 booktitle = {Advances in Neural Information Processing Systems},
 editor = {S. Bengio and H. Wallach and H. Larochelle and K. Grauman and N. Cesa-Bianchi and R. Garnett},
 pages = {},
 publisher = {Curran Associates, Inc.},
 title = {Efficient Neural Network Robustness Certification with General Activation Functions},
 url = {https://proceedings.neurips.cc/paper_files/paper/2018/file/d04863f100d59b3eb688a11f95b0ae60-Paper.pdf},
 volume = {31},
 year = {2018}
}

@book{Goodfellow2016,
    title={Deep Learning},
    author={Ian Goodfellow and Yoshua Bengio and Aaron Courville},
    publisher={MIT Press},
    note={\url{http://www.deeplearningbook.org}},
    year={2016}
}

@incollection{Li2020SoftRootSign,
  doi = {10.1007/978-3-030-60636-7_26},
  url = {https://doi.org/10.1007/978-3-030-60636-7_26},
  year = {2020},
  publisher = {Springer International Publishing},
  pages = {310--319},
  author = {Dandan Li and Yuan Zhou},
  title = {Soft-Root-Sign: A New Bounded Neural Activation Function},
  booktitle = {Pattern Recognition and Computer Vision}
}

@inproceedings{Dinu2007,
  doi = {10.1109/isie.2007.4374572},
  url = {https://doi.org/10.1109/isie.2007.4374572},
  year = {2007},
  month = jun,
  publisher = {{IEEE}},
  author = {Andrei Dinu and Marcian Cirstea},
  title = {A Digital Neural Network {FPGA} Direct Hardware Implementation Algorithm},
  booktitle = {2007 {IEEE} International Symposium on Industrial Electronics}
}

@article{Dinu2010,
  doi = {10.1109/tie.2009.2033097},
  url = {https://doi.org/10.1109/tie.2009.2033097},
  year = {2010},
  month = may,
  publisher = {Institute of Electrical and Electronics Engineers ({IEEE})},
  volume = {57},
  number = {5},
  pages = {1845--1848},
  author = {Andrei Dinu and Marcian N Cirstea and Silvia E Cirstea},
  title = {Direct Neural-Network Hardware-Implementation Algorithm},
  journal = {{IEEE} Transactions on Industrial Electronics}
}

@article{Apicella2021,
  doi = {10.1016/j.neunet.2021.01.026},
  url = {https://doi.org/10.1016/j.neunet.2021.01.026},
  year = {2021},
  month = jun,
  publisher = {Elsevier {BV}},
  volume = {138},
  pages = {14--32},
  author = {Andrea Apicella and Francesco Donnarumma and Francesco Isgr{\`{o}} and Roberto Prevete},
  title = {A survey on modern trainable activation functions},
  journal = {Neural Networks}
}

@misc{Chen2023Saturated,
  doi = {10.48550/ARXIV.2305.07537},
  url = {https://arxiv.org/abs/2305.07537},
  author = {Chen,  Junjia and Pan,  Zhibin},
  title = {Saturated Non-Monotonic Activation Functions},
  publisher = {arXiv},
  year = {2023},
  copyright = {arXiv.org perpetual,  non-exclusive license}
}

@article{Rosenblatt1958,
  doi = {10.1037/h0042519},
  url = {https://doi.org/10.1037/h0042519},
  year = {1958},
  publisher = {American Psychological Association ({APA})},
  volume = {65},
  number = {6},
  pages = {386--408},
  author = {F. Rosenblatt},
  title = {The perceptron: A probabilistic model for information storage and organization in the brain},
  journal = {Psychological Review}
}

@article{Nader2021,
  doi = {10.1145/3464384},
  url = {https://doi.org/10.1145/3464384},
  year = {2021},
  month = jun,
  publisher = {Association for Computing Machinery ({ACM})},
  volume = {1},
  number = {2},
  pages = {1--36},
  author = {Andrew Nader and Danielle Azar},
  title = {Evolution of Activation Functions: An Empirical Investigation},
  journal = {{ACM} Transactions on Evolutionary Learning and Optimization}
}

@inproceedings{Saha2018,
  title = {A New Activation Function for Artificial Neural Net Based Habitability Classification},
  url = {http://dx.doi.org/10.1109/ICACCI.2018.8554460},
  DOI = {10.1109/icacci.2018.8554460},
  booktitle = {2018 International Conference on Advances in Computing,  Communications and Informatics (ICACCI)},
  publisher = {IEEE},
  author = {Saha,  Snehanshu and Mathur,  Archana and Bora,  Kakoli and Basak,  Suryoday and Agrawal,  Surbhi},
  year = {2018},
  month = sep 
}

@article{Saha2020,
  doi = {10.1140/epjst/e2020-000098-9},
  url = {https://doi.org/10.1140/epjst/e2020-000098-9},
  year = {2020},
  month = nov,
  publisher = {Springer Science and Business Media {LLC}},
  volume = {229},
  number = {16},
  pages = {2629--2738},
  author = {Snehanshu Saha and Nithin Nagaraj and Archana Mathur and Rahul Yedida and Sneha H R},
  title = {Evolution of novel activation functions in neural network training for astronomy data: habitability classification of exoplanets},
  journal = {The European Physical Journal Special Topics}
}

@misc{Noel2021,
  doi = {10.48550/ARXIV.2111.04020},
  url = {https://arxiv.org/abs/2111.04020},
  author = {Noel,  Matthew Mithra and Bharadwaj,  Shubham and Muthiah-Nakarajan,  Venkataraman and Dutta,  Praneet and Amali,  Geraldine Bessie},
  title = {Biologically Inspired Oscillating Activation Functions Can Bridge the Performance Gap between Biological and Artificial Neurons},
  publisher = {arXiv},
  year = {2021},
  copyright = {arXiv.org perpetual,  non-exclusive license}
}

@InProceedings{Glorot2011,
  title = 	 {Deep Sparse Rectifier Neural Networks},
  author = 	 {Xavier Glorot and Antoine Bordes and Yoshua Bengio},
  booktitle = 	 {Proceedings of the Fourteenth International Conference on Artificial Intelligence and Statistics},
  pages = 	 {315--323},
  year = 	 {2011},
  editor = 	 {Geoffrey Gordon and David Dunson and Miroslav Dudík},
  volume = 	 {15},
  series = 	 {Proceedings of Machine Learning Research},
  address = 	 {Fort Lauderdale, FL, USA},
  month = 	 {4},
  publisher = 	 {PMLR},
  url = 	 {http://proceedings.mlr.press/v15/glorot11a.html},
}

@online{CS231n,
    author    = "Andrej Karpathy ",
    title     = "CS231n: Convolutional Neural Networks for Visual Recognition --- Module 2: Convolutional Neural Networks",
    url       = "https://cs231n.github.io/",
    year = 2023,
    urldate = {2023-10-11}
}

@inproceedings{Mishkin2015,
  author       = {Dmytro Mishkin and
                  Jiri Matas},
  editor       = {Yoshua Bengio and
                  Yann LeCun},
  title        = {All you need is a good init},
  booktitle    = {4th International Conference on Learning Representations, {ICLR} 2016,
                  San Juan, Puerto Rico, May 2-4, 2016, Conference Track Proceedings},
  year         = {2016},
  url          = {http://arxiv.org/abs/1511.06422},
  bibsource    = {dblp computer science bibliography, https://dblp.org}
}

@article{Mishkin2017,
  doi = {10.1016/j.cviu.2017.05.007},
  url = {https://doi.org/10.1016/j.cviu.2017.05.007},
  year = {2017},
  month = aug,
  publisher = {Elsevier {BV}},
  volume = {161},
  pages = {11--19},
  author = {Dmytro Mishkin and Nikolay Sergievskiy and Jiri Matas},
  title = {Systematic evaluation of convolution neural network advances on the Imagenet},
  journal = {Computer Vision and Image Understanding}
}

@inproceedings{He2015,
  doi = {10.1109/iccv.2015.123},
  url = {https://doi.org/10.1109/iccv.2015.123},
  year = {2015},
  month = dec,
  publisher = {{IEEE}},
  author = {Kaiming He and Xiangyu Zhang and Shaoqing Ren and Jian Sun},
  title = {Delving Deep into Rectifiers: Surpassing Human-Level Performance on {ImageNet} Classification},
  booktitle = {2015 {IEEE} International Conference on Computer Vision ({ICCV})}
}

@misc{Xu2015,
  doi = {10.48550/ARXIV.1505.00853},
  url = {https://arxiv.org/abs/1505.00853},
  author = {Xu,  Bing and Wang,  Naiyan and Chen,  Tianqi and Li,  Mu},
  title = {Empirical Evaluation of Rectified Activations in Convolutional Network},
  publisher = {arXiv},
  year = {2015},
  copyright = {arXiv.org perpetual,  non-exclusive license}
}

@article{Jin2016,
  doi = {10.1609/aaai.v30i1.10287},
  url = {https://doi.org/10.1609/aaai.v30i1.10287},
  year = {2016},
  month = feb,
  publisher = {Association for the Advancement of Artificial Intelligence ({AAAI})},
  volume = {30},
  number = {1},
  author = {Xiaojie Jin and Chunyan Xu and Jiashi Feng and Yunchao Wei and Junjun Xiong and Shuicheng Yan},
  title = {Deep Learning with S-Shaped Rectified Linear Activation Units},
  journal = {Proceedings of the {AAAI} Conference on Artificial Intelligence}
}

@InProceedings{Goodfellow2013,
  title = 	 {Maxout Networks},
  author = 	 {Goodfellow, Ian and Warde-Farley, David and Mirza, Mehdi and Courville, Aaron and Bengio, Yoshua},
  booktitle = 	 {Proceedings of the 30th International Conference on Machine Learning},
  pages = 	 {1319--1327},
  year = 	 {2013},
  editor = 	 {Dasgupta, Sanjoy and McAllester, David},
  volume = 	 {28},
  number =       {3},
  series = 	 {Proceedings of Machine Learning Research},
  address = 	 {Atlanta, Georgia, USA},
  month = 	 {17--19 Jun},
  publisher =    {PMLR},
  url = 	 {https://proceedings.mlr.press/v28/goodfellow13.html}
}

@article{Castaneda2019,
  doi = {10.1186/s40537-019-0233-0},
  url = {https://doi.org/10.1186/s40537-019-0233-0},
  year = {2019},
  month = aug,
  publisher = {Springer Science and Business Media {LLC}},
  volume = {6},
  number = {1},
  author = {Gabriel Castaneda and Paul Morris and Taghi M. Khoshgoftaar},
  title = {Evaluation of maxout activations in deep learning across several big data domains},
  journal = {Journal of Big Data}
}

@article{Hanif2020,
  doi = {10.1016/j.icte.2019.06.001},
  url = {https://doi.org/10.1016/j.icte.2019.06.001},
  year = {2020},
  month = mar,
  publisher = {Elsevier {BV}},
  volume = {6},
  number = {1},
  pages = {28--37},
  author = {Muhammad Shehzad Hanif and Muhammad Bilal},
  title = {Competitive residual neural network for image classification},
  journal = {{ICT} Express}
}

@misc{Graham2014,
  doi = {10.48550/ARXIV.1409.6070},
  url = {https://arxiv.org/abs/1409.6070},
  author = {Graham,  Benjamin},
  title = {Spatially-sparse convolutional neural networks},
  publisher = {arXiv},
  year = {2014},
  copyright = {arXiv.org perpetual,  non-exclusive license}
}

@misc{Chang2015,
  doi = {10.48550/ARXIV.1511.02583},
  url = {https://arxiv.org/abs/1511.02583},
  author = {Chang,  Jia-Ren and Chen,  Yong-Sheng},
  title = {Batch-normalized Maxout Network in Network},
  publisher = {arXiv},
  year = {2015},
  copyright = {arXiv.org perpetual,  non-exclusive license}
}

@article{Lin2018,
  doi = {10.1016/j.procs.2018.04.239},
  url = {https://doi.org/10.1016/j.procs.2018.04.239},
  year = {2018},
  publisher = {Elsevier {BV}},
  volume = {131},
  pages = {977--984},
  author = {Guifang Lin and Wei Shen},
  title = {Research on convolutional neural network based on improved Relu piecewise activation function},
  journal = {Procedia Computer Science}
}

@article{Zhang2018Multiple,
  doi = {10.1016/j.jocs.2018.07.003},
  url = {https://doi.org/10.1016/j.jocs.2018.07.003},
  year = {2018},
  month = sep,
  publisher = {Elsevier {BV}},
  volume = {28},
  pages = {1--10},
  author = {Yu-Dong Zhang and Chichun Pan and Junding Sun and Chaosheng Tang},
  title = {Multiple sclerosis identification by convolutional neural network with dropout and parametric {ReLU}},
  journal = {Journal of Computational Science}
}

@article{Jiang2018,
  doi = {10.1016/j.neucom.2017.09.056},
  url = {https://doi.org/10.1016/j.neucom.2017.09.056},
  year = {2018},
  month = jan,
  publisher = {Elsevier {BV}},
  volume = {275},
  pages = {1132--1139},
  author = {Xiaoheng Jiang and Yanwei Pang and Xuelong Li and Jing Pan and Yinghong Xie},
  title = {Deep neural networks with Elastic Rectified Linear Units for object recognition},
  journal = {Neurocomputing}
}

@inproceedings{Clevert2015,
  author       = {Djork{-}Arn{\'{e}} Clevert and
                  Thomas Unterthiner and
                  Sepp Hochreiter},
  editor       = {Yoshua Bengio and
                  Yann LeCun},
  title        = {Fast and Accurate Deep Network Learning by Exponential Linear Units
                  (ELUs)},
  booktitle    = {4th International Conference on Learning Representations, {ICLR} 2016,
                  San Juan, Puerto Rico, May 2-4, 2016, Conference Track Proceedings},
  year         = {2016},
  url          = {http://arxiv.org/abs/1511.07289},
  bibsource    = {dblp computer science bibliography, https://dblp.org}
}

@misc{Trottier2016,
  doi = {10.48550/ARXIV.1605.09332},
  url = {https://arxiv.org/abs/1605.09332},
  author = {Trottier,  Ludovic and Giguère,  Philippe and Chaib-draa,  Brahim},
  title = {Parametric Exponential Linear Unit for Deep Convolutional Neural Networks},
  publisher = {arXiv},
  year = {2016},
  copyright = {arXiv.org perpetual,  non-exclusive license}
}

@article{Dubey2022,
  doi = {10.1016/j.neucom.2022.06.111},
  url = {https://doi.org/10.1016/j.neucom.2022.06.111},
  year = {2022},
  month = sep,
  publisher = {Elsevier {BV}},
  volume = {503},
  pages = {92--108},
  author = {Shiv Ram Dubey and Satish Kumar Singh and Bidyut Baran Chaudhuri},
  title = {Activation functions in deep learning: A comprehensive survey and benchmark},
  journal = {Neurocomputing}
}

@article{Pan2020,
  doi = {10.1007/s10845-020-01538-5},
  url = {https://doi.org/10.1007/s10845-020-01538-5},
  year = {2020},
  month = jan,
  publisher = {Springer Science and Business Media {LLC}},
  volume = {31},
  number = {8},
  pages = {1825--1836},
  author = {Yuhang Pan and Yonghao Wang and Ping Zhou and Ying Yan and Dongming Guo},
  title = {Activation functions selection for {BP} neural network model of ground surface roughness},
  journal = {Journal of Intelligent Manufacturing}
}

@article{Chandra2003,
  doi = {10.1023/b:nepl.0000011137.04221.96},
  url = {https://doi.org/10.1023/b:nepl.0000011137.04221.96},
  year = {2003},
  month = dec,
  publisher = {Springer Science and Business Media {LLC}},
  volume = {18},
  number = {3},
  pages = {205--215},
  author = {Pravin Chandra},
  title = {Sigmoidal Function Classes for Feedforward Artificial Neural Networks},
  journal = {Neural Processing Letters}
}

@article{Chandra2004,
  doi = {10.1016/j.neucom.2004.04.001},
  url = {https://doi.org/10.1016/j.neucom.2004.04.001},
  year = {2004},
  month = oct,
  publisher = {Elsevier {BV}},
  volume = {61},
  pages = {429--437},
  author = {Pravin Chandra and Yogesh Singh},
  title = {An activation function adapting training algorithm for sigmoidal feedforward networks},
  journal = {Neurocomputing}
}

@article{Chandra2004Case,
  doi = {10.1016/j.neucom.2003.08.005},
  url = {https://doi.org/10.1016/j.neucom.2003.08.005},
  year = {2004},
  month = jan,
  publisher = {Elsevier {BV}},
  volume = {56},
  pages = {447--454},
  author = {Pravin Chandra and Yogesh Singh},
  title = {A case for the self-adaptation of activation functions in {FFANNs}},
  journal = {Neurocomputing}
}

@article{SinghSodhi2014,
  doi = {10.1016/j.neucom.2014.06.007},
  url = {https://doi.org/10.1016/j.neucom.2014.06.007},
  year = {2014},
  month = nov,
  publisher = {Elsevier {BV}},
  volume = {143},
  pages = {182--196},
  author = {Sartaj Singh Sodhi and Pravin Chandra},
  title = {Bi-modal derivative activation function for sigmoidal feedforward networks},
  journal = {Neurocomputing}
}

@article{SamatinNjikam2016,
  doi = {10.1007/s10489-015-0744-0},
  url = {https://doi.org/10.1007/s10489-015-0744-0},
  year = {2016},
  month = jan,
  publisher = {Springer Science and Business Media {LLC}},
  volume = {45},
  number = {1},
  pages = {75--82},
  author = {Aboubakar Nasser Samatin Njikam and Huan Zhao},
  title = {A novel activation function for multilayer feed-forward neural networks},
  journal = {Applied Intelligence}
}

@misc{Xu2016,
  doi = {10.48550/ARXIV.1602.05980},
  url = {https://arxiv.org/abs/1602.05980},
  author = {Xu,  Bing and Huang,  Ruitong and Li,  Mu},
  title = {Revise Saturated Activation Functions},
  publisher = {arXiv},
  year = {2016},
  copyright = {arXiv.org perpetual,  non-exclusive license}
}

@inproceedings{Eger2018,
  doi = {10.18653/v1/d18-1472},
  url = {https://doi.org/10.18653/v1/d18-1472},
  year = {2018},
  publisher = {Association for Computational Linguistics},
  author = {Steffen Eger and Paul Youssef and Iryna Gurevych},
  title = {Is it Time to Swish? Comparing Deep Learning Activation Functions Across {NLP} tasks},
  booktitle = {Proceedings of the 2018 Conference on Empirical Methods in Natural Language Processing}
}

@misc{Ramachandran2017,
  doi = {10.48550/ARXIV.1710.05941},
  url = {https://arxiv.org/abs/1710.05941},
  author = {Ramachandran,  Prajit and Zoph,  Barret and Le,  Quoc V.},
  title = {Searching for Activation Functions},
  publisher = {arXiv},
  year = {2017},
  copyright = {arXiv.org perpetual,  non-exclusive license}
}

@inproceedings{Kong2017,
  doi = {10.1109/ijcnn.2017.7966168},
  url = {https://doi.org/10.1109/ijcnn.2017.7966168},
  year = {2017},
  month = may,
  publisher = {{IEEE}},
  author = {Shumin Kong and Masahiro Takatsuka},
  title = {Hexpo: A vanishing-proof activation function},
  booktitle = {2017 International Joint Conference on Neural Networks ({IJCNN})}
}

@article{Elfwing2018,
  doi = {10.1016/j.neunet.2017.12.012},
  url = {https://doi.org/10.1016/j.neunet.2017.12.012},
  year = {2018},
  month = nov,
  publisher = {Elsevier {BV}},
  volume = {107},
  pages = {3--11},
  author = {Stefan Elfwing and Eiji Uchibe and Kenji Doya},
  title = {Sigmoid-weighted linear units for neural network function approximation in reinforcement learning},
  journal = {Neural Networks}
}

@article{Qin2019,
  doi = {10.1109/tie.2018.2856205},
  url = {https://doi.org/10.1109/tie.2018.2856205},
  year = {2019},
  month = may,
  publisher = {Institute of Electrical and Electronics Engineers ({IEEE})},
  volume = {66},
  number = {5},
  pages = {3814--3824},
  author = {Yi Qin and Xin Wang and Jingqiang Zou},
  title = {The Optimized Deep Belief Networks With Improved Logistic Sigmoid Units and Their Application in Fault Diagnosis for Planetary Gearboxes of Wind Turbines},
  journal = {{IEEE} Transactions on Industrial Electronics}
}

@incollection{Roy2023,
  doi = {10.1007/978-3-031-31407-0_35},
  url = {https://doi.org/10.1007/978-3-031-31407-0_35},
  year = {2023},
  publisher = {Springer Nature Switzerland},
  pages = {462--476},
  author = {Swalpa Kumar Roy and Suvojit Manna and Shiv Ram Dubey and Bidyut Baran Chaudhuri},
  title = {{LiSHT}: Non-parametric Linearly Scaled Hyperbolic Tangent Activation Function for~Neural Networks},
  booktitle = {Communications in Computer and Information Science}
}

@article{Farzad2017,
  doi = {10.1007/s00521-017-3210-6},
  url = {https://doi.org/10.1007/s00521-017-3210-6},
  year = {2017},
  month = oct,
  publisher = {Springer Science and Business Media {LLC}},
  volume = {31},
  number = {7},
  pages = {2507--2521},
  author = {Amir Farzad and Hoda Mashayekhi and Hamid Hassanpour},
  title = {A comparative performance analysis of different activation functions in {LSTM} networks for classification},
  journal = {Neural Computing and Applications}
}

@techreport{Elliott1993,
 author = {David LeRoy Elliott},
 title = {A better Activation Function for Artificial Neural Networks},
 year = {1993},
 url = {https://www.researchgate.net/publication/277299531},
 publisher = {University of Maryland},
}

@article{Bingham2022,
  doi = {10.1016/j.neunet.2022.01.001},
  url = {https://doi.org/10.1016/j.neunet.2022.01.001},
  year = {2022},
  month = apr,
  publisher = {Elsevier {BV}},
  volume = {148},
  pages = {48--65},
  author = {Garrett Bingham and Risto Miikkulainen},
  title = {Discovering Parametric Activation Functions},
  journal = {Neural Networks}
}

@incollection{Balaji2020,
  doi = {10.1007/978-3-030-55180-3_43},
  url = {https://doi.org/10.1007/978-3-030-55180-3_43},
  year = {2020},
  month = aug,
  publisher = {Springer International Publishing},
  pages = {583--597},
  author = {Srinivasan Balaji and T. Kavya and Natasha Sebastian},
  title = {Learn-Able Parameter Guided Activation Functions},
  booktitle = {Advances in Intelligent Systems and Computing}
}

@misc{Zhou2023Lau,
title={{LAU}: A novel two-parameter learnable Logmoid Activation Unit},
author={Xue-Mei Zhou and Ling-Fang Li and Xing-Zhou Zheng and Mingxing Luo},
year={2023},
url={https://openreview.net/forum?id=uwBUzlm0GS}
}

@misc{Zhou2023TwoParameter,
title={A two-parameter learnable Logmoid Activation Unit},
author={Xue-Mei Zhou and Ling-Fang Li and Xing-Zhou Zheng and Mingxing Luo},
year={2023},
url={https://openreview.net/forum?id=LcXWYmA8Ek}
}

@inproceedings{Wuraola2018,
  doi = {10.1109/ijcnn.2018.8489043},
  url = {https://doi.org/10.1109/ijcnn.2018.8489043},
  year = {2018},
  month = jul,
  publisher = {{IEEE}},
  author = {Adedamola Wuraola and Nitish Patel},
  title = {{SQNL}: A New Computationally Efficient Activation Function},
  booktitle = {2018 International Joint Conference on Neural Networks ({IJCNN})}
}

@article{Wuraola2021,
  doi = {10.1016/j.neucom.2021.02.030},
  url = {https://doi.org/10.1016/j.neucom.2021.02.030},
  year = {2021},
  month = jun,
  publisher = {Elsevier {BV}},
  volume = {442},
  pages = {73--88},
  author = {Adedamola Wuraola and Nitish Patel and Sing Kiong Nguang},
  title = {Efficient activation functions for embedded inference engines},
  journal = {Neurocomputing}
}

@InProceedings{Shang2016,
  title = 	 {Understanding and Improving Convolutional Neural Networks via Concatenated Rectified Linear Units},
  author = 	 {Shang, Wenling and Sohn, Kihyuk and Almeida, Diogo and Lee, Honglak},
  booktitle = 	 {Proceedings of The 33rd International Conference on Machine Learning},
  pages = 	 {2217--2225},
  year = 	 {2016},
  editor = 	 {Balcan, Maria Florina and Weinberger, Kilian Q.},
  volume = 	 {48},
  series = 	 {Proceedings of Machine Learning Research},
  address = 	 {New York, New York, USA},
  month = 	 {20--22 Jun},
  publisher =    {PMLR},
  url = 	 {https://proceedings.mlr.press/v48/shang16.html}
}

@inproceedings{Gupta2017,
  doi = {10.1109/iccvw.2017.119},
  url = {https://doi.org/10.1109/iccvw.2017.119},
  year = {2017},
  month = oct,
  publisher = {{IEEE}},
  author = {Anubha Gupta and Rahul Duggal},
  title = {P-{TELU}: Parametric Tan Hyperbolic Linear Unit Activation for Deep Neural Networks},
  booktitle = {2017 {IEEE} International Conference on Computer Vision Workshops ({ICCVW})}
}

@inproceedings{Qiu2018FReLU,
  doi = {10.1109/icpr.2018.8546022},
  url = {https://doi.org/10.1109/icpr.2018.8546022},
  year = {2018},
  month = aug,
  publisher = {{IEEE}},
  author = {Suo Qiu and Xiangmin Xu and Bolun Cai},
  title = {{FReLU}: Flexible Rectified Linear Units for Improving Convolutional Neural Networks},
  booktitle = {2018 24th International Conference on Pattern Recognition ({ICPR})}
}

@inproceedings{Klambauer2017,
 author = {Klambauer, G\"{u}nter and Unterthiner, Thomas and Mayr, Andreas and Hochreiter, Sepp},
 booktitle = {Advances in Neural Information Processing Systems},
 editor = {I. Guyon and U. Von Luxburg and S. Bengio and H. Wallach and R. Fergus and S. Vishwanathan and R. Garnett},
 pages = {},
 publisher = {Curran Associates, Inc.},
 title = {Self-Normalizing Neural Networks},
 url = {https://proceedings.neurips.cc/paper_files/paper/2017/file/5d44ee6f2c3f71b73125876103c8f6c4-Paper.pdf},
 volume = {30},
 year = {2017}
}

@article{Pratama2020,
  doi = {10.1007/s10489-020-01885-z},
  url = {https://doi.org/10.1007/s10489-020-01885-z},
  year = {2020},
  month = oct,
  publisher = {Springer Science and Business Media {LLC}},
  volume = {51},
  number = {3},
  pages = {1784--1801},
  author = {Kevin Pratama and Dae-Ki Kang},
  title = {Trainable activation function with differentiable negative side and adaptable rectified point},
  journal = {Applied Intelligence}
}

@misc{
Chen2021Redefining,
title={Redefining The Self-Normalization Property},
author={Zhaodong Chen and Zhao WeiQin and Lei Deng and Guoqi Li and Yuan Xie},
year={2021},
url={https://openreview.net/forum?id=gfwfOskyzSx}
}

@article{Cao2018Randomly,
  doi = {10.1016/j.neucom.2017.09.031},
  url = {https://doi.org/10.1016/j.neucom.2017.09.031},
  year = {2018},
  month = jan,
  publisher = {Elsevier {BV}},
  volume = {275},
  pages = {859--868},
  author = {Jiale Cao and Yanwei Pang and Xuelong Li and Jingkun Liang},
  title = {Randomly translational activation inspired by the input distributions of {ReLU}},
  journal = {Neurocomputing}
}

@article{Dubey2021,
  doi = {10.1007/s11042-020-10269-x},
  url = {https://doi.org/10.1007/s11042-020-10269-x},
  year = {2021},
  month = jan,
  publisher = {Springer Science and Business Media {LLC}},
  volume = {80},
  number = {15},
  pages = {23181--23206},
  author = {Shiv Ram Dubey and Soumendu Chakraborty},
  title = {Average biased {ReLU} based {CNN} descriptor for improved face retrieval},
  journal = {Multimedia Tools and Applications}
}

@inproceedings{Si2018,
  doi = {10.1109/dcas.2018.8620116},
  url = {https://doi.org/10.1109/dcas.2018.8620116},
  year = {2018},
  month = nov,
  publisher = {{IEEE}},
  author = {Jiong Si and Sarah L. Harris and Evangelos Yfantis},
  title = {A Dynamic {ReLU} on Neural Network},
  booktitle = {2018 {IEEE} 13th Dallas Circuits and Systems Conference ({DCAS})}
}

@article{Hu2019,
  doi = {10.1109/access.2019.2959036},
  url = {https://doi.org/10.1109/access.2019.2959036},
  year = {2019},
  publisher = {Institute of Electrical and Electronics Engineers ({IEEE})},
  volume = {7},
  pages = {180409--180416},
  author = {Xiaobin Hu and Peifeng Niu and Jianmei Wang and Xinxin Zhang},
  title = {A Dynamic Rectified Linear Activation Units},
  journal = {{IEEE} Access}
}

@article{Godin2018,
  doi = {10.1016/j.patrec.2018.09.006},
  url = {https://doi.org/10.1016/j.patrec.2018.09.006},
  year = {2018},
  month = dec,
  publisher = {Elsevier {BV}},
  volume = {116},
  pages = {8--14},
  author = {Fr{\'{e}}deric Godin and Jonas Degrave and Joni Dambre and Wesley De Neve},
  title = {Dual Rectified Linear Units ({DReLUs}): A replacement for tanh activation functions in Quasi-Recurrent Neural Networks},
  journal = {Pattern Recognition Letters}
}

@article{Tang2018,
  doi = {10.1016/j.neucom.2017.07.061},
  url = {https://doi.org/10.1016/j.neucom.2017.07.061},
  year = {2018},
  month = jan,
  publisher = {Elsevier {BV}},
  volume = {273},
  pages = {37--46},
  author = {Zhimin Tang and Linkai Luo and Hong Peng and Shaohui Li},
  title = {A joint residual network with paired {ReLUs} activation for image super-resolution},
  journal = {Neurocomputing}
}

@inproceedings{Hu2018,
  doi = {10.1109/fskd.2018.8687140},
  url = {https://doi.org/10.1109/fskd.2018.8687140},
  year = {2018},
  month = jul,
  publisher = {{IEEE}},
  author = {He Hu},
  title = {{vReLU} Activation Functions for Artificial Neural Networks},
  booktitle = {2018 14th International Conference on Natural Computation,  Fuzzy Systems and Knowledge Discovery ({ICNC}-{FSKD})}
}

@article{VallsPrez2023,
  doi = {10.1016/j.engappai.2023.105863},
  url = {https://doi.org/10.1016/j.engappai.2023.105863},
  year = {2023},
  month = apr,
  publisher = {Elsevier {BV}},
  volume = {120},
  pages = {105863},
  author = {Iv{\'{a}}n Vall{\'{e}}s-P{\'{e}}rez and Emilio Soria-Olivas and Marcelino Mart{\'{\i}}nez-Sober and Antonio J. Serrano-L{\'{o}}pez and Joan Vila-Franc{\'{e}}s and Juan G{\'{o}}mez-Sanch{\'{\i}}s},
  title = {Empirical study of the modulus as activation function in computer vision applications},
  journal = {Engineering Applications of Artificial Intelligence}
}

@misc{Berngardt2023,
  doi = {10.48550/ARXIV.2304.11758},
  url = {https://arxiv.org/abs/2304.11758},
  author = {Berngardt,  Oleg I.},
  title = {Improving Classification Neural Networks by using Absolute activation function (MNIST/LeNET-5 example)},
  publisher = {arXiv},
  year = {2023},
  copyright = {Creative Commons Attribution 4.0 International}
}

@article{Macdo2019,
  doi = {10.1016/j.eswa.2019.01.066},
  url = {https://doi.org/10.1016/j.eswa.2019.01.066},
  year = {2019},
  month = jun,
  publisher = {Elsevier {BV}},
  volume = {124},
  pages = {271--281},
  author = {David Mac{\^{e}}do and Cleber Zanchettin and Adriano L.I. Oliveira and Teresa Ludermir},
  title = {Enhancing batch normalized convolutional networks using displaced rectifier linear units: A systematic comparative study},
  journal = {Expert Systems with Applications}
}

@inproceedings{Godfrey2019,
  doi = {10.1109/smc.2019.8913972},
  url = {https://doi.org/10.1109/smc.2019.8913972},
  year = {2019},
  month = oct,
  publisher = {{IEEE}},
  author = {Luke B. Godfrey},
  title = {An Evaluation of Parametric Activation Functions for Deep Learning},
  booktitle = {2019 {IEEE} International Conference on Systems,  Man and Cybernetics ({SMC})}
}

@inproceedings{Mahdavi2013,
 author = {Mahdavi, Mehrdad and Zhang, Lijun and Jin, Rong},
 booktitle = {Advances in Neural Information Processing Systems},
 editor = {C.J. Burges and L. Bottou and M. Welling and Z. Ghahramani and K.Q. Weinberger},
 pages = {},
 publisher = {Curran Associates, Inc.},
 title = {Mixed Optimization for Smooth Functions},
 url = {https://proceedings.neurips.cc/paper_files/paper/2013/file/f73b76ce8949fe29bf2a537cfa420e8f-Paper.pdf},
 volume = {26},
 year = {2013}
}

@inproceedings{Yang2013,
 author = {Yang, Tianbao},
 booktitle = {Advances in Neural Information Processing Systems},
 editor = {C.J. Burges and L. Bottou and M. Welling and Z. Ghahramani and K.Q. Weinberger},
 pages = {},
 publisher = {Curran Associates, Inc.},
 title = {Trading Computation for Communication: Distributed Stochastic Dual Coordinate Ascent},
 url = {https://proceedings.neurips.cc/paper_files/paper/2013/file/dc912a253d1e9ba40e2c597ed2376640-Paper.pdf},
 volume = {26},
 year = {2013}
}

@InProceedings{Zhang2013,
  title = 	 {O(logT) Projections for Stochastic Optimization of Smooth and Strongly Convex Functions},
  author = 	 {Zhang, Lijun and Yang, Tianbao and Jin, Rong and He, Xiaofei},
  booktitle = 	 {Proceedings of the 30th International Conference on Machine Learning},
  pages = 	 {1121--1129},
  year = 	 {2013},
  editor = 	 {Dasgupta, Sanjoy and McAllester, David},
  volume = 	 {28},
  number =       {3},
  series = 	 {Proceedings of Machine Learning Research},
  address = 	 {Atlanta, Georgia, USA},
  month = 	 {17--19 Jun},
  publisher =    {PMLR},
  url = 	 {https://proceedings.mlr.press/v28/zhang13e.html}
}

@article{Bergou2022,
        doi = {10.1287/ijoo.2022.0072},
        url = {https://doi.org/10.1287/ijoo.2022.0072},
        year = {2022},
        month = oct,
        publisher = {Institute for Operations Research and the Management Sciences ({INFORMS})},
        volume = {4},
        number = {4},
        pages = {403--425},
        author = {El Houcine Bergou and Youssef Diouane and Vladimír Kunc and Vyacheslav Kungurtsev and Cl{\'{e}}ment W. Royer},
        title = {A Subsampling Line-Search Method with Second-Order Results},
        journal = {{INFORMS} Journal on Optimization}
    }

@article{Nanni2022,
  doi = {10.3390/s22166129},
  url = {https://doi.org/10.3390/s22166129},
  year = {2022},
  month = aug,
  publisher = {{MDPI} {AG}},
  volume = {22},
  number = {16},
  pages = {6129},
  author = {Loris Nanni and Sheryl Brahnam and Michelangelo Paci and Stefano Ghidoni},
  title = {Comparison of Different Convolutional Neural Network Activation Functions and Methods for Building Ensembles for Small to Midsize Medical Data Sets},
  journal = {Sensors}
}

@article{Adem2022,
  doi = {10.1016/j.eswa.2022.117583},
  url = {https://doi.org/10.1016/j.eswa.2022.117583},
  year = {2022},
  month = oct,
  publisher = {Elsevier {BV}},
  volume = {203},
  pages = {117583},
  author = {Kemal Adem},
  title = {Impact of activation functions and number of layers on detection of exudates using circular Hough transform and convolutional neural networks},
  journal = {Expert Systems with Applications}
}

@inproceedings{Basirat2020,
  doi = {10.1109/wacv45572.2020.9093485},
  url = {https://doi.org/10.1109/wacv45572.2020.9093485},
  year = {2020},
  month = mar,
  publisher = {{IEEE}},
  author = {Mina Basirat and Peter M. Roth},
  title = {L*{ReLU}: Piece-wise Linear Activation Functions for Deep Fine-grained Visual Categorization},
  booktitle = {2020 {IEEE} Winter Conference on Applications of Computer Vision ({WACV})}
}

@InProceedings{Hayou2019,
  title = 	 {On the Impact of the Activation function on Deep Neural Networks Training},
  author =       {Hayou, Soufiane and Doucet, Arnaud and Rousseau, Judith},
  booktitle = 	 {Proceedings of the 36th International Conference on Machine Learning},
  pages = 	 {2672--2680},
  year = 	 {2019},
  editor = 	 {Chaudhuri, Kamalika and Salakhutdinov, Ruslan},
  volume = 	 {97},
  series = 	 {Proceedings of Machine Learning Research},
  month = 	 {09--15 Jun},
  publisher =    {PMLR},
  url = 	 {https://proceedings.mlr.press/v97/hayou19a.html}
}

@misc{Hayou2018,
  doi = {10.48550/ARXIV.1805.08266},
  url = {https://arxiv.org/abs/1805.08266},
  author = {Hayou,  Soufiane and Doucet,  Arnaud and Rousseau,  Judith},
  title = {On the Selection of Initialization and Activation Function for Deep Neural Networks},
  publisher = {arXiv},
  year = {2018},
  copyright = {arXiv.org perpetual,  non-exclusive license}
}

@article{Kiliarslan2021,
  doi = {10.1016/j.eswa.2021.114805},
  url = {https://doi.org/10.1016/j.eswa.2021.114805},
  year = {2021},
  month = jul,
  publisher = {Elsevier {BV}},
  volume = {174},
  pages = {114805},
  author = {Serhat Kili{\c{c}}arslan and Mete Celik},
  title = {{RSigELU}: A nonlinear activation function for deep neural networks},
  journal = {Expert Systems with Applications}
}

@inproceedings{Gu2019,
  doi = {10.1109/iccv.2019.00429},
  url = {https://doi.org/10.1109/iccv.2019.00429},
  year = {2019},
  month = oct,
  publisher = {{IEEE}},
  author = {Shuhang Gu and Wen Li and Luc Van Gool and Radu Timofte},
  title = {Fast Image Restoration With Multi-Bin Trainable Linear Units},
  booktitle = {2019 {IEEE}/{CVF} International Conference on Computer Vision ({ICCV})}
}

@inproceedings{Liu2019NaturalLogarithmRectified,
  doi = {10.1109/iccc47050.2019.9064398},
  url = {https://doi.org/10.1109/iccc47050.2019.9064398},
  year = {2019},
  month = dec,
  publisher = {{IEEE}},
  author = {Yang Liu and Jianpeng Zhang and Chao Gao and Jinghua Qu and Lixin Ji},
  title = {Natural-Logarithm-Rectified Activation Function in Convolutional Neural Networks},
  booktitle = {2019 {IEEE} 5th International Conference on Computer and Communications ({ICCC})}
}

@article{Wang2019ReLTanh,
  doi = {10.1016/j.neucom.2019.07.017},
  url = {https://doi.org/10.1016/j.neucom.2019.07.017},
  year = {2019},
  month = oct,
  publisher = {Elsevier {BV}},
  volume = {363},
  pages = {88--98},
  author = {Xin Wang and Yi Qin and Yi Wang and Sheng Xiang and Haizhou Chen},
  title = {{ReLTanh}: An activation function with vanishing gradient resistance for {SAE}-based {DNNs} and its application to rotating machinery fault diagnosis},
  journal = {Neurocomputing}
}

@misc{Nicolae2018,
  doi = {10.48550/ARXIV.1809.09534},
  url = {https://arxiv.org/abs/1809.09534},
  author = {Nicolae,  Andrei},
  title = {PLU: The Piecewise Linear Unit Activation Function},
  publisher = {arXiv},
  year = {2018},
  copyright = {arXiv.org perpetual,  non-exclusive license}
}

@article{Li2013,
  doi = {10.1007/s13042-013-0198-9},
  url = {https://doi.org/10.1007/s13042-013-0198-9},
  year = {2013},
  month = sep,
  publisher = {Springer Science and Business Media {LLC}},
  volume = {5},
  number = {1},
  pages = {73--83},
  author = {Jin-Cheng Li and Wing W. Y. Ng and Daniel S. Yeung and Patrick P. K. Chan},
  title = {Bi-firing deep neural networks},
  journal = {International Journal of Machine Learning and Cybernetics}
}

@article{Liew2016,
  doi = {10.1016/j.neucom.2016.08.037},
  url = {https://doi.org/10.1016/j.neucom.2016.08.037},
  year = {2016},
  month = dec,
  publisher = {Elsevier {BV}},
  volume = {216},
  pages = {718--734},
  author = {Shan Sung Liew and Mohamed Khalil-Hani and Rabia Bakhteri},
  title = {Bounded activation functions for enhanced training stability of deep neural networks on visual pattern recognition problems},
  journal = {Neurocomputing}
}

@article{Wang2015Effective,
  doi = {10.1007/s00778-015-0391-4},
  url = {https://doi.org/10.1007/s00778-015-0391-4},
  year = {2015},
  month = jul,
  publisher = {Springer Science and Business Media {LLC}},
  volume = {25},
  number = {1},
  pages = {79--101},
  author = {Wei Wang and Xiaoyan Yang and Beng Chin Ooi and Dongxiang Zhang and Yueting Zhuang},
  title = {Effective deep learning-based multi-modal retrieval},
  journal = {The {VLDB} Journal}
}

@article{Ying2019,
  doi = {10.1109/access.2019.2928442},
  url = {https://doi.org/10.1109/access.2019.2928442},
  year = {2019},
  publisher = {Institute of Electrical and Electronics Engineers ({IEEE})},
  volume = {7},
  pages = {101633--101640},
  author = {Yao Ying and Jianlin Su and Peng Shan and Ligang Miao and Xiaolian Wang and Silong Peng},
  title = {Rectified Exponential Units for Convolutional Neural Networks},
  journal = {{IEEE} Access}
}

@misc{Barron2017,
  doi = {10.48550/ARXIV.1704.07483},
  url = {https://arxiv.org/abs/1704.07483},
  author = {Barron,  Jonathan T.},
  title = {Continuously Differentiable Exponential Linear Units},
  publisher = {arXiv},
  year = {2017},
  copyright = {arXiv.org perpetual,  non-exclusive license}
}

@article{Li2018Improving,
  doi = {10.1016/j.neucom.2018.01.084},
  url = {https://doi.org/10.1016/j.neucom.2018.01.084},
  year = {2018},
  month = aug,
  publisher = {Elsevier {BV}},
  volume = {301},
  pages = {11--24},
  author = {Yang Li and Chunxiao Fan and Yong Li and Qiong Wu and Yue Ming},
  title = {Improving deep neural network with Multiple Parametric Exponential Linear Units},
  journal = {Neurocomputing}
}

@article{Pham2019,
  doi = {10.1016/j.compmedimag.2019.101647},
  url = {https://doi.org/10.1016/j.compmedimag.2019.101647},
  year = {2019},
  month = oct,
  publisher = {Elsevier {BV}},
  volume = {77},
  pages = {101647},
  author = {Chi-Hieu Pham and Carlos Tor-D{\'{\i}}ez and H{\'{e}}l{\`{e}}ne Meunier and Nathalie Bednarek and Ronan Fablet and Nicolas Passat and Fran{\c{c}}ois Rousseau},
  title = {Multiscale brain {MRI} super-resolution using deep 3D convolutional networks},
  journal = {Computerized Medical Imaging and Graphics}
}

@INPROCEEDINGS{Godfrey2015,
author = {Luke. B. Godfrey and Michael. S. Gashler},
booktitle = {2015 7th International Joint Conference on Knowledge Discovery, Knowledge Engineering and Knowledge Management (IC3K)},
title = {A continuum among logarithmic, linear, and exponential functions, and its potential to improve generalization in neural networks},
year = {2015},
volume = {},
isbn = {978-9-8975-8164-9},
pages = {481-486},
url = {https://ieeexplore.ieee.org/document/7526959/},
publisher = {IEEE Computer Society},
address = {Los Alamitos, CA, USA},
month = {nov}
}

@inproceedings{Grelsson2018,
  doi = {10.1109/icpr.2018.8545104},
  url = {https://doi.org/10.1109/icpr.2018.8545104},
  year = {2018},
  month = aug,
  publisher = {{IEEE}},
  author = {Bertil Grelsson and Michael Felsberg},
  title = {Improved Learning in Convolutional Neural Networks with Shifted Exponential Linear Units ({ShELUs})},
  booktitle = {2018 24th International Conference on Pattern Recognition ({ICPR})}
}

@article{Zhang2019Research,
  author = {Tao Zhang and Jian Yang and Wen-ai Song and Chao-feng Song},
title = {Research on Improved Activation Function TReLU},
year = {2019},
journal = {Journal of Chinese Computer Systems},
volume = {40},
number = {1},
eid = {58},
pages = {58-63},
url = {http://xwxt.sict.ac.cn/EN/Y2019/V40/I1/58},
}

@article{Qiumei2019,
  doi = {10.1109/access.2019.2948112},
  url = {https://doi.org/10.1109/access.2019.2948112},
  year = {2019},
  publisher = {Institute of Electrical and Electronics Engineers ({IEEE})},
  volume = {7},
  pages = {151359--151367},
  author = {Zheng Qiumei and Tan Dan and Wang Fenghua},
  title = {Improved Convolutional Neural Network Based on Fast Exponentially Linear Unit Activation Function},
  journal = {{IEEE} Access}
}

@article{Schraudolph1999,
  doi = {10.1162/089976699300016467},
  url = {https://doi.org/10.1162/089976699300016467},
  year = {1999},
  month = may,
  publisher = {{MIT} Press - Journals},
  volume = {11},
  number = {4},
  pages = {853--862},
  author = {Nicol N. Schraudolph},
  title = {A Fast,  Compact Approximation of the Exponential Function},
  journal = {Neural Computation}
}

@article{Kim2020,
  doi = {10.1016/j.neucom.2020.03.051},
  url = {https://doi.org/10.1016/j.neucom.2020.03.051},
  year = {2020},
  month = sep,
  publisher = {Elsevier {BV}},
  volume = {406},
  pages = {253--266},
  author = {Daeho Kim and Jinah Kim and Jaeil Kim},
  title = {Elastic exponential linear units for convolutional neural networks},
  journal = {Neurocomputing}
}

@article{Cheng2020,
  doi = {10.1016/j.neunet.2020.02.012},
  url = {https://doi.org/10.1016/j.neunet.2020.02.012},
  year = {2020},
  month = may,
  publisher = {Elsevier {BV}},
  volume = {125},
  pages = {281--289},
  author = {Qishang Cheng and HongLiang Li and Qingbo Wu and Lei Ma and King Ngi Ngan},
  title = {Parametric Deformable Exponential Linear Units for deep neural networks},
  journal = {Neural Networks}
}

@article{Yu2020,
  doi = {10.1109/access.2020.2987829},
  url = {https://doi.org/10.1109/access.2020.2987829},
  year = {2020},
  publisher = {Institute of Electrical and Electronics Engineers ({IEEE})},
  volume = {8},
  pages = {72727--72741},
  author = {Yongbin Yu and Kwabena Adu and Nyima Tashi and Patrick Anokye and Xiangxiang Wang and Mighty Abra Ayidzoe},
  title = {{RMAF}: Relu-Memristor-Like Activation Function for Deep Learning},
  journal = {{IEEE} Access}
}

@inproceedings{Basirat2019,
  doi = {10.3217/978-3-85125-663-5-41},
  url = {https://openlib.tugraz.at/download.php?id=5d09dba127371&location=medra},
  author = {{Basirat,  Mina } and {Jammer,  Alexandra } and {Roth,  Peter M. }},
  title = {The Quest for the Golden Activation Function},
  publisher = {Verlag der Technischen Universit\"{a}t Graz},
  booktitle = {Proceedings of ARW \& OAGM Workshop 2019},
  year = {2019}
}

@misc{Basirat2018,
  doi = {10.48550/ARXIV.1808.00783},
  url = {https://arxiv.org/abs/1808.00783},
  author = {Basirat,  Mina and Roth,  Peter M.},
  title = {The Quest for the Golden Activation Function},
  publisher = {arXiv},
  year = {2018},
  copyright = {arXiv.org perpetual,  non-exclusive license}
}

@misc{Courbariaux2016,
  doi = {10.48550/ARXIV.1602.02830},
  url = {https://arxiv.org/abs/1602.02830},
  author = {Courbariaux,  Matthieu and Hubara,  Itay and Soudry,  Daniel and El-Yaniv,  Ran and Bengio,  Yoshua},
  title = {Binarized Neural Networks: Training Deep Neural Networks with Weights and Activations Constrained to +1 or -1},
  publisher = {arXiv},
  year = {2016},
  copyright = {arXiv.org perpetual,  non-exclusive license}
}

@misc{Alcaide2018,
  doi = {10.48550/ARXIV.1801.07145},
  url = {https://arxiv.org/abs/1801.07145},
  author = {Alcaide,  Eric},
  title = {E-swish: Adjusting Activations to Different Network Depths},
  publisher = {arXiv},
  year = {2018},
  copyright = {arXiv.org perpetual,  non-exclusive license}
}

@article{Chieng2018,
  doi = {10.26555/ijain.v4i2.249},
  url = {https://doi.org/10.26555/ijain.v4i2.249},
  year = {2018},
  month = jul,
  publisher = {Universitas Ahmad Dahlan,  Kampus 3},
  volume = {4},
  number = {2},
  pages = {76},
  author = {Hock Hung Chieng and Noorhaniza Wahid and Ong Pauline and Sai Raj Kishore Perla},
  title = {Flatten-T Swish: a thresholded {ReLU}-Swish-like activation function for deep learning},
  journal = {International Journal of Advances in Intelligent Informatics}
}

@incollection{Badiger2022,
  doi = {10.1007/978-981-16-8862-1_59},
  url = {https://doi.org/10.1007/978-981-16-8862-1_59},
  year = {2022},
  publisher = {Springer Singapore},
  pages = {905--919},
  author = {Manjunatha Badiger and Jose Alex Mathew},
  title = {Retrospective Review of Activation Functions in Artificial Neural Networks},
  booktitle = {Proceedings of Third International Conference on Communication,  Computing and Electronics Systems}
}

@misc{Jagtap2022,
  doi = {10.48550/ARXIV.2209.02681},
  url = {https://arxiv.org/abs/2209.02681},
  author = {Jagtap,  Ameya D. and Karniadakis,  George Em},
  title = {How important are activation functions in regression and classification? A survey,  performance comparison,  and future directions},
  publisher = {arXiv},
  year = {2022},
  copyright = {Creative Commons Attribution 4.0 International}
}

@article{Jagtap2022Deep,
  doi = {10.1016/j.neucom.2021.10.036},
  url = {https://doi.org/10.1016/j.neucom.2021.10.036},
  year = {2022},
  month = jan,
  publisher = {Elsevier {BV}},
  volume = {468},
  pages = {165--180},
  author = {Ameya D. Jagtap and Yeonjong Shin and Kenji Kawaguchi and George Em Karniadakis},
  title = {Deep Kronecker neural networks: A general framework for neural networks with adaptive activation functions},
  journal = {Neurocomputing}
}

@article{Nayef2021,
  doi = {10.1007/s11042-021-11593-6},
  url = {https://doi.org/10.1007/s11042-021-11593-6},
  year = {2021},
  month = oct,
  publisher = {Springer Science and Business Media {LLC}},
  volume = {81},
  number = {2},
  pages = {2065--2094},
  author = {Bahera H. Nayef and Siti Norul Huda Sheikh Abdullah and Rossilawati Sulaiman and Zaid Abdi Alkareem Alyasseri},
  title = {Optimized leaky {ReLU} for handwritten Arabic character recognition using convolution neural networks},
  journal = {Multimedia Tools and Applications}
}

@misc{Patwardhan2018,
  doi = {10.48550/ARXIV.1805.08878},
  url = {https://arxiv.org/abs/1805.08878},
  author = {Patwardhan,  Narendra and Ingalhalikar,  Madhura and Walambe,  Rahee},
  title = {ARiA: Utilizing Richard's Curve for Controlling the Non-monotonicity of the Activation Function in Deep Neural Nets},
  publisher = {arXiv},
  year = {2018},
  copyright = {arXiv.org perpetual,  non-exclusive license}
}

@article{Richards1959,
  doi = {10.1093/jxb/10.2.290},
  url = {https://doi.org/10.1093/jxb/10.2.290},
  year = {1959},
  publisher = {Oxford University Press ({OUP})},
  volume = {10},
  number = {2},
  pages = {290--301},
  author = {Francis John Richards},
  title = {A Flexible Growth Function for Empirical Use},
  journal = {Journal of Experimental Botany}
}

@article{Erturul2018,
  doi = {10.1016/j.neunet.2018.01.007},
  url = {https://doi.org/10.1016/j.neunet.2018.01.007},
  year = {2018},
  month = mar,
  publisher = {Elsevier {BV}},
  volume = {99},
  pages = {148--157},
  author = {\"{O}mer Faruk Ertu{\u{g}}rul},
  title = {A novel type of activation function in artificial neural networks: Trained activation function},
  journal = {Neural Networks}
}

@inproceedings{Manessi2018,
  doi = {10.1109/icpr.2018.8545362},
  url = {https://doi.org/10.1109/icpr.2018.8545362},
  year = {2018},
  month = aug,
  publisher = {{IEEE}},
  author = {Franco Manessi and Alessandro Rozza},
  title = {Learning Combinations of Activation Functions},
  booktitle = {2018 24th International Conference on Pattern Recognition ({ICPR})}
}

@incollection{Stfeld2020,
  doi = {10.1007/978-3-030-52243-8_4},
  url = {https://doi.org/10.1007/978-3-030-52243-8_4},
  year = {2020},
  publisher = {Springer International Publishing},
  pages = {37--50},
  author = {Leon Ren{\'{e}} S\"{u}tfeld and Flemming Brieger and Holger Finger and Sonja F\"{u}llhase and Gordon Pipa},
  title = {Adaptive Blending Units: Trainable Activation Functions for Deep Neural Networks},
  booktitle = {Advances in Intelligent Systems and Computing}
}

@inproceedings{Wang2018LookUp,
  doi = {10.1109/wacv.2018.00139},
  url = {https://doi.org/10.1109/wacv.2018.00139},
  year = {2018},
  month = mar,
  publisher = {{IEEE}},
  author = {Min Wang and Baoyuan Liu and Hassan Foroosh},
  title = {Look-Up Table Unit Activation Function for Deep Convolutional Neural Networks},
  booktitle = {2018 {IEEE} Winter Conference on Applications of Computer Vision ({WACV})}
}

@inproceedings{Waseem2021,
  doi = {10.1109/tensymp52854.2021.9551000},
  url = {https://doi.org/10.1109/tensymp52854.2021.9551000},
  year = {2021},
  month = aug,
  publisher = {{IEEE}},
  author = {Shaik Mohammed Waseem and Alavala Venkata Suraj and Subir Kumar Roy},
  title = {Accelerating the Activation Function Selection for Hybrid Deep Neural Networks {\textendash} {FPGA} Implementation},
  booktitle = {2021 {IEEE} Region 10 Symposium ({TENSYMP})}
}

@inproceedings{Wang2020Wide,
  doi = {10.1109/wacv45572.2020.9093436},
  url = {https://doi.org/10.1109/wacv45572.2020.9093436},
  year = {2020},
  month = mar,
  publisher = {{IEEE}},
  author = {Min Wang and Baoyuan Liu and Hassan Foroosh},
  title = {Wide Hidden Expansion Layer for Deep Convolutional Neural Networks},
  booktitle = {2020 {IEEE} Winter Conference on Applications of Computer Vision ({WACV})}
}

@article{Maguolo2021,
  doi = {10.1016/j.eswa.2020.114048},
  url = {https://doi.org/10.1016/j.eswa.2020.114048},
  year = {2021},
  month = mar,
  publisher = {Elsevier {BV}},
  volume = {166},
  pages = {114048},
  author = {Gianluca Maguolo and Loris Nanni and Stefano Ghidoni},
  title = {Ensemble of convolutional neural networks trained with different activation functions},
  journal = {Expert Systems with Applications}
}

@article{Adu2021,
  doi = {10.1002/ima.22638},
  url = {https://doi.org/10.1002/ima.22638},
  year = {2021},
  month = aug,
  publisher = {Wiley},
  volume = {32},
  number = {1},
  pages = {123--143},
  author = {Kwabena Adu and Yongbin Yu and Jingye Cai and Isaac Asare and Jennifer Quahin},
  title = {The influence of the activation function in a capsule network for brain tumor type classification},
  journal = {International Journal of Imaging Systems and Technology}
}

@inproceedings{Haq2021,
	author = {Haq, Zeeshan Ali and Jaffery, Zainul Abdin},
	title = {Impact of activation functions and number of layers on the classification of fruits using CNN},
	year = {2021},
	booktitle = {Proceedings of the 2021 8th International Conference on Computing for Sustainable Global Development, INDIACom 2021},
	pages = {227 -- 231},
  publisher = {{IEEE}},
	doi = {10.1109/INDIACom51348.2021.00040},
	url = {https://ieeexplore.ieee.org/document/9441246},
}

@inproceedings{Klabjan2019,
  doi = {10.1109/bigdata47090.2019.9006069},
  url = {https://doi.org/10.1109/bigdata47090.2019.9006069},
  year = {2019},
  month = dec,
  publisher = {{IEEE}},
  author = {Diego Klabjan and Mark Harmon},
  title = {Activation Ensembles for Deep Neural Networks},
  booktitle = {2019 {IEEE} International Conference on Big Data (Big Data)}
}

@misc{Goyal2019,
  doi = {10.48550/ARXIV.1906.09529},
  url = {https://arxiv.org/abs/1906.09529},
  author = {Goyal,  Mohit and Goyal,  Rajan and Lall,  Brejesh},
  title = {Learning Activation Functions: A new paradigm for understanding Neural Networks},
  publisher = {arXiv},
  year = {2019},
  copyright = {arXiv.org perpetual,  non-exclusive license}
}

@inproceedings{Eisenach2017,
title={Nonparametrically Learning Activation Functions in Deep Neural Nets},
author={Carson Eisenach and Zhaoran Wang and Han Liu},
booktitle={International Conference on Learning Representations Workshops},
year={2017},
url={https://openreview.net/forum?id=H1wgawqxl}
}

@inproceedings{Vercellino2017,
author = {Conner Joseph Vercellino and William Yang Wang},
title = {Hyperactivations for Activation Function Exploration},
year = {2017},
booktitle = {Proceedings of the 3st International Conference on Neural Information Processing Systems},
location = {Long Beach, California, USA},
series = {NIPS'17},
url={http://metalearning.ml/2017/papers/metalearn17_vercellino.pdf},
}

@article{Mishra2017,
  doi = {10.1016/j.asoc.2017.09.002},
  url = {https://doi.org/10.1016/j.asoc.2017.09.002},
  year = {2017},
  month = dec,
  publisher = {Elsevier {BV}},
  volume = {61},
  pages = {983--994},
  author = {Akash Mishra and Pravin Chandra and Udayan Ghose and Sartaj Singh Sodhi},
  title = {Bi-modal derivative adaptive activation function sigmoidal feedforward artificial neural networks},
  journal = {Applied Soft Computing}
}

@inproceedings{Dugas2000,
 author = {Dugas, Charles and Bengio, Yoshua and B\'{e}lisle, Fran\c{c}ois and Nadeau, Claude and Garcia, Ren\'{e}},
 booktitle = {Advances in Neural Information Processing Systems},
 editor = {T. Leen and T. Dietterich and V. Tresp},
 pages = {},
 publisher = {MIT Press},
 title = {Incorporating Second-Order Functional Knowledge for Better Option Pricing},
 url = {https://proceedings.neurips.cc/paper_files/paper/2000/file/44968aece94f667e4095002d140b5896-Paper.pdf},
 volume = {13},
 year = {2000}
}

@inproceedings{HaoZheng2015,
  doi = {10.1109/ijcnn.2015.7280459},
  url = {https://doi.org/10.1109/ijcnn.2015.7280459},
  year = {2015},
  month = jul,
  publisher = {{IEEE}},
  author = {Hao Zheng and  Zhanlei Yang and  Wenju Liu and  Jizhong Liang and  Yanpeng Li},
  title = {Improving deep neural networks using softplus units},
  booktitle = {2015 International Joint Conference on Neural Networks ({IJCNN})}
}

@inproceedings{Ratnawati2020,
  doi = {10.1063/5.0023872},
  url = {https://doi.org/10.1063/5.0023872},
  year = {2020},
  publisher = {{AIP} Publishing},
  author = {Dian Eka Ratnawati and  Marjono and  Widodo and Syaiful Anam},
  title = {Comparison of activation function on extreme learning machine ({ELM}) performance for classifying the active compound},
  booktitle = {{SYMPOSIUM} {ON} {BIOMATHEMATICS} 2019 ({SYMOMATH} 2019)}
}

@inproceedings{Hara2015,
  doi = {10.1109/ijcnn.2015.7280578},
  url = {https://doi.org/10.1109/ijcnn.2015.7280578},
  year = {2015},
  month = jul,
  publisher = {{IEEE}},
  author = {Kazuyuki Hara and Daisuke Saito and Hayaru Shouno},
  title = {Analysis of function of rectified linear unit used in deep learning},
  booktitle = {2015 International Joint Conference on Neural Networks ({IJCNN})}
}

@inproceedings{Pandey2023,
  doi = {10.1109/icict57646.2023.10134464},
  url = {https://doi.org/10.1109/icict57646.2023.10134464},
  year = {2023},
  month = apr,
  publisher = {{IEEE}},
  author = {Gaurav Kumar Pandey and Sumit Srivastava},
  title = {{ResNet}-18 comparative analysis of various activation functions for image classification},
  booktitle = {2023 International Conference on Inventive Computation Technologies ({ICICT})}
}

@article{Pishchik2023,
  doi = {10.20944/preprints202301.0463.v1},
  url = {https://doi.org/10.20944/preprints202301.0463.v1},
  year = {2023},
  month = jan,
  publisher = {{MDPI} {AG}},
  author = {Evgenii Pishchik},
  title = {Trainable Activations for Image Classification}
}

@misc{Liao2020,
  doi = {10.48550/ARXIV.2004.13271},
  url = {https://arxiv.org/abs/2004.13271},
  author = {Liao,  Zhaohe},
  title = {Trainable Activation Function in Image Classification},
  publisher = {arXiv},
  year = {2020},
  copyright = {arXiv.org perpetual,  non-exclusive license}
}

@article{Zhao2017,
  doi = {10.1007/s10489-017-1028-7},
  url = {https://doi.org/10.1007/s10489-017-1028-7},
  year = {2017},
  month = sep,
  publisher = {Springer Science and Business Media {LLC}},
  volume = {48},
  number = {7},
  pages = {1707--1720},
  author = {Huizhen Zhao and Fuxian Liu and Longyue Li and Chang Luo},
  title = {A novel softplus linear unit for deep convolutional neural networks},
  journal = {Applied Intelligence}
}

@article{Varshney2021,
  doi = {10.1007/s11760-021-01863-z},
  url = {https://doi.org/10.1007/s11760-021-01863-z},
  year = {2021},
  month = feb,
  publisher = {Springer Science and Business Media {LLC}},
  volume = {15},
  number = {6},
  pages = {1323--1330},
  author = {Munender Varshney and Pravendra Singh},
  title = {Optimizing nonlinear activation function for convolutional neural networks},
  journal = {Signal,  Image and Video Processing}
}

@incollection{Dureja2019,
  doi = {10.1007/978-981-13-6772-4_103},
  url = {https://doi.org/10.1007/978-981-13-6772-4_103},
  year = {2019},
  publisher = {Springer Singapore},
  pages = {1179--1190},
  author = {Aman Dureja and Payal Pahwa},
  title = {Analysis of Nonlinear Activation Functions for Classification Tasks Using Convolutional Neural Networks},
  booktitle = {Lecture Notes in Electrical Engineering}
}

@incollection{Vargas2021,
  doi = {10.1007/978-3-030-85713-4_4},
  url = {https://doi.org/10.1007/978-3-030-85713-4_4},
  year = {2021},
  publisher = {Springer International Publishing},
  pages = {33--43},
  author = {V{\'{\i}}ctor Manuel Vargas and David Guijo-Rubio and Pedro Antonio Guti{\'{e}}rrez and C{\'{e}}sar Herv{\'{a}}s-Mart{\'{\i}}nez},
  title = {{ReLU}-Based Activations: Analysis and~Experimental Study for Deep Learning},
  booktitle = {Advances in Artificial Intelligence}
}

@inproceedings{Seo2017,
  doi = {10.1109/pimrc.2017.8292678},
  url = {https://doi.org/10.1109/pimrc.2017.8292678},
  year = {2017},
  month = oct,
  publisher = {{IEEE}},
  author = {Jihoon Seo and Juyul Lee and Keunyoung Kim},
  title = {Activation functions of deep neural networks for polar decoding applications},
  booktitle = {2017 {IEEE} 28th Annual International Symposium on Personal,  Indoor,  and Mobile Radio Communications ({PIMRC})}
}

@inproceedings{Xu2018Novel,
  doi = {10.1109/icdsba.2018.00079},
  url = {https://doi.org/10.1109/icdsba.2018.00079},
  year = {2018},
  month = sep,
  publisher = {{IEEE}},
  author = {Chen Xu and Jie Huang and Sheng-peng Wang and An-qing Hu},
  title = {A Novel Parameterized Activation Function in Visual Geometry Group},
  booktitle = {2018 2nd International Conference on Data Science and Business Analytics ({ICDSBA})}
}

@incollection{Sun2019,
  doi = {10.1007/978-3-030-25128-4_164},
  url = {https://doi.org/10.1007/978-3-030-25128-4_164},
  year = {2019},
  month = jul,
  publisher = {Springer International Publishing},
  pages = {1326--1335},
  author = {Kelei Sun and Jiaming Yu and Li Zhang and Zhiheng Dong},
  title = {A Convolutional Neural Network Model Based on Improved Softplus Activation Function},
  booktitle = {Advances in Intelligent Systems and Computing}
}

@inproceedings{Misra2020,
  url = {https://www.bmvc2020-conference.com/conference/papers/paper_0928.html},
  year = {2020},
  month = sep,
  publisher = {{BMVC}},
  author = {Diganta Misra},
  title = {Mish: A Self Regularized Non-Monotonic Activation Function},
  booktitle = {The 31st British Machine Vision Conference}
}

@inproceedings{Hu2018SqueezeAndExcitation,
  doi = {10.1109/cvpr.2018.00745},
  url = {https://doi.org/10.1109/cvpr.2018.00745},
  year = {2018},
  month = jun,
  publisher = {{IEEE}},
  author = {Jie Hu and Li Shen and Gang Sun},
  title = {Squeeze-and-Excitation Networks},
  booktitle = {2018 {IEEE}/{CVF} Conference on Computer Vision and Pattern Recognition}
}

@misc{Iandola2016,
  doi = {10.48550/ARXIV.1602.07360},
  url = {https://arxiv.org/abs/1602.07360},
  author = {Iandola,  Forrest N. and Han,  Song and Moskewicz,  Matthew W. and Ashraf,  Khalid and Dally,  William J. and Keutzer,  Kurt},
  title = {SqueezeNet: AlexNet-level accuracy with 50x fewer parameters and <0.5MB model size},
  publisher = {arXiv},
  year = {2016},
  copyright = {arXiv.org perpetual,  non-exclusive license}
}

@inproceedings{Zhang2018ShuffleNet,
  doi = {10.1109/cvpr.2018.00716},
  url = {https://doi.org/10.1109/cvpr.2018.00716},
  year = {2018},
  month = jun,
  publisher = {{IEEE}},
  author = {Xiangyu Zhang and Xinyu Zhou and Mengxiao Lin and Jian Sun},
  title = {{ShuffleNet}: An Extremely Efficient Convolutional Neural Network for Mobile Devices},
  booktitle = {2018 {IEEE}/{CVF} Conference on Computer Vision and Pattern Recognition}
}

@misc{Carlile2017,
  doi = {10.48550/ARXIV.1710.09967},
  url = {https://arxiv.org/abs/1710.09967},
  author = {Carlile,  Brad and Delamarter,  Guy and Kinney,  Paul and Marti,  Akiko and Whitney,  Brian},
  title = {Improving Deep Learning by Inverse Square Root Linear Units (ISRLUs)},
  publisher = {arXiv},
  year = {2017},
  copyright = {arXiv.org perpetual,  non-exclusive license}
}

@inproceedings{Wang2021ScaledYOLOv4,
  doi = {10.1109/cvpr46437.2021.01283},
  url = {https://doi.org/10.1109/cvpr46437.2021.01283},
  year = {2021},
  month = jun,
  publisher = {{IEEE}},
  author = {Chien-Yao Wang and Alexey Bochkovskiy and Hong-Yuan Mark Liao},
  title = {Scaled-{YOLOv}4: Scaling Cross Stage Partial Network},
  booktitle = {2021 {IEEE}/{CVF} Conference on Computer Vision and Pattern Recognition ({CVPR})}
}

@article{Lin2020,
  doi = {10.1109/tpami.2018.2858826},
  url = {https://doi.org/10.1109/tpami.2018.2858826},
  year = {2020},
  month = feb,
  publisher = {Institute of Electrical and Electronics Engineers ({IEEE})},
  volume = {42},
  number = {2},
  pages = {318--327},
  author = {Tsung-Yi Lin and Priya Goyal and Ross Girshick and Kaiming He and Piotr Dollar},
  title = {Focal Loss for Dense Object Detection},
  journal = {{IEEE} Transactions on Pattern Analysis and Machine Intelligence}
}

@misc{Bochkovskiy2020,
  doi = {10.48550/ARXIV.2004.10934},
  url = {https://arxiv.org/abs/2004.10934},
  author = {Bochkovskiy,  Alexey and Wang,  Chien-Yao and Liao,  Hong-Yuan Mark},
  title = {YOLOv4: Optimal Speed and Accuracy of Object Detection},
  publisher = {arXiv},
  year = {2020},
  copyright = {arXiv.org perpetual,  non-exclusive license}
}

@inproceedings{Nguyen2021,
  doi = {10.1109/icsse52999.2021.9538437},
  url = {https://doi.org/10.1109/icsse52999.2021.9538437},
  year = {2021},
  month = aug,
  publisher = {{IEEE}},
  author = {Anh Nguyen and Khoa Pham and Dat Ngo and Thanh Ngo and Lam Pham},
  title = {An Analysis of State-of-the-art Activation Functions For Supervised Deep Neural Network},
  booktitle = {2021 International Conference on System Science and Engineering ({ICSSE})}
}

@misc{Hendrycks2016,
  doi = {10.48550/ARXIV.1606.08415},
  url = {https://arxiv.org/abs/1606.08415},
  author = {Hendrycks,  Dan and Gimpel,  Kevin},
  title = {Gaussian Error Linear Units (GELUs)},
  publisher = {arXiv},
  year = {2016},
  copyright = {arXiv.org perpetual,  non-exclusive license}
}

@misc{Lee2023GELU,
  doi = {10.48550/ARXIV.2305.12073},
  url = {https://arxiv.org/abs/2305.12073},
  author = {Lee,  Minhyeok},
  title = {GELU Activation Function in Deep Learning: A Comprehensive Mathematical Analysis and Performance},
  publisher = {arXiv},
  year = {2023},
  copyright = {Creative Commons Attribution 4.0 International}
}

@inproceedings{Kang2022,
  doi = {10.1109/icvr55215.2022.9848068},
  url = {https://doi.org/10.1109/icvr55215.2022.9848068},
  year = {2022},
  month = may,
  publisher = {{IEEE}},
  author = {Jian Kang and Rui Liu and Yijing Li and Qian Liu and Pengfei Wang and Qiang Zhang and Dongsheng Zhou},
  title = {An Improved 3D Human Pose Estimation Model Based on Temporal Convolution with Gaussian Error Linear Units},
  booktitle = {2022 8th International Conference on Virtual Reality ({ICVR})}
}

@article{Wong2022,
  doi = {10.1007/s12065-022-00795-y},
  url = {https://doi.org/10.1007/s12065-022-00795-y},
  year = {2022},
  month = nov,
  publisher = {Springer Science and Business Media {LLC}},
  author = {Kit Wong and Rolf Dornberger and Thomas Hanne},
  title = {An analysis of weight initialization methods in connection with different activation functions for feedforward neural networks},
  journal = {Evolutionary Intelligence}
}

@misc{Yu2019Symmetrical,
  doi = {10.48550/ARXIV.1911.03925},
  url = {https://arxiv.org/abs/1911.03925},
  author = {Yu,  Chao and Su,  Zhiguo},
  title = {Symmetrical Gaussian Error Linear Units (SGELUs)},
  publisher = {arXiv},
  year = {2019},
  copyright = {arXiv.org perpetual,  non-exclusive license}
}

@inproceedings{Su2017,
 author = {Su, Qinliang and Liao, xuejun and Carin, Lawrence},
 booktitle = {Advances in Neural Information Processing Systems},
 editor = {I. Guyon and U. Von Luxburg and S. Bengio and H. Wallach and R. Fergus and S. Vishwanathan and R. Garnett},
 publisher = {Curran Associates, Inc.},
 title = {A Probabilistic Framework for Nonlinearities in Stochastic Neural Networks},
 url = {https://proceedings.neurips.cc/paper_files/paper/2017/file/35936504a37d53e03abdfbc7318d9ec7-Paper.pdf},
 volume = {30},
 year = {2017}
}

@misc{Shridhar2019,
  doi = {10.48550/ARXIV.1905.10761},
  url = {https://arxiv.org/abs/1905.10761},
  author = {Shridhar,  Kumar and Lee,  Joonho and Hayashi,  Hideaki and Mehta,  Purvanshi and Iwana,  Brian Kenji and Kang,  Seokjun and Uchida,  Seiichi and Ahmed,  Sheraz and Dengel,  Andreas},
  title = {ProbAct: A Probabilistic Activation Function for Deep Neural Networks},
  publisher = {arXiv},
  year = {2019},
  copyright = {arXiv.org perpetual,  non-exclusive license}
}

@inproceedings{Berradi2018,
  doi = {10.1145/3230905.3230956},
  url = {https://doi.org/10.1145/3230905.3230956},
  year = {2018},
  month = may,
  publisher = {{ACM}},
  author = {Yassine Berradi},
  title = {Symmetric Power Activation Functions for Deep Neural Networks},
  booktitle = {Proceedings of the International Conference on Learning and Optimization Algorithms: Theory and Applications}
}

@article{LpezRubio2019,
  doi = {10.1007/s11063-018-09974-4},
  url = {https://doi.org/10.1007/s11063-018-09974-4},
  year = {2019},
  month = jan,
  publisher = {Springer Science and Business Media {LLC}},
  volume = {50},
  number = {1},
  pages = {121--147},
  author = {Ezequiel L{\'{o}}pez-Rubio and Francisco Ortega-Zamorano and Enrique Dom{\'{\i}}nguez and Jos{\'{e}} Mu{\~{n}}oz-P{\'{e}}rez},
  title = {Piecewise Polynomial Activation Functions for Feedforward Neural Networks},
  journal = {Neural Processing Letters}
}

@InProceedings{Li2016,
  title = 	 {Multi-Bias Non-linear Activation in Deep Neural Networks},
  author = 	 {Li, Hongyang and Ouyang, Wanli and Wang, Xiaogang},
  booktitle = 	 {Proceedings of The 33rd International Conference on Machine Learning},
  pages = 	 {221--229},
  year = 	 {2016},
  editor = 	 {Balcan, Maria Florina and Weinberger, Kilian Q.},
  volume = 	 {48},
  series = 	 {Proceedings of Machine Learning Research},
  address = 	 {New York, New York, USA},
  month = 	 {20--22 Jun},
  publisher =    {PMLR},
  url = 	 {https://proceedings.mlr.press/v48/lia16.html}
}

@article{Fakhoury2022,
  doi = {10.1016/j.neunet.2022.04.029},
  url = {https://doi.org/10.1016/j.neunet.2022.04.029},
  year = {2022},
  month = aug,
  publisher = {Elsevier {BV}},
  volume = {152},
  pages = {332--346},
  author = {Daniele Fakhoury and Emanuele Fakhoury and Hendrik Speleers},
  title = {{ExSpliNet}: An interpretable and expressive spline-based neural network},
  journal = {Neural Networks}
}

@article{Kapoor2021,
  doi = {10.1007/s13369-020-05207-w},
  url = {https://doi.org/10.1007/s13369-020-05207-w},
  year = {2021},
  month = jan,
  publisher = {Springer Science and Business Media {LLC}},
  volume = {46},
  number = {10},
  pages = {9451--9464},
  author = {Divneet Singh Kapoor and Amit Kumar Kohli},
  title = {Adaptive-Slope Squashing-Function-Based {ANN} for {CSI} Estimation and Symbol Detection in {SFBC}-{OFDM} System},
  journal = {Arabian Journal for Science and Engineering}
}

@article{Simos2021,
  doi = {10.1007/s00521-021-05787-0},
  url = {https://doi.org/10.1007/s00521-021-05787-0},
  year = {2021},
  month = mar,
  publisher = {Springer Science and Business Media {LLC}},
  volume = {33},
  number = {16},
  pages = {10227--10233},
  author = {T. E. Simos and Ch. Tsitouras},
  title = {Efficiently inaccurate approximation of hyperbolic tangent used as transfer function in artificial neural networks},
  journal = {Neural Computing and Applications}
}

@article{Bohra2020,
  doi = {10.1109/ojsp.2020.3039379},
  url = {https://doi.org/10.1109/ojsp.2020.3039379},
  year = {2020},
  publisher = {Institute of Electrical and Electronics Engineers ({IEEE})},
  volume = {1},
  pages = {295--309},
  author = {Pakshal Bohra and Joaquim Campos and Harshit Gupta and Shayan Aziznejad and Michael Unser},
  title = {Learning Activation Functions in Deep (Spline) Neural Networks},
  journal = {{IEEE} Open Journal of Signal Processing}
}

@article{Neumayer2023,
  title = {Approximation of Lipschitz Functions Using Deep Spline Neural Networks},
  volume = {5},
  ISSN = {2577-0187},
  url = {http://dx.doi.org/10.1137/22M1504573},
  DOI = {10.1137/22m1504573},
  number = {2},
  journal = {SIAM Journal on Mathematics of Data Science},
  publisher = {Society for Industrial & Applied Mathematics (SIAM)},
  author = {Neumayer,  Sebastian and Goujon,  Alexis and Bohra,  Pakshal and Unser,  Michael},
  year = {2023},
  month = may,
  pages = {306–322}
}

@misc{Ducotterd2022,
  doi = {10.48550/ARXIV.2210.16222},
  url = {https://arxiv.org/abs/2210.16222},
  author = {Ducotterd,  Stanislas and Goujon,  Alexis and Bohra,  Pakshal and Perdios,  Dimitris and Neumayer,  Sebastian and Unser,  Michael},
  title = {Improving Lipschitz-Constrained Neural Networks by Learning Activation Functions},
  publisher = {arXiv},
  year = {2022},
  copyright = {arXiv.org perpetual,  non-exclusive license}
}

@article{Parhi2020,
  doi = {10.1109/lsp.2020.3027517},
  url = {https://doi.org/10.1109/lsp.2020.3027517},
  year = {2020},
  publisher = {Institute of Electrical and Electronics Engineers ({IEEE})},
  volume = {27},
  pages = {1779--1783},
  author = {Rahul Parhi and Robert D. Nowak},
  title = {The Role of Neural Network Activation Functions},
  journal = {{IEEE} Signal Processing Letters}
}

@inproceedings{Sivri2022,
  doi = {10.1109/ecai54874.2022.9847486},
  url = {https://doi.org/10.1109/ecai54874.2022.9847486},
  year = {2022},
  month = jun,
  publisher = {{IEEE}},
  author = {Talya Tumer Sivri and Nergis Pervan Akman and Ali Berkol},
  title = {Multiclass Classification Using Arctangent Activation Function and Its Variations},
  booktitle = {2022 14th International Conference on Electronics,  Computers and Artificial Intelligence ({ECAI})}
}

@article{TmerSivri2023,
  title = {The Impact of Irrationals on the Range of Arctan Activation Function for Deep Learning Models},
  url = {http://dx.doi.org/10.20944/preprints202305.1245.v1},
  DOI = {10.20944/preprints202305.1245.v1},
  publisher = {MDPI AG},
  author = {T\"{u}mer-Sivri,  Talya and Pervan-Akman,  Nergis and Berkol,  Ali},
  year = {2023},
  month = may 
}

@article{Lai2023,
  doi = {10.3390/math11143081},
  url = {https://doi.org/10.3390/math11143081},
  year = {2023},
  month = jul,
  publisher = {{MDPI} {AG}},
  volume = {11},
  number = {14},
  pages = {3081},
  author = {Derek Ka-Hei Lai and Ethan Shiu-Wang Cheng and Bryan Pak-Hei So and Ye-Jiao Mao and Sophia Ming-Yan Cheung and Daphne Sze Ki Cheung and Duo Wai-Chi Wong and James Chung-Wai Cheung},
  title = {Transformer Models and Convolutional Networks with Different Activation Functions for Swallow Classification Using Depth Video Data},
  journal = {Mathematics}
}

@article{Jiang2022,
  doi = {10.3390/electronics11223799},
  url = {https://doi.org/10.3390/electronics11223799},
  year = {2022},
  month = nov,
  publisher = {{MDPI} {AG}},
  volume = {11},
  number = {22},
  pages = {3799},
  author = {Yuanyuan Jiang and Jinyang Xie and Dong Zhang},
  title = {An Adaptive Offset Activation Function for {CNN} Image Classification Tasks},
  journal = {Electronics}
}

@article{Aziznejad2020,
  doi = {10.1109/tsp.2020.3014611},
  url = {https://doi.org/10.1109/tsp.2020.3014611},
  year = {2020},
  publisher = {Institute of Electrical and Electronics Engineers ({IEEE})},
  volume = {68},
  pages = {4688--4699},
  author = {Shayan Aziznejad and Harshit Gupta and Joaquim Campos and Michael Unser},
  title = {Deep Neural Networks With Trainable Activations and Controlled Lipschitz Constant},
  journal = {{IEEE} Transactions on Signal Processing}
}

@inproceedings{Lane1990,
 author = {Lane, Stephen and Flax, Marshall and Handelman, David and Gelfand, Jack},
 booktitle = {Advances in Neural Information Processing Systems},
 editor = {R.P. Lippmann and J. Moody and D. Touretzky},
 pages = {},
 publisher = {Morgan-Kaufmann},
 title = {Multi-Layer Perceptrons with B-Spline Receptive Field Functions},
 url = {https://proceedings.neurips.cc/paper_files/paper/1990/file/94f6d7e04a4d452035300f18b984988c-Paper.pdf},
 volume = {3},
 year = {1990}
}

@article{Vaicaitis2022,
  doi = {10.1109/tmtt.2022.3210034},
  url = {https://doi.org/10.1109/tmtt.2022.3210034},
  year = {2022},
  month = nov,
  publisher = {Institute of Electrical and Electronics Engineers ({IEEE})},
  volume = {70},
  number = {11},
  pages = {4910--4915},
  author = {Andrius Vaicaitis and John Dooley},
  title = {Segmented Spline Curve Neural Network for Low Latency Digital Predistortion of {RF} Power Amplifiers},
  journal = {{IEEE} Transactions on Microwave Theory and Techniques}
}

@article{Kumar2014,
  doi = {10.36478/ijscomp.2014.377.385},
  url = {https://medwelljournals.com/abstract/?doi=ijscomp.2014.377.385},
  year = {2014},
  publisher = {Medwell Publications},
  volume = {9},
  number = {6},
  pages = {377--385},
  author = {R. Ganesh Kumar and Y.S. Kumaraswamy},
  title = {Spline Activated Neural Network for Classifying Cardiac Arrhythmia},
  journal = {International Journal of Soft Computing}
}

@incollection{Tezel2007,
  doi = {10.1007/978-3-540-74819-9_1},
  url = {https://doi.org/10.1007/978-3-540-74819-9_1},
  publisher = {Springer Berlin Heidelberg},
  pages = {1--8},
  year = {2007},
  author = {G\"{u}lay Tezel and Y\"{u}ksel \"{O}zbay},
  title = {A New Neural Network with Adaptive Activation Function for Classification of {ECG} Arrhythmias},
  booktitle = {Lecture Notes in Computer Science}
}

@inproceedings{Campolucci1996,
  doi = {10.1109/melcon.1996.551220},
  url = {https://doi.org/10.1109/melcon.1996.551220},
  publisher = {{IEEE}},
  author = {P. Campolucci and F. Capperelli and S. Guarnieri and F. Piazza and A. Uncini},
  title = {Neural networks with adaptive spline activation function},
  year = {1996},
  booktitle = {Proceedings of 8th Mediterranean Electrotechnical Conference on Industrial Applications in Power Systems,  Computer Science and Telecommunications ({MELECON} 96)}
}

@incollection{Uncini2002,
  doi = {10.1007/3-540-45808-5_19},
  url = {https://doi.org/10.1007/3-540-45808-5_19},
  year = {2002},
  publisher = {Springer Berlin Heidelberg},
  pages = {168--177},
  author = {Aurelio Uncini},
  title = {Sound Synthesis by Flexible Activation Function Recurrent Neural Networks},
  booktitle = {Neural Nets}
}

@incollection{Mayer2001,
  doi = {10.1007/3-540-45329-6_10},
  url = {https://doi.org/10.1007/3-540-45329-6_10},
  year = {2001},
  publisher = {Springer Berlin Heidelberg},
  pages = {63--73},
  author = {Helmut A. Mayer and Roland Schwaiger},
  title = {Evolution of Cubic Spline Activation Functions for Artificial Neural Networks},
  booktitle = {Progress in Artificial Intelligence}
}

@inproceedings{Mayer2002,
  doi = {10.1109/ijcnn.2002.1007787},
  url = {https://doi.org/10.1109/ijcnn.2002.1007787},
  publisher = {{IEEE}},
  year = {2002},
  author = {Helmut A. Mayer and Roland Schwaiger},
  title = {Differentiation of neuron types by evolving activation function templates for artificial neural networks},
  booktitle = {Proceedings of the 2002 International Joint Conference on Neural Networks. {IJCNN}{\textquotesingle}02 (Cat. No.02CH37290)}
}

@inproceedings{Hagg2017,
  doi = {10.1145/3071178.3071275},
  url = {https://doi.org/10.1145/3071178.3071275},
  year = {2017},
  month = jul,
  publisher = {{ACM}},
  author = {Alexander Hagg and Maximilian Mensing and Alexander Asteroth},
  title = {Evolving parsimonious networks by mixing activation functions},
  booktitle = {Proceedings of the Genetic and Evolutionary Computation Conference}
}

@article{Knezevic2023,
  doi = {10.1109/access.2022.3232064},
  url = {https://doi.org/10.1109/access.2022.3232064},
  year = {2023},
  publisher = {Institute of Electrical and Electronics Engineers ({IEEE})},
  volume = {11},
  pages = {284--299},
  author = {Karlo Knezevic and Juraj Fulir and Domagoj Jakobovic and Stjepan Picek and Marko Durasevic},
  title = {{NeuroSCA}: Evolving Activation Functions for Side-Channel Analysis},
  journal = {{IEEE} Access}
}

@inproceedings{Liu1996,
  doi = {10.1109/icec.1996.542681},
  url = {https://doi.org/10.1109/icec.1996.542681},
  publisher = {{IEEE}},
  author = {Y. Liu and X. Yao},
  year = {1996},
  title = {Evolutionary design of artificial neural networks with different nodes},
  booktitle = {Proceedings of {IEEE} International Conference on Evolutionary Computation}
}

@inproceedings{Cui2022,
  doi = {10.1109/cibcb55180.2022.9863054},
  url = {https://doi.org/10.1109/cibcb55180.2022.9863054},
  year = {2022},
  month = aug,
  publisher = {{IEEE}},
  author = {Peiyu Cui and Kay C. Wiese},
  title = {{EvoDNN} - Evolving Weights,  Biases,  and Activation Functions in a Deep Neural Network},
  booktitle = {2022 {IEEE} Conference on Computational Intelligence in Bioinformatics and Computational Biology ({CIBCB})}
}

@inproceedings{Cui2019,
  doi = {10.1109/cec.2019.8789964},
  url = {https://doi.org/10.1109/cec.2019.8789964},
  year = {2019},
  month = jun,
  publisher = {{IEEE}},
  author = {Peiyu Cui and Boris Shabash and Kay C. Wiese},
  title = {{EvoDNN} - An Evolutionary Deep Neural Network with Heterogeneous Activation Functions},
  booktitle = {2019 {IEEE} Congress on Evolutionary Computation ({CEC})}
}

@misc{Marchisio2018,
  doi = {10.48550/ARXIV.1811.03980},
  url = {https://arxiv.org/abs/1811.03980},
  author = {Marchisio,  Alberto and Hanif,  Muhammad Abdullah and Rehman,  Semeen and Martina,  Maurizio and Shafique,  Muhammad},
  title = {A Methodology for Automatic Selection of Activation Functions to Design Hybrid Deep Neural Networks},
  publisher = {arXiv},
  year = {2018},
  copyright = {arXiv.org perpetual,  non-exclusive license}
}

@article{Vijayaprabakaran2022,
  doi = {10.1016/j.jksuci.2020.04.015},
  url = {https://doi.org/10.1016/j.jksuci.2020.04.015},
  year = {2022},
  month = jun,
  publisher = {Elsevier {BV}},
  volume = {34},
  number = {6},
  pages = {2637--2650},
  author = {Vijayaprabakaran, K. and Sathiyamurthy, K.},
  title = {Towards activation function search for long short-term model network: A differential evolution based approach},
  journal = {Journal of King Saud University - Computer and Information Sciences}
}

@incollection{ONeill2018,
  doi = {10.1007/978-3-030-03991-2_56},
  url = {https://doi.org/10.1007/978-3-030-03991-2_56},
  year = {2018},
  publisher = {Springer International Publishing},
  pages = {616--629},
  author = {Damien O'Neill and Bing Xue and Mengjie Zhang},
  title = {Co-evolution of Novel Tree-Like {ANNs} and Activation Functions: An Observational Study},
  booktitle = {{AI} 2018: Advances in Artificial Intelligence}
}

@inproceedings{Papavasileiou2017,
  doi = {10.1109/ssci.2017.8285328},
  url = {https://doi.org/10.1109/ssci.2017.8285328},
  year = {2017},
  month = nov,
  publisher = {{IEEE}},
  author = {Evgenia Papavasileiou and Bart Jansen},
  title = {The importance of the activation function in {NeuroEvolution} with {FS}-{NEAT} and {FD}-{NEAT}},
  booktitle = {2017 {IEEE} Symposium Series on Computational Intelligence ({SSCI})}
}

@inproceedings{Solazzi2000Neural,
  doi = {10.1109/icassp.2000.860155},
  url = {https://doi.org/10.1109/icassp.2000.860155},
  publisher = {{IEEE}},
  year = {2000},
  author = {M. Solazzi and A. Uncini and F. Piazza},
  title = {Neural equalizer with adaptive multidimensional spline activation functions},
  booktitle = {2000 {IEEE} International Conference on Acoustics,  Speech,  and Signal Processing. Proceedings (Cat. No.00CH37100)}
}

@inproceedings{Piazza1993,
  doi = {10.1109/ijcnn.1993.716806},
  url = {https://doi.org/10.1109/ijcnn.1993.716806},
  publisher = {{IEEE}},
  year = {1993},
  author = {F. Piazza and A. Uncini and M. Zenobi},
  title = {Neural networks with digital {LUT} activation functions},
  booktitle = {Proceedings of 1993 International Conference on Neural Networks ({IJCNN}-93-Nagoya,  Japan)}
}

@article{Fiori2002,
  doi = {10.1016/s0893-6080(01)00105-8},
  url = {https://doi.org/10.1016/s0893-6080(01)00105-8},
  year = {2002},
  month = jan,
  publisher = {Elsevier {BV}},
  volume = {15},
  number = {1},
  pages = {85--94},
  author = {Simone Fiori},
  title = {Hybrid independent component analysis by adaptive {LUT} activation function neurons},
  journal = {Neural Networks}
}

@inproceedings{Kuzuya2022,
  doi = {10.1109/smc53654.2022.9945478},
  url = {https://doi.org/10.1109/smc53654.2022.9945478},
  year = {2022},
  month = oct,
  publisher = {{IEEE}},
  author = {Naoki Kuzuya and Tomoharu Nagao},
  title = {Designing B-spline-based Highly Efficient Neural Networks for {IoT} Applications on Edge Platforms},
  booktitle = {2022 {IEEE} International Conference on Systems,  Man,  and Cybernetics ({SMC})}
}

@inproceedings{Tripathi2021,
  doi = {10.1109/asiancon51346.2021.9544754},
  url = {https://doi.org/10.1109/asiancon51346.2021.9544754},
  year = {2021},
  month = aug,
  publisher = {{IEEE}},
  author = {Raghuvendra Pratap Tripathi and Manish Tiwari and Amit Dhawan and Anand Sharma and Sumit Kumar Jha},
  title = {A Survey on Efficient Realization of Activation Functions of Artificial Neural Network},
  booktitle = {2021 Asian Conference on Innovation in Technology ({ASIANCON})}
}

@inproceedings{Bouguezzi2021,
  doi = {10.1109/ssd52085.2021.9429506},
  url = {https://doi.org/10.1109/ssd52085.2021.9429506},
  year = {2021},
  month = mar,
  publisher = {{IEEE}},
  author = {Safa Bouguezzi and Hassene Faiedh and Chokri Souani},
  title = {Hardware Implementation of Tanh Exponential Activation Function using {FPGA}},
  booktitle = {2021 18th International Multi-Conference on Systems,  Signals {\&} Devices ({SSD})}
}

@inproceedings{Li2018AnEfficient,
  doi = {10.1109/icivc.2018.8492754},
  url = {https://doi.org/10.1109/icivc.2018.8492754},
  year = {2018},
  month = jun,
  publisher = {{IEEE}},
  author = {Lin Li and Shengbing Zhang and Juan Wu},
  title = {An Efficient Hardware Architecture for Activation Function in Deep Learning Processor},
  booktitle = {2018 {IEEE} 3rd International Conference on Image,  Vision and Computing ({ICIVC})}
}

@article{Tsai2015,
  doi = {10.1109/tcsii.2015.2456531},
  url = {https://doi.org/10.1109/tcsii.2015.2456531},
  year = {2015},
  month = nov,
  publisher = {Institute of Electrical and Electronics Engineers ({IEEE})},
  volume = {62},
  number = {11},
  pages = {1073--1077},
  author = {Chang-Hung Tsai and Yu-Ting Chih and Wing Hung Wong and Chen-Yi Lee},
  title = {A Hardware-Efficient Sigmoid Function With Adjustable Precision for a Neural Network System},
  journal = {{IEEE} Transactions on Circuits and Systems {II}: Express Briefs}
}

@inproceedings{Namin2009,
  doi = {10.1109/iscas.2009.5118213},
  url = {https://doi.org/10.1109/iscas.2009.5118213},
  year = {2009},
  month = may,
  publisher = {{IEEE}},
  author = {Ashkan Hosseinzadeh Namin and Karl Leboeuf and Roberto Muscedere and Huapeng Wu and Majid Ahmadi},
  title = {Efficient hardware implementation of the hyperbolic tangent sigmoid function},
  booktitle = {2009 {IEEE} International Symposium on Circuits and Systems}
}

@inproceedings{Pogiri2022,
  doi = {10.1109/ises54909.2022.00090},
  url = {https://doi.org/10.1109/ises54909.2022.00090},
  year = {2022},
  month = dec,
  publisher = {{IEEE}},
  author = {Revathi Pogiri and Samit Ari and K K Mahapatra},
  title = {Design and {FPGA} Implementation of the {LUT} based Sigmoid Function for {DNN} Applications},
  booktitle = {2022 {IEEE} International Symposium on Smart Electronic Systems ({iSES})}
}

@article{Shakiba2021,
  doi = {10.1109/tie.2020.3034856},
  url = {https://doi.org/10.1109/tie.2020.3034856},
  year = {2021},
  month = nov,
  publisher = {Institute of Electrical and Electronics Engineers ({IEEE})},
  volume = {68},
  number = {11},
  pages = {10856--10867},
  author = {Fatemeh Mohammadi Shakiba and MengChu Zhou},
  title = {Novel Analog Implementation of a Hyperbolic Tangent Neuron in Artificial Neural Networks},
  journal = {{IEEE} Transactions on Industrial Electronics}
}

@article{Xie2020,
  doi = {10.1109/tvlsi.2020.3015391},
  url = {https://doi.org/10.1109/tvlsi.2020.3015391},
  year = {2020},
  month = dec,
  publisher = {Institute of Electrical and Electronics Engineers ({IEEE})},
  volume = {28},
  number = {12},
  pages = {2540--2550},
  author = {Yusheng Xie and Alex Noel Joseph Raj and Zhendong Hu and Shaohaohan Huang and Zhun Fan and Miroslav Joler},
  title = {A Twofold Lookup Table Architecture for Efficient Approximation of Activation Functions},
  journal = {{IEEE} Transactions on Very Large Scale Integration ({VLSI}) Systems}
}

@inproceedings{MiaoKang2005,
  doi = {10.1109/icnnb.2005.1614681},
  url = {https://doi.org/10.1109/icnnb.2005.1614681},
  publisher = {{IEEE}},
  year = {2005},
  author = {Miao Kang and D. Palmer-Brown},
  title = {An Adaptive Function Neural Network ({ADFUNN}) Classifier},
  booktitle = {2005 International Conference on Neural Networks and Brain}
}

@inproceedings{Kang2007,
  doi = {10.1109/ijcnn.2007.4371406},
  url = {https://doi.org/10.1109/ijcnn.2007.4371406},
  year = {2007},
  month = aug,
  publisher = {{IEEE}},
  author = {Miao Kang and Dominic Palmer-Brown},
  title = {A Multi-layer {ADaptive} {FUnction} Neural Network ({MADFUNN}) for Letter Image Recognition},
  booktitle = {2007 International Joint Conference on Neural Networks}
}

@misc{Farhadi2019,
  doi = {10.48550/ARXIV.1901.09849},
  url = {https://arxiv.org/abs/1901.09849},
  author = {Farhadi,  Farnoush and Nia,  Vahid Partovi and Lodi,  Andrea},
  title = {Activation Adaptation in Neural Networks},
  publisher = {arXiv},
  year = {2019},
  copyright = {arXiv.org perpetual,  non-exclusive license}
}

@article{Li2019PowerNet,
  doi = {10.48550/ARXIV.1909.05136},
  url = {https://arxiv.org/abs/1909.05136},
  author = {Li,  Bo and Tang,  Shanshan and Yu,  Haijun},
  title = {PowerNet: Efficient Representations of Polynomials and Smooth Functions by Deep Neural Networks with Rectified Power Units},
  publisher = {arXiv},
  year = {2019},
  copyright = {arXiv.org perpetual,  non-exclusive license}
}

@InProceedings{Telgarsky2017,
  title = 	 {Neural Networks and Rational Functions},
  author =       {Matus Telgarsky},
  booktitle = 	 {Proceedings of the 34th International Conference on Machine Learning},
  pages = 	 {3387--3393},
  year = 	 {2017},
  editor = 	 {Precup, Doina and Teh, Yee Whye},
  volume = 	 {70},
  series = 	 {Proceedings of Machine Learning Research},
  month = 	 {06--11 Aug},
  publisher =    {PMLR},
  url = 	 {https://proceedings.mlr.press/v70/telgarsky17a.html},
}

@inproceedings{Molina2020,
title={Padé Activation Units: End-to-end Learning of Flexible Activation Functions in Deep Networks},
author={Alejandro Molina and Patrick Schramowski and Kristian Kersting},
booktitle={International Conference on Learning Representations},
year={2020},
url={https://openreview.net/forum?id=BJlBSkHtDS}
}

@incollection{Brezinski2002,
  doi = {10.1007/978-1-4613-0261-2_4},
  url = {https://doi.org/10.1007/978-1-4613-0261-2_4},
  year = {2002},
  publisher = {Springer {US}},
  pages = {87--134},
  author = {Claude Brezinski},
  title = {Pad{\'{e}} Approximations},
  booktitle = {Computational Aspects of Linear Control}
}

@article{Brezinski1996,
  doi = {10.1016/0168-9274(95)00110-7},
  url = {https://doi.org/10.1016/0168-9274(95)00110-7},
  year = {1996},
  month = mar,
  publisher = {Elsevier {BV}},
  volume = {20},
  number = {3},
  pages = {299--318},
  author = {C. Brezinski},
  title = {Extrapolation algorithms and Pad{\'{e}} approximations: a historical survey},
  journal = {Applied Numerical Mathematics}
}

@inproceedings{Boulle2020,
 author = {Boulle, Nicolas and Nakatsukasa, Yuji and Townsend, Alex},
 booktitle = {Advances in Neural Information Processing Systems},
 editor = {H. Larochelle and M. Ranzato and R. Hadsell and M.F. Balcan and H. Lin},
 pages = {14243--14253},
 publisher = {Curran Associates, Inc.},
 title = {Rational neural networks},
 url = {https://proceedings.neurips.cc/paper_files/paper/2020/file/a3f390d88e4c41f2747bfa2f1b5f87db-Paper.pdf},
 volume = {33},
 year = {2020}
}

@inproceedings{Chen2018Rational,
  doi = {10.1109/icdm.2018.00021},
  url = {https://doi.org/10.1109/icdm.2018.00021},
  year = {2018},
  month = nov,
  publisher = {{IEEE}},
  author = {Zhiqian Chen and Feng Chen and Rongjie Lai and Xuchao Zhang and Chang-Tien Lu},
  title = {Rational Neural Networks for Approximating Graph Convolution Operator on Jump Discontinuities},
  booktitle = {2018 {IEEE} International Conference on Data Mining ({ICDM})}
}

@incollection{Trimmel2022,
  doi = {10.1007/978-3-031-20044-1_41},
  url = {https://doi.org/10.1007/978-3-031-20044-1_41},
  year = {2022},
  publisher = {Springer Nature Switzerland},
  pages = {722--738},
  author = {Martin Trimmel and Mihai Zanfir and Richard Hartley and Cristian Sminchisescu},
  title = {{ERA}: Enhanced Rational Activations},
  booktitle = {Lecture Notes in Computer Science}
}

@article{Apicella2019,
  doi = {10.1016/j.neucom.2019.08.065},
  url = {https://doi.org/10.1016/j.neucom.2019.08.065},
  year = {2019},
  month = dec,
  publisher = {Elsevier {BV}},
  volume = {370},
  pages = {1--15},
  author = {Andrea Apicella and Francesco Isgr{\`{o}} and Roberto Prevete},
  title = {A simple and efficient architecture for trainable activation functions},
  journal = {Neurocomputing}
}

@misc{Chen2020Dynamic,
  doi = {10.48550/ARXIV.2003.10027},
  url = {https://arxiv.org/abs/2003.10027},
  author = {Chen,  Yinpeng and Dai,  Xiyang and Liu,  Mengchen and Chen,  Dongdong and Yuan,  Lu and Liu,  Zicheng},
  title = {Dynamic ReLU},
  publisher = {arXiv},
  year = {2020},
  copyright = {arXiv.org perpetual,  non-exclusive license}
}

@misc{Minhas2019,
  doi = {10.48550/ARXIV.1912.12187},
  url = {https://arxiv.org/abs/1912.12187},
  author = {Minhas,  Fayyaz ul Amir Afsar and Asif,  Amina},
  title = {Learning Neural Activations},
  publisher = {arXiv},
  year = {2019},
  copyright = {arXiv.org perpetual,  non-exclusive license}
}

@article{Scardapane2019,
  doi = {10.1016/j.neunet.2018.11.002},
  url = {https://doi.org/10.1016/j.neunet.2018.11.002},
  year = {2019},
  month = feb,
  publisher = {Elsevier {BV}},
  volume = {110},
  pages = {19--32},
  author = {Simone Scardapane and Steven Van Vaerenbergh and Simone Totaro and Aurelio Uncini},
  title = {Kafnets: Kernel-based non-parametric activation functions for neural networks},
  journal = {Neural Networks}
}

@article{Kiliarslan2022,
  doi = {10.1007/s00521-022-07211-7},
  url = {https://doi.org/10.1007/s00521-022-07211-7},
  year = {2022},
  month = apr,
  publisher = {Springer Science and Business Media {LLC}},
  volume = {34},
  number = {16},
  pages = {13909--13923},
  author = {Serhat Kili{\c{c}}arslan and Mete Celik},
  title = {{KAF}+{RSigELU}: a nonlinear and kernel-based activation function for deep neural networks},
  journal = {Neural Computing and Applications}
}

@misc{Chen2020AReLU,
  doi = {10.48550/ARXIV.2006.13858},
  url = {https://arxiv.org/abs/2006.13858},
  author = {Chen,  Dengsheng and Li,  Jun and Xu,  Kai},
  title = {AReLU: Attention-based Rectified Linear Unit},
  publisher = {arXiv},
  year = {2020},
  copyright = {arXiv.org perpetual,  non-exclusive license}
}

@inproceedings{Kang2021,
  doi = {10.21437/asvspoof.2021-13},
  url = {https://doi.org/10.21437/asvspoof.2021-13},
  year = {2021},
  month = sep,
  publisher = {{ISCA}},
  author = {Woo Hyun Kang and Jahangir Alam and Abderrahim Fathan},
  title = {Investigation on activation functions for robust end-to-end spoofing attack detection system},
  booktitle = {2021 Edition of the Automatic Speaker Verification and Spoofing Countermeasures Challenge}
}

@inproceedings{Gupta2021,
  doi = {10.1117/12.2582176},
  url = {https://doi.org/10.1117/12.2582176},
  year = {2021},
  month = feb,
  publisher = {{SPIE}},
  author = {Prateek Gupta and Hanna Siebert and Mattias P. Heinrich and Kumar T. Rajamani},
  editor = {Bennett A. Landman and Ivana I{\v{s}}gum},
  title = {{DA}-{AR}-Net: an attentive activation based Deformable auto-encoder for group-wise registration},
  booktitle = {Medical Imaging 2021: Image Processing}
}

@misc{Delfosse2021,
  doi = {10.48550/ARXIV.2102.09407},
  url = {https://arxiv.org/abs/2102.09407},
  author = {Delfosse,  Quentin and Schramowski,  Patrick and Mundt,  Martin and Molina,  Alejandro and Kersting,  Kristian},
  title = {Adaptive Rational Activations to Boost Deep Reinforcement Learning},
  publisher = {arXiv},
  year = {2021},
  copyright = {Creative Commons Attribution 4.0 International}
}

@article{ElJaafari2020,
  doi = {10.1007/s11760-020-01746-9},
  url = {https://doi.org/10.1007/s11760-020-01746-9},
  year = {2020},
  month = jul,
  publisher = {Springer Science and Business Media {LLC}},
  volume = {15},
  number = {2},
  pages = {241--246},
  author = {Ilyas El Jaafari and Ayoub Ellahyani and Said Charfi},
  title = {Parametric rectified nonlinear unit ({PRenu}) for convolution neural networks},
  journal = {Signal,  Image and Video Processing}
}

@article{Chai2022,
  doi = {10.3390/sym14051027},
  url = {https://doi.org/10.3390/sym14051027},
  year = {2022},
  month = may,
  publisher = {{MDPI} {AG}},
  volume = {14},
  number = {5},
  pages = {1027},
  author = {Enhui Chai and Wei Yu and Tianxiang Cui and Jianfeng Ren and Shusheng Ding},
  title = {An Efficient Asymmetric Nonlinear Activation Function for Deep Neural Networks},
  journal = {Symmetry}
}

@incollection{Douge2021,
  doi = {10.1007/978-3-030-70866-5_15},
  url = {https://doi.org/10.1007/978-3-030-70866-5_15},
  year = {2021},
  publisher = {Springer International Publishing},
  pages = {237--244},
  author = {Khalid Douge and Aissam Berrahou and Youssef Talibi Alaoui and Mohammed Talibi Alaoui},
  title = {A Self-gated Activation Function {SINSIG} Based on the Sine Trigonometric for Neural Network Models},
  booktitle = {Machine Learning for Networking}
}

@article{Vargas2023,
  doi = {10.1109/tnnls.2021.3105444},
  url = {https://doi.org/10.1109/tnnls.2021.3105444},
  year = {2023},
  month = mar,
  publisher = {Institute of Electrical and Electronics Engineers ({IEEE})},
  volume = {34},
  number = {3},
  pages = {1478--1488},
  author = {V{\'{\i}}ctor Manuel Vargas and Pedro Antonio Guti{\'{e}}rrez and Javier Barbero-G{\'{o}}mez and C{\'{e}}sar Herv{\'{a}}s-Mart{\'{\i}}nez},
  title = {Activation Functions for Convolutional Neural Networks: Proposals and Experimental Study},
  journal = {{IEEE} Transactions on Neural Networks and Learning Systems}
}

@article{Ciuparu2020,
  doi = {10.1016/j.neucom.2019.12.014},
  url = {https://doi.org/10.1016/j.neucom.2019.12.014},
  year = {2020},
  month = apr,
  publisher = {Elsevier {BV}},
  volume = {384},
  pages = {376--388},
  author = {Andrei Ciuparu and Adriana Nagy-D{\u{a}}b{\^{a}}can and Raul C. Mure{\c{s}}an},
  title = {Soft++, a multi-parametric non-saturating non-linearity that improves convergence in deep neural architectures},
  journal = {Neurocomputing}
}

@article{Chieng2020,
  doi = {10.32890/jict.20.1.2021.9267},
  url = {https://doi.org/10.32890/jict.20.1.2021.9267},
  year = {2020},
  publisher = {{UUM} Press,  Universiti Utara Malaysia},
  volume = {20},
  author = {Hock Hung Chieng and Noorhaniza Wahid and Pauline Ong},
  title = {Parametric Flatten-T swish: an adaptive nonlinear activation function for deep learning},
  journal = {Journal of Information and Communication Technology}
}

@article{Zhang2023Fully,
  doi = {10.1155/2023/1539436},
  url = {https://doi.org/10.1155/2023/1539436},
  year = {2023},
  month = jun,
  publisher = {Hindawi Limited},
  volume = {2023},
  pages = {1--9},
  author = {Wei Zhang and Zhi Han and Xiai Chen and Baichen Liu and Huidi Jia and Yandong Tang},
  editor = {Qiang Wu},
  title = {Fully Kernected Neural Networks},
  journal = {Journal of Mathematics}
}

@inproceedings{Yang2018Square,
  doi = {10.1109/icalip.2018.8455590},
  url = {https://doi.org/10.1109/icalip.2018.8455590},
  year = {2018},
  month = jul,
  publisher = {{IEEE}},
  author = {Xiaoyu Yang and Yufei Chen and Haiquan Liang},
  title = {Square Root Based Activation Function in Neural Networks},
  booktitle = {2018 International Conference on Audio,  Language and Image Processing ({ICALIP})}
}

@incollection{Sharma2020ANovel,
  doi = {10.1007/978-981-15-5827-6_10},
  url = {https://doi.org/10.1007/978-981-15-5827-6_10},
  year = {2020},
  publisher = {Springer Singapore},
  pages = {120--130},
  author = {Ochin Sharma},
  title = {A Novel Activation Function in Convolutional Neural Network for Image Classification in Deep Learning},
  booktitle = {Data Science and Analytics}
}

@article{Deepthi2023,
  doi = {10.1007/s11227-023-05466-y},
  url = {https://doi.org/10.1007/s11227-023-05466-y},
  year = {2023},
  month = jun,
  publisher = {Springer Science and Business Media {LLC}},
  author = {M. Deepthi and G. N. V. R. Vikram and Pamulapati Venkatappareddy},
  title = {Development of a novel activation function based on Chebyshev polynomials: an aid for classification and denoising of images},
  journal = {The Journal of Supercomputing}
}

@article{Venkatappareddy2021,
  doi = {10.1016/j.dsp.2021.103093},
  url = {https://doi.org/10.1016/j.dsp.2021.103093},
  year = {2021},
  month = aug,
  publisher = {Elsevier {BV}},
  volume = {115},
  pages = {103093},
  author = {Pamulapati Venkatappareddy and Jayanth Culli and Siddharth Srivastava and Brejesh Lall},
  title = {A Legendre polynomial based activation function: An aid for modeling of max pooling},
  journal = {Digital Signal Processing}
}

@article{Zhang2023GaussianType,
  doi = {10.1016/j.neucom.2022.10.082},
  url = {https://doi.org/10.1016/j.neucom.2022.10.082},
  year = {2023},
  month = jan,
  publisher = {Elsevier {BV}},
  volume = {518},
  pages = {95--110},
  author = {Yun Zhang and Qinglong Hua and Haotian Wang and Zhenyuan Ji and Yong Wang},
  title = {Gaussian-type activation function with learnable parameters in complex-valued convolutional neural network and its application for {PolSAR} classification},
  journal = {Neurocomputing}
}

@article{Ornek2022,
  doi = {10.1016/j.neucom.2022.04.010},
  url = {https://doi.org/10.1016/j.neucom.2022.04.010},
  year = {2022},
  month = jul,
  publisher = {Elsevier {BV}},
  volume = {492},
  pages = {23--33},
  author = {B\"{u}lent Nafi \"{O}rnek and Salih Berkan Aydemir and Timur D\"{u}zenli and Bilal \"{O}zak},
  title = {Some remarks on activation function design in complex extreme learning using Schwarz lemma},
  journal = {Neurocomputing}
}

@article{Hua2021,
  doi = {10.1117/1.jrs.15.026510},
  url = {https://doi.org/10.1117/1.jrs.15.026510},
  year = {2021},
  month = may,
  publisher = {{SPIE}-Intl Soc Optical Eng},
  volume = {15},
  number = {02},
  author = {Qinglong Hua and Yun Zhang and Yicheng Jiang and Huilin Mu},
  title = {Gaussian-type activation function for complex-valued {CNN} and its application in polar-{SAR} image classification},
  journal = {Journal of Applied Remote Sensing}
}

@article{Kim2002,
  doi = {10.1023/a:1016359216961},
  url = {https://doi.org/10.1023/a:1016359216961},
  year = {2002},
  publisher = {Springer Science and Business Media {LLC}},
  volume = {32},
  number = {1/2},
  pages = {29--43},
  author = {Taehwan Kim and T\"{u}lay Adali},
  journal = {The Journal of {VLSI} Signal Processing},
  title = {Fully Complex Multi-Layer Perceptron Network for Nonlinear Signal Processing},
}

@article{Savitha2009,
  doi = {10.1016/j.neucom.2009.06.004},
  url = {https://doi.org/10.1016/j.neucom.2009.06.004},
  year = {2009},
  month = oct,
  publisher = {Elsevier {BV}},
  volume = {72},
  number = {16-18},
  pages = {3771--3781},
  author = {R. Savitha and S. Suresh and N. Sundararajan and P. Saratchandran},
  title = {A new learning algorithm with logarithmic performance index for complex-valued neural networks},
  journal = {Neurocomputing}
}

@article{Huang2008,
  doi = {10.1016/j.neucom.2007.07.025},
  url = {https://doi.org/10.1016/j.neucom.2007.07.025},
  year = {2008},
  month = jan,
  publisher = {Elsevier {BV}},
  volume = {71},
  number = {4-6},
  pages = {576--583},
  author = {Guang-Bin Huang and Ming-Bin Li and Lei Chen and Chee-Kheong Siew},
  title = {Incremental extreme learning machine with fully complex hidden nodes},
  journal = {Neurocomputing}
}

@article{Savitha2012,
  doi = {10.1016/j.neucom.2011.05.036},
  url = {https://doi.org/10.1016/j.neucom.2011.05.036},
  year = {2012},
  month = feb,
  publisher = {Elsevier {BV}},
  volume = {78},
  number = {1},
  pages = {104--110},
  author = {R. Savitha and S. Suresh and N. Sundararajan and H.J. Kim},
  title = {A fully complex-valued radial basis function classifier for real-valued classification problems},
  journal = {Neurocomputing}
}

@article{Hu2020,
  doi = {10.1016/j.neucom.2020.02.006},
  url = {https://doi.org/10.1016/j.neucom.2020.02.006},
  year = {2020},
  month = nov,
  publisher = {Elsevier {BV}},
  volume = {416},
  pages = {1--11},
  author = {Jin Hu and Haidong Tan and Chunna Zeng},
  title = {Global exponential stability of delayed complex-valued neural networks with discontinuous activation functions},
  journal = {Neurocomputing}
}

@article{Tan2018,
  doi = {10.1016/j.neucom.2017.11.047},
  url = {https://doi.org/10.1016/j.neucom.2017.11.047},
  year = {2018},
  month = jan,
  publisher = {Elsevier {BV}},
  volume = {275},
  pages = {2681--2701},
  author = {Manchun Tan and Desheng Xu},
  title = {Multiple \ce{\mu}-stability analysis for memristor-based complex-valued neural networks with nonmonotonic piecewise nonlinear activation functions and unbounded time-varying delays},
  journal = {Neurocomputing}
}

@incollection{Kuroe2003,
  doi = {10.1007/3-540-44989-2_117},
  url = {https://doi.org/10.1007/3-540-44989-2_117},
  year = {2003},
  publisher = {Springer Berlin Heidelberg},
  pages = {985--992},
  author = {Yasuaki Kuroe and Mitsuo Yoshid and Takehiro Mori},
  title = {On Activation Functions for Complex-Valued Neural Networks {\textemdash} Existence of Energy Functions},
  booktitle = {Artificial Neural Networks and Neural Information Processing {\textemdash} {ICANN}/{ICONIP} 2003}
}

@article{Ozdemir2011,
  doi = {10.1016/j.cnsns.2011.03.005},
  url = {https://doi.org/10.1016/j.cnsns.2011.03.005},
  year = {2011},
  month = dec,
  publisher = {Elsevier {BV}},
  volume = {16},
  number = {12},
  pages = {4698--4703},
  author = {Necati \"{O}zdemir and Beyza B. {\.{I}}skender and Nihal Y{\i}lmaz \"{O}zg\"{u}r},
  title = {Complex valued neural network with M\"{o}bius activation function},
  journal = {Communications in Nonlinear Science and Numerical Simulation}
}

@inproceedings{Celebi2019,
  doi = {10.36287/setsci.4.6.050},
  url = {https://doi.org/10.36287/setsci.4.6.050},
  year = {2019},
  month = jul,
  publisher = {{SETSCI}},
  author = {Mehmet {\c{C}}elebi and Murat Ceylan},
  title = {The New Activation Function for Complex Valued Neural Networks: Complex Swish Function},
  booktitle = {4th International Symposium on Innovative Approaches in Engineering and Natural Sciences Proceedings}
}

@article{Scardapane2020,
  doi = {10.1109/tetci.2018.2872600},
  url = {https://doi.org/10.1109/tetci.2018.2872600},
  year = {2020},
  month = apr,
  publisher = {Institute of Electrical and Electronics Engineers ({IEEE})},
  volume = {4},
  number = {2},
  pages = {140--150},
  author = {Simone Scardapane and Steven Van Vaerenbergh and Amir Hussain and Aurelio Uncini},
  title = {Complex-Valued Neural Networks With Nonparametric Activation Functions},
  journal = {{IEEE} Transactions on Emerging Topics in Computational Intelligence}
}

@article{Lee2022ComplexValued,
  doi = {10.1109/jas.2022.105743},
  url = {https://doi.org/10.1109/jas.2022.105743},
  year = {2022},
  month = aug,
  publisher = {Institute of Electrical and Electronics Engineers ({IEEE})},
  volume = {9},
  number = {8},
  pages = {1406--1426},
  author = {ChiYan Lee and Hideyuki Hasegawa and Shangce Gao},
  title = {Complex-Valued Neural Networks: A Comprehensive Survey},
  journal = {{IEEE}/{CAA} Journal of Automatica Sinica}
}

@article{Jagtap2023,
  doi = {10.1615/jmachlearnmodelcomput.2023047367},
  url = {https://doi.org/10.1615/jmachlearnmodelcomput.2023047367},
  year = {2023},
  publisher = {Begell House},
  volume = {4},
  number = {1},
  pages = {21--75},
  author = {Ameya D. Jagtap and George Em Karniadakis},
  title = {How Important Are Activation Functions in Regression and Classification? A Survey, Performance Comparison, and Future Directions},
  journal = {Journal of Machine Learning for Modeling and Computing}
}

\end{document}